\documentclass[10pt]{article}% [11pt]    % DRAFT at Arxiv

\usepackage{amsmath,amssymb,amsfonts,epsfig,amsfonts}
\usepackage{cite}
\usepackage{graphicx}
\usepackage{subfigure}
\usepackage{float}
\usepackage{url}
\usepackage{amsfonts,amssymb,amsmath,bm,paralist,theorem,cite,ifthen,color,amsmath}
\usepackage{theorem}
\usepackage{epsfig}
\usepackage{graphicx}
\usepackage{array}
\usepackage{caption}
\usepackage{calc}
\usepackage{longtable}
\usepackage{mathrsfs}
\usepackage{extarrows}  % For long equals
\usepackage{hyperref}
 \usepackage{txfonts}  % from TIT IA paper
\makeatletter

\title{ \bf Guaranteed Matrix Completion via Non-convex Factorization
} \vskip 1cm
\author{ %[Full Paper]  \\
Ruoyu Sun\thanks{R. Sun is with the Department of Electrical and Computer Engineering, University of Minnesota, Minneapolis, MN 55455. Email: sunxx394@umn.edu.} \text{ }and
 Zhi-Quan Luo\thanks{Z.-Q. Luo is with the Chinese University of HongKong, Shenzhen, China. He is also affiliated with
the Department of Electrical and Computer Engineering, University of Minnesota, Minneapolis, MN 55455.
Email: luozq@cuhk.edu.cn. }
\\
First version: Nov 28 2014;  This version: Oct 09 2016. 
}
\date{}
\newtheorem{lemma}{Lemma}[section]
\newtheorem{prop}{Proposition}[section]
\newtheorem{thm}{Theorem}[section]
\newtheorem{Def}{Definition}[section]
\newtheorem{coro}{Corollary}[section]

\newtheorem{claim}{Claim}[section]

\newcommand{\st}{{\rm s.t.}}
\newcommand{\tr}{{\rm Tr}}

\newcommand{\bP}{\mathcal{P}}
\newcommand{\rmax}{\rm max}
\newcommand{\rmin}{\rm min}

\newcommand{\rspan}{\mathrm{span}}
\newcommand{\red}{ \color{red}}

\newcommand{\black}{ \color{black}}
\newcommand{\dR}{\mathbb{R}}

\begin{document}
%\ninept
%
\maketitle

\begin{abstract}
Matrix factorization is a popular approach for large-scale matrix completion.
% and constitutes a basic component of many solutions for Netflix Prize competition.
% In this approach, the unknown low-rank matrix is expressed as the product of two much smaller matrices so that the low-rank property is automatically fulfilled.
The optimization formulation based on matrix factorization can be solved very efficiently by standard algorithms in practice.
However, due to the non-convexity caused by the factorization model, there is a limited theoretical understanding of this formulation.
In this paper, we establish a theoretical guarantee for the factorization formulation to correctly recover the underlying low-rank matrix.
In particular, we show that under similar conditions to those in previous works,
many standard optimization algorithms converge to the global optima of a factorization formulation, and recover the true low-rank matrix.
We study the local geometry of a properly regularized factorization formulation and prove that any stationary point in a certain local region is globally optimal.
A major difference of our work from the existing results is that we do not need resampling in either the algorithm or its analysis.
Compared to other works on nonconvex optimization, one extra difficulty lies in analyzing nonconvex constrained optimization when
the constraint (or the corresponding regularizer) is not ``consistent'' with the gradient direction. 
One technical contribution is the perturbation analysis for non-symmetric matrix factorization.

\end{abstract}

\thispagestyle{empty}
%\newpage
%\tableofcontents
%\newpage
%\newpage

%%%%%%%%%%%%%%%%%%%%%%%%%%%%%%%%%%%%%%%%%%%%%%%%%%%%%%%%%%%%%%%%%%%%%%%%%%%%%
%%%%%%%%%%%----------------   section 1 Introduction --------%%%%%%%%%%%%%%%%
%%%%%%%%%%%%%%%%%%%%%%%%%%%%%%%%%%%%%%%%%%%%%%%%%%%%%%%%%%%%%%%%%%%%%%%%%%%%%
\section{Introduction}
 In the era of big data, there has been an increasing need for handling the enormous amount of data generated by mobile devices, sensors, online merchants, social networks, etc. Exploiting low-rank structure of the data matrix is a powerful method to deal with ``big data''. One prototype example is the low rank matrix completion problem in which the goal is to recover an unknown low rank matrix $M \in \mathbb{R}^{m \times n}$ for which only a subset
  of its entries $M_{ij}, (i,j)\in \Omega \subseteq \{1,2,\dots, m \}\times \{ 1,2, \dots, n\}$ are specified. Matrix completion has found numerous applications in
  various fields such as recommender systems \cite{koren2009matrix}, computer vision \cite{chen2004recovering} and system identification \cite{liu2009interior},
  to name a few.
%  video denoising  % network anomaly detection.
  % in which the goal is to predict the missing entries in a movie rating matrix given only a subset of movie ratings. This problem can be viewed as a low rank matrix completion problem under a reasonable assumption that the rating matrix is approximately low rank.
  % One prototype example is the NetFlix problem in which the goal is to predict the missing entries in a movie rating matrix given only a subset of movie ratings. This problem can be viewed as a low rank matrix completion problem under a reasonable assumption that the rating matrix is approximately low rank.

% Matrix completion can be formulated as an optimization problem where the objective function is to
% In general, there are of an optimization formulation for the matrix completion problem is how to impose the low rank structure,
% The two most popular approaches to impose low rank structure are based on nuclear norm and matrix factorization respectively.
There are two popular approaches to impose the low-rank structure: the nuclear norm based approach and the matrix factorization (MF) based approach.
  In the first approach, the whole matrix is the optimization variable and the nuclear norm (denoted as $\| \cdot \|_*$) of this matrix variable,
  which can be viewed as a convex approximation of its rank, serves as the objective function or a regularization term.
For the matrix completion problem, the nuclear norm based formulation becomes either a linearly constrained minimization problem  \cite{candes2009exact}
\begin{equation}\label{NucMin}
\min_{Z \in \mathbb{R}^{m \times n}}  \quad \|Z\|_*, \quad
\st  \quad Z_{ij} = M_{ij}, \ \forall \ (i,j)\in \Omega,
\end{equation}
a quadratically constrained minimization problem
\begin{equation}\label{NucMin,noisy}
\min_{Z \in \mathbb{R}^{m \times n}}  \quad \|Z\|_*, \quad
\st  \sum_{(i,j)\in \Omega}(Z_{ij} - M_{ij})^2 \leq \epsilon,
\end{equation}
or a regularized unconstrained problem
\begin{equation}\label{NucMin,Regularized}
\min_{Z \in \mathbb{R}^{m \times n}}  \quad \|Z\|_* + \lambda \sum_{(i,j)\in \Omega}(Z_{ij} - M_{ij})^2.
\end{equation}
On the theoretical side, it has been shown that given a rank-$r$ matrix $M$ satisfying an incoherence condition, solving (\ref{NucMin}) will exactly reconstruct $M$ with high probability provided that $O( r (m+n)\log^2(m+n) )$ entries are uniformly randomly revealed \cite{candes2009exact,candes2010power,gross2011recovering,recht2011simpler}.
This result was later generalized to noisy matrix completion, whereby the optimization formulation \eqref{NucMin,noisy} is adopted \cite{candes2010matrix}.
Using a different proof framework, reference \cite{negahban2012restricted} provided theoretical guarantee for a variant of the formulation \eqref{NucMin,Regularized}.
%While these results are proved by constructing dual certificates, a method that has been used in compressive sensing since \cite{candes2005decoding},
%reference \cite{negahban2012restricted} applied the proof framework based on restricted
%strong convexity \cite{negahban2009unified} to the matrix completion problem and provided theoretical guarantee for a variant of the formulation \eqref{NucMin,Regularized}.
On the computational side, problems \eqref{NucMin} and \eqref{NucMin,noisy} can be reformulated as a semidefinite program (SDP) and solved to global optima
by standard SDP solvers when the matrix dimension is smaller than 500. % such as SeDuMi \cite{SeDuMi} and SDPT3 \cite{SDPT3}
To solve problems with larger size, researchers have developed first order algorithms,
including the SVT (singular value thresholding) algorithm for the formulation \eqref{NucMin} \cite{cai2010singular},
and several variants of the proximal gradient method for the formulation \eqref{NucMin,Regularized} \cite{ma2011fixed,toh2010accelerated} .
Although linear convergence of the proximal gradient method has been established for the formulation \eqref{NucMin,Regularized}{\black under certain conditions}  \cite{agarwal2012fast,hou2013linear}, the per-iteration cost of computing SVD (Singular Value Decomposition) may increase rapidly as
the dimension of the problem increases, making these algorithms rather slow or even useless for problems of huge size.
The other major drawback is the memory requirement of storing a large $m$ by $n$ matrix. % Add a remark.
%under restricted strong convexity condition \cite{agarwal2012fast} and for the formulation \eqref{NucMin,Regularized} without any condition on the unknown matrix $M$ \cite{hou2013linear}.
% The computational cost of these algorithms mainly lies in SVD (Singular Value Decomposition) at each iteration, which may increase rapidly as
% the dimension of the problem increases.

 In the second approach, the unknown rank $r$ matrix is expressed as the product of two much smaller matrices $XY^T$,
 where  $X \in \mathbb{R}^{m \times r}, Y \in \mathbb{R}^{n \times r}$, so that the low-rank requirement is automatically fulfilled.
 Such a matrix factorization model has long been used in PCA (principle component analysis) and many other applications \cite{singh2008unified}.
It has gained great popularity in the recommender systems field and served as the basic building block of many competing algorithms for the Netflix Prize
 \cite{koren2009matrix,takacs2007major} due to several reasons.  % compared to the nuclear norm based approach
 First, the compact representation of the unknown matrix greatly reduces the per-iteration computation cost as well as the storage space
 (requiring essentially linear storage of $O((m+n)r)$ for small $r$). Second, the per-iteration computation cost is rather small and people have found in practice that huge size optimization problems based on the factorization model can be solved very fast.
Third, as elaborated in \cite{koren2009matrix}, the factorization model can be easily modified to incorporate additional application-specific requirements.

 A popular factorization based formulation for matrix completion takes the form of an unconstrained regularized square-loss minimization problem \cite{koren2009matrix}:
\begin{equation}\label{P0}
\begin{split}
\mathrm{P}0: \quad \min_{X\in \mathbb{R}^{m\times r},Y \in \mathbb{R}^{n\times r}} \quad & \frac{1}{2}\sum_{(i,j)\in \Omega} [M_{ij} - (XY^T)_{ij}]^2 + \lambda(\| X\|_F^2 + \|Y \|_F^2).
\end{split}
\end{equation}
There are a few variants of this formulation: the coefficient $\lambda$ can be zero \cite{keshavan2012efficient,jain2013low,hardt2013understanding,hardt2014fast} or different for each row of $X,Y$ \cite{zhou2008large}; each square loss term $[M_{ij} - (XY^T)_{ij}]^2$ can have different weights \cite{koren2009matrix};
an additional matrix variable $Z\in \mathbb{R}^{n\times r}$ can be introduced \cite{wen2012solving}.
Problem \eqref{P0} is a non-convex fourth-order polynomial optimization problem, and can be solved to stationary points by standard nonlinear optimization algorithms such as gradient descent method, alternating minimization \cite{koren2009matrix,zhou2008large,jain2013low,hardt2013understanding} and
SGD (stochastic gradient descent) \cite{funk2006netflix,paterek2007improving,takacs2007major,koren2009matrix}.
Alternating minimization is easily parallelizable but has higher per-iteration computation cost than SGD;
in contrast, SGD requires little computation per iteration, but its parallelization is challenging.
Recently several parallelizable variants of the SGD \cite{gemulla2011large,recht2013parallel,zhuang2013fast} and
variants of the block coordinate descent method with very low per-iteration cost \cite{pilaszy2010fast,yu2012scalable} have been developed.
% Rsearchers have developed parallel algorithms with very low per-iteration cost such as various types of SGD %\cite{gemulla2011large,recht2013parallel,zhuang2013fast},  % enjoy both parallelizability and parallel or distributed
% and the CCD++ algorithm \cite{yu2012scalable}.
%The CCD++ algorithm belongs to the class of block coordinate descent method; as a comparison, alternating minimization in the context of matrix completion usually refers to a two-block coordinate descent method.
% and coordinate descent algorithm \cite{yu2012scalable}.
% In particular, reference \cite{yu2012scalable} proposed CCD++ algorithm which is a coordinate descent algorithm that updates coordinates in a specific order (as a comparison, alternating minimization in the context of matrix completion usually refers to a two-block coordinate descent method).}
 Some of these algorithms have been tested in distributed computation platforms and can achieve good performance
  and high efficiency, solving very large problems with more than a million rows and columns in just a few minutes.
  % within a few minutes for very large data sets with more than a million rows and columns.

{\black
\subsection{Our contributions}   }
Despite the great empirical success, the theoretical understanding of the algorithms for the factorization based formulation is fairly limited. % for matrix completion
More specifically, the fundamental question of whether these algorithms (including many recently proposed ones) can recover the true low-rank matrix
%(or converge to the global optima of the non-convex optimization problem)
remains largely open.
In this paper, we partially answer this question by showing that under similar conditions to those used in previous works,
many standard optimization algorithms for a factorization based formulation (see \eqref{P1}) indeed converge to the true low-rank matrix (see Theorem \ref{major theorem}).
Our result applies to a large class of algorithms including gradient descent, SGD and many block coordinate descent type methods such as two-block alternating minimization and block coordinate gradient descent. We also show the linear convergence of some of these algorithms (see Theorem \ref{theorem 2: linear convergence} and Corollary \ref{coro: linear convergence for other algorithms}).

To the best of our knowledge, our result is the first one that analyzes the geometry of matrix factorization in
Euclidean space for matrix completion. %As a result, we provide exact recovery guarantee for many standard  algorithms.
 % such as gradient descent and block coordinate gradient descent.
In addition, our result also provides the first recovery guarantee for alternating minimization without resampling (i.e.\ without using independent samples in different iterations).
Below we elaborate these two contributions in light of the existing works. % We summarize below our contributions in light of the existing works for matrix completion.

1) We analyze the local geometry
of the matrix factorization formation (in Euclidean space).
%Our result provides a validation of the formulation  rather than a validation of a single algorithm.
We argue that the success of many algorithms attributes mostly (or at least partially) to the geometry of the problem, rather than the specific algorithms being used.
The geometrical property we establish is that the local gradient direction $-\nabla f(x)$ is aligned with the global descent direction $x^* - x$.
For the classical matrix factorization formulation $\| M - XY^T \|_F^2 $, we develop a novel perturbation analysis to deal with the ambiguity of the factorization. 
For the sampling loss $\|\bP_{\Omega}(M - XY^T) \|_F^2$,
an incoherence regularizer (or constraint) is needed,
which causes an extra difficulty of analyzing nonconvex constrained optimization.
Unfortunately, projection to the constraint (or the gradient of the regularizer) is not aligned with the global direction,
and we add one more regularizer to ``correct'' the local descent direction. 
A high-level lesson is that regularization may change the geometry of the problem.

%As a result, it is easy to apply our result to many recently proposed algorithms (e.g. \cite{yu2012scalable,hastie2014matrix}), which
%are variants of classical optimization methods. Moreover, our result may be easily extended to cover new optimization algorithms that have not yet been applied
%to matrix completion, such as
%some recently proposed variants of SGD  \cite{johnson2013accelerating} \cite{roux2012stochastic}
%(we believe that the verification of these variants are not too much harder than
%their original versions, as done in Claim \ref{claim of step 1, orthogonization}).
 % (they do not require diminishing step sizes for convergence). which exhibit much better convergence behavior than vanilla SGD
% We expect that SVRG can achieve better performance than vanilla SGD in matrix completion problem.
% significantly outperformed existing SGD methods.

2) Our result applies to the standard forms of the algorithms (though our optimization formulation is a bit different),
 which do not require the additional resampling scheme used in other works \cite{keshavan2012efficient,jain2013low,hardt2013understanding,hardt2014fast}.
% The algorithms we analyze are deterministic algorithms and provide exact recovery of the unknown matrix.
% In addition, they are closer to the algorithms used in practice, in which the samples will be used multiple times. Theoretically speaking,
We obtain a sample complexity bound that is independent of the recovery error $\epsilon$,
while all previous sample complexity bounds for the matrix factorization based formulation (in Euclidean space) depend on $\epsilon$.
%In other words, we prove the \emph{exact recovery} of  the original matrix, while previous works for matrix factorization only prove the \emph{inexact recovery}.
% It is common to see the appearance of $\log(1/\epsilon )$ in the computational complexity of iterative methods
% (e.g. in solving linear system of equations, in which case the best bound independent of $\log(1/\epsilon)$ is unknown).
% However, for the sample complexity, the additional factor of $\log(1/\epsilon)$ is mostly due to technical reasons.
There is a subtle theoretical issue for the resampling scheme; see more discussions in Section \ref{sec: Comparison with other non-convex schemes} and \cite[Sec. 1.5.3]{sun2015thesis}.
 % and Appendix \ref{appen: discussion of random models}.

% if they start from certain specific initial point.
%More precisely, we prove the following result: consider a variant of problem (\ref{P0}) and any algorithm that converges to stationary points of this problem, then starting from some specific initial point this algorithm must converge to the global optima (thus reconstruct $M$ exactly) with high probability
%under some standard assumptions (the number of random observed entries is sufficiently large and $M$ is incoherent).

\vspace{0.5cm}
\subsection{Related works }\label{sec: Comparison with other non-convex schemes}
% (the coupling of the indicator variables)
% In the original proof of   One of the main techniques that lead to the simplification of the proof is the . which requires intricate tools to handel
% Along an independent line, the field of recommendation systems focuses on
% Although the factorization model was already very popular for huge problems \cite{koren2009matrix},
%and alternating minimization and SGD are two most popular methods

\textbf{Factorization models.}
The first recovery guarantee for the factorization based matrix completion is provided in \cite{keshavan2010matrix},
where Keshavan, Montanari and Oh considered a factorization model in Grassmannian manifold and showed that
 the matrix can be recovered by a proper initialization and a gradient descent method on Grassmannian manifold.
% %------delete since too complex -----------------
% {\blue In their formulation, the unknown rank $r$ matrix is expressed as the product of three smaller matrices $XSY^T$ with $X \in \dR^{m \times r}, XX^{T} = mI $, $Y \in \dR^{n \times r}, YY^{T}= nI$, and the objective function $ F(X,Y) \triangleq \min_{S \in \mathbb{R}^{r \times r} } \sum_{(i,j)\in \Omega}[M_{ij} - (XSY^T)_{ij} ]^2 $. Since this $F(X,Y)$ depends on $X,Y$ only through their column spaces, it
%   can be viewed as a function on Grassmannian manifolds (a Grassmannian manifold consists of linear subspaces of a certain dimension).
%   The resulting optimization problem $\min_{X,Y} F(X,Y)$ is in fact a two-level optimization problem. }
% ({\red ??? Question: if delete this blue part, then the role of $S$ and the drawback of "3-factor model" claimed later is unclear.  }
%    %-----------------------
  Besides being quite complicated, this model is not as flexible as the factorization model in Euclidean space, % to incorporate application-specific features,
  and it is not easy to solve by many advanced large-scale optimization algorithms. % in Euclidean space.
  Moreover, most algorithms in Grassmann manifold require line search, and little is known about the convergence rate.

The factorization model in Euclidean space was first analyzed in an unpublished work \cite{keshavan2012efficient} of Keshavan
\footnote{  Reference \cite{keshavan2012efficient} is a PhD thesis that discusses various algorithms including the algorithm proposed in \cite{keshavan2010matrix}
  and alternating minimization. In this paper when we refer to \cite{keshavan2012efficient}, we are only referring to
  \cite[Ch. 5]{keshavan2012efficient} which presents resampling-based alternating minimization and the corresponding result.},
as well as a later work of Jain et al. \cite{jain2013low}.
Both works considered  alternating minimization with resampling scheme, a special variant of the original alternating minimization.
%There are a few very subtle issues with the resampling scheme,  and we will discuss these issues in the next few paragraphs.
%% some of which have not been explicitly recognized in the literature,
%These two works used the same algorithm and a similar proof framework (reducing the distance between the column spaces of the iterates and the true factors), though the proof details and the sample complexity bounds are different.
The sample complexity bounds were later improved by Hardt \cite{hardt2013understanding} and Hardt and Wooters \cite{hardt2014fast},
where in the latter work, notably, the authors devised an algorithm with a corresponding sample complexity bound independent of the condition number.
However, these improvements are obtained for more sophisticated versions of resampling-based alternating minimization,
not the typical alternating minimization algorithm.

\textbf{Resampling.}
The issues of resampling have been discussed in a recent  work on phase retrieval by Cand{\`e}s et al.\cite{candes2014phase}.
% Here we make some additional comments; and
We will point out a subtle theoretical issue not mentioned in \cite{candes2014phase}, as well as some other practical issues.

The resampling scheme (a.k.a. golfing scheme \cite{gross2011recovering}) can be used at almost no cost for the nuclear norm approach \cite{gross2009quantum,gross2011recovering,recht2011simpler},
but for the alternating minimization it causes many issues.
At first, it may seem that for both approaches resampling is a cheap way to get around a common difficulty: the dependency of the iterates on the sample set.
 % is a common difficulty encountered in analyzing nuclear norm approach and the factorization model.
However, there is a crucial difference: for the nuclear norm approach, resampling is just a proof technique used in a ``conceptual'' algorithm for constructing the dual certificate, while for the alternating minimization, resampling is used in the actual algorithm.
This difference causes some issues of resampling-based alternating minimization at conceptual, practical and theoretical levels.

1) Gap between theory and algorithm.  Algorithmically, an easy resampling scheme is to
% resampling-based alternating minimization in \cite{keshavan2012efficient,jain2013low}
randomly partition the given set $\Omega$ into non-overlapping subsets $\Omega_k, k=1,\dots, L$, as proposed in \cite{keshavan2012efficient,jain2013low}
\footnote{The description in \cite{jain2013low} has some ambiguity and
 it might refer to the scheme of sampling $\Omega_k$'s with replacement; anyhow, under this model $\Omega_k$'s are still dependent.
  See \cite[Sec. 1.5.3]{sun2015thesis} for more discussions.  }. % Appendix \ref{appen: discussion of random models}.
  However, the results in \cite{keshavan2012efficient,jain2013low,hardt2013understanding,hardt2014fast} actually require
  a generative model of independent $\Omega_k$'s, instead of sampling $\Omega_k$'s based on a given $\Omega$.
Therefore, the results in \cite{keshavan2012efficient,jain2013low,hardt2013understanding,hardt2014fast}
do not directly apply to the partition based resampling scheme that is easy to use.
See \cite[Sec. 1.5.3]{sun2015thesis} for more discussions on this subtle issue.

% moreover, it is not clear whether there exists a simple resampling scheme based on a given $\Omega$ to which these results apply.
%This issue has been discussed in \cite[Section D]{hardt2014fast};
%we will provide more detailed discussions in Appendix \ref{appen: discussion of random models}.
  % to be statistically independent (possibly overlapping),
% However, these $\Omega_k$'s are not statistically independent, and do not satisfy the requirement

% One issue is that resampling-based alternating minimization in \cite{keshavan2012efficient,jain2013low}
% uses sampling \emph{without} replacement \footnote{The description in \cite{jain2013low} has some ambiguity and
% it might refer to random model 3 rather than random model 1; anyhow, both models have not been validated. See more discussions in Appendix \ref{appen: discussion of random models}.  } but the results in \cite{keshavan2012efficient,jain2013low} require sampling \emph{with} replacement.
 % To be more precise, \cite{jain2013low} wrote
 %''Partition $\Omega$ into $2T+ 1$ subsets $\Omega_0,¡¤¡¤¡¤, \Omega_{2T}$ with each
%element of $\Omega$ belonging to one of the $\Omega$ with equal probability (sampling with replacement)''.
%There is a contradiction here: if $\Omega $ is partitioned into $ \Omega_0 \cup \dots \cup \Omega_{2T}$,
%then it is sampling \emph{without} replacement (see ``random model 1'').
%For the sampling with replacement model, see ``random model 3'' in Appendix \ref{appen: discussion of random models}.
%Other works, such as \cite{keshavan2010matrix} and \cite{hardt2013understanding}
% },

This issue has been discussed by Hardt and Wooters in \cite[Appendix D]{hardt2014fast},
and they proposed a new resampling scheme  \cite[Algorithm 6]{hardt2014fast} to which the results in \cite{keshavan2012efficient,jain2013low,hardt2013understanding,hardt2014fast} can apply,
provided that the generative model of $\Omega$ is exactly known.
In practice, the underlying generative model of $\Omega$ is usually unknown, in which case the scheme \cite[Algorithm 6]{hardt2014fast} does not work.
%  it is not clear whether it is even possible to generate independent $\Omega_k$'s based on a given $\Omega$ from an unknown distribution.
% It is not clear whether there exists a resampling scheme based on a given $\Omega$ from an unknown distribution to which these results apply.
In contrast, the classical results in \cite{candes2009exact,candes2010power,recht2011simpler,gross2011recovering} and our result herein are robust to the generative model of $\Omega$: these results actually state that for an overwhelming portion of $\Omega$ with a given size, one can recover $M$ through a certain algorithm, thus for many reasonable probability distributions of $\Omega$ a high probability result holds.

2) Impracticality.
As argued previously, assuming a generative model of $\Omega_k$'s is not practical since $\Omega$ is usually given.
For given $\Omega$, the only known validated resampling scheme \cite[Algorithm 6]{hardt2014fast}, besides not being robust to the underlying generative model of $\Omega$, might be  a bit  complicated to use in practice.  Even the simple resampling scheme of partitioning $\Omega$ (which has not been validated yet) is rather unrealistic since each sample is used only once during the algorithm.
% First, each sample is used only once during the algorithm, which might not be desirable in practice
% since data are usually limited. %  and practitioners would prefer to use them repeatedly in the algorithm.
%Second, different accuracy requirements will lead to different pre-partition of the samples, and thus different forms of the algorithm.
%If the algorithm has produced an estimate of $M$ and one asks for a more accurate estimate, then one has to re-partition $\Omega$ and re-run the algorithm from the beginning.

 % since depends on the accuracy requirement. (even though it is more practical than random model 1)
% and exact recovery requires a sequence of algorithms;
3) Inexact recovery. %  or infinite sample complexity.
 A theoretical consequence of the resampling scheme is that the required sample complexity $|\Omega|$ becomes dependent on the desired accuracy $\epsilon$, and goes to infinity as $\epsilon$ goes to zero. This is different from the classical results (and ours) where exact reconstruction only requires finite samples.
While it is common to see the dependency of \emph{time complexity} on the accuracy $\epsilon$, it is relatively uncommon to see the dependency of \emph{sample complexity} on $\epsilon $.

{\black
In a recent work \cite{jain2014fast} the authors have managed to remove the dependency of the required sample size on $\epsilon$ by using a singular value projection algorithm.
However, \cite{jain2014fast} considers a matrix variable of the same size as the original matrix, which requires significantly more memory than
% may not have the same advantage in memory and modeling flexibility as provided by
 the matrix factorization approach considered in this paper.
Moreover, it requires resampling at a number of iterations (though not all), which may suffer from the same issues we mentioned earlier.
% is difficult to implement. % and impractical.
The resampling is also required in the recent work of \cite{de2014global}; see \cite[Sec. 1.5.3]{sun2015thesis} for more discussions.
}
% Moreover, the algorithm in \cite{jain2014fast} still uses independent samples in a number of iterations (though not all iterations),

%

{\red

}

\textbf{Other works on non-convex formulations.}
Non-convex formulation has also been studied for the phase retrieval problem in some recent works \cite{netrapalli2013phase,candes2014phase}.
% Similar to the works on matrix completion, these works
These works provide theoretical guarantee for some algorithms specially tailored to certain non-convex formulations and with specific initializations.
The major difference between \cite{netrapalli2013phase} and \cite{candes2014phase} is that
the former requires independent samples in each iteration, while the latter  uses the same samples throughout in the proposed algorithm.
% does not need this assumption and
As mentioned earlier, such a difference also exists between all previous works on alternating minimization for matrix completion \cite{keshavan2012efficient,jain2013low,hardt2013understanding,hardt2014fast} and our work.
% For more discussions on the trick of using independent samples, we refer the readers to \cite{candes2014phase}.
% These works establish the global convergence of non-convex formulation
%  In particular, reference \cite{netrapalli2013phase} establishes
% Finally, we would like to mention another line of works on

Finally, we note that there is a growing list of works on the theoretical guarantee of non-convex formulations for various problems, such as sparse regression
(e.g. \cite{zhang2012general,loh2013regularized,fan2014strong}), sparse PCA \cite{yuan2013truncated,wang2014nonconvex}, robust PCA \cite{netrapalli2014non} and EM (Expected-Maximization) algorithm \cite{balakrishnan2014statistical, wang2014high}.
We emphasize several aspects that distinguish our paper from other recent works on non-convex optimization.
First, our paper is one of the first to analyze the (local) geometry of the problem. Second, we deal with non-symmetric matrix factorization which has a more bizarre geometry than symmetric matrix factorization and some other models. 
Third, one difficulty of our problem essentially lies in nonconvex constrained optimization (though we consider the closely related regularized form). 

%Many more works emerge after the appearance of the first version of this paper,
%such as dictionary learning
% The techniques used in these works, however, seem to be quite different from those in the current work.

{\black  % ## for Luo to check
\subsection{Proof Overview and Techniques}\label{sec: proof overview}

  \textbf{Basic idea: local geometry.} The very first question is what kind of property can ensure global convergence for non-convex optimization.
We will establish a local geometrical property of a regularized objective such that any stationary point in a local region is globally optimal.
This is achieved in three steps:
% Intuitively, the following framework is enough:
% (i) the objective function is locally convex around global minima; (ii) all iterates stay in the local region.
% However, there are a few obstacles to apply this framework to our problem:
 (i) study the local geometry
 of the fully observed objective $\| M - XY^T \|_F^2$;
(ii) study the local geometry
of the matrix completion objective
$ \| \bP_{\Omega}(M - XY^T)\|_F^2$;
% the local region may be very small;
(iii) study the local geometry
of a regularized objective.
% constraining iterates in a local region changes the optimality condition.  % by projection causes some extra issue.
Next, we will discuss the difficulties involved in each step and describe how we address these difficulties.
 % This basic framework is the foundation of our proof  (in fact, isolated local minimum suffices)

\vspace{0.3cm}
\textbf{Local geometry of $\|M - XY^T \|_F^2$}.
We start by considering a simple case that $M$ is fully observed and the objective function is $f(X,Y) = \|M - XY^T \|_F^2$.
What is the geometrical landscape of this function?
In the simplest case $m=n=r = 1$ and $f(x,y) = (xy - 1)^2$, the set of stationary points is $\{ (x,y) \mid xy = 1 \} \cup \{ (0,0) \}$, in which $(0,0)$ is a saddle point and the curve $xy = 1$ consists of global optima.
We plot the function around the curve $xy=1$ in the positive orthant in Figure \ref{FigLocalGeo}.
\begin{figure}[ht]
  \centering
{ \includegraphics[width=7cm,height=5cm]{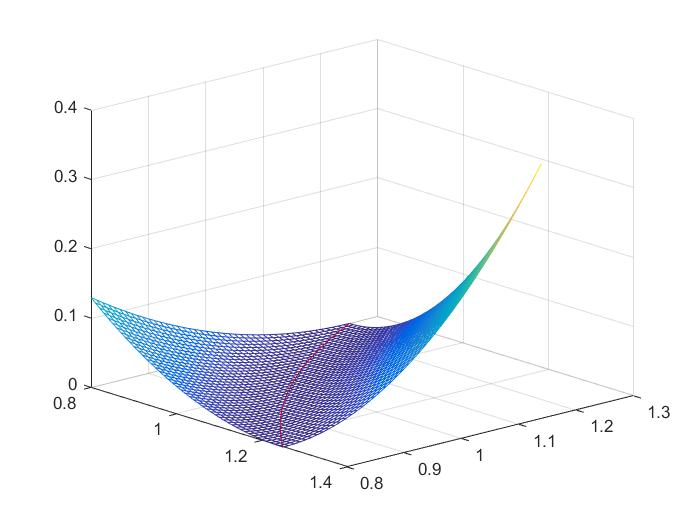} }
  \vspace{-.2cm}
   \caption{
   The plot of function $f(x,y) = (xy-1)^2$
   around the set of global optima $xy = 1$
   in the positive orthant.
   The bottom of this bowl shape
   is a hyperbola $xy = 1$.
   }  \label{FigLocalGeo} \vspace{-.3cm}
\end{figure}

Clearly a certain geometrical property prevents bad local minima in the neighborhood of the global optima, but what kind of property?
We emphasize that the property can \textit{not} be local convexity because the set of global optima is non-convex in $\dR^2$.
% This property should be related to the convexity of the function $(1 - z)^2$.
% and this convexity is partially preserved
% when $z$ is reparameterized into $z = xy$.
% Since $f(x,y)$ is invariant under transformation $(x,y) \rightarrow (x q, y q^{-1})$,
% One conceptual solution is
% to view $f(x,y) = (1-xy)^2$ as a function defined over a quotient space $(\dR \times \dR \backslash \dR^* )$ where $\dR^* = R\backslash \{ 0\} $.
Due to the intrinsic symmetry that
$f(x,y) = f(xq, yq^{-1})$, only the product $z = xy$ affects the value of $f$.
We hope that the strong convexity of $(1-z)^2$ can be partially preserved
when $z$ is reparameterized into $z = xy$.
It turns out we can prove the following local convexity-type property:
for any $(x,y)$ such that $xy$ is close to $1$ and $|x|,|y|$ are upper bounded, there exists
$uv = 1$ such that
$$
 \langle \nabla f(x,y), (x,y) - (u,v) \rangle \geq c \|(x,y) - (u,v) \|^2. $$
An interpretation is that the
negative gradient direction $-\nabla f$ should be
aligned with the global direction $(u,v) - (x,y)$; a convex function has a similar property, but the difference is that here the global direction is adjusted according  to the position of $(x,y)$.

%because of the symmetry.

For general $m,n,r$, the geometrical landscape is probably much more complicated
than the scalar case.
% we can view the function $f(X,Y) = \frac{1}{2}\| \bP_{\Omega}(M - XY^T)\|^2$
%as defined over a quotient manifold $\text{GL}(m , r) \times GL(n , r)/\mathrm{GL}(r, r)$,
% the element of which is an equivalent class $c(Z) = \{ (X,Y) \mid XY^T = Z \}$, where $\mathrm{GL}(k, r)$ is the general linear group consisting of $k \times r$ matrices
% \footnote{While this manifold-type perspective seems intuitive, it does not lead to a concrete proof partially because
%that quotient space is not easy to analyze.}.
Nevertheless, we can still prove that the convexity
of $\| M - Z\|^2$ is partially preserved when reparameterizing $Z$ as $Z = XY$.
The exact expression is a variant of
\eqref{LSC,1st step} which we will discuss
in more detail later.
% Then the function $f(X,Y)$ has a unique global minimum $c(M)$ -- thus isolated -- on the quotient manifold, making LSC on the quotient manifold possible.
% \footnote{The notion of (geodesic) convexity on a Riemann manifold is tricky. It is well known that any convex function on a compact
% manifold is a constant function.}.
% Nevertheless, it is not clear how to define a natural metric on the quotient manifold,
% and we still have to perform the analysis in the Euclidean space $\dR^{m\times r} \times \dR^{n \times r}$.
Technically, we need to connect the Euclidean space and the quotient manifold
via ``coupled perturbation analysis'': given $X,Y$ such that $\| XY^T - M \|_F$ is small,
find decomposition $M = UV^T$ such that $U,V$ are close to $X$ and $Y$ respectively (a simpler version of Proposition
\ref{prop1}).
The difference from traditional perturbation analysis of Wedin \cite{wedin1972perturbation} (i.e.\ if two matrices are close then
their row/column spaces are close) is that in \cite{wedin1972perturbation} the row/column spaces are fixed  % and have closed-form expression,
while in our problem $U,V$ are up to our choice.
%As a result, \cite{wedin1972perturbation} only requires a ``verification'' proof that bounds a given error,
%while we need a ``constructive'' proof that designs a factorization $M = UV^T$ and shows it works.
%Naive factorizations of $M$ such as SVD does not work; in fact, we need to factorize $M = UV^T$ according to the structure of $X$ and $Y$.
%For Proposition \ref{prop1}, utilizing a coarse structure of $X ,Y $ is enough.

\vspace{0.3cm}
\textbf{Local geometry of $ \| \bP_{\Omega}(M-XY^T)\|_F^2$}.
Let us come back to the original matrix completion problem, in which an additional sampling operator $\bP_{\Omega}$ is introduced.
Similarly, we hope that
$f_{\Omega}(Z) = \frac{1}{2} \|\bP_{\Omega}(M - Z) \|^2$ is strongly convex and this strong convexity can be partially preserved after reparametrization $Z = XY^T$. However, one issue is that the function $f_{\Omega}(Z)$ is possibly non-strongly-convex (though still convex).
In fact, if $f_{\Omega}$ is locally strongly convex around $M$, then we should have
$$
f_{\Omega}(Z) - f_{\Omega}(M) \geq O( \|Z - M \|_F^2 ), \forall \ Z \text{ close to }M.
$$
Assuming $Z$ is rank-$r$, this inequality can be rewritten as
 \begin{equation}\label{LSC}
 \|\bP_{\Omega}(M - XY^T) \|_F^2 \geq  C p \| M- XY^T \|_F^2, \ \forall (X,Y) \in K(\delta) ,
 \end{equation}
where $K(\delta)$ is a neighborhood of $M$ defined as $\{ (X,Y) \mid  \| XY^T - M \|_F \leq \delta \} $ and $C$ is a numerical constant.
We wish \eqref{LSC} to hold with high probability (w.h.p.) for random $\Omega$ in which each position in $M$ is chosen with probability $p$.
This inequality is closely related to matrix RIP (restricted isometry property) in \cite{candes2010matrix} (see equation (III.4) therein).
If $X,Y$ are independent of $\Omega$, then
\eqref{LSC} follows easily from the concentration inequalities.  Unfortunately, if $X,Y$ are chosen arbitrarily instead of independently from $\Omega$, the bound \eqref{LSC} may fail to hold.

%\textbf{Size of local region.} Convexity in a ``local'' region is not enough; the size of the region matters because
% a small region would make it difficult to find an initial point in that region. How large should the size of the local region be?

A solution, as employed in \cite{keshavan2010matrix}, is to utilize a random graph lemma in \cite{feige2005spectral}
 which provides a bound on $\|\bP_{\Omega}(A) \|_F $ for any rank-$1$ matrix $A$ (possibly dependent on $\Omega$).
This lemma, combined with another probability result in \cite{candes2009exact}, implies a bound on $ \| \bP_{\Omega}(M - XY^T) \|_F $. However, this bound is not good enough since it only leads to \eqref{LSC} when $ \delta = O(1/n) $.
The underlying reason is that the bound given by the random graph lemma is actually quite loose if
 $X $ or $Y$ have unbalanced rows, i.e. certain row has large norm.
One solution is to force the iterates to have bounded row norms (a.k.a. incoherent), by adding a constraint or regularizer.
With the incoherence requirement on $X,Y$, now \eqref{LSC} can be shown to be hold for $ \delta = O(1) $, or more precisely, $\delta = O(\Sigma_{\min})$, where $\Sigma_{\min}$ is the minimum eigenvalue of $M$.  With such a $\delta$, it is possible to find an initial point in the region $K(\delta)$.
% Then a properly chosen initial point falls into this region.

In summary, although $f_{\Omega}(Z) =
\frac{1}{2} \|\bP_{\Omega}( Z - M ) \|_F^2 $ is possibly non-strongly-convex, by restricting to
an incoherent neighborhood of $M$ it is ``relative'' strongly convex (called ``relative'' since we fix $M$ in \eqref{LSC}).
More specifically, we have
that w.h.p.
  \begin{equation}\label{LSC,1st step}
 \|\bP_{\Omega}(M - XY^T) \|_F^2 \geq  C p \| M- XY^T \|_F^2, \ \forall (X,Y) \in \mathcal{B} \triangleq K(\delta) \cap K_1.
\end{equation}
where $K_1 $ denotes the set of $(X,Y) $ with bounded row norms.
Note that this inequality also implies
that global optimally in $\mathcal{B}$ leads to exact recovery; or equivalently,
zero training error leads to zero generalization error.

Having established the geometry
of $f_{\Omega}(Z)$, we can use the same technique for the fully observed case to show the local geometry  \footnote{
For illustration purpose, we present a two-step approach: first establish a geometrical property of $f_{\Omega}(Z)$, then extend the property to $f_{\Omega}(XY^T)$.
However, our current proof does not follow the two-step approach but directly establish the property of $f_{\Omega}(XY^T)$.
%If the two steps are independent, then we may study other functions $f$ instead of $f_{\Omega}$.
%% satisfies a geometrical property.
%However, we can directly prove the property of $f_{\Omega}(XY^T)$
In fact, although we establish the property
of $f_{\Omega}(Z)$ in Claim \ref{lemma: P Omega and P has relation},
the proof of this claim is very similar to the proof of
\eqref{intro, LSC 1st}.
} of  $$ F(X,Y) \triangleq
f_{\Omega}(X,Y) = \frac{1}{2} \| \bP_{\Omega}(M - XY^T) \|_F^2 .$$

More specifically, we can prove that for any $(X,Y) \in \mathcal{B}$, there exists $(U,V) \in
\mathcal{X}^* = \{ (U,V) \in \dR^{m \times r} \times \dR^{n \times r} \mid UV^T = M  \} $ such that
% for any $(X,Y)$ in a certain  % Lemma \ref{lemma main; about local convexity} is
\begin{equation}\label{intro, LSC 1st}
  \langle \nabla_X F(X,Y), X- U \rangle
   +  \langle \nabla_Y F(X,Y), Y - V \rangle \geq  c (\| X - U\|_F^2 + \| Y - V\|_F^2 ).
\end{equation}
% where $F(X,Y) =\frac{1}{2} \|\bP_{\Omega}(M - XY^T) \|_F^2  $.
% is the objective function, and $\mathcal{X}^* = \{ (U,V) \mid UV^T = M, U \in \dR^{m \times r}, V \in \dR^{n \times r} \}  $ can be viewed as
%the set of global optimizers.
Denoting $\bm x = (X,Y), \bm x^* = (U,V)$ and utilizing $\nabla F(\bm x^*) = 0$,
 \eqref{intro, LSC 1st} becomes
 \begin{equation}\label{intro, LSC 1st simplified}
   \forall \ \bm x \in \mathcal{B}, \ \exists \ \bm x^* \in \mathcal{X}^*, \ \text{ s.t.} \  \langle \nabla F(\bm x) - \nabla F(\bm x^*), \bm x - \bm x^* \rangle \geq c \| \bm x - \bm x^* \|^2.
\end{equation}
It links the local optimality measure $\| \nabla F(\bm x)\| $ with the global optimality measure
 $\text{dist}(\bm x, \mathcal{X}^*) = \min_{\bm x^* \in \mathcal{X}^*} \|\bm x - \bm x^* \|$, and implies that any stationary point of $F$ in $\mathcal{B} $  is a global minimum.
 % satisfies $ \text{dist}(\bm \hat{x}, \mathcal{X}^*) = 0 $, i.e. $\hat{x}$

% The problem property \eqref{intro, local relative convexity} is related to ``local strong convexity'' but not quite the same.
If \eqref{intro, LSC 1st simplified} holds for arbitrary $\bm x, \bm x^*$ then $F$ would be strongly convex in $\bm x$.
Let us emphasize again two differences
of \eqref{intro, LSC 1st simplified} with local strong convexity:
i) since $\bm x^*$ is not arbitrary but has to be one global minimum, \eqref{intro, LSC 1st simplified}
indicates local ``relative convexity'' of $F$;
ii) due to the ambiguity of factorization, $\bm x^*$ should be chosen according to $\bm x$, thus \eqref{intro, LSC 1st simplified}
indicates local relative convexity up to a group transformation (it might be conceptually helpful to view it as a property in the quotient manifold, but we do not explicitly exploit its structure).

%In a high level, the effect of random sampling on the local geometry is that the size of the local region is shrunk to $O(1/n)$. By adding a restriction on
%the maximum row-norm, the size of the local region is increased to $O(1)$.

% if there is only one global minimizer $\bm x^*$ and we restrict $\bm x$ to be close to $\bm x^*$, then \eqref{intro, LSC 1st simplified} can be viewed as ``relative LSC'' (LSC relative to $\bm x^*$).
% In our problem $\mathcal{X}^*$ is a nonconvex uncountable set, thus \eqref{intro, LSC 1st simplified} is not LSC -- at least in Euclidean space.
% As mentioned before, conceptually we can view it as relative LSC in the quotient manifold.
%

%To summarize, up to now we resolve two issues to establish a geometrical property:
%(a) LSC does not hold in any region in the Euclidean space, thus we view the function as in the quotient manifold. To connect the
%Euclidean space and the quotient manifold, we establish coupled perturbation analysis.
%(b) The LSC region is too small because of the dependence of iterates on the samples. Adding row-norm constraints enlarges the local region
%so that a properly chosen initial point falls into that region.

\vspace{0.3cm}
\textbf{Local geometry with regularizers/constraints.}
The property \eqref{intro, LSC 1st simplified} is still not desirable.
 The original purpose of studying geometry is to show there is no spurious ``1st order local-min'' (point that satisfies 1st order optimality conditions).
To establish the geometrical property with sampling,
 we restrict to an incoherent set $K_1$, but this restriction changes the meaning of the 1st order local-min.
 In fact, to ensure
the iterates stay in the incoherent region $K_1$, we need to solve a constrained optimization problem $\min_{\bm x \in K_1} F(\bm x)$ or a regularized problem
$\min_{\bm x}  F(\bm x) + G_1(\bm x)$ where $G_1$ is a regularizer forcing $\bm x$ to be in $K_1$.
% The first approach seems difficult, so we adopt the second approach.  % As the simulation in \cite{sun2015thesis} shows, while
Standard optimization algorithms converge to the KKT points of
$ \min_{\bm x \in K_1} F(\bm x) $ or the stationary points of $F + G_1$, which
may not be the stationary points of
$F$.
The property \eqref{intro, LSC 1st simplified} only implies any stationary point of $F$ in $\mathcal{B}$ is globally optimal.

 % : it should be the KKT point with possibly non-vanishing gradient.
% for unconstrained optimization the 1st order point has a vanishing gradient, but
% Standard optimization algorithms
% usually converge to the KKT point, not a point with vanishing gradient.
 % \footnote{
% For a specific algorithm, it is possible to prove convergence to a point in the interior of the feasible region
% in which case $\nabla F = 0$ (e.g. in related works \cite{chen2015fast,zheng2016convergence} appeared after this work)..

%\footnote{A stationary point of $F$ in $\mathcal{B} = K_1 \cap K(\delta)$ may be found by  specific algorithm for the current problem though.
% Here we are interested in an algorithm-independent analysis which has minimal assumptions on the algorithm, thus
% we do not strive to show an algorithm can find a stationary point of $F$.
}

% It remains to guarantee that all iterates stay in $\mathcal{B}$.  We can prove that the spectral method plus clipping the large rows can provide an initial point in $\mathcal{B}$, thus the challenge is to prevent the iterates from leaving $\mathcal{B}$.

% There are two possible approaches: (i) prove that some algorithms for minimizing $F$ naturally produce iterates in $\mathcal{B}$;
%We aim to separate the geometry and the algorithm, thus choose a possibly more difficult approach of analyzing
%the stationary points of $F + G_1$.}.

\begin{figure}[ht]
  \centering
{ \includegraphics[width=4.5cm,height=3.5cm]{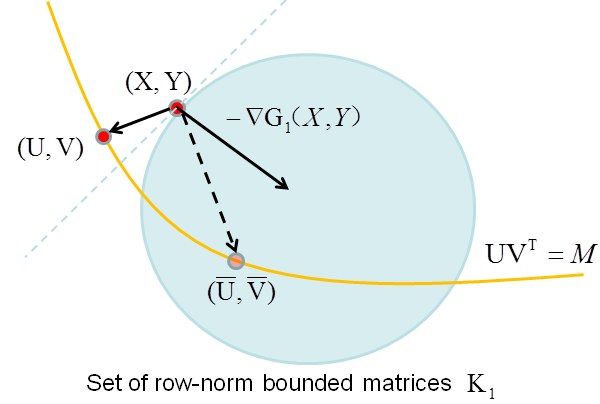} }
  \vspace{-.2cm}
   \caption{ {\small Illustration of why a single regularizer $G_1$ is not enough. The requirement \eqref{G1 bound analysis} means $-\nabla G_1(\bm x) = -\nabla G_1(X,Y)$ is positively correlated with $\bm x^* -  \bm x$.
   This holds if we could pick some $\bm x^* = (\bar{U}, \bar{V})$ lying in the row-bounded region $K_1$.
    However, we need to choose $\bm x^* = (U,V)$ in the hyperbolic space $\{(U,V) \mid UV^T = M \}$ that is close to $(X,Y)$.
    The figure indicates that such a $(U,V)$ may be outside of $K_1$, and $(U,V) - (X,Y)$ may be negatively correlated with
    $ -\nabla G_1(X,Y)$. }
   }  \label{FigProjK} \vspace{-.3cm}
\end{figure}

We shall focus on the regularized problem $\min F+G_1$; the constrained problem
$\min_{\bm x\in K_1} F$ is similar.
% \footnote{%For the constrained problem, standard algorithms converge to a point $\bm \bar{x}$ that satisfies the KKT condition.
% We will chose $G_1$ to be a penalty function, and it is easy to argue that the KKT condition for the constrained problem is closely related to $\nabla F + \nabla G_1 = 0$.
%The technical challenge for the regularized problem and the constrained problem is quite similar.
Because of the extra regularizer, the property \eqref{intro, LSC 1st simplified} is not enough.  We need to
 prove a result similar to \eqref{intro, LSC 1st simplified}, but with $\nabla F$ replaced by $\nabla F + \nabla G_1$:
 \begin{equation}\label{intro, LSC 2nd simplified}
   \forall \ \bm x \in \mathcal{B}, \ \exists \ \bm x^* \in \mathcal{X}^*, \ \text{s.t.} \  \langle \nabla F(\bm x) +
   \nabla G_1(\bm x), \bm x - \bm x^* \rangle \geq c \| \bm x - \bm x^* \|^2.
\end{equation}
If it happens to be the case that
\begin{equation}\label{G1 bound analysis}
\langle \nabla G_1(\bm x) , \bm x - \bm x^* \rangle \geq 0 ,
\end{equation} then combining with the existing result \eqref{intro, LSC 1st simplified} we are done;
unfortunately, we do not know how to prove \eqref{G1 bound analysis}.    % and we suspect it is wrong based on the simulation.
Intuitively, \eqref{G1 bound analysis} means that  $-\nabla G_1(\bm x)$, which is almost the same direction as the projection to the incoherent region $K_1$, is positively correlated with the global direction $\bm x^* - \bm x$.
At first sight, this seems trivially true
because for any point $ \bar{ \bm x} \in K_1 $
we have
$\langle \nabla G_1(\bm x), \bm x -  \bar{\bm x} \rangle \geq 0$
 (as illutrated in Fig. \ref{FigProjK}).
However, a rather strange issue is that $\bm x^*$ is chosen to be a point in $\{(U,V) \mid UV^T = M \}$ that is close to $\bm x$,
thus there is no guarantee that $\bm x^*$ lies in $K_1$.
An underlying reason is that the global optimum set $\{(U,V) \mid UV^T = M \}$ is unbounded and thus not a subset of $K_1$.
If we enforce $(U,V)$ to be in $K_1$, we may not be able to find $(U,V)$ that is close enough to $(X,Y)$.
% To resolve this issue, we add a regularizer to further bound the norms of $X,Y$.
%   In a high level, this issue is a side effect of the two techniques:
 % we try to prove a geometrical property
 % in a quotient manifold
 % but we have to impose the row-norm constraints in the Euclidean space.

Technically, the issue is that $(U,V)$ chosen in Proposition \ref{prop1}
have row-norms bounded above by quantities proportional to the norms of $X,Y$, and can be higher than the row-norms of $X,Y$ (threshold of $K_1$).
To resolve this issue, we add an extra regularizer $G_2(X,Y)$ to force $(X,Y)$ to lie in $K_2$, a set of matrix pairs with bounded norms. % to be bounded. % Note that the row-norm bounds on $X,Y$ already imply a bound on the norm of $X,Y$, but here we require an even stringent bound.
This extra bound makes $\langle \nabla G_1(\bm x) , \bm x - \bm x^* \rangle \geq 0 $ straightforward to prove,
but a similar issue arises: now we need to prove \eqref{intro, LSC 1st simplified} for $F + G_1 + G_2$ instead of $F$.
Again, it suffices to prove that for any $\bm x \in K(\delta) \cap K_1 \cap K_2$ there exists $\bm x^*$ such that
\begin{equation}\label{G2 bound analysis}
\langle \nabla G_2(\bm x) , \bm x - \bm x^* \rangle \geq 0 .
\end{equation}
This is what we prove as outlined next.
% If this inequality did not hold, we might need additional regularizers and this process would continue;
% fortunately, it turns out that \eqref{G2 bound analysis} holds.
% We will briefly discuss how to prove \eqref{G2 bound analysis} next.
% This almost completes the proof.

\vspace{0.3cm}
\textbf{Constrained perturbation analysis.}
The desired inequality \eqref{G2 bound analysis} is implied by the following condition on $U,V$:
$ \| U\|_F \leq \| X\|_F, \| V\|_F \leq \|Y \|_F$  when $\| X\|_F, \| Y\|_F$ are large.
Recall that previously we try to find $U,V$ that are close to $X,Y$; see Proposition \ref{prop1}.
Now we need to impose extra constraints on $U,V$, giving rise to Proposition \ref{prop2}.
The extra constraints make the perturbation analysis significantly more involved; % Such a ``constrained perturbation analysis''
% Even for the simple case $r=2$, the preliminary analysis in Appendix \ref{appen: analysis of prop2', simplfied}
% shows that two different kind of operations are needed.
in fact, we apply a sophisticated iterative procedure to construct the factorization $M = UV^T$.
The main steps of the proof are briefly given in Appendix \ref{appen: analysis of prop2', simplfied}.

One crucial component of our proof can be viewed as the perturbation analysis for ``preconditioning''.
Roughly speaking, the basic problem is: given an $r \times r$ matrix $ \hat{X} $ with a large condition number,
find another matrix $\hat{U}$ with the same Frobenius norm as $\hat{X}$ but smaller inverse Frobenious norm (i.e. $\|\hat{U}^{-1} \|_F \leq \frac{1}{1 - \delta} \| \hat{X}^{-1}\|_F $).
In other words, we want to reduce $\sum_{i=1}^r \frac{1}{\sigma_i^2}$ with $\sum_{i=1}^r \sigma_i^2 $ fixed,
where $\sigma_i$'s are all singular values.
Intuitively, by reducing $\sum_{i=1}^r \frac{1}{\sigma_i^2}$ we reduce the discrepancy of singular values.
This process is somewhat similar to preconditioning in numerical algebra that reduces the gap between the largest and smallest eigenvalue.
The precise statement of the basic problem and its relation with the key technical result Proposition \ref{prop2} are provided in Appendix \ref{appen: precond relation}.

\vspace{0.3cm}
\textbf{Algorithm requirements}.
%The regularizers can help keep the iterates stay in $K_1 \cap K_2$.
% To guarantee that the iterates stay in the desired local region,
% it is not enough to have a descent algorithm since the decrease of $\|\bP_{\Omega}(M - X_k Y_k^T) \|$
% does not imply the same in $\| M - X_k Y_k^T \|_F $. % when $()$
We provide three conditions and show that if an algorithm satisfies either of them, then with specific initialization the iterates will stay in the desired basin (see Proposition \ref{prop: K(delta) condition}).
%The first condition covers GD and SGD with certain standard step size rules.
%% and it says that the function value of any point in the line segment between any two consecutive iterates is small.
%% Intuitively, the iterates will stay in the desired basin if the algorithm generates decreasing function values; for some technical reasons (due to the regularizer), we require a slightly stronger condition that the function evaluated along the line segment of any two consecutive iterates are small.
%% This condition is satisfies by, for example, the gradient descent method with constant step size.
%% Note that this condition does not require the algorithm to be descent, thus can cover SGD.
%The second condition covers alternating minimization type methods (a.k.a. BCD type methods); in particular, this condition covers inexact update
%or minimizing a surrogate function at each subproblem.
%The third condition covers any algorithm that generates iterates in a certain pre-defined bounded region,
%which bears some similarity with the classical trust region method in optimization (see, e.g. \cite{yuan2000review}).
A special case of the third condition has been used in \cite{keshavan2010matrix} for Grassmann manifold optimization.
% Here, we provide a much more general optimization framework by introducing the first two conditions that
Together, these three conditions cover a wide spectrum of algorithms including GD, SGD and block coordinate descent type methods.

\vspace{0.3cm}
\textbf{Proof outline.}
%The basic idea is to prove a  geometrical property of the objective function around the global optima, thus starting from a good initial point a locally convergent algorithm will converge to the global optima.
The overall proof can be divided into two parts: the geometrical property (Lemma \ref{lemma main; about local convexity}) and the algorithm property (Lemma \ref{lemma of incoherent neighborhood}).
    % (i.e.\ the true matrix; see Lemma \ref{lemma main; about local convexity})
For the geometrical property, Lemma \ref{lemma main; about local convexity} states that the regularized objective function $F+G_1 + G_2$ enjoys some nice geometrical property in a certain local region around the global optima, thus there is no other stationary point in this region.
For the algorithm property, Lemma \ref{lemma of incoherent neighborhood} states that starting from an easily computable initial point,
many standard algorithms generate a sequence that are inside the desired region and these algorithms also converge to stationary points.
Since these stationary points must be global optima by Lemma \ref{lemma main; about local convexity}, we obtain that these algorithms converge to the global optima.

\subsection{Other Remarks}

\textbf{Difference with previous works.}
% Let us briefly discuss the difference of \cite{keshavan2010matrix} for Grassmannian optimization and \cite{keshavan2012efficient,jain2013low,hardt2013understanding,hardt2014fast} for resampling-based algorithms.
% where $Z$ is a certain matrix related to the iterates, a major difficulty is due to the dependency
As discussed earlier, one major challenge is to bound $\bP_{\Omega}(A)$ when $A$ may be dependent on $\Omega$. One simple strategy
 as adopted in \cite{keshavan2012efficient,jain2013low,hardt2013understanding,hardt2014fast} is to use a resampling scheme to decouple $A$ and the observation set.
% and the subsequent analysis can be relatively easy.
This strategy artificially avoids this difficulty, and causes a few issues discussed earlier in Section \ref{sec: Comparison with other non-convex schemes}. Another strategy, as employed in \cite{keshavan2010matrix}, is to use a random graph lemma in \cite{feige2005spectral}.

We apply the random graph lemma of \cite{feige2005spectral}
when extending the local geometry of
$\|M - XY^T \|_F^2 $ to
$ \| \bP_{\Omega}( M - XY^T ) \|_F^2 $.
The difference of our work with \cite{keshavan2010matrix} is that
we study the local geometry in Euclidean space (and, indirectly, the geometry of the quotient manifold),
which is quite different from the local geometry in Grassmann manifold studied in \cite{keshavan2010matrix}.
%  The dependency of iterates on $\Omega$ is just the first barrier, which we will overcome using . There are other difficulties besides the probability tools. with the factorization model in Euclidean space
Technically, the complications of the proof in \cite{keshavan2010matrix} % (besides the validation of initialization)
% are mostly due to the Grassmann manifold model. It includes
are mostly due to heavy computation of various quantities in Grassmann manifold;
 in addition, much effort is spent in estimating the terms related to the extra factor $S$ which enables the decoupling of $X$ and $Y$ (\cite{keshavan2010matrix} actually uses a three-factor decomposition $XSY^T$).
For our problem, one difficulty is to ``pull back'' the distance in the quotient manifold to the Euclidean space, by the coupled perturbation analysis. Another difficulty is to align the gradient of the regularizer with the global direction (this is not an issue for Grassman manifold), which requires a more sophisticated perturbation analysis.
The difficulties have been discussed in detail in Section \ref{sec: proof overview}.

% and the main technical results we establish are Proposition \ref{prop1} and Proposition \ref{prop2}.
% , as shown by our construction \eqref{def of U,V}. where $X$ and $Y$ are well-conditioned, where $X$ and $Y$ are not well-conditioned
%For Proposition \ref{prop2}, it turns out that we need an iterative procedure to construct the factorization $M = UV^T$;
%moreover, the preliminary analysis in Appendix \ref{appen: analysis of prop2', simplfied} illustrates that a simple one-step construction probably does not work
%and a sophisticated iterative procedure is necessary.

\vspace{0.2cm}
\textbf{Symmetric PSD or rank-1 case}.
The symmetric PSD (positive semi-definite) case or the rank-1 case are easier to deal with, because in the 3-step study of the local geometry the third step is not necessary.
 %  since the third issue mentioned previously is no longer an issue.
When $M$ is rank-1 (possibly non-symmetric),
the regularizer $G_2(\cdot)$ may still be needed, but Proposition \ref{prop2} is trivial
since its assumptions cannot hold for $r=1$.
When $M$ is symmetric PSD, a popular approach is to use a symmetric factorization $M = XX^T$ instead of the non-symmetric factorization,
and the loss function becomes $\| \bP_{\Omega}(M - XX^T)\|_F^2 $.
The same proof in our paper can be translated to this symmetric PSD case, except that the third step is not necessary. In fact, it is possible to show that \eqref{G1 bound analysis} holds without any additional requirement on $\bm x$.
As a result, the regularizer $G_2$ and a major technical result Proposition \ref{prop2} are not needed.
In both the symmetric PSD and rank-1 case, we only need to establish the intermediate result
\eqref{intro, LSC 1st} and the proof can be greatly simplified.
 Stronger sample complexity and time complexity bounds may be established in these two cases.

% The geometrical property \eqref{intro, LSC 1st simplified} is of the form $\angle \nabla F(\bm x), \bm x - \bm x^*  \rangle \geq  $.
%For general machine learning problems, even proving convergence
%A closely related subject is identifiability: do the observed entries uniquely define a low-rank matrix?
%The relation \eqref{LSC} implies that there is a unique rank-$r$ matrix in a certain local region that matches the observations in $\Omega$.
%Very tricky!! Is global optimizer enough???
%That being said, rigorously speaking convergence to global optimizers

% Note that identifiability is implicitly established in
 % the generalization error $\|M - XY^T\|_F $ is bounded by the training error $\|\bP_{\Omega}(M - XY^T)\|_F $ (times some constant)
% thus if the training error is zero then the generalization error is also zero.

% Minimizing the training error does not necessarily lead to

\vspace{0.2cm}
\textbf{Simulation Results}
The regularizers are introduced due to theoretical purposes; interestingly, they turn out to be helpful in the numerical experiments (the comments below
are extracted from the thesis \cite[Chapter 2]{sun2015thesis}).

First, the simulation suggests that the imbalance of the rows of $X $ or $Y$ is an important issue for matrix completion in practice, a phenomenon not reported before to our knowledge.
The table in Figure 2.10 of \cite{sun2015thesis} shows that
when $|\Omega|$ is small, in
all successful instances the iterates are balanced, while in all failed instances the iterates are unbalanced.
This contrast occurs for many standard algorithms such as AltMin,GD and SGD.

Second, adding only the regularizer $G_1$ helps, but not too much. Adding an extra regularizer $G_2$ can push the sample complexity
to be very close to the fundamental limit, at  least for the synthetic Gaussian data.
These experiments seem to indicate that the new regularizers do change the geometry of the problem.

 \vspace{0.2cm}
 \textbf{Necessity of incoherence?}
 While our regularizers are helpful when $|\Omega|$ is small,
 an open question is whether the
row-norm requirement is needed for the local geometry when $|\Omega|$ is large.
We observe that the row-norms can be automatically controlled by standard algorithms for the
synthetic Gaussian data when there are, say, $5rn$ samples for $n \times n$ matrices.
% However, we do not know whether this phenomenon is universal for any incoherent low-rank matrix $M$.
%Even if this is universal,
There are two possible explanations (assuming a large $|\Omega|$):
(i) the local geometrical property \eqref{intro, LSC 1st} holds without the incoherence requirement;
(ii) \eqref{intro, LSC 1st} still requires incoherence, but there is an unknown mechanism for many algorithms to control
the row-norms.
% (possibly utilizing other properties of $M$).

To exclude the first possibility, we need to find $(X,Y)\in K(\delta)$ such that $\nabla F(X,Y) = 0$ but
 $XY^T \neq M $; since \eqref{intro, LSC 1st} holds, such $(X,Y)$ must have unbalanced row-norms. Such an example would validate the necessity of the incoherence restriction for the local geometry.
Note that the necessity of incoherence for the local geometry is different from the necessity of an incoherence regularizer/constraint for a specific algorithm.
% We have been focused on the geometrical aspect of the problem rather than a specific algorithm. Nevertheless, i
Even if the local geometry requires incoherence, it remains an interesting question why many algorithms can automatically control row-norms when $|\Omega|$ is large.

\subsection{Notations and organization}
\textbf{Notations.}
Throughout the paper, $M \in \mathbb{R}^{m\times n}$ denotes the unknown data matrix we want to recover, and $r \ll \min \{m,n \}$ is the rank of $M$.
The SVD of $M$ is $M = \hat{U}\Sigma\hat{V}^T$, where $\hat{U} \in \mathbb{R}^{m\times r}, \hat{V} \in \mathbb{R}^{n\times r}$  and $\Sigma \in \mathbb{R}^{r\times r}$ is a diagonal matrix with diagonal entries $\Sigma_1 \geq \Sigma_2 \geq \dots \geq \Sigma_r$.
We denote the maximum and minimum singular value as $\Sigma_{\rmax}$ and $ \Sigma_{\rmin}$, respectively, and denote
$\kappa \triangleq \Sigma_{\rmax}/\Sigma_{\rmin}$ as the condition number of $M$. % We assume $\kappa$ is bounded as $n \rightarrow \infty.$
Define $\alpha = m/n$, which is assumed to be bounded away from 0 and $\infty$ as $n \longrightarrow \infty$. Without loss of generality, assume
 $m\geq n$, then $\alpha \geq 1$.

Define the short notations $[m]\triangleq\{1,2,\dots,m\},[n]\triangleq\{1,2,\dots,n\}$. Let $\Omega \subseteq [m]\times[n]$ be the set of observed positions, i.e.\  $\{ M_{ij} \mid (i,j)\in \Omega\}$ is the set of all observed entries of $M$,
  and define $p \triangleq \frac{|\Omega|}{mn}${\black which can be viewed as the probability that each entry is observed.}
  % the average number of observed entries per row or column.
  % Denote $\Omega^c$ as the relative complement of $\Omega$ in $[m] \times [n]$.
  For a linear subspace $\mathcal{S}$, denote $\mathcal{P}_{\mathcal{S}}$ as the projection
 onto $\mathcal{S}$. By a slight abuse of notation, we denote $\mathcal{P}_{\Omega}$ as the projection onto the subspace $\{W \in \mathbb{R}^{m\times n}: W_{i,j}=0, \forall (i,j)\notin \Omega \}$.
In other words, $\mathcal{P}_{\Omega}(A) $ is a matrix where the entries in $\Omega$ are the same as $A$ while the entries outside of $\Omega$ are zero.
% Denote $\mathcal{P}_{\Omega^c} = \mathcal{I} - \mathcal{P}_{\Omega}$, where $\mathcal{I}$ is the identity operator.

For a vector $x \in \mathbb{R}^n, $ denote $\|x \|$ as its Euclidean norm. For a matrix $X,$ denote $\| X\|_F$ as its Frobenius norm, and $\|X \|_2$ as its spectral norm (i.e.\ the largest singular value).
Denote $\sigma_{\rmax}(X), \sigma_{\rmin}(X)$ as the largest and smallest singular values of $X$, respectively.
Let $X^{\dag}$ denote the pseudo inverse of a matrix $X$.
The standard inner product between
vectors or matrices are written as $\langle x,y\rangle$ or $\langle X,Y \rangle,$ respectively.
Denote $A^{(i)}$ as the $i$th row of a matrix A.
We will use $C,C_1,C_T,C_d,$ etc. to denote universal numerical constants.

\textbf{Organization.} The rest of the paper is organized as follows. In Section \ref{sec£ºform and algorithms} we introduce the problem formulation and four typical algorithms.
% (only consider several standard algorithms).
 In Section \ref{sec: main result}, we present the main results and the main lemmas used in the proofs of these results.
% The proof of the two lemmas are given in Section \ref{sec: proof of lemma 1} and Section \ref{sec: proof of lemma 2} respectively.
% The proof of the first lemma depends on two main technical results Proposition \ref{prop1} and Proposition \ref{prop2}, the proofs of which are given in the  appendix.
The proof of the two lemmas used in proving Theorem \ref{major theorem} are given in Section \ref{sec: proof of lemma 1}
  and Section \ref{sec: proof of lemma 2} respectively.
 The proof of the first lemma depends on two ``coupled perturbation analysis'' results Proposition \ref{prop1} and Proposition \ref{prop2}, the proofs of which are given in Appendix \ref{appen: proof of Prop 1} and Appendix \ref{section of proof of Prop 2} respectively.
The proof of a lemma used in proving Theorem \ref{theorem 2: linear convergence} is given in Appendix \ref{appen: cost-to-go lemma proof}.

\section{Problem Formulation and Algorithms}\label{sec£ºform and algorithms}
\subsection{Assumptions}\label{sec: assumptions}
\textbf{Incoherence condition.} The incoherence condition for the matrix completion problem is first introduced by Cand\`{e}s and Recht in \cite{candes2009exact} and has become a standard assumption for low-rank matrix recovery problems (except a few recent works such as \cite{chen2014coherent,bhojanapalli2014universal}).
We will define an incoherence condition for an $m\times n$ matrix $M$ which is the same as that in \cite{keshavan2010matrix}.
% and then define an incoherence condition for an arbitrary $m \times r$ or $n \times r$ matrix.
% and slightly different from the version in \cite{candes2009exact} (different $A_1$ conditions).

\begin{Def}\label{incoherence def of M}
We say a matrix $M = \hat{U}\Sigma\hat{V}^T$ (compact SVD of $M$) is $\mu$-incoherent if: % $(\mu_0,\mu_1) $
\begin{equation}\label{incoherence cond}
 \sum_{k=1}^r \hat{U}_{ik}^2 \leq \frac{\mu r}{m}, \quad  \sum_{k=1}^r \hat{V}_{jk}^2 \leq \frac{\mu r}{n}, \quad 1\leq i \leq m,1\leq j \leq n.
\end{equation}
%\begin{equation}\nonumber
%\rA_1:|M_{ij} | \leq \mu_1 \Sigma_{\max} \frac{\sqrt{ r}}{\sqrt{mn} }, \quad \forall \; i,j.  % $(\mu_0,\mu_1) $
%\end{equation}
%Here, $\Sigma_{\max}$ is the maximum singular value of $M$.
\end{Def}

% When $\mu_0,\mu_1$ are large enough, $\rA_0,\rA_1$ will always hold; thus we assume $\mu_0, \mu_1$ are chosen to be the smallest among all
% possible $\mu_0, \mu_1$'s.
It can be shown that $\mu \in [1, \frac{ \max\{m,n \} }{r}]$.  % \mu_1 \in [\frac{1}{\kappa}, \frac{ \max\{m,n \} }{r} ]
For some popular random models for generating $M$, the incoherence condition holds with a parameter scaling as $\sqrt{r\log n}$ (see \cite{keshavan2010matrix}). In this paper, we just assume that $M$ is $\mu$-incoherent.{\black
Note that the incoherence condition  implies that $\hat{U}, \hat{V}$ have bounded row norm.  % $\rA_0$
Throughout the paper, we also use the terminology ``incoherent'' to (imprecisely) describe $m\times r$ or $n \times r$ matrices that have bounded row norm (see the definition of set $K_1$ in \eqref{def of K_1, K_2}). }

\textbf{Random sampling model.}
In the statement of the results in this paper, the probability is taken with respect to the uniform random model of $\Omega \subseteq [m]\times [n]$
with fixed size $|\Omega| = S$ (i.e.\ $\Omega $ is generated uniformly at random  from set
$\{\Omega'\subseteq [m]\times[n] : \text{ the size of } \Omega' \text{ is } S \}$ ).
We remark that this model is ``equivalent to'' a  Bernolli model that each entry of $M$
is included into $\Omega$ independently with probability $p = \frac{S}{mn}$
in the sense that if the success of an algorithm holds for the Bernolli model with a certain $p$ with high probability,
then the success also holds for the uniform random model with $|\Omega| = pmn$ with high probability (see \cite{candes2009exact}
 or \cite[Sec. 1D]{keshavan2010matrix} for more details).
 Thus in the proofs we will instead use the Bernolli model.

\subsection{ Problem formulation }
% For technical reasons,
We consider a variant of (P0) with incoherence-control regularizers. %  which can be viewed as penalty functions.
In particular, we introduce two types of regularization terms besides the square loss function:
the first type is designed to force the iterates $X_k, Y_k$ to be incoherent (i.e.\ with bounded row norm),
and the second type is designed to upper bound the norm of $X_k$ and $Y_k$.
Note that (P0) is related to the Lagrangian method, while
our regularizer is based on the penalty function method for constrained optimization problems.
We can also view the regularizer $ \lambda( \| X\|_F^2 + \| Y\|_F^2 )$ as a ``soft regularizer'', and
our new regularizer as a ``hard regularizer''.
The advantage of the hard regularizer is that it does not distort the optimal solution.

% Note that adding these regularizers
{\black Our regularizers are smooth functions with simple gradients, thus the algorithms for our formulation have similar per-iteration computation cost as the algorithms for the formulation without regularizers.
In the numerical experiments, we find that when $|\Omega|$ is large, the iterates are always incoherent and bounded, and
our algorithms are the same as the traditional algorithms for the unregularized formulation;
when $|\Omega|$ is relatively small, the traditional algorithms may produce high error, and our regularizer becomes active
 and significantly reduce the error. %  these algorithms . % work.
 In some sense, our algorithms for the new formulation are ``better'' versions of the traditional algorithms,
  and our theoretical results can be viewed as a validation of the traditional algorithms in the ``large-$|\Omega|$ regime''
  and a validation of the modified algorithm in the ``small-$\Omega$'' regime.
  Preliminary simulation results show that many algorithms for the proposed formulation
  can recover the matrix when $|\Omega|$ is very close to the fundamental limit, significantly improving upon the traditional algorithms;
   see \cite[Chapter 3]{sun2015thesis}.

%The reason is that our regularizers are of penalty-function type and do not affect the algorithms when the iterates do not violate the desired constraints.
%In fact, our regularizers (or constraints) serve as a ``safeguard'':
% when $p$ is large enough that traditional algorithms (e.g. AltMin, SGD) successfully recover $M$, our regularizers (or constraints) are inactive
% and our algorithms are the same as the traditional algorithms; when the traditional algorithms fail, our regularizers become active
% and make the algorithms work.

 }
% the first type of regularization terms is designed  (the produced solution in the $k$-th iteration)
% to force $X_k, Y_k$ (generated by the algorithms in $k$-th iteration) to be incoherent, i.e.\ the norm of each row of $X_k, Y_k$ is bounded above.
 % (by Definition \ref{def, incoherence of X}).
% The second type of regularization terms is added to upper bound the norm of $X_k$ and $Y_k$ (the same purpose as the regularization term $\lambda(\| X\|_F^2 + \| Y\|_F^2)$ in \eqref{P0}).
%Reference \cite{keshavan2010matrix} also used the first type of regularizers, but not the second type; this difference with our work
%is mainly because their formulation forced $X,Y$ to have a fixed norm.

%Now we briefly explain why we add the second type of regularization terms.
%In the algorithm of \cite{KMO09}, a low rank matrix $W$ is decomposed as $XSY^T$, where $X^TX= mI, Y^TY= nI$.
%In contrast, a low rank matrix $W$ is decomposed as $XY^T$ in our algorithm, where $X \in \mathbb{R}^{m \times r}$ and $Y \in \mathbb{R}^{n \times r}$ can be any matrix. One drawback of this decomposition is that $X_k$ and $Y_k$ could become unbounded and ill-conditioned due to the lack of constraints on the norms of $X_k$ and $Y_k$. For example, when $M = diag(1,1)$ (i.e.\ a diagonal matrix with diagonal entries $1,1$), it is possible that $X_k = diag(L, 1/L), Y_k =diag(1/L, L)$, where $L$ is an arbitrarily large number.
%To keep $X_k$ and $Y_k$ bounded and well-conditioned, we introduce a penalty function on the norms of $X_k$ and $Y_k$ as a regularization term.

The regularization function $G$ is defined as follows:
 \begin{equation}\label{regularized function}
 \begin{split}
 G(X,Y) \triangleq \rho \sum_{i=1}^m G_0\left(\frac{3\| X^{(i)} \|^2  } {2\beta_1^2} \right) + \rho \sum_{j=1}^n G_0\left(\frac{3 \| Y^{(j)} \|^2 } {2\beta_2^2} \right)    \\
 + \rho G_0\left(\frac{3\| X\|_F^2 } {2\beta_T^2} \right) + \rho G_0\left(\frac{3\| Y \|_F^2 } {2\beta_T^2} \right),
 \end{split}
 \end{equation}
where $A^{(i)}$ denotes the $i$th row of a matrix A,
 \begin{equation}\label{G_0 expression}
 G_0(z) \triangleq I_{[1,\infty]}(z) (z-1)^2 = \max\{ 0, z-1 \}^2,    % $$ (e^{(z-1)^2}-1), $$
 \end{equation}
 \begin{equation}\label{beta 1 beta T def}
  \begin{split}
  \beta_T \triangleq \sqrt{C_T r\Sigma_{\max}}, \; \beta_1 \triangleq \beta_T \sqrt{ \frac{3\mu r}{m}  } {\black = \sqrt{C_T r\Sigma_{\max}} \sqrt{ \frac{3\mu r}{m}  } } , \\
   \; \beta_2 \triangleq \beta_T \sqrt{ \frac{3\mu r}{n}   } {\black =\sqrt{C_T r\Sigma_{\max}}  \sqrt{ \frac{3\mu r}{n}   }  }.
    \end{split}
  \end{equation}
 % \textbf{ Remark: I define $\beta_1,\beta_2$ in terms of $\beta_T$ because I want to emphasize $\beta_1, \beta_2$ are about the average row norm of
%  a matrix with total norm $\beta_T$,
%  i.e.\ in the order of $\beta_T \sqrt{1/n}$. Also, the expressions of $\beta_1,\beta_2$ are only used very few times, and all of their appearances
%  are related to $\beta_T$. In other words, the exact expressions of $\beta_1,\beta_2$ are not needed.}
Here,{\black $I_{\mathcal{C}}$ is the indicator function of a set $\mathcal{C}$, i.e.\ $I_{\mathcal{C}}(z) $ equals $1 $ when $z \in \mathcal{C}$ and $0$ otherwise.} $\rho$ is a constant specified shortly.
Throughout the paper, $\delta$ and $\delta_0$ are defined as
\begin{equation}\label{delta definition throughout}
\begin{split}
\delta       \triangleq \frac{\Sigma_{\rmin}}{C_d r^{1.5} \kappa },  \quad
\delta_0     \triangleq \frac{\delta}{6},
\end{split}
\end{equation}
where $C_d$ is some numerical constant. The coefficient $\rho$ is defined as (a larger $\rho$ also works)
\begin{equation}\label{rho definition throughout}
\rho \triangleq \frac{  2 p \delta_0^2} {G_0(3/2)} = 8p \delta_0^2.
\end{equation}
The numerical constant $C_T >5$ will be specified in the proof of our main result. The parameter $\beta_T$ is chosen to be of the same order as $\|\hat{U}\Sigma^{1/2}\|_F$ and $\|\hat{V}\Sigma^{1/2}\|_F$, and
 % (later, we will show $(X,Y) = (\hat{U}\Sigma^{1/2},\hat{V}\Sigma^{1/2})$ is optimal to the regularized problem).
 $\beta_1, \beta_2$ are chosen to be of the same order as $\sqrt{r} \|( \hat{U}\Sigma^{1/2})^{(i)}\|, \sqrt{r}\|( \hat{V}\Sigma^{1/2})^{(j)}\| $.
 The additional factor $\sqrt{3r}$ is due to technical consideration (to prove (\ref{Ui<=Xi, introduce 3r in beta1})).
Our regularizer $G$ involves $\Sigma_{\max}$ and $\mu$ which depend on the unknown matrix $M$; in practice,
we can estimate $\Sigma_{\max} $ by $ c_1 \sqrt{ \frac{ \|\bP_{\Omega}(M) \|_F^2 }{ p r} }$, and estimate $\mu$ by
$ c_2 \frac{\sqrt{mn}}{r \Sigma_{\max}} \max_{(i,j)\in \Omega} |M_{ij}|  $  (according to \eqref{Mij bound by mu})
where $c_1, c_2$ are numerical constants to tune.
% and due to   Proposition \ref{prop1} and Proposition \ref{prop2}).

It is easy to verify that $G_0$ is continuously differentiable.
The choice of function $G_0$ is not unique; in fact, we can choose any $G_0$ that satisfies the following requirements: a) $G_0$ is convex and continuously differentiable; b) $G_0(z)=0, z\in [0,1]$. % increases rapidly to $\infty$ in $ [1,\infty]$.
In \cite{keshavan2010matrix}, $G_0$ is chosen as $G_0(z) = I_{[1,\infty]}(z) (e^{(z-1)^2}-1)$, which also satisfies these two requirements.
Choosing different $G_0$ does not affect the proof except the change of numerical constants (which depend on $G_0(3/2), G_0^{\prime}(3/2), G_0^{\prime \prime}(3/2) $). %  but we prefer the choice of $G_0$ given by \eqref{G_0 expression} (explained slightly later).
Note that the requirement of $G_0$ being non-decreasing and convex guarantees the convexity of $G(X,Y)$.
In fact, according to the well-known result that the composition of a non-decreasing convex function and a convex function is a convex function,
and notice that  $\|X^{(i)} \|^2, \| Y^{(j)}\|^2, \| X\|_F^2, \| Y\|_F^2$ are convex,
we have that each component of $G$ is convex and thus $G$ is convex.
% the augmented Lagrangian term   (see the classical optimization reference \cite{bertsekas1999nonlinear}).

% The regularization coefficient $\rho$ will be determined later.
%  chosen according to the choice of $G_0$.

 Denote the square loss term in (P0) as $F(X,Y)  \triangleq \sum_{(i,j)\in \Omega} [M_{ij} - (XY^T)_{ij}]^2  = \|\bP_{\Omega}(M - XY^T)\|_F^2$.
  Replacing the objective function of (P0) by $ \tilde{F}(X,Y)  \triangleq F(X,Y) + G(X,Y)$,
% Replacing the regularization term $\lambda(\| X\|_F^2 + \| Y\|_F^2)$ in \eqref{P0} by  $G(X,Y)$ defined in \eqref{regularized function},
% $ \sum_{(i,j)\in \Omega} [M_{ij} - (XY^T)_{ij}]^2 + \lambda(\| X\|_F^2 + \|Y \|_F^2)$
we obtain the following problem:
\begin{equation}\label{P1}
\begin{split}
\mathrm{P}1: \quad \min_{  X\in \mathbb{R}^{m\times r},Y \in \mathbb{R}^{n\times r}} \quad &
  \frac{1}{2}\|\bP_{\Omega}(M - XY^T)\|_F^2 + G(X,Y).
% s.t \quad & Z_{ij} = M_{ij},(i,j)\in \Omega.
\end{split}
\end{equation}
%We choose $G_0$ as in \eqref{G_0 expression} not only because it is simple, but also because it can be interpreted as penalty terms for
%For $G_0$ given in \eqref{G_0 expression},
We remark that (P1) can be interpreted as the penalized version of the following constrained problem (see, e.g. \cite{zangwill1967non})
% (see, e.g., \cite[Sec. 4.2.1]{bertsekas1999nonlinear} for the quadratic penalty function method)
 \begin{equation}\label{constrained formulation}
 \begin{split}
   \min_{X,Y} & \quad \frac{1}{2}\|\bP_{\Omega}(M - XY^T)\|_F^2  , \\
  \st \quad &  \| X\|_F^2 \leq \frac{2}{3}\beta_T^2, \quad \| Y\|_F^2 \leq \frac{2}{3}\beta_T^2; \\
       & \| X^{(i)}\|^2 \leq \frac{2}{3}\beta_1^2, \ \forall \ i,  \quad \| Y^{(j)}\|^2 \leq \frac{2}{3}\beta_2^2, \ \forall \ j.
       \end{split}
  \end{equation}
 To illustrate this, note that the constraint $ f_1(X) \triangleq \frac{3\| X\|_F^2}{2\beta_T^2}  - 1 \leq 0 $ corresponds to the penalty term
  $\rho G_0(f_1(X)+1) =  \rho \max\{0, f_1(X)\}^2 $ which appears as the third term in $G(X,Y)$, and similarly other constraints correspond to other
  terms in $G(X,Y)$.
  % can be treated in a similar way.
  In other words, the regularization function $G(X,Y)$ is just a penalty function for the constraints of the problem \eqref{constrained formulation}.
% The choice of $G_0$ as in \eqref{G_0 expression},
  The function $\max\{0, \cdot \}^2$ is a popular choice for the penalty function in optimization (see, e.g. \cite{zangwill1967non}),
  which motivates our choice of $G_0$ in \eqref{G_0 expression}.
  Our result can be extended to cover the algorithms for the constrained version \eqref{constrained formulation},
  or a partially regularized formulation (e.g. only penalize the violation of the constraint
   $\| X\|_F^2 \leq \frac{2}{3}\beta_T^2, \| Y\|_F^2 \leq \frac{2}{3}\beta_T^2$).

%\begin{equation}
%\begin{split}
%  F(X,Y) & \triangleq \sum_{(i,j)\in \Omega} [M_{ij} - (XY^T)_{ij}]^2  = \|\bP_{\Omega}(M - XY^T)\|_F^2, \\
%  \tilde{F}(X,Y)  &   \triangleq F(X,Y) + G(X,Y).
%\end{split}
%\end{equation}

%\begin{equation}
%\begin{split}
%P1: \quad \min_{Z\in \mathbb{R}^{m\times n}, X\in \mathbb{R}^{m\times r},Y \in \mathbb{R}^{n\times r}} \quad &   \tilde{F}(X,Y,Z) = \frac{1}{2}\|Z - XY^T\|_F^2 + G(X,Y) \\
%s.t \quad & Z_{ij} = M_{ij},(i,j)\in \Omega.
%\end{split}
%\end{equation}

It is easy to check that the optimal value of (P1) is zero and $(X,Y) = (\hat{U}\Sigma^{1/2},\hat{V}\Sigma^{1/2})$ is an optimal solution to (P1),
provided that $M$ is $\mu$-incoherent.
In fact, since $\tilde{F}$ is a nonnegative function, we only need to show $\tilde{F}(X,Y)  = 0 $ for this choice of $(X,Y)$.
As $XY^T = M $ implies $\|\bP_{\Omega}(M - XY^T)\|_F^2 = 0$, we only need to show $G(X,Y) = G(\hat{U}\Sigma^{1/2},\hat{V}\Sigma^{1/2})$ equals zero.
In the expression of $G(X,Y)$, the third and fourth terms $G_0(\frac{3\| X\|_F^2 } {2\beta_T^2} )$ and $G_0(\frac{3\| Y \|_F^2 } {2\beta_T^2} ) $ equal zero
because $\|X\|_F^2 = \|Y \|_F^2 \leq r \Sigma_{\rmax} < \frac{2}{3}\beta_T^2$.
The first and second terms $\sum_i G_0(\frac{3\| X^{(i)} \|^2  } {2\beta_1^2} )$ and $\sum_j G_0(\frac{3 \| Y^{(j)} \|^2 } {2\beta_2^2} ) $ equal zero
because $\|X^{(i)}\|^2 \leq \Sigma_{\rmax} \| \hat{U}^{(i)} \|^2 \leq \Sigma_{\rmax} \frac{\mu  r}{m} \leq \frac{2}{3} \beta_1^2, $ for all $i$ and, similarly, $\|Y^{(j)}\|^2 \leq  \frac{2}{3} \beta_2^2, $ for all $j$,{\black where we have used the incoherence condition \eqref{incoherence cond}.}
This verifies our previous claim that the ``hard regularizer'' $G(X,Y)$ does not distort the optimal solution of the original formulation.
% Since $M$ is $(\mu ,\mu_1)$ incoherent, we have
%Thus, $G_0(\frac{3\| X^{(i)} \|^2  } {2\beta_1^2} )= G_0(\frac{3 \| Y^{(j)} \|^2 } {2\beta_2^2} ) = 0.$ Therefore, $G(X,Y) = 0$ since each term of $G(X,Y)$ equals zero.

One commonly used assumption in the optimization literature is that the gradient of the objective function is Lipschitz continuous.
For any positive number $\beta$, define a bounded set
  \begin{equation}\label{Gamma def}
   \Gamma(\beta) \triangleq \{ (X,Y)| X \in \dR^{m\times r}, Y \in \dR^{n\times r},  \|X\|_F \leq \beta,  \|Y\|_F \leq \beta \}.
   \end{equation}
   The following result shows that this assumption (Lipschitz continuous gradients) holds for our objective function within a bounded set.
\begin{claim}\label{claim: Lip constant}
Suppose $\beta_0 \geq \beta_T$ and
 \begin{equation}\label{Lip constant}
  L(\beta_0) \triangleq 4\beta_0^2 + 54 \rho \frac{\beta_0^2}{\beta_1^4}  .
  \end{equation}
 Then $\nabla \tilde{F}(X,Y)$ is Lipschitz continuous over the set $\Gamma(\beta_0)$ with Lipschitz constant $L(\beta_0)$, i.e.\   \begin{equation}\nonumber
  \begin{split}
     \| \nabla \tilde{F}(X,Y) - \nabla \tilde{F}( U, V) \|_F \leq L(\beta_0) \| (X,Y) - (U,V) \|_F, \;\;  \\
      \forall (X,Y), (U,V) \in \Gamma(\beta_0),
      \end{split}
  \end{equation}
  where  $\| (X,Y) - (U,V) \|_F =  \sqrt{ \| X - U \|_F^2 + \| Y - V \|_F^2 } $.
\end{claim}
The proof of Claim \ref{claim: Lip constant} is given in Appendix \ref{appen: proof of Claim of Lip constant}.

%Now, we are ready to present our algorithm and global convergence result.
%As mentioned before, two modifications need to be made to Algorithm 1. First, we want a block coordinate descent algorithm
%for problem (P1) instead of problem $P0$. Thus, at each step, $X_k$ and $Y_k$ are computed by minimizing $\tilde{F}$ instead of $F$. Second, we choose the initial point as in Remark \ref{requirement of initial point}. Algorithm 2 is summarized in the following table.

%\emph{Algorithm 2:}
%\begin{equation}
%\begin{split}
%\text{Initialization}:  &\quad \text{Choose}(X_0,Y_0,Z_0), s.t. (X_0,Y_0) \text{ is given by Remark \ref{requirement of initial point}}, Z_0 = \bP_{\Omega^c}(X_0Y_0^T) + \bP_{\Omega}(M).  \\
%\text{k-th iteration}: & \quad X_k \longleftarrow \arg \min_X \tilde{F}(X,Y_{k-1},Z_{k-1}),  \\
%                       & \quad Y_k \longleftarrow \arg \min_Y \tilde{F}(X_k,Y,Z_{k-1}), \\
%                           & \quad  Z_k \longleftarrow \bP_{\Omega^c}(X_kY_k^T) + \bP_{\Omega}(M).\\ %X_kY_k^T + \bP_{\Omega}(M-X_kY_k^T). \\
%\end{split}
%\end{equation}

\subsection{Row-scaled Spectral Initialization}
% Since problem (P1) is non-convex, it may have multiple local optima. To guarantee convergence to global optima,
 % For technical reasons,
 Our results require the initial point to be close enough to the global optima.
% Now, we describe the second modification to Algorithm 1: to choose a good initial point.
To be more precise,  we want the initial point to be in an incoherent neighborhood of the original matrix $M$ (this neighborhood will be specified later).
Special initialization is also required in other works on non-convex formulations \cite{keshavan2010matrix,keshavan2012efficient,jain2013low,hardt2013understanding,hardt2014fast,netrapalli2013phase,candes2014phase}.
%We first describe the definition of the incoherent neighborhood $K_1 \cap K_2 \c$
%We will show that $(X_0,Y_0)$ given by the procedure $\textsc{Initialize}$
% We require the initial point to be incoherent, bounded above and close enough to $M$.
% In fact, such initial point can be found by ``trimming and projection'' \cite{KMO09}.

We will show that such an initial point can be found through a simple procedure.
This procedure consists of two steps: first, using the spectral method (see, e.g. \cite{keshavan2010matrix}),
we obtain $M_0 =\hat{ X}_0 \hat{Y}_0^T$ which is close to $M$; second, %  similar to \cite[Remark 6.2]{keshavan2010matrix},
 we scale the rows of $(\hat{ X}_0, \hat{Y}_0)$ to make it incoherent (i.e.\ with bounded row-norm).  %(though the notion of incoherence is slightly different in our paper).
%Although we adopt a method similar to that of \cite{keshavan2010matrix}, our proof  % from that of \cite{keshavan2010matrix}
%requires a new result Proposition \ref{prop1}.
%The extra second step is due to our stronger requirements on the initial point: compared to \cite{KMO09}, we have add requirements that
%the initial point is incoherent and
% $\frac{1}{p}\bP_{\Omega}(M)$
{\black
Denote the best rank-$r$ approximation of a matrix $A$ as $ \mathrm{P}_r(  A ) $.
%which can be computed as
%\begin{equation}\label{best rank r def}
%  \mathrm{P}_r( A )  = \sum_{i=1}^r \sigma_i^{\prime} x_i y_i^T,
%  \text{ if the SVD of } A \text{ is } \sum_{i=1}^n \sigma_i^{\prime} x_i y_i^T.
%\end{equation}
Define an operation $\mathrm{SVD}_r$ that maps a matrix $A$ to the SVD components $(X,D,Y)$ of its best rank-$r$ approximation $\mathrm{P}_r(  A )$, i.e.\ %  of its best rank-$r$ approximation, i.e.\ % (i.e.\ $XDY^T$ is the SVD of $ \mathrm{P}_r( A )$).
\begin{equation}\label{SVD_r def}
%\begin{split}
 \mathrm{SVD}_r( A ) \triangleq    (X, D, Y), \text{ where } XDY^T \text{ is compact SVD of } \mathrm{P}_r(  A ).  % \mathrm{P}_r( A ).
  \end{equation}

%  I define a separate $SVD_r$ is to simplify the notation in the table.
%  Compute $(\bar{X}_0, D_0, \bar{Y}_0 ) = \mathrm{SVD}( Q_r ( \frac{1}{p} \bP_{\Omega}(M) ) )$,
% Then $\mathrm{SVD}(  \mathrm{P}_r (  A ) )$ gives the SVD components of the best rank-$r$
% approximation of $A$. % $\mathrm{P}_r(\frac{1}{p} \bP_{\Omega}(M))$.   \frac{1}{p} \bP_{\Omega}(M) )
The initialization procedure is given in Table \ref{table of Initialization}. The property of the initial point generated by this procedure will
be presented in Claim \ref{claim: initial point properties}.
}

  {\black In the numerical experiments, we find that  the proposed initialization is not better than random initialization if we use the proposed
  formulation with the incoherence-control regularizer.
   In contrast, for traditional formulations (either unregularized
  or with a regularizer $\lambda(\| X\|_F^2 + \| Y\|_F^2)$) the proposed initialization does lead to better recovery performance (lower sample complexity).
%  when $|\Omega| $ is relatively small.
  We also notice that the row-scaling step is crucial for this improvement since simply initializing via the spectral method does not help too much.
  See \cite[Chapter 3]{sun2015thesis} for the simulation results and  discussions.  }
% More formally, $\mathrm{SVD}_r$ is defined as
%\begin{equation}\label{SVD_r def}
%\begin{split}
%    \mathrm{SVD}_r( \frac{1}{p}\bP_{\Omega}(M) ) =    &  (\bar{X}_0, D_0, \bar{Y}_0 ),  \\
%    \text{where }    \bar{X}_0 = [x_1,\dots, x_r] \in \dR^{m \times r}, & D_0 = \mathrm{Diag}(\sigma_1^{\prime}, \dots, \sigma_r^{\prime}) \in \dR^{r \times r}, \bar{Y}_0 = [y_1,\dots, y_r] \in \dR^{n \times r}.  \\
%  \end{split}
%\end{equation}
%Here, $\mathrm{Diag}(\sigma_1^{\prime}, \dots, \sigma_r^{\prime})$ denotes the diagonal matrix with diagonal entries $\sigma_1^{\prime},\dots,\sigma_r^{\prime} $.

\begin{table*}[htbp]\caption{Initialization procedure (\textsc{Initialize})}\label{table of Initialization}
\begin{tabular}{p{400pt}}
\hline
\textbf{Input}:  $\bP_{\Omega}(M)$,  target rank $r$, target row norm bounds $\beta_1, \beta_2$. \\ % sample probability $p = |\Omega|/(mn)$,
\hline
\textbf{Algorithm} \textsc{Initialize}($\bP_{\Omega}(M), p, r$). \\
\hline
 $\text{} \quad$ 1. {\black Compute $(\bar{X}_0, D_0, \bar{Y}_0 ) = \mathrm{SVD}_r \left(  \frac{1}{p} \bP_{\Omega}(M) \right) $, as defined in  \eqref{SVD_r def}. } \\  % \left( \mathrm{P}_r and \eqref{best rank r def}
 $\text{} \quad \quad \; $  Compute $\hat{X}_0 = \bar{X}_0 D_0^{1/2}, \hat{Y}_0 = \bar{Y}_0 D_0^{1/2} $ .    \\
 $\text{} \quad$ 2. For each row of $\hat{X}_0$ (resp.$\ \hat{Y}_0$) with norm larger than $ \sqrt{\frac{2}{3}}\beta_1 $ (resp.$\ \sqrt{\frac{2}{3}}\beta_2$),
 % {\blue Remark: I'm not sure I correctly understand your modification of adding slash and space between ``resp.'' and ``$\sqrt{\frac{2}{3}}\beta_2$''; is the space too large or too small? I remove the space between the two in the text, and add a slash and space insider the dollar signs. }
 scale it to make the norm of this row equal  $ \sqrt{\frac{2}{3}}\beta_1 $ (resp.$\ \sqrt{\frac{2}{3}}\beta_2$) to obtain $X_0, Y_0$, i.e.\   \begin{equation}\label{initial point rescale}
\begin{split}
X_0^{(i)} =   \frac{\hat{X}_0^{(i)}}{\|\hat{X}_0^{(i)} \|}  \min \left\{ \|\hat{X}_0^{(i)} \|, \sqrt{\frac{2}{3 }} \beta_1 \right\}, i=1,\dots,m.
 \\  Y_0^{(j)} =  \frac{\hat{Y}_0^{(j)}}{\|\hat{Y}_0^{(j)} \|}  \min \left\{\|\hat{Y}_0^{(j)} \|, \sqrt{\frac{2}{3 }} \beta_2 \right\}, j=1,\dots,n.
 \end{split}
\end{equation}
  \\
\hline
\textbf{Output} $X_0 \in \dR^{m \times r}, Y_0 \in \dR^{n \times r} . $ \\
\hline
\end{tabular}
\end{table*}

% As $C_0$ increases, $C_d$ will increase linearly with $\sqrt{C_0}.$ However, the distance is still $\mathcal{O}(\Sigma_{\rmin})$ and cannot be made
% arbitrarily close to zero.

\subsection{Algorithms}
Our result applies to many standard algorithms such as gradient descent, SGD and block coordinate descent type methods (including alternating minimization, block coordinate gradient descent, block successive upper bound minimization, etc.).
%{\red Due to the introduction of soft regularizers, these standard algorithms have different forms than }
We will describe several typical algorithms in this subsection.

The gradient $\nabla \tilde{F} = \nabla F + \nabla G = (\nabla_X F + \nabla_X G, \nabla_Y F + \nabla_Y G  )$ can be easily computed as follows:
\begin{equation}\label{grad of F,G expression}
\begin{split}
  \nabla_X F(X,Y) & = \bP_{\Omega}(XY^T - M) Y,   \\
  \nabla_Y F(X,Y) & =  \bP_{\Omega}(XY^T - M)^T X,   \\
    \nabla_X G(X,Y) & = \rho \sum_{i=1}^m G_0^{\prime}(\frac{3\|  X^{(i)} \|^2  } {2\beta_1^2} ) \frac{3 \bar{X}^{(i)} } {\beta_1^2} + \rho G_0^{\prime}(\frac{3\| X\|_F^2 } {2\beta_T^2} ) \frac{3 X}{\beta_T^2},    \\
      \nabla_Y G(X,Y) & = \rho \sum_{j=1}^n G_0^{\prime}(\frac{3 \|Y^{(j)}  \|^2 } {2\beta_2^2} ) \frac{3 \bar{Y}^{(j)}  }{\beta_2^2} +
      \rho G_0^{\prime}(\frac{3\| Y \|_F^2 }  {2\beta_T^2} ) \frac{3 Y }{\beta_T^2},
\end{split}
\end{equation}
where $ G_0^{\prime}(z) =I_{[1,\infty]}(z) 2(z-1)  $, % e^{(z-1)^2}
and $\bar{X}^{(i)}$ (resp. $\bar{Y}^{(j)}$) denotes a matrix with the $i$-th (resp. $j$-th) row being $X^{(i)} $ (resp. $Y^{(j)}$) and the other rows being zero.

We first present a gradient descent algorithm in Table \ref{table of Algorithm 1}.
There are many choices of stepsizes such as constant stepsize, exact line search, limited line search, diminishing stepsize and Armijo rule \cite{bertsekas1999nonlinear}. {\color{black} We present three stepsize rules here: constant stepsize, restricted Armijo rule and restricted line search (the latter two are the variants of
 Armijo rule and exact line search).
Note that the restricted line search rule is similar to that used in \cite{keshavan2010matrix} for the gradient descent method over Grassmannian manifolds.
To simplify the notations, we denote  $\bm x_k(\eta) \triangleq( X_k(\eta), Y_k(\eta) )$
 and $d(\bm x_k(\eta), \bm x_0) \triangleq \sqrt{\| X_{k}(\eta) - X_0 \|_F^2
     + \| Y_{k}(\eta) - Y_0 \|_F^2}. $  }
%\footnote{For other stepsize rules with extra constraint
% $\sqrt{\| X_{k}(\eta) - X_0 \|_F^2  + \| Y_{k}(\eta) - Y_0 \|_F^2} \leq \frac{4\delta}{5}$, we can also prove the global convergence by an argument similar to % that in \cite[Sec. VI.D]{keshavan2010matrix}. } .
% The convergence to stationary points by these variants require  }.
 % two of them here. The second stepsize rule was also used in \cite{keshavan2010matrix} for the gradient descent algorithm over Grassmanian manifold.
 % two of them here.
% will prove all three choices of stepsizes work.
\begin{table*}[htbp]\caption{Algorithm 1 (Gradient descent)}\label{table of Algorithm 1}
\begin{tabular}{p{400pt}}
%\hline
%Algorithm 1 (Gradient descent) \\
\hline
Initialization: $ (X_0,Y_0) \longleftarrow \textsc{Initialize}(\bP_{\Omega}(M), p, r ) $.  \\
The $k$-th iteration:  \\
$ \quad \quad \quad \quad \quad \quad  X_k \longleftarrow X_{k}(\eta_k) \triangleq  X_{k-1} -  \eta_k \nabla_X \tilde{F}(X_{k-1}, Y_{k-1}), $ \\
 % TO RECOVER    %
$ \quad \quad \quad \quad \quad \quad  Y_k \longleftarrow Y_{k}(\eta_k) \triangleq  Y_{k-1} -  \eta_k \nabla_Y \tilde{F}(X_{k-1}, Y_{k-1}), $ \\
$ \quad$  where the stepsize $\eta_k$ is chosen according to one of the following rules: \\ % the following rule (constant stepsize):  \\  %
  $ \quad \quad  $ a) Constant stepsize:  $ \eta_k = \eta \leq \bar{\eta}_1, \ \forall \ k $ ($\bar{\eta}_1$ is a constant defined by
   \eqref{eta 1 def} in Appendix \ref{appen: prove Claim of algorithm property}).    \\  %  TO BE RECOVERED:
   $ \quad \quad$ {\color{black}b) Restricted Armijo rule: Let $\sigma \in (0,1), \xi \in (0,1), s_0  $ be fixed scalars. } \\
            $ \quad \quad \quad \quad  $    b1) Find the smallest nonnegative integer $i $ such that  $  d(\bm x_k(\xi^{i} s_0 ) , \bm x_0 ) \leq 5\delta/6 $ and \\
             $ \text{ } \quad \quad
            \quad \quad \quad \quad \quad \quad \tilde{F}(\bm x_k( \xi^{i} s_0) ) \leq \tilde{F}(\bm x_{k-1} ) - \sigma \xi^{i} s_0
            \| \nabla \tilde{F}(\bm x_{k-1})\|_F^2 $.  \\
            $ \quad \quad \quad \quad $    b2) Let $\eta_k =  \xi^{i} s_0$.  \\
  $ \quad \quad$   {\color{black} c) Restricted line search: $ \eta_k = \arg \min_{\eta \in \dR, d(\bm x_k(\eta) , \bm x_0 ) \leq 5\delta/6 } \tilde{F}(\bm x_k(\eta)) . $  }
    \\
\hline
\end{tabular}
\end{table*}
% $ \quad \quad$ b) Diminishing stepsize: $\eta_k = \eta \leq 1/L. $  \\
% a)  a initial guess of stepsize $s $

AltMin (alternating minimization) belongs to the class of block coordinate descent (BCD) type methods.
%While the original BCD algorithm cyclically updates each block of variables by solving the subproblem exactly,
One can update the blocks in different orders (e.g. cyclic \cite{tseng2001convergence,
sun2015improved,sun2016worst}, randomized {\cite{nesterov2012efficiency}} or parallel) and solve the subproblem inexactly.
Commonly used inexact BCD type algorithms include BCGD (block coordinate gradient descent, which updates each variable by a single gradient step \cite{nesterov2012efficiency}) and BSUM (block successive upper bound minimization, which updates each variable by minimizing an upper bound
of the objective function \cite{razaviyayn2013unified}).
BCD-type methods have been widely used in engineering (e.g. \cite{baligh2014cross,BScluster}).  % sun2013long,
% (e.g. perform one gradient step , or minimizing an upper bound of the objective function \cite{razaviyayn2013unified})
%,  lead to different algorithms
% (although \cite{yu2012scalable} considers the problem \eqref{P0}, similar algorithms can be applied to \eqref{P1}).
 In the context of matrix completion, Hastie et al. \cite{hastie2014matrix} proposed an algorithm
 that could be viewed as a BSUM algorithm.
 Just considering different choices of the blocks will lead to different algorithms for the matrix completion problem \cite{yu2012scalable}.
 % ((although \cite{yu2012scalable} considers (P0), similar algorithms can be applied to (P1)).
% Therefore, the class of BCD type methods contains more than, say, twenty different algorithms.
Our result applies to many BCD type methods, including the two-block alternating minimization, BCGD and BSUM.
% We think that our results can be generalized to other versions such as parallel update and/or randomized order.
While it is not very interesting to list all possible algorithms to which our results are applicable,
we just present two specific algorithms for illustration.

The first BCD type algorithm we present is (two-block) AltMin, which, in the context of matrix completion,
usually refers to the algorithm that alternates between $X$ and $Y$ by updating one factor at a time with the other factor fixed.
Although the overall objective function is non-convex, each subproblem of $X$ or $Y$ is convex and thus can be solved efficiently.
The details are given in Table \ref{table of Algorithm 2, AltMin}.
% Each iteration, we need to solve a convex problem to update $X$ or $Y$, which can be done efficiently.
% by fixing one factor and minimizing the objective function with respect to anther factor.

\begin{table*}[htbp]\caption{Algorithm 2 (Two-block Alternating Minimization)}\label{table of Algorithm 2, AltMin}
\begin{tabular}{p{400pt}}
%\hline
%Algorithm 2 (Alternating minimization) \\
\hline
Initialization: $ (X_0,Y_0) \longleftarrow \textsc{Initialize}(\bP_{\Omega}(M), p, r ) $.  \\
The $k$-th iteration:  \\
$ \quad \quad \quad \quad \quad \quad  X_k \longleftarrow \arg \min_X \tilde{F}(X,Y_{k-1}), $ \\
$ \quad \quad \quad \quad \quad \quad  Y_k \longleftarrow \arg \min_Y \tilde{F}(X_{k-1},Y).  $ \\
\hline
\end{tabular}
\end{table*}

 For the case without the regularization term $G(X,Y)$, the objective function becomes $F(X,Y)$ and
  is quadratic with respect to $X$ or $Y$. Thus $X_k,Y_k$ have closed form update. Suppose $X^T = (x_1, \dots, x_m)$ and $Y^T = (y_1, \dots, y_n)$, where $x_i, y_j \in \dR^{r \times 1}$.
Then $ (x_1^*,\dots, x_m^*) \triangleq (\arg \min_X F(X,Y))^T $ and $ (y_1^*,\dots, y_n^*) \triangleq (\arg \min_Y F(X,Y) )^T $
are given by
\begin{equation}\label{ALS update}
\begin{split}
  x_i^* = (\sum_{j \in \Omega_{i}^{x} } y_j y_j^T )^{\dag} (\sum_{j\in \Omega_i^{x}} M_{ij} y_j ), \; i=1,\dots, m, \\
  y_j^* = (\sum_{i \in \Omega_j^y } x_i x_i^T )^{\dag} (\sum_{i\in \Omega_j^{y}} M_{ij} x_i ), \; j=1,\dots, n,
  \end{split}
\end{equation}
where $\Omega_i^x = \{j \mid (i,j) \in \Omega \}, \Omega_j^y = \{i \mid (i,j) \in \Omega \}$, and $A^{\dag}$ denotes the pseudo inverse of a matrix $A$.
For our problem with the regularization term $G(X,Y)$, we no longer have closed form update of $X_k, Y_k$.
One way to solve the convex subproblems is to start from the solution given in \eqref{ALS update} and then apply the gradient descent method until convergence.
The details for solving $ \min_X \tilde{F}(X,Y) $ is given in Table \ref{table of AltMin subproblem} (the stepsize can be chosen by one of the standard rules
of the gradient descent method), and the other subproblem $\min_Y \tilde{F}(X,Y) $ can be solved in a similar fashion.

Theoretically speaking, AltMin for our formulation (P1) is not as efficient as the vanilla AtlMin for (P0) since an extra inner loop is needed to solve the subproblem.
However, we remark that in the regimes of $|\Omega|$ that the vanilla AltMin works,
 the least square solution $X$ (resp.\ $Y$) is always bounded and incoherent (empirical observation),
 in which case the regularizer $G$ is inactive; therefore, the gradient updates in Table \ref{table of AltMin subproblem} do not happen.
In the regimes of $|\Omega|$ that the vanilla AltMin fails, $G$ is active and the gradient updates do happen;
however, instead of solving the subproblem exactly, one could perform one gradient step and the algorithm becomes the popular variant BCGD \cite{nesterov2012efficiency}. Our main result of exact recovery still holds for BCGD (the proof for Algorithm 3 in Claim \ref{Algo 1-3 satisfy conditions} can be applied to BCGD since BCGD is a special case of BSUM).

%% ----------- Delete on May 22, 2015; see modified version above -----
%%--------------Delete due to space; ADD BACK IN 11/26--------------
%Theoretically speaking, alternating minimization for our formulation (P1) is not as efficient as that for (P0) since an extra inner loop is needed to solve the subproblem.
%% due to the regularization term $G(X,Y)$.
%However, we remark that in many practical scenarios the gradient updates in Table \ref{table of AltMin subproblem} are unnecessary and the least square solution $X$ given in the initialization step is already the minimizer of $\tilde{F}(X,Y)$. The reason is that oftentimes the least square solution $X$ (resp.\ $Y$) is both bounded and incoherent,
%i.e.\ $\| X \|_F \leq \sqrt{2/3} \beta_T $ and $ \| X^{(i)}\| \leq  \sqrt{2/3} \beta_1 $, which implies $ \nabla_X G(X,Y) = 0 $ and thus $\nabla_X \tilde{F}(X,Y) = 0$ (since the least square solution is obtained by solving $\nabla_X F(X,Y)=0$).
%Moreover, instead of solving the subproblem exactly, one could just perform one gradient step and the algorithm becomes the popular variant BCGD \cite{nesterov2012efficiency}. Our main result of exact recovery still holds for BCGD (the proof for Algorithm 3 in Claim \ref{Algo 1-3 satisfy conditions} can be applied to BCGD since BCGD is a special case of BSUM).
% % (because the proof for Algorithm 1 in Claim \ref{Algo 1-3 satisfy conditions} can be applied to BCGD).
%% and omitted in this paper).
%% also solve the subproblem inexactly (e.g. perform one gradient update).
%
%% ----------------------------

\begin{table*}[htbp]\caption{Solving subproblem of Algorithm 2}\label{table of AltMin subproblem}
\begin{tabular}{p{400pt}}
\hline
Solving subproblem of Algorithm 2: $ \min_X \tilde{F}(X,Y) $. \\
\hline
Input: $Y = (y_1, \dots, y_n) \in \dR^{n \times r}$. \\
Initialization: $X = (x_1,\dots, x_m)$, where $x_i = (\sum_{j \in \Omega_{i}^{x} } y_j y_j^T )^{\dag} (\sum_{j\in \Omega_i^{x}} M_{ij} y_j ), \; i=1,\dots, m$,  \\
Repeat:  \\
$ \quad \quad \quad \quad \quad \quad  X \longleftarrow  X -  \eta \nabla_X \tilde{F}(X, Y), $ \\
Until Stopping criterion is met. \\
\hline
\end{tabular}
\end{table*}

In the second BCD type algorithm called row BSUM, we update the rows of $X$ and $Y$ cyclically by minimizing an upper bound of the objective function; see Table \ref{table of Algorithm 3, RowCD}. %, and we call this algorithm
The extra terms $ \frac{\lambda_0}{2} \| X^{(i)} - X_{k-1}^{(i)} \|^2 $ or $\frac{\lambda_0}{2}\| Y^{(j)} - Y_{k-1}^{(j)} \|^2 $
are added to make the subproblems strongly convex, which help prove convergence to stationary points.
Such a technique has also been used in the alternating least square algorithm for tensor decomposition \cite{razaviyayn2013unified}.
Note that for the two-block BCD algorithm, convergence to stationary points can be guaranteed even when the subproblems are not strongly convex \cite{grippo2000convergence},
thus in Algorithm 2 we do not add the extra terms.
The benefit of cyclically updating the rows is that each subproblem can be solved efficiently using a simple binary search; see
Appendix \ref{appen: Algorithm 3 subproblem} for the details.
We remark again that instead of solving the subproblem exactly, one could just perform one gradient step to update each row of $X$ and $Y$ (with $\lambda = 0$)
and our result still holds.
%Recall that in alternating minimization, we update all rows of $X$ or $Y$ simultaneously in one iteration.
%For the case $\rho = 0$ (i.e.\ no regularization terms), the two algorithms are the same since $F$ is decomposable across each $X^{(i)}$ and $Y^{(j)}$.
%For the general case $\rho >0$, $\tilde{F}$ is not decomposable due to the coupling of different rows in the terms $\rho G_0(\frac{3\| X\|_F^2}{2\beta_T^2})$
%and $\rho G_0(\frac{3\| Y\|_F^2}{2\beta_T^2})$.
% One way to solve the convex subproblems is to start from the least square solution and then perform gradient update until convergence.

\begin{table*}[htbp]\caption{Algorithm 3 (Row BSUM)}\label{table of Algorithm 3, RowCD}
\begin{tabular}{p{400pt}}
%\hline
%Algorithm 3 (Row BSUM) \\
\hline
Initialization: $ (X_0,Y_0) \longleftarrow \textsc{Initialize}(\bP_{\Omega}(M), p, r ) $.  \\
Parameter: $\lambda_0 > 0$. \\
The $k$-th loop:  \\
$ \quad \quad$ For $i$ = 1 to $m$: \\
$ \quad   X_k^{(i)} \longleftarrow \arg \min_{X^{(i)}} \tilde{F}(X_{k}^{(1)}, \dots,X_{k}^{(i-1)}, X^{(i)}, X_{k-1}^{(i+1)}
  \dots, X_{k-1}^{(m)}, Y_{k-1}) + \frac{\lambda_0}{2} \| X^{(i)} - X_{k-1}^{(i)} \|^2,  $ \\
$ \quad \quad$ For $j$ = 1 to $n$:  \\
$ \quad  Y_k^{(j)} \longleftarrow \arg \min_{Y^{(j)}} \tilde{F}(X_{k}, Y_{k}^{(1)}, \dots, Y_{k}^{(j-1)}, Y^{(j)}
, Y_{k-1}^{(j+1)},\dots, Y_{k-1}^{(m)} ) + \frac{\lambda_0}{2}\| Y^{(j)} - Y_{k-1}^{(j)} \|^2  . $ \\
\hline
\end{tabular}
\end{table*}

The fourth algorithm we present is SGD (stochastic gradient descent) \cite{funk2006netflix,koren2009matrix} tailored for our problem (P1).
In the optimization literature, this algorithm for minimizing the sum of finitely many functions is more commonly referred to as ``incremental gradient method'', while SGD represents the algorithm for minimizing the expectation of a function; nevertheless, in this paper we follow the convention in the computer science literature and still call it ``SGD''.
In SGD, at each iteration we pick a component function and perform a gradient update.
Similar to the BCD type methods where the blocks can be chosen in different orders, one can pick the component functions in a cyclic order, in an essentially
cyclic order, or in a random order (either sampling with replacement or without replacement).
In practice, the version of sampling without replacement converges much faster than the version of sampling with replacement (see \cite[Chapter 2]{sun2015thesis} for simulation results). In general, the understanding of sampling without replacement for optimization algorithms is quite limited (see, e.g., \cite{sun2015expected} for one example of such analysis).

 In this paper we only consider the cyclic order, and use a standard stepsize rule for SGD \cite{luo1994analysis,bertsekas2000gradient} which requires
the stepsizes $\{ \eta_k \}$ to go to zero as $k\rightarrow \infty$, but neither too fast nor too slow (this choice guarantees convergence to stationary points even for nonconvex problems). One such choice of stepsizes is $\eta_k = O(1/k)$.
We remark that our results also apply to other versions of SGD with different update orders or stepsize rules as long as they converge to stationary points.

To apply SGD to our problem, we decompose the objective function $\tilde{F}(X,Y)$ as follows:
$$
  \tilde{F}(X,Y) = \sum_{(i,j)\in \Omega} F_{ij}(X,Y) + \sum_{i=1}^m G_{1i}(X) + \sum_{j=1}^n G_{2j}(Y) + G_3(X) + G_4(Y)
                 = \sum_{k = 1}^{ |\Omega| + m + n +2 } f_{k}(X,Y),
$$
where the component functions
\begin{equation}\label{component fn def}
\begin{split}
F_{ij}(X,Y) = [(XY^T - M)_{ij}]^2  & = [ (X^{(i)})^T Y^{(j)} -  M_{ij} ]^2, \; (i,j) \in \Omega,   \\
G_{1i}(X) = \rho G_0(\frac{3\| X^{(i)} \|^2  } {2\beta_1^2} ) , 1\leq i \leq m, \;\;
  &  G_{2j}(Y) = \rho G_0(\frac{3 \| Y^{(j)} \|^2 } {2\beta_2^2} ) , 1\leq j \leq n,  \\
 G_3(X) = \rho G_0(\frac{3\| X\|_F^2 } {2\beta_T^2} ) ,  \;\;
 &  G_4(Y) =  \rho G_0(\frac{3\| Y \|_F^2 } {2\beta_T^2} )
\end{split}
\end{equation}
and $ \{ f_k(X,Y)\}_{k=1}^{|\Omega| + m + n +2}$ denotes the collection of all component functions.
With these definitions, the SGD algorithm is given in Table \ref{table of Algorithm 4, SGD}.

% with an extra requirement that all stepsizes are bounded above by a constant. diminishing stepsize rule which is a standard choice

\begin{table*}[htbp]\caption{Algorithm 4 (SGD)}\label{table of Algorithm 4, SGD}
\begin{tabular}{p{400pt}}
%\hline
%Algorithm 4 (SGD) \\
\hline
Initialization: $ (X_0,Y_0) \longleftarrow \textsc{Initialize}(\bP_{\Omega}(M), p, r ) $.  \\
Parameters: $\eta_k, k=0,1,\dots$ satisfying $\sum_k \eta_k = \infty, \sum_k \eta_k^2 < \eta_{\mathrm{sum}}$ and $0 < \eta_k \leq \bar{\eta}$, \\
$ \text{ } \quad \quad \quad \quad \quad  $ where $\eta_{\mathrm{sum}} $ and $\bar{\eta}$ are constants specified in Appendix \ref{appen: prove Claim of algorithm property}. \\
 % for suitably chosen $\bar{\eta}$.  \\
The $(k+1)$-th loop:  \\
$ \quad \quad$ $X_{k,0} \longleftarrow X_{k}, \quad Y_{k,0} \longleftarrow Y_{k}. $  \\
$ \quad \quad$ For $i$ = 1 to $|\Omega| + m + n +2$ : \\
$ \quad \quad \quad \quad \quad \quad  X_{k,i} \longleftarrow  X_{k,i-1} -  \eta_k \nabla_X f_i(X_{k, i-1}, Y_{k, i-1}), $ \\
$ \quad \quad \quad \quad \quad \quad  Y_{k,i} \longleftarrow  Y_{k,i-1} -  \eta_k \nabla_Y f_i(X_{k,i-1}, Y_{k,i-1}). $ \\
$ \quad \quad$ End  \\
$ \quad \quad$ $ X_{k+1} \longleftarrow X_{k,|\Omega| + m + n +2 }, \quad Y_{k+1} \longleftarrow Y_{k,|\Omega| + m + n +2}. $ \\
\hline
\end{tabular}
\end{table*}

\section{Main Results }\label{sec: main result}
%For technical reasons, we replace the original objective function $F(X,Y,Z) = \frac{1}{2}\|Z - XY^T\|_F^2$ by a regularized version $\tilde{F}(X,Y,Z) = F(X,Y,Z) + G(X,Y),$ which will be defined later.
 % another way to state the theorem: M is generated randomly such that it is $(\mu_0,\mu_1)$ incoherent with high probability.
 % In following way, M is assumed to be "deterministic", and "high prob." is because of the randomness of "\Omega".
The main result of this paper is that Algorithms 1-4 (standard optimization algorithms) will converge to the global optima of problem (P1) given in \eqref{P1} and reconstruct $M$ exactly with high probability, provided that the number of revealed entries is large enough.
Similar to the results for nuclear norm minimization \cite{candes2009exact,candes2010power, recht2011simpler,gross2011recovering}, the probability is taken with respect to the random choice of $\Omega$, and the result also applies to a uniform random model of $\Omega$.

%, and ``with probability $99\%$'' means that out of all possible sets $\Omega$ with a given size, $99\%$ of them can lead to exact reconstruction by Algorithm 1-4 \emph{for sure}.
% We will prove this theorem in Section \ref{section: proof of main thm, and lemmas}.

%--------------Main result --------------------------------
\begin{thm}\label{major theorem}(Exact Recovery)                    %  and let $\mu = max\{\mu_0,\mu_1\}$
Assume a rank-$r$ matrix $M \in \dR^{m \times n}$ is $\mu$-incoherent. Suppose the condition number of $M$ is $\kappa$ and{\black $\alpha = m/n \geq 1$.}
 Then there exists a numerical constant $C_0$ such that: if $\Omega$ is uniformly generated at random with size
\begin{equation}\label{Omega bound}
  |\Omega| \geq C_0 \alpha n r \kappa^2 \max\{ \mu \log n, \sqrt{\alpha} \mu^2 r^{6}  \kappa^4 \},  % |\Omega| \geq C_0 \alpha^{\frac{3}{2}}  \mu^2 r^7 \kappa^4 n \log n,
\end{equation}
then with probability at least $1 -2/n^4$, each of Algorithms 1-4 reconstructs $M$ exactly.
Here, we say an algorithm reconstructs $M$ if  each limit point $(X^*, Y^*)$ of the sequence $\{X_k,Y_k\}$ generated by this algorithm satisfies $X^* (Y^*)^T =  M$.
\end{thm}
%assume all singular values of $M$ lie in the region $[\Sigma_{\rmin},\Sigma_{\rmax}],$ where $\Sigma_{\rmin},\Sigma_{\rmax}$ are bounded away from $0$ and $\infty$

{\black This result shows that although \eqref{P1} is a non-convex optimization problem, many standard algorithms can converge to the global optima
with certain initialization.  }
{\black
Different from all previous works on alternating minimization for matrix completion,
our result does not require the algorithm to use independent samples in different iterations.
To the best of our knowledge, our result is the first one that provides theoretical guarantee for alternating minimization without resampling.
In addition, this result also provides the first \emph{exact} recovery guarantee for many algorithms such as gradient descent, SGD and BSUM.
}

As demonstrated in \cite{candes2009exact} (and proved in \cite[Theorem 1.7]{candes2010power}), $O( n r \log n)$ entries are the minimum requirement to recover the original matrix:
$O(nr)$ is the number of degrees of freedom of a rank $r$ matrix $M$, and the additional $\log n$ factor is due to the coupon collector effect \cite{candes2009exact}.
For $r = O(1) $ and $\kappa$ bounded, Theorem \ref{major theorem} is order optimal in terms of the sample complexity since  only $O(n \log n )$ entries are needed to exactly recover $M$. % provides the optimal sample complexity in the sense that
For $r = \mathcal{O}( \log n)$, however,
our result is suboptimal by a polylogarithmic factor.
The initialization has contributed $r^4 \kappa^4 $ to the sample complexity bound, and we expect that using other initialization procedures (e.g.
the one proposed in \cite{hardt2013understanding}) can reduce the exponents of $r$ and $\kappa$.

{\black
Theorem \ref{major theorem} only establishes the convergence, but not the convergence speed.
With some extra effort, we can prove the linear convergence of the gradient descent method (see Theorem \ref{theorem 2: linear convergence} below).
Again, this result can be extended beyond the gradient descent method.
In fact, by a standard optimization argument, we can prove the linear convergence of
any algorithm that satisfies ``sufficient decrease'' (i.e. $\tilde{F}(\bm x^k)
- \tilde{F}(\bm x^{k+1}) \geq O( \| \nabla \tilde{F}(\bm x^k) \|_F^2 )$) and the requirements in Lemma \ref{lemma of incoherent neighborhood};
see Corollary \ref{coro: linear convergence for other algorithms}.
Many first order methods, including alternating type methods (e.g. BCGD, two-block BCD), can be shown to have the sufficient decrease property
under mild conditions. For space reason, we do not verify all the methods considered in this paper, but only present the linear convergence result for the gradient descent method.
The proof of Theorem \ref{theorem 2: linear convergence} is given in Section \ref{sec: proof of thm 2}.
%(it is pronot hard to extend it to other algorithms).
% We do not present the time complexity bounds in this version for space reason.

\begin{thm}\label{theorem 2: linear convergence}(Linear convergence)                    %  and let $\mu = max\{\mu_0,\mu_1\}$
Under the same condition of Theorem \ref{major theorem}, with probability at least $1 -2/n^4$, Algorithm 1a (gradient descent with constant stepsize) converges linearly; more precisely, the sequence $\{X_k,Y_k\}$ generated by Algorithm 1a satisfies
\begin{equation}\label{linear convergence property}
  \tilde{F}(X_k, Y_k ) \leq (1 - \frac{1 }{2} \eta_1 \xi )^k,
\end{equation}
where $ \xi = \frac{1 }{C_g r^5 \kappa^3 } p \Sigma_{\min} $ (here $C_g$ is a numerical constant), $\eta_1$ is the stepsize and
$\eta_1 \xi < 1$.
\end{thm}

The linear convergence will immediately lead to a time complexity of $\tilde{O}(\text{poly}(n) \log \frac{1}{\epsilon})$ for achieving any $\epsilon$-optimal
solution, where the $\tilde{O}$ notation hides factors polynomial in $r,\kappa, \alpha$.
We conjecture that the time complexity bound can be improved to $\tilde{O}( |\Omega| \log(1/ \epsilon) )$ as observed in practice.
 However, finding the optimal time complexity bound is not the focus of this paper, and is left as future work.
% but we are not able to prove this by the current techniques.

The above result shows that $\tilde{F}(X_k, Y_k)$ converges to zero at a linear speed.
Note that  $\tilde{F}(X,Y) = 0$ (global convergence) only implies $\bP_{\Omega}(M - X Y^T) = 0$,
not necessarily $M = XY^T$ (exact reconvery). % there is a difference between global convergence and recovering $M$:
 The following lemma implies that with high probability (for random $\Omega$) the global convergence implies the exact recovery.
In fact, it shows that the observed loss $\|\bP_{\Omega}(M-XY^T) \|_F^2 $ is on the order of the recovery error $p\|M-XY^T \|_F^2$ if $(X,Y)$ lies in an incoherent neighborhood of $M$.
As discussed in the introduction, this lemma
can also be viewed as a geometrical property of $f_{\Omega}(Z) = \|\bP_{\Omega}(M-Z) \|_F^2$
in a local incoherent region (view
$\bP_{\Omega}(Z-M) $ as the gradient of $f_{\Omega}(Z)$).
\begin{claim}\label{lemma: P Omega and P has relation}
% There exist numerical constants $C_0,C_1$ such that the following happens.  Assume $\Omega$ is uniformly randomly generated with
% size $|\Omega|$ satisfying \eqref{Omega bound} and assume $\delta \leq \sqrt{C_1/C_0} \frac{\Sigma_{\min}}{r^4 \kappa^3 }$.
Under the same condition of Theorem \ref{major theorem}, with probability at least $1-1/(2n^4)$, we have
% For any $(X,Y)\in K_1 \cap K_2 \cap K(\delta)$,
\begin{equation}\label{RSC of P_Omega}
\begin{split}
\frac{1}{3} p\|M-XY^T \|_F^2 \leq \|\bP_{\Omega}(M-XY^T) \|_F^2 \leq 2 p\|M-XY^T \|_F^2, \\
 \quad \forall (X,Y) \in K_1\cap K_2 \cap K(\delta).
 \end{split}
\end{equation}
\end{claim}
 The proof of this claim is given in Appendix \ref{appen: RSC proof}.
 This result is a simple corollary of several intermediate bounds established in the proof of Lemma \ref{lemma main; about local convexity}.
 % in Section \ref{section: proof of main thm, and lemmas}.
% can be viewed as an intermediate step in proving Theorem \ref{major theorem}.

}

% {\black Compared to the result of \cite{keshavan2010matrix}, our exponent of $r$ is larger while the exponent of $\kappa$ is smaller. It is not clear whether the large exponent of $r$ is due to the weakness of the proof technique or the nature of the problem.}
% our exponents of $r$ increase from 2 to 10, and the exponents of $\kappa$ increases from 6 to 8.

\subsection{ Proof of Theorem \ref{major theorem} and main lemmas }\label{section: proof of main thm, and lemmas}
To prove Theorem \ref{major theorem}, we only need to prove two lemmas which describe the local geometry of the regularized objective in (P1)
and the properties of the algorithms respectively.
Roughly speaking, the first lemma shows that any stationary point of (P1) in a certain region is globally optimal, and the second lemma shows that
each of Algorithms 1-4 converges to stationary points in that region. This region can be viewed as an ``incoherent neighborhood'' of $M$, and can be formally
defined as $K_1 \cap K_2 \cap K(\delta)$, where $K_1, K_2$ are defined as
 \begin{equation}\label{def of K_1, K_2}
 \begin{split}
 K_1 & \triangleq \{ (X,Y)| X \in \dR^{m\times r}, Y \in \dR^{n\times r}, \|X^{(i)} \| \leq \beta_1, \|Y^{(j)} \| \leq \beta_2 , \forall i,j \},  \\
 K_2 & \triangleq \{ (X,Y)| X \in \dR^{m\times r}, Y \in \dR^{n\times r},  \|X\|_F \leq \beta_T,  \|Y\|_F \leq \beta_T \}.
  \end{split}
 \end{equation}
 and $K(\delta)$ is defined as
\begin{equation}\label{def of K(d)}
 K( \delta) \triangleq \{ (X,Y)| X \in \dR^{m\times r}, Y \in \dR^{n\times r}, \|M-XY^T \|_F \leq \delta \}.
 \end{equation}
 Note that $K_2 = \Gamma(\beta_T)$ by our definition of $\Gamma$ in \eqref{Gamma def}.
As mentioned in Section \ref{sec: assumptions}, we only need to consider a Bernolli model of $\Omega$ where each entry is included into $\Omega$
with probability $p = \frac{ S }{mn}$, where $S $ satisfies \eqref{Omega bound}.

The first lemma describes the local geometry and implies that any stationary point $(X,Y)$ in $ K_1 \cap K_2 \cap K(\delta) $ satisfies $XY^T = M$.
The main steps to derive this geometrical property is described
in Section \ref{sec: proof overview}.
The formal proof will be given in Section \ref{sec: proof of lemma 1}.
% implies that any stationary point $(X^*, Y^*)$ of (P1) that is close to $M$, incoherent and bounded above satisfies $X_*Y_*^T = M$.
 %---------------Lemma 2-----------------------
\begin{lemma}\label{lemma main; about local convexity}
There exist numerical constants $C_0,C_d$ such that the following holds. Assume $\delta$ is defined by \eqref{delta definition throughout} and
$\Omega$ is generated by a Bernolli model with expected cardinality $S$ satisfying \eqref{Omega bound}
 (i.e. $S$ is lower bounded by the right hand side of \eqref{Omega bound}).
%\begin{equation}\label{bound on Omega}
%|\Omega| \geq C_0 \sqrt{\alpha} nr \kappa^2 \max\{ \mu \log n, \mu^2 r^{11} \sqrt{\alpha} \kappa^6 \},
%\end{equation}
%\begin{equation}\label
%\delta \triangleq \sqrt{C_1/C_0} \frac{\Sigma_{\rmin}}{r^4 \kappa^3}.
%\end{equation}
 Then, with probability at least $1-1/n^4$, the following holds:
for all $(X,Y) \in K_1 \cap K_2 \cap K(\delta),$ there exist $U\in \mathbb{R}^{m \times r},\mathbb{V}\in R^{n \times r},$ such that
$UV^T =M$ and
\begin{equation}\label{local relative convexity}
\langle \nabla_X \tilde{F}(X,Y), X-U \rangle + \langle \nabla_Y \tilde{F}(X,Y), Y-V \rangle \geq \frac{p}{4} \|M-XY^T \|_F^2.
\end{equation}
\end{lemma}

The second lemma describes the properties of the algorithms we presented.
Throughout the paper,
``under the same condition of Lemma \ref{lemma main; about local convexity}'' means
``assume $\delta$ is defined by \eqref{delta definition throughout} and $\Omega$ is generated by a Bernolli model with expected cardinality $S$ satisfying \eqref{Omega bound}, where $C_0, C_d$ are the same numerical constants as those in Lemma \ref{lemma main; about local convexity}''.
The proof of Lemma \ref{lemma of incoherent neighborhood} will be given in Section \ref{sec: proof of lemma 2}.
%----------Lemma 2-------------
\begin{lemma}\label{lemma of incoherent neighborhood}
Under the same conditions of Lemma \ref{lemma main; about local convexity}, with probability at least $1-1/n^4$, the sequence $(X_k,Y_k)$ generated by either of Algorithms 1-4 has the following properties: \\
(a) Each limit point of $(X_k, Y_k)$ is a stationary point of (P1). \\
(b) $ (X_k, Y_k) \in K_1 \cap K_2 \cap K(\delta), \; \forall k \geq 0$.
% (c)  $(X_k, Y_k) \in K(\delta), \forall k \geq 0$.
\end{lemma}

Intuitively, $\|X_k^{(i)} \|, \|Y_k^{(j)} \|,\|X_k\|_F, \|Y_k\|_F  $  are bounded because of the regularization terms we introduced and that the objective function is decreasing, and $\|M - X_kY_k^T \|_F $ is bounded because the objective function is decreasing (however, the intuition is not enough and the proof requires some extra effort).
In Section \ref{sec: proof of lemma 2} we provide some easily verifiable conditions for Property (b) to hold (see Proposition \ref{prop: K(delta) condition}), so that Lemma \ref{lemma of incoherent neighborhood} and Theorem \ref{major theorem} can be extended to other algorithms.

With these two lemmas, the proof of Theorem \ref{major theorem} is quite straightforward and presented below.

\emph{Proof of Theorem \ref{major theorem}}:
Consider any limit point $(X_*,Y_*)$ of sequence $\{(X_k,Y_k)\}$ generated by either of Algorithms 1-4.
According to Property (a) of Lemma \eqref{lemma of incoherent neighborhood}, $(X_*,Y_*)$ is a stationary point of problem (P1), i.e.\
$
\nabla_X \tilde{F}(X_*,Y_*) = 0,  \nabla_Y \tilde{F}(X_*,Y_*) =0. $
According to Property (b) of Lemma \ref{lemma of incoherent neighborhood}, with probability at least $1-1/n^4$,
$(X_k,Y_k) \in K_1 \cap K_2 \cap K(\delta)$ for all $k$, implying $(X_*,Y_*) \in K_1 \cap K_2 \cap K(\delta)$.
% since $K_1 \cap K_2 \cap K(\delta)$ is a closed set.
 Then we can apply Lemma \ref{lemma main; about local convexity} by plugging $(X,Y) = (X^*,Y^*)$ into \eqref{local relative convexity} to conclude that with probability at least $1 - 2/n^4$, $\|M-X_*Y_*^T \|_F \leq 0$, i.e.\ $X_*Y_*^T = M$.
 {\black $\Box$ } % \qed % $\Box$

{\black
Remark: Note that $X_*Y_*^T = M$ does not necessarily imply the global optimality of $(X_*, Y_*)$ since we have not proved $ G(X_*, Y_*) = 0 $.
% We can further prove that $(X_*,Y_*)$ is a global minimizer of $\tilde{F}$, where $(X_*, Y_*)$ is a limit point of the sequence $\{(X_k,Y_k)\}$ generated by % either of Algorithm 1-4.
Nevertheless, the global optimality can be easily proved using a different version of Lemma \ref{lemma main; about local convexity} (see the discussion before
Lemma \ref{lemma: cost-to-go}); in other words, Theorem \ref{major theorem} can be slightly strengthened to ``Algorithm 1-4 converge to the global optima of problem (P1)'', instead of ``Algorithm 1-4 recover $M$''.

% The global optimality can be proved using a different version of Lemma \ref{lemma main; about local convexity}; .
% In the above proof, we have already shown that $X_* Y_*^T = M $;
%%There is still a small gap between $X_*Y_*^T = M$ and $(X_*, Y_*)$ being global optimal since we have not proved $ G(X_*, Y_*) = 0 $.
%next we prove that $G(X_*, Y_*)$ has to be zero. In fact, $ X_* Y_*^T = M  $ implies
%  $ \nabla F(X_*, Y_*) = (\bP_{\Omega}(X_* Y_*^T - M)Y_* , \bP_{\Omega}(X_* Y_*^T - M)^T X_* ) = 0 $.
%  Since  $(X_*,Y_*)$ is a stationary point of problem (P1), we have
% $0 = \nabla \tilde{F}(X_*,Y_*) =  \nabla F(X_*,Y_*) + \nabla G(X_*,Y_*) $. Combining with $ \nabla F(X_*, Y_*) = 0$ we get
% $ \nabla G(X_*,Y_*) = 0 $. According to Claim \ref{cost-to-go G' and G claim} in Appendix \ref{appen: cost-to-go estimate}, we get $G(X_*, Y_*) = 0$.
% Therefore, $\tilde{F}(X_*, Y_*) = 0$, which means that $(X_*, Y_*)$ is a global minimizer of $\tilde{F}$.
 }

% {\blue For some reason, slash QED does not work; I try to add amsthm

The same argument can be used to show a more general result than Theorem \ref{major theorem}, as stated in
the following corollary.
\begin{coro}\label{coro: convergence for other algorithms}
Under the same conditions of Theorem \ref{major theorem},
any algorithm satisfying Properties (a) and (b) in Lemma \ref{lemma of incoherent neighborhood} reconstructs $M$ exactly
 with probability at least $1-2/n^4$.
\end{coro}

{\black
\subsection{ Proof of Theorem \ref{theorem 2: linear convergence} }\label{sec: proof of thm 2}
}
The proof of Theorem \ref{theorem 2: linear convergence} applies a standard framework for first order methods: the convergence rate (or iteration complexity) can be derived from the ``cost-to-go estimate'' and the ``sufficient descent'' condition.
For instance, the linear convergence $f(\bm x_k) - f^*  \leq (1 - c_1 c_2)^k$ is a direct corollary of
the cost-to-go estimate $  \| \nabla f(\bm x_k)  \|^2 \geq c_1 [f(\bm x_k) - f^*] $ and the
sufficient descent condition $ f(\bm x_{k}) - f(\bm x_{k+1}) \geq c_2 \| \nabla f(\bm x_k) \|^2 $,
where $f^* $ is the minimum value of $f$, and $c_1, c_2$ are certain constants.
We remark that using other optimization frameworks may lead to stronger time complexity bounds; this is left as future work.
% which, when combined with the sufficient decrease, leads to the linear convergence of $f(\bm x_k)$ to $f(\bm x^*)$.

For our problem, a variant of Lemma \ref{lemma main; about local convexity} can be viewed as the cost-to-go estimate; see Lemma \ref{lemma: cost-to-go} below.
One difference with Lemma \ref{lemma main; about local convexity} is the following: for a stationary point $(X_*,Y_*)$ that $\nabla \tilde{F}(X_*,Y_*) = 0$, Lemma \ref{lemma: cost-to-go} implies $\tilde{F}(X_*, Y_*) = 0$ (global optimality), but Lemma \ref{lemma main; about local convexity} implies $M = X_* Y_*^T $ (exact recovery).
The relation between these two lemmas is that Lemma \ref{lemma: cost-to-go} is a direct consequence of  \eqref{enhanced local convexity}, a slightly stronger version of Lemma \ref{lemma main; about local convexity}.
The main difficulties of proving the two lemmas are the same and
lie in Proposition \ref{prop1} and Proposition \ref{prop2};
see the formal proof in Appendix \ref{appen: cost-to-go lemma proof}.

\begin{lemma}\label{lemma: cost-to-go}(Cost-to-go estimate)
Under the same conditions of Lemma \ref{lemma main; about local convexity}, with probability at least $1-1/n^4$, the following holds:
 % there exist $U\in \mathbb{R}^{m \times r},\mathbb{V}\in R^{n \times r},$ such that
% $UV^T =M$ and
\begin{equation}\label{cost to go eq.}
\| \nabla  \tilde{F}(X,Y) \|_F^2  \geq  \xi \tilde{F}(X,Y), \ \forall \ (X,Y) \in K_1 \cap K_2 \cap K(\delta),  % \frac{p}{4} \|M-XY^T \|_F^2.
\end{equation}
where $ \xi =  \frac{1 }{C_g r^5 \kappa^3 } p \Sigma_{\min} $ (here $C_g \geq 1$ is a numerical constant).
\end{lemma}

% We stress that although cost-to-go estimate is widely used in convex optimization, for non-convex problems the cost-to-go estimate is quite rare.

The following claim shows that Algorithm 1a satisfies the sufficient descent condition.
It is easy to prove: it is well known that for minimizing a function (possibly non-convex) with Lipschitz continuous gradient, the gradient descent method with constant step-size satisfies the sufficient decrease condition.
\begin{claim}\label{claim: sufficient decrease}(Sufficient descent)
 For the sequence $\bm x_k = (X_k, Y_k)$ generated by Algorithm 1a (gradient descent with constant stepsize), we have
 \begin{equation}\label{sufficient decrease eq}
  \tilde{F}(\bm x_{k }) -  \tilde{F}(\bm x_{k + 1})  \geq \frac{\eta_1 }{2} \| \nabla \tilde{F}(\bm x_k ) \|_F^2,
 \end{equation}
 where $\eta_1$ is the stepsize bounded above by $\bar{\eta_1}$ defined in \eqref{eta 1 def}.
\end{claim}

% In the proof of Claim \ref{Algo 1-3 satisfy conditions}, let $\lambda = 1$ in \eqref{descent bound for GD} we immediately obtain \eqref{sufficient decrease eq}, which proves Claim \ref{claim: sufficient decrease}.

The linear convergence can be easily derived from Lemma \ref{lemma main; about local convexity} and Claim \ref{claim: sufficient decrease}.
For completeness, we present the proof below.

\emph{Proof of Theorem \ref{theorem 2: linear convergence}}:
According to Property (b) of Lemma \ref{lemma of incoherent neighborhood}, with probability at least $1-1/n^4$,
$(X_k,Y_k) \in K_1 \cap K_2 \cap K(\delta)$ for all $k$.
According to Lemma \ref{lemma: cost-to-go} and Claim \ref{claim: sufficient decrease}, we have (with probability at least $1 - 2/n^4$)
$$
\tilde{F}(\bm x_{k }) -  \tilde{F}(\bm x_{k + 1}) \geq \frac{\eta_1 }{2} \| \nabla \tilde{F}(\bm x_k ) \|_F^2
\geq \frac{\eta_1 }{2} \xi  \tilde{F}( \bm x_k ), \ \forall k.
$$
This relation can be rewritten as
\begin{equation}\label{one step decrease of F}
  \tilde{F}(\bm x_{k+1 }) \leq (1 - \frac{1 }{2} \eta_1 \xi ) \tilde{F}(\bm x_{k }), \ \forall \ k.
\end{equation}
The stepsize $\eta_1$ can be bounded as $ 0< \eta_1 \leq \bar{\eta}_1 \overset{\eqref{eta 1 first bound}}{\leq} \frac{ 1 }{4 \beta_T^2 } = \frac{1 }{4 C_T r \Sigma_{\max}} \leq  \frac{1 }{ \Sigma_{\max}}$. Since $ 0< \xi = \frac{1 }{C_g r^5 \kappa^3 } p \Sigma_{\min} \leq \Sigma_{\min} $,
 we have $ 0< \eta_1 \xi  \leq \frac{ \Sigma_{\min} }{ \Sigma_{\max}} \leq 1$, which implies $0 < 1 - \frac{1 }{2} \eta_1 \xi < 1$.
 Then the relation \eqref{one step decrease of F} leads to
$$
  \tilde{F}(\bm x_{k }) \leq (1 - \frac{1 }{2} \eta_1 \xi )^k \tilde{F}(\bm x_{0 }), \ \forall \ k,
$$
which finishes the proof.  $\Box$

The same argument can be used to show a more general result than Theorem \ref{theorem 2: linear convergence}, as stated in
the following corollary.
\vspace{-0.5cm}
\begin{coro}\label{coro: linear convergence for other algorithms}
Under the same conditions of Theorem \ref{major theorem},
any algorithm satisfying Properties (a) and (b) in Lemma \ref{lemma of incoherent neighborhood} and the sufficient decrease condition
\eqref{sufficient decrease eq} has the linear convergence property, i.e. generates a sequence $(X_k, Y_k)$ that satisfies \eqref{linear convergence property}.
\end{coro}

%%%%%%%%%%%%%%%%%%%%%%%%%%%%%%%%%%%%%%%%%%%%%%%%%%%%%%%%%%%%%%%%%%%%%%%%%%%%%%%%%%%%%%%%%%%%%%%%%%%%
%%%%%%%%%%% section 3  Proof of Lemma 2.2 %%%%%%%%%%%%%%%%%%%%%%%%%%%%%%%%%%%%%%%%%%%%%%%%%%%%%%%%%%
%%%%%%%%%%%%%%%%%%%%%%%%%%%%%%%%%%%%%%%%%%%%%%%%%%%%%%%%%%%%%%%%%%%%%%%%%%%%%%%%%%%%%%%%%%%%%%%%%%%%%%
\section{Proof of Lemma \ref{lemma main; about local convexity}}\label{sec: proof of lemma 1} % and Several key lemmas}
  % The proof of Lemma \ref{lemma main; about local convexity} is the main technical part of this paper.
   In Section \ref{subsec: prelim}, we will show that to prove Lemma \ref{lemma main; about local convexity}, we only need to construct $U,V$ to satisfy three inequalities that $\| \bP_{\Omega}((U-X)(V-Y)^T)\|_F$ and $\|((U-X)(V-Y)^T)\|_F$ are bounded above and $\langle \nabla_X G, X-U \rangle + \langle \nabla_Y G, Y-V \rangle$ is bounded below. In Section \ref{subsec: define U,V} we describe two propositions that specify the choice of $U,V$, and then we show that such $U,V$ satisfy the three desired inequalities in Section \ref{subsec: define U,V} and subsequent subsections.
 %  (the proofs will be given in Appendix)
%   We also show that $\|((U-X)(V-Y)^T)\|_F$ is bounded above according to our choice of $U,V$.
%After that, we show that $\| \bP_{\Omega}((U-X)(V-Y)^T)\|_F$ is bounded above (subsection \ref{subsec: bound P_Omega((U-X)(V-Y))})
%and $\langle \nabla_X G, X-U \rangle + \langle \nabla_Y G, Y-V \rangle $ is bounded below (subsection \ref{subsec: bound phi_G}), which completes the proof of Lemma \ref{lemma main; about local convexity}.

%-------------------------------------------------------------
%----------- subsection 3.1 Preliminary analysis----------------------------------
%---------------------------------------------------------------
\subsection{Preliminary analysis}\label{subsec: prelim}
Since $(X,Y) \in K(\delta)$, we have
\begin{equation}\label{d def}
d \triangleq \|M-XY^T \|_F \leq \delta \overset{\eqref{delta definition throughout}}{=} \frac{\Sigma_{\rmin}}{ C_d r^{1.5} \kappa } . % \leq \sqrt{\frac{C_1}{C_0} } \frac{\Sigma_{\rmin}}{r^5 \kappa^4}.
\end{equation}
% A factorization $M = UV^T$ such that
To ensure \eqref{local relative convexity} holds,
%\begin{equation}\label{error bound}
%\langle \nabla_X \tilde{F}(X,Y), X-U \rangle + \langle \nabla_Y \tilde{F}(X,Y), Y-V \rangle \geq \frac{p}{8} d^2.
%\end{equation}
%Obviously $(\ref{error bound})$ is implied by
we only need to ensure that the following two inequalities hold:
\begin{subequations}
\begin{align}
\phi_F = \langle \nabla_X F, X-U \rangle + \langle \nabla_Y F, Y-V \rangle \geq \frac{p}{4} d^2,    \label{phi F}  \\
\phi_G = \langle \nabla_X G, X-U \rangle + \langle \nabla_Y G, Y-V \rangle \geq 0.  \label{phi G}  % -\frac{p}{8} d^2.
\end{align}
\end{subequations}

%Recall that $F(X,Y) = \frac{1}{2}\|\bP_{\Omega}(M - XY^T)\|_F^2.$ We have $\nabla_X F = (XY^T-Z)Y = \bP_{\Omega}(XY^T-M)Y,$  $\nabla_Y F = (XY^T-Z)^TX =  (\bP_{\Omega}(XY^T-M))^T X.$

Define
\begin{equation}\label{a def}
a \triangleq U(Y-V)^T + (X-U)V^T, \quad b \triangleq (U-X)(V-Y)^T.
\end{equation}
 Then
 \begin{equation}\nonumber
 XY^T - M = a + b, \;\; (X-U)Y^T + X(Y-V)^T = a + 2b .
 \end{equation}
Using the expressions of $\nabla_X F, \nabla_Y F$ in \eqref{grad of F,G expression}, we bound $\phi_F$ as follows:
\begin{equation}\label{phi F expression}
\begin{split}
\phi_F = & \langle \nabla_X F, X-U \rangle + \langle \nabla_Y F, Y-V \rangle \\
     =  & \langle \bP_{\Omega}(XY^T - M), (X-U)Y^T + X(Y-V)^T \rangle \\
     = &  \langle \bP_{\Omega}(a + b), \bP_{\Omega}(a + 2b) \rangle \\
     = &  \| \bP_{\Omega}(a)\|_F^2 + 2\|\bP_{\Omega}(b)\|_F^2 + 3 \langle \bP_{\Omega}(a), \bP_{\Omega}(b) \rangle \\
     \geq & \| \bP_{\Omega}(a)\|_F^2 + 2\|\bP_{\Omega}(b)\|_F^2  - 3 \| \bP_{\Omega}(a)\|_F \|\bP_{\Omega}(b)\|_F.
%     = &   \| \bP_{\Omega}(a)\|_F ( \| \bP_{\Omega}(a)\|_F - 3\| \bP_{\Omega}(b)\|_F). \\
\end{split}
\end{equation}
The reason to decompose $M-XY^T$ as $a + b$ is the following.
In order to  bound $\| \bP_{\Omega}(M-XY^T) \|_F$, we notice $E(\bP_{\Omega}(M-XY^T)) = p (M-XY^T)$ and
wish to prove $ \| \bP_{\Omega}(M-XY^T) \|_F^2 \approx O(p d^2). $ However,
$ \| \bP_{\Omega}(A) \|_F $ could be as large as $ \| A\|_F$
if the matrix $A$ is not independent of the random subset $\Omega$ (e.g. choose $A $ s.t. $A = \bP_{\Omega}(A) $).
This issue can be resolved by decomposing $XY^T - M$ as $a + b$ and bounding $\| \bP_{\Omega}(a) \|_F$ and $\| \bP_{\Omega}(b)\|_F$ separately.
In fact, $\|\bP_{\Omega}(a)\|_F$ can be bounded because $ a $ lies in a space spanned by the matrices with the same row space or column space as $M$, which is independent of $\Omega$ (Theorem 4.1 in \cite{candes2009exact}).
$\| \bP_{\Omega}(b) \|_F$ can be bounded according to a random graph lemma of \cite{feige2005spectral, keshavan2010matrix},
which requires $U,V,X,Y$ to be incoherent{\black (i.e.\ have bounded row norm).}

We claim that \eqref{phi F} is implied by the following two inequalities:
\begin{subequations} % \label{P Omega (b) bound}
\begin{align}
& \| \bP_{\Omega}(b)\|_F = \| \bP_{\Omega}((U-X)(V-Y)^T)\|_F  \leq \frac{1}{5}\sqrt{p}d; \label{P Omega (b) bound}  \\ % \frac{\sqrt{2p}}{18
& \|b \|_F = \|(U-X)(V-Y)^T\|_F  \leq \frac{1}{10}d.  \label{bound b}
\end{align}
\end{subequations}
In fact, assume (\ref{P Omega (b) bound}) and (\ref{bound b}) are true, we prove $\phi_F \geq p d^2/4  $ as follows.
By $XY^T - M = a + b$ we have % Inequality (\ref{bound b}) implies
\begin{equation}\label{a bound}
\|a \|_F \geq \|M-XY^T \|_F - \|b \|_F \overset{\eqref{bound b}}{\geq} \frac{9}{10}d.
\end{equation}
{\black
Recall that the SVD of $M$ is $M = \hat{U} \Sigma \hat{V}^T$ and $M$ satisfies the incoherence condtion \eqref{incoherence cond}.
It follows from $M= UV^T = \hat{U} \Sigma \hat{V}^T $ that $M, U, \hat{U}$ have the same column space, thus there exists some matrix $ B_1 \in \dR^{r \times r}$
such that $U = \hat{U} B_1 $; similarly, there exists $ B_2 \in \dR^{r \times r} $ such that $V = \hat{V} B_2$.
Therefore, by the definition of $a$ in \eqref{a def} we have
\begin{equation}
a \in \mathcal{T} \triangleq \{ \hat{U} W_2^T + W_1 \hat{V}^T \mid W_1 \in \dR^{m \times r}, W_2 \in \dR^{n \times r}  \}.
\end{equation}  % Theorem 3.4 in \cite{recht2011simpler} (or
By  Theorem 4.1 in \cite{candes2009exact}, for $|\Omega|$ satisfying \eqref{Omega bound} with large enough $C_0$, we have that with probability at least $1 - 1/(2n^4)$, $ \| \bP_{\mathcal{T}} \bP_{\Omega} \bP_{\mathcal{T}} (a) - p \bP_{\mathcal{T}} (a)  \|_F \leq \frac{1}{6} p\| a\|_F $
(note that this bound holds uniformly for all $a \in \mathcal{T}$, thus also holds when $a$ is dependent on $\Omega$).  Since $a \in \mathcal{T}$, this inequality can be simplified to
\begin{equation}\label{Candes, Recht result}
\| \bP_{\mathcal{T}} \bP_{\Omega}  (a) - p a  \|_F \leq \frac{1}{6} p\| a\|_F.
\end{equation}
Following the analysis of \cite[Corollary 4.3]{candes2009exact}, we have
\begin{equation}\label{P Omega (a) bound, interm}
\begin{split}
\| \bP_{\Omega}(a)\|_F^2 = \| \bP_{\Omega}\bP_{\mathcal{T}}(a)\|_F^2  &  = \langle a ,  \bP_{\mathcal{T}} \bP_{\Omega}^2 \bP_{\mathcal{T}}(a) \rangle
 = \langle a ,  \bP_{\mathcal{T}} \bP_{\Omega} (a) \rangle  \\
 & = \langle a, p a \rangle + \langle a, \bP_{\mathcal{T}} \bP_{\Omega}(a) - p a \rangle.
\end{split}
\end{equation}
The absolute value of the second term can be bounded as $$ | \langle a, \bP_{\mathcal{T}} \bP_{\Omega}(a) - p a \rangle | \leq
\|a \|_F \| \bP_{\mathcal{T}} \bP_{\Omega}(a) - p a \|_F \overset{\eqref{Candes, Recht result}}{\leq}  \frac{1}{6} p\| a\|_F^2 ,$$
which implies $ - \frac{1}{6} p\| a\|_F^2 \leq \langle a, \bP_{\mathcal{T}} \bP_{\Omega}(a) - p a \rangle \leq \frac{1}{6} p\| a\|_F^2 $.
Substituting into \eqref{P Omega (a) bound, interm}, we obtain that with probability at least $1 - 1/(2n^4)$,
\begin{equation}\label{P Omega (a) bound, upper and low}
 \frac{5}{6}\| a\|_F^2   \leq   \| \bP_{\Omega}(a)\|_F^2 \leq \frac{7}{6}\| a\|_F^2.
\end{equation}
}
The first inequality of the above relation implies
\begin{equation}\label{P Omega (a) bound}
 \| \bP_{\Omega}(a)\|_F^2  \geq \frac{5}{6}\| a\|_F^2  \overset{\eqref{a bound}}{\geq} \frac{27}{40} p d^2.
 \end{equation}
According to \eqref{phi F expression} and the bounds \eqref{P Omega (a) bound} and \eqref{P Omega (b) bound}, we have
 $\phi_F/(pd^2) \geq \frac{27}{40}  + 2 (\frac{1}{5})^2 - \frac{3}{5} \sqrt{\frac{27}{40}} \geq \frac{1}{4} $,
 which proves \eqref{phi F}.
 %  \| \bP_{\Omega}(a)\|_F ( \| \bP_{\Omega}(a)\|_F - 3\| \bP_{\Omega}(b)\|_F) \geq \frac{\sqrt{2p}}{3}d(\frac{\sqrt{2p}}{3}d - \frac{3\sqrt{2p}}{18}d) = \frac{1}{9}pd^2.$

In summary, to find a factorization $M = UV^T$ such that \eqref{local relative convexity} holds, we only need to ensure that
the factorization satisfies
(\ref{bound b}), (\ref{P Omega (b) bound}) and (\ref{phi G}).{\black In the following three subsections, we will show that such a factorization $M = UV^T $ exists.
Specifically, $U,V$ will be defined in Table \ref{table: def of U,V}
and the three desired inequalities will be proved in Corollary \ref{coro: summary of U,V, extra}, Proposition \ref{upper bound of P((U-X)(V-Y))} and Claim \ref{claim of phi >= 0} respectively.  }
 % We will show that such a factorization exists  (see ).

%-------------------------------------------------------------
%----------- subsection 3.2 How to define $U,V$ and some technical results----------------------------------
%---------------------------------------------------------------
\subsection{Definitions of $U,V$ and key technical results}\label{subsec: define U,V}

%To complete the proof of Lemma \ref{lemma2}, we only need to prove that there exist $U,V$ that satisfying (\ref{P Omega (b) bound}), (\ref{bound b}) and (\ref{phi G}) simultaneously.

We construct $U,V$ according to two propositions, which will be stated in this subsection and proved in the appendix.
The first proposition states that if $XY^T$ is close to $M$, then there exists a factorization $M=UV^T$ such that $U$ (resp. $V$) is close to $X$ (resp. $Y$), and $U,V$ are incoherent. Roughly speaking, this proposition shows the continuity of the factorization map $ Z = XY^T \mapsto (X,Y) $ near a low-rank matrix $M$.
{\black The condition $X,Y \in K_1 \cap K_2 \cap K(\delta)$ and \eqref{delta definition throughout} implies that
$ d \triangleq  \|M-XY^T \|_F    \leq \delta = \frac{\Sigma_{\min}}{C_d r^{1.5} \kappa  } $ and $ \|X \|_F \leq \beta_T, \|Y \|_F \leq \beta_T $,
thus for large enough $C_d$, the assumptions of Proposition \ref{prop1} hold. Similarly, the assumptions of the other results in this subsection also hold.}

%------key result 1: prop1------------
\begin{prop}\label{prop1}
Suppose $M \in \dR^{m \times n}$ is a rank-$r$ matrix with $\Sigma_{\max}$ ($\Sigma_{\min}$) being the largest (smallest) non-zero singular value, and $M$ is $\mu$-incoherent.  There exists a numerical constant $C_T$ such that the following holds:
% There exists numerical constant $C_d$ such that: if
If
\begin{subequations}\label{cond of X,Y, Prop 1}
\begin{align}
d \triangleq  \|M-XY^T \|_F     \leq & \frac{\Sigma_{\min}}{11 r  } , \label{cond a) of X,Y, Prop 1} \\
\| X\|_F  \leq \beta_T,  \quad    \| Y\|_F & \leq  \beta_T, \label{cond b) of X,Y, Prop 1}
\end{align}
\end{subequations}
where $\beta_T = \sqrt{C_T r\Sigma_{\rmax}}$,
%\newline (1) $M$ is $\mu$-incoherent;
%\newline (2) $\|M-XY^T \|_F = d \leq \frac{\Sigma_{\rmin}}{5};$
% \newline (3) $\|X \|_F \leq \beta_T,\| Y\|_F \leq \beta_T;$
then there exist $U \in \mathbb{R}^{m\times r}, V\in \mathbb{R}^{n \times r}$ such that
\begin{subequations}\label{req of U,V, Prop 1}
\begin{align}
  UV^T  & = M ,   \label{req a) of U,V, Prop 1} \\
 \|U \|_F  & \leq   {\black  (1 - \frac{d}{\Sigma_{\min} } )  } \|X \|_F ,  \label{req b) of U,V, Prop 1} \\
 \|U-X \|_F \leq \frac{ 6 \beta_T}{ 5 \Sigma_{\min}}d, & \;\;\|V-Y \|_F \leq \frac{ 3 \beta_T}{ \Sigma_{\min}}d ,
 \label{req c) of U,V, Prop 1}   \\
 \| U^{(i)}\|^2 \leq \frac{r\mu }{m}\beta_T^2,   & \; \; \| V^{(j)}\|^2 \leq \frac{3r\mu }{2n}\beta_T^2 .    \label{req d) of U,V, Prop 1}
\end{align}
\end{subequations}
%\newline  Then there exist $U\in \mathbb{R}^{m \times r},V\in \mathbb{R}^{n \times r}$ such that
%\newline (a) $UV^T =M$;
%\newline (b) $\|U-X \|_F \leq \frac{5\beta_T}{4\Sigma_{\rmin}}d, \|V-Y \|_F \leq \frac{5\beta_T}{2\Sigma_{\rmin}}d$;
%\newline (c) $\| U^{(i)}\|^2 \leq \frac{3r\mu }{2m}\beta_T^2, \| V^{(j)}\|^2 \leq \frac{3r\mu }{2n}\beta_T^2 $;
%\newline (d) $\| U\|_F \leq \| X\|_F.$
\end{prop}

%\begin{prop}\label{prop1}
%Suppose
%\newline (1) $M$ is $(\mu ,\mu_1)$ incoherent;
%\newline (2) $\|M-XY^T \|_F = d \leq \frac{\Sigma_{\rmin}}{5};$
% \newline (3) $\|X \|_F \leq \beta_T,\| Y\|_F \leq \beta_T;$
%\newline  Then there exist $U\in \mathbb{R}^{m \times r},V\in \mathbb{R}^{n \times r}$ such that
%\newline (a) $UV^T =M$;
%\newline (b) $\|U-X \|_F \leq \frac{5\beta_T}{4\Sigma_{\rmin}}d, \|V-Y \|_F \leq \frac{5\beta_T}{2\Sigma_{\rmin}}d$;
%\newline (c) $\| U^{(i)}\|^2 \leq \frac{3r\mu }{2m}\beta_T^2, \| V^{(j)}\|^2 \leq \frac{3r\mu }{2n}\beta_T^2 $;
%\newline (d) $\| U\|_F \leq \| X\|_F.$
%\end{prop}

%Requirement (b) guarantees that $U,V$ satisfy (\ref{bound b}): the bounds on $\|U-X \|_F,\|Y-V \|_F$ will immediately give a bound on $\|(U-X)(Y-V)^T \|_F.$
%Requirement (c) and (d) are used to prove that $U,V$ satisfy (\ref{P Omega (b) bound}) and (\ref{phi G}).% Requirement (c) is used to prove(\ref{P Omega (b) bound}); requirement (c) and (d) are used to prove (\ref{phi G}). %However, requirement (d) is not strong enough to prove (\ref{phi G}).

     The proof of Proposition \ref{prop1} is given in Appendix \ref{appen: proof of Prop 1}.
% the condition \eqref{cond b) of X,Y, Prop 1} is reasonable: $XY^T$ can be an arbitrary factorization and

Remark 1: A symmetric result that switches $X,U$ and $Y,V$ in the above proposition holds: under the conditions of Proposition (\ref{prop1}), there exist $U,V$ satisfying \eqref{req of U,V, Prop 1} with $U,V$ reversed, i.e.\  $UV^T = M$, $ \| V\|_F (1 - \frac{d}{\Sigma_{\min} } ) \leq \|Y \|_F$, $ \|U-X \|_F \leq \frac{ 3 \beta_T}{ \Sigma_{\rmin}}d, \|V-Y \|_F \leq \frac{ 6 \beta_T}{ 5 \Sigma_{\rmin}}d$,
 and $ \| U^{(i)}\|^2 \leq \frac{3r\mu }{2m}\beta_T^2,  \| V^{(j)}\|^2 \leq \frac{r\mu }{n}\beta_T^2  $.
 % requirements (a),(b),(c) of Proposition \ref{prop1}, and $ \| V\|_F \leq \|Y \|_F.$

{\black
Remark 2: To prove Theorem \ref{major theorem} (convergence), we only need $ \|U \|_F   \leq \|X \|_F $;
here the slightly stronger requirement $ \|U \|_F   \leq (1 - \frac{d}{\Sigma_{\min} } )  \|X \|_F $ is for the purpose of proving Theorem \ref{theorem 2: linear convergence} (linear convergence).  }

% Roughly speaking, this proposition shows that if two low-rank matrices $M$ and $XY^T$ are close enough, then there exists a factorization $M = UV^T$
% so that the factors $U,V$ are close to $X,Y$ respectively.
% We would like to mention that
Remark 3: Without the incoherence assumption on $M$,  by the same proof we can show that there still exist $U,V$ satisfying \eqref{req a) of U,V, Prop 1}
and \eqref{req c) of U,V, Prop 1}, i.e.\ $M = UV^T$ and $U, V$ are close to $X,Y$ respectively. % (ignore \eqref{req b) of U,V, Prop 1} for the moment).
Such a result bears some similarity with the classical perturbation theory for singular value decomposition \cite{wedin1972perturbation}.
In particular, \cite{wedin1972perturbation} proved that for two low-rank matrices\footnote{The result in \cite{wedin1972perturbation} also covered the case of two approximately low-rank matrices, but we only consider the case of exact low-rank matrices here.} that are close, the spaces spanned by the left (resp.\ right) singular vectors of the two matrices are also close. Note that the singular vectors themselves may be very sensitive to perturbations
and no such perturbation bounds can be established (see \cite[Sec. 6]{stewart1998perturbation}).
The difference of our work with the classical perturbation theory is that we do not consider SVD of two matrices; instead, we allow one matrix to have an arbitrary factorization, and the factorization of the other matrix can be chosen accordingly. %  to that of the first matrix.
Since we do not have any restriction on the factorization $XY^T$ (except the dimensions) and the norms of $X$ and $Y$
can be arbitrarily large, the distance between two corresponding factors has to be proportional to the norm of one single factor, which explains the
coefficient $\beta_T$ in \eqref{req c) of U,V, Prop 1}.

Unfortunately, Proposition \ref{prop1} is not strong enough to prove $\phi_G \geq 0$ when both $\|X \|_F$ and $\|Y \|_F$ are large (see an analysis in Section \ref{subsec: bound phi_G}).{\black To resolve this issue, we need to prove the second proposition in which there is an additional assumption that both $\| X\|_F $ and $\| Y\|_F$
are large, and an additional requirement that both $\| U\|_F$ and $\| V\|_F$ are bounded (by the norms of original factors $\| X\|_F$
and $\| Y\|_F$ respectively).}
More specifically, the proposition states that if $M$ is close to $XY^T$, and both $\| X\|_F$ and $\| Y\|_F$ are large, then there is a factorization
$M=UV^T$ such that $U$ (resp.$\ V$) is close to $X$ (resp.$\ Y$), and $\|U\|_F \leq \|X \|_F, \|V \|_F \leq \| Y\|_F$.
For the purpose of proving linear convergence, we prove a slightly stronger result that $ \|V \|_F \leq (1 - d/\Sigma_{\min})\| Y\|_F$ .
{\black
The previous result Proposition \ref{prop1} can be viewed as a perturbation analysis for an arbitrary factorization, while
Proposition \ref{prop2} can be viewed as an enhanced perturbation analysis for a constrained factorization.
Although Proposition \ref{prop2} is just a simple variant of Proposition \ref{prop1}, it seems to require a much more involved proof than Proposition \ref{prop1}.
See the formal proof of Proposition \ref{prop2} in Appendix \ref{section of proof of Prop 2}.   }
\vspace{0cm}
% which is a variant of Proposition \ref{prop1}.  In this sense,
%  Compared to Proposition \ref{prop1}, in the second proposition
% We will prove Proposition \ref{prop2} . }
%---key result 2: prop2-------------------------
\begin{prop}\label{prop2}
Suppose $M \in \dR^{m \times n}$ is a rank-$r$ matrix with $\Sigma_{\max}$ ($\Sigma_{\min}$) being the largest (smallest) non-zero singular value, and $M$ is $\mu$-incoherent.  There exist numerical constants $C_d, C_T$ such that the following holds:
if 
\begin{subequations}\label{cond of X,Y, Prop 2}
\begin{align}
d \triangleq  \|M-XY^T \|_F    & \leq \frac{\Sigma_{\min}}{C_d r  } , \label{cond a) of X,Y, Prop 2} \\
\sqrt{\frac{2}{3}}\beta_T \leq \| X\|_F  \leq \beta_T,  &  \quad \sqrt{\frac{2}{3}}\beta_T \leq \| Y\|_F \leq \beta_T, \label{cond b) of X,Y, Prop 2}
\end{align}
\end{subequations}
where $\beta_T = \sqrt{C_T r\Sigma_{\rmax}}$, then there exist $U \in \mathbb{R}^{m\times r}, V\in \mathbb{R}^{n \times r}$ such that
\begin{subequations}\label{req of U,V, Prop 2}
\begin{align}
  UV^T  & = M ,  \label{req a) of U,V, Prop 2} \\
  \|U \|_F \leq \|X \|_F, \; &  \; \|V \|_F \leq (1 - \frac{d}{\Sigma_{\min} } ) \|Y \|_F,  \label{req b) of U,V, Prop 2} \\
 \| U-X\|_F \| V-Y\|_F  &   \leq    65 \sqrt{r} \frac{\beta_T^2}{\Sigma_{\min}^2}  d^2 ,  \quad  \nonumber \\
  \max \{ \| U - X \|_F,\| V - Y \|_F \} & \leq \frac{17}{2} \sqrt{r} \frac{\beta_T}{\Sigma_{\min}}   d,
 \label{req c) of U,V, Prop 2}   \\                   % previously without linear convergence, 65--> 25; 17/2 --> 16/3
 \| U^{(i)}\|^2 \leq  \frac{r\mu }{m}\beta_T^2,    \; &  \; \| V^{(j)}\|^2 \leq \frac{r\mu }{n}\beta_T^2 . \label{req d) of U,V, Prop 2}
\end{align}
\end{subequations}

\end{prop}

Remark: A symmetric result that switches $X,U$ and $Y,V$ in the above proposition still holds;
the only change is that \eqref{req b) of U,V, Prop 2} will become
 $ \|U \|_F \leq (1 - \frac{d}{\Sigma_{\min} } ) \|X \|_F,  \; \|V \|_F \leq  \|Y \|_F $.
 It is easy to prove a variant of the above proposition in which \eqref{req b) of U,V, Prop 2} is changed to
 $ \|U \|_F \leq (1 - \frac{d}{ 2 \Sigma_{\min} } ) \|X \|_F,  \|V \|_F \leq (1 - \frac{d}{2 \Sigma_{\min} } ) \|Y \|_F $;
 in other words, the asymmetry of $X,U$ and $Y,V$ in \eqref{req b) of U,V, Prop 2} is artificial. Nevertheless, Proposition \ref{prop2}
 is enough for our purpose.
% All conditions and requirements are symmetric to $X,U$ and $Y,V$, except \eqref{req b) of U,V, Prop 2}.

%\begin{prop}\label{prop2}
%There exists numerical constant $C_d$ such that: if
%\newline  (1) $M$ is $(\mu ,\mu_1)$ incoherent;
%\newline (2) $\|M-XY^T \|_F = d < \frac{\Sigma_{\min}}{C_d r^{1.5} \kappa^{1.5}};$
%\newline (3)$\sqrt{\frac{2}{3}}\beta_T \leq \| X\|_F  \leq \beta_T, \sqrt{\frac{2}{3}}\beta_T \leq \| Y\|_F \leq \beta_T$, where $\beta_T = \sqrt{C_T r\Sigma_{\rmax}}$;
%\newline then there exists $U \in \mathbb{R}^{m\times r}, V\in \mathbb{R}^{n \times r}$ such that
%\newline (a) $UV^T = M$;
%\newline (b) $\| U-X\|_F \leq 11 \frac{\beta_T}{\Sigma_{\min}}  \kappa d , \| V-Y\|_F \leq 11 \frac{\beta_T}{\Sigma_{\min}} \sqrt{r}  \kappa d ;$
%\newline (c) $\| U^{(i)}\|^2 \leq \frac{r\mu }{m}\beta_T^2, \| V^{(j)}\|^2 \leq \frac{r\mu }{n}\beta_T^2 $;
%\newline (d) $\|U \|_F \leq \|X \|_F$, $\|V \|_F \leq \|Y \|_F$.
%\end{prop}

{\black
Throughout the proof of Lemma \ref{lemma main; about local convexity}, $U,V$ are defined in Table \ref{subsec: define U,V}.
\begin{table*}[htbp]\caption{Definition of $U,V$}\label{table: def of U,V}
\begin{tabular}{p{400pt}}
\hline
Definition of $U,V$ in different cases \\
\hline
 Case 1: $ \| X\|_F \leq \| Y \|_F $.   \\
 % $\text{} \quad$ Case 1.1:  $\|Y \|_F < \sqrt{ \frac{2}{3}} \beta_T$. Define $U,V$ according to Proposition \ref{prop1};  \\ \leq \| Y \|_F
 $\text{} \quad \quad$ Case 1.1 : $\|X \|_F < \sqrt{ \frac{2}{3}} \beta_T $.  Define $U,V$ according to
the symmetrical result of \\
$\text{} \quad \quad\quad \quad \quad \quad \quad $ Proposition \ref{prop1}, i.e.\ $U,V$  satisfy \eqref{req of U,V, Prop 1} with $X,U$ and $Y,V$ reversed. \\
 $\text{} \quad \quad$ Case 1.2:  $ \|X \|_F, \| Y\|_F \in [ \sqrt{\frac{2}{3}}\beta_T, \beta_T ]$. Define $U,V$ according to Proposition \ref{prop2}.  \\
Case 2: $ \| Y \|_F < \| X \|_F $.  \\
  $\text{} \quad \quad $ Similar to Case 1 but with the roles of $X,U$ and $Y,V$ reversed.   \\
  % $\text{} \quad \quad $ Case 3.1: $ \|XY^T - M \|_F \geq   \frac{\Sigma{\rmin} }{C r^{3} \kappa^3 } $. Define $U,V$ according to Proposition \ref{prop1}. \\
 %  $\text{} \quad \quad $ Case 3.2: $ \|XY^T - M \|_F <  \frac{\Sigma{\rmin} }{C r^{3} \kappa^3 } $.  Define $U,V$ according to Propostion \ref{prop2}. \\
\hline
\end{tabular}
\end{table*}

%\newline (1)
%\newline (2)If  $ \| Y\|_F \in [ \sqrt{\frac{2}{3}}\beta_T, \beta_T ]$, and $\|X \|_F < \sqrt{ \frac{2}{3}} \beta_T,$ define $U,V$ according to
%the symmetrical result of Proposition \ref{prop1}, i.e.\ $U,V$ satisfy the requirement (a),(b),(c) of Proposition \ref{prop1}, and $ \| V\|_F \leq \|Y \|_F.$
%\newline (3)If  $ \|X \|_F, \| Y\|_F \in [ \sqrt{\frac{2}{3}}\beta_T, \beta_T ]$, define $U,V$ according to Propostion \ref{prop2}.
%in Proposition \ref{prop1} and \ref{prop2}. Based on whether $\| X\|_F, \| Y\|_F$ are less than $\sqrt{\frac{2}{3}}\beta_T$, $U,V$ are defined differently.

According to Proposition \ref{prop1} and Proposition \ref{prop2} (and their symmetric results), the properties of $U,V$ defined in Tabel \ref{table: def of U,V} are summarized in the following corollary. For simplicity, we only present the case that $  \| X\|_F \leq \| Y \|_F$;
in the other case that $ \| X\|_F > \| Y \|_F $, a symmetric result of Corollary \ref{coro: summary of U,V} holds.
\begin{coro}\label{coro: summary of U,V}
Suppose $d \triangleq \|XY^T - M \|_F  \leq  \frac{\Sigma_{\min}}{C_d r }$ and $\| X\|_F \leq \| Y \|_F$,
then $U,V$ defined in Table \ref{table: def of U,V} satisfy: % have the following properties:
\begin{subequations}
\begin{align}
   \quad  UV^T &  = M ; \label{summary of U,V (a)}   \\
 \quad \| U-X\|_F \| V-Y\|_F  &   \leq    65 \sqrt{r} \frac{\beta_T^2}{\Sigma_{\rmin}^2}  d^2;  \nonumber \\
 \quad  \max \{ \| U - X \|_F,\| V - Y \|_F \}&  \leq \frac{17}{2} \sqrt{r} \frac{\beta_T}{\Sigma_{\min}}   d,
 %\| U-X\|_F \leq 11 \frac{\beta_T}{\Sigma_{\min}}  \kappa d ,  & \;\; \| V-Y\|_F \leq 11 \frac{\beta_T}{\Sigma_{\min}} \sqrt{r}  \kappa d ;
  \label{summary of U,V (b)}  \\
   \quad  \| U^{(i)}\|^2 \leq \frac{3}{2} \frac{r\mu }{m}\beta_T^2, & \;\; \| V^{(j)}\|^2 \leq \frac{3}{2} \frac{r\mu }{n}\beta_T^2  ;
    \label{summary of U,V (c)}  \\
       \| V \|_F \leq (1 - \frac{d}{\Sigma_{\min}})\| Y \|_F; \; \text{ if }  & \| X\|_F > \sqrt{\frac{2}{3}}\beta_T ,  \text{ then } \| U\|_F \leq  \|X \|_F .   \label{summary of U,V (d)}
    % \| U\|_F \leq (1 - \frac{d}{\Sigma_{\min}})\|X \|_F;  \\
   %   \text{if }  \| Y \|_F > \sqrt{\frac{2}{3}}\beta_T \geq \| X \|_F, & \text{ then }  \| V \|_F \leq (1 - \frac{d}{\Sigma_{\min}})\| Y \|_F;  \\
    %  \text{if }  \min \{ \| X \|_F, \| Y\|_F \} > \sqrt{\frac{2}{3}}\beta_T ,   &
%       \text{ then }  \| U\|_F \leq  \|X \|_F, \| V \|_F \leq (1 - \frac{d}{\Sigma_{\min}})\| Y \|_F.  \label{summary of U,V (e2)}
  % \quad  \text{If }  \| X\|_F > \sqrt{\frac{2}{3}}\beta_T,  \text{ then }   \| U\|_F \leq \|X \|_F; & \quad
%    \text{if }  \| Y\|_F > \sqrt{\frac{2}{3}}\beta_T,   \text{ then }   \| V\|_F \leq \|Y \|_F.   \label{summary of U,V (d)}  \\
%   \quad  \text{If } \max \{ \| X\|_F, \| Y\|_F \} > \sqrt{\frac{2}{3}}\beta_T,  &  \text{ then }
%   \min \left\{ \frac{\| U\|_F}{\|X \|_F}, \frac{\| V\|_F}{\| Y \|_F} \right \} \leq 1 - \frac{d}{\Sigma_{\min}}.  \label{summary of U,V (e)}
\end{align}
\end{subequations}
\end{coro}

}

{\black
In \eqref{summary of U,V (b)}, we bound
 $ \|U-X \|_F \|V-Y \|_F$ by $O(d^2)$  with a rather complicated coefficient, but to prove \eqref{bound b}
we need a bound $O(d)$ with a coefficient $1/10$.
Under a slightly stronger condition on $d$ than that of Corollary \ref{coro: summary of U,V}, which still holds for $(X,Y) \in K(\delta)$ with $\delta$ defined in \eqref{delta definition throughout}, we can prove the bound \eqref{bound b} by \eqref{summary of U,V (b)}. }
 % $ \|U-X \|_F \|V-Y \|_F$ is bounded by $O(d^2)$ in \eqref{summary of U,V (b)},
 % but the coefficient before $d^2$ is rather nasty;  utilize the condition that $d$ is small enough to
% and we will eliminate the coefficient }
\begin{coro}\label{coro: summary of U,V, extra}
There exists a numerical constant $C_d$ such that if
\begin{equation}\label{d bound by d0^2; bound b}
 d \triangleq  \|M-XY^T \|_F  \leq   \frac{\Sigma_{\rmin} }{C_d r^{1.5} \kappa },
 \end{equation}
then $U,V$ defined in Table \ref{table: def of U,V} satisfy \eqref{bound b}.
%\begin{equation}\label{summary of U,V, extra U-X,V-Y}
%   \|(U-X)(V-Y)^T \|_F   \leq \frac{1}{10}d.    % \| U-X\|_F \leq 11  \frac{\beta_T}{\Sigma_{\min}} r  \kappa d, \;\; \| V-Y\|_F \leq 11  \frac{\beta_T}{\Sigma_{\min}} r \kappa d.
% \end{equation}
\end{coro}

\emph{Proof of Corollary \ref{coro: summary of U,V, extra}:}
{\black According to \eqref{summary of U,V (b)} }, we have
\begin{align*}
\|U-X \|_F \|V-Y \|_F \leq
 65 \frac{\beta_T^2 }{\Sigma_{\rmin}^2 } \sqrt{r}  d^2  = 65 C_T r^{1.5}  \frac{\Sigma_{\rmax}}{\Sigma_{\rmin}^2} d^2 \\
 = 65 C_T r^{1.5} \kappa \frac{d}{\Sigma_{\rmin}} d
 \leq \frac{1}{10} d ,
\end{align*}
  where the last inequliaty follows from \eqref{d bound by d0^2; bound b} with $C_d \geq 650 C_T. $  $\Box$ % $\Box$

% Thus, we only need to prove (\ref{P Omega (b) bound}) and (\ref{phi G}).
In the next two subsections, we will use the properties in Corollary \ref{coro: summary of U,V} to prove (\ref{P Omega (b) bound}) and (\ref{phi G}).
% Property (d) is crucial for the proof of (\ref{phi G}) and we will discuss this issue more later.

\subsection{Upper bound on $\| \bP_{\Omega}((U-X)(V-Y)^T)\|_F$ }\label{subsec: bound P_Omega((U-X)(V-Y))}
The following result states that for $U,V$ defined in Table \ref{table: def of U,V}, (\ref{P Omega (b) bound}) holds. % $\| \bP_{\Omega}( (U-X)(V-Y)^T) \|_F^2 $
%--------Major Prop: Prop.3: upper bound P((U-X)(V-Y)) -----------------
\begin{prop}\label{upper bound of P((U-X)(V-Y))}
%There exist numerical constants $C_0,C_d$ such that the following happens. Assume $\Omega$ is uniformly randomly generated with
%\begin{equation}\label{bound on Omega}
%|\Omega| \geq C_0 \sqrt{\alpha} nr \kappa^2 \max\{ \mu \log n, \mu^2 r^11 \sqrt{\alpha} \kappa^6 \}.
%\end{equation}
%Assume $\delta \leq \frac{\Sigma_{\min}}{C_d r^4 \kappa^3}$.
Under the same conditions as Lemma \ref{lemma main; about local convexity}, with probability at least $ 1-1/(2 n^4)$, the following is true. For any $(X,Y)\in K_1 \cap K_2 \cap K(\delta)$ and $U,V$ defined in Table \ref{table: def of U,V}, we have
\begin{equation}\label{P Omega (b) bound, formal}
\| \bP_{\Omega}( (U-X)(V-Y)^T) \|_F^2 \leq \frac{p}{25}\|M-XY^T \|_F^2.
\end{equation}
\end{prop}

\emph{Proof of  Proposition \ref{upper bound of P((U-X)(V-Y))}:}
% To prove Proposition \ref{upper bound of P((U-X)(V-Y))},
We need the following random graph lemma \cite[Lemma 7.1]{keshavan2010matrix}. %, which is restated below.
\begin{lemma}\label{random graph lemma}
There exist numerical constants $C_0,C_1$ such that if $|\Omega| \geq C_0 \sqrt{\alpha}n \log n$, then with probability at least $ 1-1/(2 n^4)$,
for all $x\in \mathbb{R}^m, y\in \mathbb{R}^n$,
\begin{equation}\label{random graph ineq.}
 \sum_{(i,j)\in \Omega} x_i y_j \leq C_1 p \|x \|_1 \| y\|_1 + C_1 \alpha^{ \frac{3}{4}} \sqrt{np} \|x\|_2 \|y \|_2.
\end{equation}
\end{lemma}

% Define $d = \|M-XY^T \|_F \leq \delta \leq \frac{\Sigma_{\min}}{C_d r^4 \kappa^3}.$
%Suppose $\bar{Z} D \bar{W}^T $ is the SVD of $b= (U-X)(V-Y)^T$ and let $Z = \bar{Z} D^{1/2} $ and $W = \bar{W} D^{1/2}$.
%Then
%\begin{equation}\label{Z,W bound}
%\| Z\|_F = \| W \|_F = \sum_i D_{ii} \leq \sqrt{r \sum_i D_{ii}^2 } =\sqrt{r} \| b \|_F  .
%\end{equation}
%Denote $b_{ij}$ as the $(i,j)$-th entry of matrix $b$, then
%\begin{equation}
%b_{ij} = (Z^{(i)})^T W^{(j)} = (U^{(i)} - X^{(i)})^T (V^{(j)} - Y^{(j)}) .
%\end{equation}
%By \eqref{summary of U,V (c)} in Corollary \ref{coro: summary of U,V} and the fact $(X,Y) \in K_1$ (which implies $\|X^{(i)} \|\leq \beta_1$, $\|Y^{(j)} \|\leq \beta_2$), we have
%\begin{equation}
%\begin{split}
%\| U^{(i)} - X^{(i)} \| & \leq \|U^{(i)}\| + \| X^{(i)} \| \leq \sqrt{\frac{6r\mu }{5m}}\beta_T + \beta_1 \leq \sqrt{8} \sqrt{\frac{r\mu }{m}}\beta_T,  \\
%\| V^{(j)} - Y^{(j)} \| & \leq \sqrt{8} \sqrt{\frac{r\mu }{n}}\beta_T .
%\end{split}
%\end{equation}
% because \sqrt{6/5} + \sqrt{3} < \sqrt{8}

Let $Z = U-X, W = V-Y$ and $z_i = \|Z^{(i)} \|^2$, $w_j = \|W^{(j)} \|^2$. We have
\begin{equation}\label{(U-X)(V-Y) first estimate} \begin{split}
\| \bP_{\Omega}( (U-X)(V-Y)^T) \|_F^2 & = \sum_{(i,j)\in \Omega} (ZW^T)_{ij}^2 \\
                                      & \leq \sum_{(i,j)\in \Omega} \|Z^{(i)} \|^2 \|W^{(j)} \|^2 = \sum_{(i,j)\in \Omega} z_i w_j.
\end{split}
\end{equation}

Invoking Lemma \ref{random graph lemma}, we have
\begin{equation}\label{inequality obtained by random graph lemma directly}
\begin{split}
 \| \bP_{\Omega}( (U-X)(V-Y)^T) \|_F^2 & \leq C_1 p \|z \|_1 \| w\|_1 + C_1 \alpha^{\frac{3}{4}} \sqrt{np} \|z\|_2 \|w \|_2.   % \\
 %                                      & \leq \frac{1}{100} p d^2 + C_4 \alpha^{3/4} \sqrt{np} \|z\|_2 \|w \|_2,
 \end{split}
 \end{equation}

Analogous to the proof of \eqref{bound b} in Corollary \ref{coro: summary of U,V, extra}, we can prove that
$  % \begin{equation}\label{z w 1 norm bound}
\| U-X\|_F \|V - Y \|_F \leq d/(10 \sqrt{C_1})
$ % \end{equation}
 for large enough $C_d$ (in fact, $C_d \geq 650 C_T \sqrt{C_1}$ suffices).
% Note that we have used an inequality  that is stronger than
% which can be proved similar
Therefore, we have % property \eqref{summary of U,V (b)} in Corollary \ref{coro: summary of U,V}, i.e.\
 \begin{equation}\label{bound on ||z||_1,||w||_1}
\| z \|_1 \|w \|_1 = \|Z \|_F^2 \| W\|_F^2  = \|U-X\|_F^2 \|V-Y \|_F^2 \leq \frac{1}{100C_1}d^2.
\end{equation}

 % in Corollary \ref{coro: summary of U,V}.
%\begin{equation}\label{bound on ||z||_1,||w||_1}
%\| z \|_1 = \|Z \|_F^2 = \|U-X\|_F^2 \leq (11 \frac{\beta_T}{\Sigma_{\min}} r  \kappa d)^2, \\
% \|w \|_1 = \|V-Y \|_F^2 \leq (11 \frac{\beta_T}{\Sigma_{\min}} r  \kappa d)^2.
%\end{equation}

We still need to bound $\| z\|_2$ and $\|w \|_2.$ We have
\begin{equation}\label{bound on ||z||_2}
\begin{split}
\| z\|_2 = \sqrt{\sum_i \| Z^{(i)}\|^4} & \leq \sqrt{\max_i{\| Z^{(i)}\|^2}  \sum_j \| Z^{(j)}\|^2} \\
                                        & \leq \max_i (\| U^{(i)}\| + \| X^{(i)}\|)  \|U-X \|_F \\
                                        & \leq  ( \sqrt{\frac{3r\mu }{2m}}\beta_T + \beta_1) \|U-X \|_F \\
                                        & \leq  \sqrt{8} \sqrt{\frac{r\mu }{m}}\beta_T \|U-X \|_F. \\
                                        %  \sqrt{\frac{r\mu }{m}} 11 \frac{\beta_T^2}{\Sigma_{\min}}  \kappa d \\
                                        % & \leq  11 \sqrt{8} C_T \sqrt{\frac{\mu }{m}} r^{1.5} \kappa^2 d.
\end{split}
\end{equation}
Here, the third inequliaty follows from the property \eqref{summary of U,V (c)} in Corollary \ref{coro: summary of U,V} and the condition $(X,Y) \in K_1$ (which implies $\|X^{(i)} \|\leq \beta_1$), and the fourth inequliaty follows from the definition of $\beta_1$ in \eqref{beta 1 beta T def}.
% and the fourth inequliaty follows from Property (b) in Corollary \ref{coro: summary of U,V}.
Similarly,
\begin{equation}\label{bound on ||w||_2}
\begin{split}
\| w\|_2 \leq & \max_j(\| V^{(j)}\| + \| Y^{(j)}\|)  \| V -Y \|_F \\
         \leq &   \sqrt{8} \sqrt{\frac{r\mu }{n}}\beta_T \|V -  Y \|_F.  %\sqrt{8} \sqrt{\frac{r\mu }{n}} 11 \frac{\beta_T^2}{\Sigma_{\min}} r \kappa d.
\end{split}
\end{equation}
Multiplying \eqref{bound on ||z||_2} and \eqref{bound on ||w||_2}, we get
\begin{align*}
 \| z\|_2 \| w\|_2 \leq 8 \frac{r\mu }{{\sqrt{mn}}}\beta_T^2 \|U-X \|_F \|V -  Y \|_F \overset{\eqref{summary of U,V (b)}}{\leq}
 8 \frac{r\mu }{\sqrt{mn}}\beta_T^2 65 \sqrt{r} \frac{\beta_T^2}{\Sigma_{\rmin}^2}  d^2  \\
 \overset{\eqref{beta 1 beta T def}}{=} 520 C_T^2 \frac{1}{{\sqrt{mn}}} \mu  r^{3.5} \kappa^2 d^2.
\end{align*}
Thus the second term in \eqref{inequality obtained by random graph lemma directly} can be bounded as
\begin{equation}\label{P(b) second term bound}
C_1 \alpha^{ \frac{3}{4} } \sqrt{np} \|z\|_2 \|w \|_2 \leq 520 C_1 C_T^2  \frac{\alpha^{\frac{3}{4}} \sqrt{np}}{\sqrt{mn}} \mu  r^{3.5} \kappa^2  d^2 \leq \frac{3}{100} pd^2,
\end{equation}
where the last inequality is equivalent to
$520^2 C_1^2 C_T^4 \alpha^{\frac{3}{2}} \mu ^2 r^{7} \kappa^4  \leq \frac{9}{100^2} |\Omega|/n $, which holds due to \eqref{Omega bound} with large enough numerical constant $C_0$.
Plugging \eqref{bound on ||z||_1,||w||_1} and \eqref{P(b) second term bound} into (\ref{inequality obtained by random graph lemma directly}), we get $\| \bP_{\Omega}( (U-X)(V-Y)^T) \|_F^2 \leq \frac{p}{25}d^2 = \frac{p}{25}\|M-XY^T \|_F^2$. $\Box$ % $\Box$
% which proves Proposition \ref{upper bound of P((U-X)(V-Y))}

%-------------------------------------------------------------
%----------- subsection 3.3 Proof of $\phi_G \geq 0$----------------------------------
%---------------------------------------------------------------
\subsection{Lower bound on $\phi_G$}\label{subsec: bound phi_G}
In this subsection, we prove the following claim. % that $U,V$ defined in Table \ref{table: def of U,V} satisfy (\ref{phi G}), i.e.\
\begin{claim}\label{claim of phi >= 0}
$U,V$ defined in Table \ref{table: def of U,V} satisfy (\ref{phi G}), i.e.\  $\phi_G = \langle \nabla_X G, X-U \rangle + \langle \nabla_Y G, Y-V \rangle \geq 0$.
\end{claim}
  %  for $U,V$ defined in Table \ref{table: def of U,V}.
  % by Corollary \ref{coro: summary of U,V}.
% Recall that $  G(X,Y) = \rho \sum_{i=1}^m G_0(\frac{3\| X^{(i)} \|^2 } { 2\beta_1^2} ) + \rho \sum_{j=1}^n G_0(\frac{3 \| Y^{(j)} \|^2 } {2\beta_2^2} )
     %      + \rho G_0(\frac{3\| X\|_F^2 } {2\beta_T^2} ) + \rho G_0(\frac{3\| Y \|_F^2 } {2\beta_T^2} ). $
\emph{Proof of  Claim \ref{claim of phi >= 0}:}

By the expressions of $\nabla_X G, \nabla_Y G$ in \eqref{grad of F,G expression}, we have
\begin{equation}\label{phi G expression}
\begin{split}
\phi_G = \langle \nabla_X G,   X-U \rangle + \langle \nabla_Y G, Y-V \rangle &  = \\
 \rho \sum_{i=1}^m G_0'(\frac{3\| X^{(i)} \|^2 } {2\beta_1^2} )\frac{3}{\beta_1^2}\langle X^{(i)}, X^{(i)}- U^{(i)} \rangle
 & + \rho G_0'(\frac{3\| X\|_F^2 } {2\beta_T^2} ) \frac{3}{\beta_T^2}\langle X, X-U \rangle \\
   + \rho \sum_{j=1}^n G_0'(\frac{3 \| Y^{(j)} \|^2 } {2\beta_2^2} ) \frac{3}{\beta_2^2}\langle Y^{(j)}, Y^{(j)}- V^{(j)} \rangle
          &  + \rho G_0'(\frac{3\| Y \|_F^2 } {2\beta_T^2} )\frac{3}{\beta_T^2}\langle Y, Y-V \rangle,
\end{split}
\end{equation}
where $ G_0^{\prime}(z) =I_{[1,\infty]}(z) 2(z-1)  $.

%Therefore, we only need to prove
%\newline $h_1 = G_0'(\frac{3\| X^{(i)} \|^2 } {2\beta_1^2}) \langle X^{(i)}, X^{(i)}- U^{(i)} \rangle \geq 0,$ $ h_2 = G_0'(\frac{3\| X\|_F^2 } {2\beta_T^2} ) \langle X, X-U \rangle \geq 0,$
%\newline $h_2 = G_0'(\frac{3\| Y^{(j)} \|^2 } {2\beta_1^2}) \langle Y^{(j)}, Y^{(j)}- V^{(j)} \rangle \geq 0 $, $ h_4 = G_0'(\frac{3\| Y \|_F^2 } {2\beta_T^2}) \langle Y, Y-V \rangle \geq 0. $

Firstly, we prove
\begin{subequations}\label{h_1, h_2 >=0}
\begin{align}
h_{1i} \triangleq G_0'(\frac{3\| X^{(i)} \|^2 } {2\beta_1^2}) \frac{3}{\beta_1^2} \langle X^{(i)}, X^{(i)}- U^{(i)} \rangle \geq 0, \ \forall \ i,  \label{h_1 >=0}  \\
h_{3j} \triangleq G_0'(\frac{3\| Y^{(j)} \|^2 } {2\beta_2^2}) \frac{3}{\beta_2^2} \langle Y^{(j)}, Y^{(j)}- V^{(j)} \rangle \geq 0, \ \forall \ j.  \label{h_2 >=0}
\end{align}
\end{subequations}
We only need to prove \eqref{h_1 >=0}; the proof of \eqref{h_2 >=0} is similar. We consider two cases.

 Case 1: $ \|X^{(i)} \|^2 \leq \frac{2\beta_1^2}{3}.$ Note that $\frac{3\| X^{(i)} \|^2 } {2\beta_1^2} \leq 1$ implies $G_0'(\frac{3\| X^{(i)} \|^2 } {2\beta_1^2})=0$, thus $h_{1i} = 0$.

 Case 2: $\|X^{(i)} \|^2 > \frac{2\beta_1^2}{3}.$ %We need to prove $\langle X^{(i)}, X^{(i)}- U^{(i)} \rangle \geq 0.$
By Corollary \ref{coro: summary of U,V} and the fact that $\beta_1^2 = \beta_T^2  \frac{3\mu r}{m} $, we have
 \begin{equation}\label{Ui<=Xi, introduce 3r in beta1}
 \|U^{(i)} \|^2 \leq \frac{3r\mu }{2m}\beta_T^2 \leq \frac{2\beta_1^2}{3} < \|X^{(i)} \|^2.
 \end{equation}
 As a result, $\langle X^{(i)}, X^{(i)} \rangle = \|X^{(i)}\| \|X^{(i)}\|  > \|X^{(i)}\| \|U^{(i)}\| \geq \langle X^{(i)}, U^{(i)} \rangle,$ which implies $\langle X^{(i)}, X^{(i)}- U^{(i)} \rangle \geq 0$. Combining this inequality with the fact that $G_0'(\frac{3\| X^{(i)} \|^2 } {2\beta_1^2}) \geq 0,$ we get $h_{1i} \geq 0.$

Secondly, we prove
\begin{equation}\label{h_2, h_4 >=0}
\begin{split}
     h_2 + h_4  & \geq 0,  \\ %- \frac{pd^2}{8}, \\
\text{where } \quad   h_2 \triangleq  G_0'(\frac{3\| X\|_F^2 } {2\beta_T^2} ) \frac{3}{\beta_T^2}\langle X, X-U \rangle,
& \quad      \\
h_4 \triangleq G_0'(\frac{3\| Y \|_F^2 } {2\beta_T^2} )\frac{3}{\beta_T^2}\langle Y, Y-V \rangle. & \quad
\end{split}
\end{equation}

Without loss of generality, we can assume $  \| X\|_F \leq \| Y\|_F, $ and we will apply Corollary \ref{coro: summary of U,V} to prove \eqref{h_2, h_4 >=0}.
If $\| Y\|_F <\| X\|_F$, we can apply a symmetric result of Corollary \ref{coro: summary of U,V} to prove \eqref{h_2, h_4 >=0}.
We further consider three cases.
% We consider several different cases, similar to Table \ref{table: def of U,V}.

Case 1: $ \| X\|_F \leq \|Y \|_F  \leq \sqrt{\frac{2}{3} } \beta_T.$
In this case $G_0'(\frac{3\| X\|_F^2 } {2\beta_T^2} ) = G_0'(\frac{3\| Y \|_F^2 } {2\beta_T^2} ) = 0 $, which implies $h_2 = h_4 = 0$, thus $\eqref{h_2, h_4 >=0}$ holds.

Case 2: $ \|X \|_F \leq \sqrt{\frac{2}{3} } \beta_T < \|Y \|_F .$
 Then $G_0'(\frac{3\| X\|_F^2 } {2\beta_T^2} ) = 0 $, which implies $h_2 = 0$. % thus $\eqref{h_2, h_4 >=0}$ holds.
By \eqref{summary of U,V (d)} in Corollary \ref{coro: summary of U,V} we have $ \| V\|_F \leq \| Y\|_F $,
which implies $\langle Y, Y \rangle   \geq \|Y \|_F \|V\|_F \geq \langle Y, V \rangle$,
i.e.\ $\langle Y , Y - V \rangle \geq 0$. Combined with the nonnegativity of $G_0^{\prime}(\cdot)$, we get $h_4 \geq 0$.
Thus $h_2 + h_4  = h_4 \geq 0. $ % \geq -pd^2/8.$

Case 3: $  \sqrt{\frac{2}{3} } \beta_T < \| X\|_F \leq \|Y \|_F $.
By \eqref{summary of U,V (d)} in Corollary \ref{coro: summary of U,V}, we have $ \| U\|_F \leq \|X \|_F$ and $\| V \|_F \leq \| Y\|_F$.
Similar to the argument in Case 2 we can prove $h_2 \geq 0, h_4 \geq 0$ and \eqref{h_2, h_4 >=0} follows.

In all three cases, we have proved \eqref{h_2, h_4 >=0}, thus \eqref{h_2, h_4 >=0} holds.
% Combining \eqref{h_1, h_2 >=0} and \eqref{h_2, h_4 >=0}, we obtain

We conclude that for $U, V$ defined in Table \ref{table: def of U,V},
$$ \phi_G \overset{ \eqref{phi G expression}}{=} \rho \left(  \sum_i h_{1i}  + \sum_j h_{3j} + h_2 + h_4 \right) \overset{\eqref{h_1, h_2 >=0},\eqref{h_2, h_4 >=0}}{\geq} 0 ,$$
which finishes the proof of Claim \ref{claim of phi >= 0}.
$ \quad \quad \Box$

%-----------------Discussion of why Prop 2 is necessary----------------------------------------------
{\black
 Remark: Based on the above proof, we can explain why Proposition \ref{prop1} is not enough to prove $ \phi_G \geq 0$. % which motivates Proposition \ref{prop2}. % and why Proposition \ref{prop2} is necessary.
Note that $ h_2 = 0 $ when $ \| X\|_F >  \sqrt{\frac{2}{3}}\beta_T $ and $ h_4 = 0 $ when $ \| Y \|_F >  \sqrt{\frac{2}{3}}\beta_T $.
To prove $h_2 \geq 0, h_4 \geq 0,$ it suffices to prove:
(i) $\|U \|_F \leq \| X\|_F$ when $\| X\|_F > \sqrt{\frac{2}{3}}\beta_T $;
(ii)  $\|V \|_F \leq \| Y\|_F $ when $ \| Y\|_F > \sqrt{\frac{2}{3}}\beta_T $.
%\newline If $\| X\|_F > \sqrt{\frac{2}{3}}\beta_T, $ then $ \| U\|_F \leq \|X \|_F. $
%\newline If $\| Y\|_F > \sqrt{\frac{2}{3}}\beta_T, $ then $ \| V\|_F \leq \|Y \|_F. $
For the choice of $U,V$ in Proposition \ref{prop1}, we have $\|U \|_F \leq \| X\|_F$, but
there is no guarantee that (ii) holds. Similarly, for the choice of $U,V$ in the symmetric result of Proposition \ref{prop1},
we have $\|V \|_F \leq \| Y\|_F $, but there is no guarantee that (i) holds.
Thus, Proposition \ref{prop1} is not enough to prove $\phi_G \geq 0$.
To guarantee that (i) and (ii) hold simultaneously, we need a complementary result for the case
 $\| X\|_F > \sqrt{\frac{2}{3}}\beta_T , \| Y\|_F > \sqrt{\frac{2}{3}}\beta_T $. This motivates our Proposition \ref{prop2}.
%%% It turns out that the proof of Proposition \ref{prop2} is much more complicated than that of Proposition \ref{prop1}.

%  $\| V \|_F \leq \| Y\|_F$ when $\| Y\|_F > \sqrt{\frac{2}{3}}\beta_T.$
% In fact, we can construct a counterexample that $U,V$ defined in the proof of Proposition \ref{prop1} do not have property (d) in Corollary \ref{coro: summary of U,V}.

%
%However, in some cases, it is impossible to choose $U,V$ such that $\|U \|_F \leq \| X\|_F$, $\|V \|_F \leq \| Y\|_F$ simultaneously. One simple example is that $r=1, \Sigma=1, M = 1+d.$

% How to resolve this difficulty? In fact, to prove $h_2 \geq 0, h_4 \geq 0,$ we only need to prove property (d) in Corollary \ref{coro: summary of U,V}:
}

%---------------------------------------------------------------

%= \frac{\sin(\theta)}{\sin( \theta + \frac{\pi}{2} -\alpha ) } \|Y_j H_j \|

%\subsubsection{Finish the proof of Lemma \ref{lemma of step 2 and 3}}

%%%%%%%%%%%%%%%%%%%%%%%%%%%%%%%%%%%%%%%%%%%%%%%%%%%%%%%%%
%%%%%%%%%%% section 5  Proof of Lemma 3 %%%%%%%%%%%%%%
%%%%%%%%%%%%%%%%%%%%%%%%%%%%%%%%%%%%%%%%%%%%%%%%%%%%%%%%%%
% \section{Proof of Technical Results for Theorem \ref{theorem 2: linear convergence}}

%%%%%%%%%%%%%%%%%%%%%%%%%%%%%%%%%%%%%%%%%%%%%%%%%%%%%%%%%
%%%%%%%%%%% section 4  Proof of Lemma 1 %%%%%%%%%%%%%%
%%%%%%%%%%%%%%%%%%%%%%%%%%%%%%%%%%%%%%%%%%%%%%%%%%%%%%%%%%
\section{Proof of Lemma \ref{lemma of incoherent neighborhood}}\label{sec: proof of lemma 2}
Property (a) in Lemma \ref{lemma of incoherent neighborhood} (convergence to stationary points) is a basic requirement for many reasonable algorithms and can be proved using classical results in optimization,
so the difficulty mainly lies in how to prove Property (b).
We will give some easily verifiable conditions for Property (b) to hold
and then show that Algorithms 1-4 satisfy these conditions.
This proof framework can be used to extend Theorem \ref{major theorem} to many other algorithms.

% \label{appen: prove lemma of incoherent neighborhood} % and Several key lemmas}
% In this section, we will prove Lemma \ref{lemma of incoherent neighborhood}.
%---------------------------------
% Before providing the conditions for $(X_t, Y_t)$ to be in $K_1 \cap K_2 \cap K(\delta)$, % s if the algorithm satisfies

% The latter part can be proved when the algorithm satisfies some properties along the update direction.
% At the end of this section,
The following claim states that Algorithms 1-4 satisfy Property (a). % in Lemma \ref{lemma of incoherent neighborhood},
The proof of this claim is given in Appendix \ref{appen: local convergence proof}.
\begin{claim}\label{claim: stationary point convergence}
Suppose $\Omega$ satisfies \eqref{RSC of P_Omega}, then each limit point of the sequence generated by Algorithms 1-4 is a stationary point of problem (P1).
\end{claim}
% all have the following property:

% We then prove that Algorithm 1-3 satisfies the two properties in Lemma \ref{lemma of incoherent neighborhood}.
% To provide conditions for $(X_k,Y_k)$ to lie
For Property (b), we first show that the initial point $(X_0, Y_0)$ lies in an incoherent neighborhood $(\sqrt{\frac{2}{3}} K_1) \cap (\sqrt{\frac{2}{3}} K_2) \cap K_{ \delta_0 }$, where %$K_1,  K_2, K(d)$ are defined as in \eqref{def of K_1, K_2} and \eqref{def of K(d)}, and
 $c K_i $ denotes the set $ \{ (cX, cY) \mid (X,Y) \in K_i \} , i=1,2. $
The proof of Claim \ref{claim: initial point properties} will be given in Appendix \ref{appen: initialization proof}.
The purpose of proving $(X_0, Y_0) \in (\sqrt{\frac{2}{3}} K_1) \cap (\sqrt{\frac{2}{3}} K_2)$ rather than $ (X_0, Y_0) \in K_1 \cap K_2$ is to guarantee that $G(X_0, Y_0) = 0$, where $G$ is the regularizer defined in (\ref{regularized function}).
\begin{claim}\label{claim: initial point properties}
%There exist numerical constants $C_0,C_d$ such that the following happens. Suppose each entry of $M$ is observed independently
%randomly with probability
%{\black
%\begin{equation}\label{p bound 1st time, claim of initialization}
%  p \geq C_0 \alpha^{2/3} r^{12} \kappa^{8} \frac{1}{m}
%\end{equation}  }
% Suppose the sample set $\Omega$ satisfies \eqref{RSC of P_Omega}, and $\delta, \delta_0$ are defined as in \eqref{delta definition throughout}.
Under the same condition of Lemma \ref{lemma main; about local convexity}, with probability at least $1 - 1/(2n^4)$,
$(X_0,Y_0)$ given by the procedure $\textsc{Initialize}$ belongs to $(\sqrt{\frac{2}{3}} K_1) \cap (\sqrt{\frac{2}{3}} K_2) \cap K_{ \delta_0 }$,
where $\delta_0$ is defined by \eqref{delta definition throughout},
i.e.\ \newline (a) $\|X_0^{(i)} \| \leq \sqrt{\frac{2}{3}}\beta_1, i =1,2,\dots,m; \;\; \|Y_0^{(j)} \| \leq \sqrt{\frac{2}{3}} \beta_2,j=1,\dots,n;$
\newline (b) $\|X_0 \|_F \leq \sqrt{\frac{2}{3}} \beta_T, \ \|Y_0 \|_F \leq \sqrt{\frac{2}{3}} \beta_T; $
\newline (c) $\| M - X_0Y_0^T\|_F \leq \delta_0.$  % In addition, such $(X_0,Y_0)$ can be computed within time $\mathcal{O}(|\Omega|r \log n),$
\end{claim}

% note: C_4 can be replaced by \frac{1}{3},C_5 by 2.

% Property (a) consists of two parts: $(X_k, Y_k)$ are incoherent and bounded (i.e.\ lies in $K_1 \cap K_2$) and
% $(X_k, Y_k)$ lies in a neighborhood $K(\delta)$.

% relies on Theorem 1.1 in \cite{keshavan2010matrix} and

% Property $(a)$ says that $X_0, Y_0$ are incoherent; property $(b)$ gives an upper bound on $\|X_0\|_F, \|Y_0\|_F$.
% means that $X_0Y_0^T$ is close to $M$; as the distance $\| M - X_0Y_0^T\|_F$ is $O(\Sigma_{\rmin})$, by ``close'' we mean that
% the coefficient of $\Sigma_{\rmin}$ is sufficiently small.

The next result provides some general conditions for $(X_t, Y_t)$ to lie in $K_1 \cap K_2 \cap K(\delta)$.
To simplify the notations, denote $ \bm x_t \triangleq (X_t,Y_t ) $ and $$ \bm u^* \triangleq (\hat{U}\Sigma^{1/2},\hat{V}\Sigma^{1/2}) ,$$
where $\hat{U}\Sigma \hat{V} $ is the SVD of $M$. Recall that $\tilde{F}(\bm u^*) = 0$ (proved in the paragraph after \eqref{constrained formulation}).
{\black
We say a function $\psi(\bar{\bm x}, \bm \Delta; \lambda)$ is a convex tight upper bound of $\tilde{F}(\bm x)$ along the direction $\bm \Delta$ at $\bar{\bm x}$ if
\begin{subequations}\label{BSUM requirement}
\begin{align}
 \psi(\bar{\bm x}, \bm \Delta; \lambda )  \text{ is convex over } \lambda & \in \dR ;   \label{BSUM requirement a}  \\
\psi( \bar{\bm x},  \bm \Delta; \lambda ) \geq \tilde{F}(\bar{\bm x} + \lambda \bm \Delta ), \; \forall \; \lambda \in \dR;  & \quad \psi(\bar{\bm x}, \bm \Delta; 0 )  = \tilde{F}(\bar{\bm x}).   \label{BSUM requirement b}
\end{align}
\end{subequations}
For example,
 $\psi(\bar{\bm x},  \bm \Delta; \lambda ) = \tilde{F}( \bar{\bm x} + \lambda \bm \Delta ) $
  satisfies \eqref{BSUM requirement} for either $\bm \Delta = ( X, 0 )$ or $\bm \Delta = (0, Y)$,
  where $X \in \dR^{m \times r}$ and $Y \in \dR^{n \times r}$ are arbitrary matrices.
  %--------------------------------------
 This definition is motivated by the block successive upper bound minimization method \cite{razaviyayn2013unified}.
 %  here we do not require $\psi$ to have the same derivative as $\tilde{F}$ at $\bm \bar{x}$;
 % nevertheless, those requirements may be necessary for proving convergence to stationary points.
 %---------------------------------------
% $\psi_2(\bar{\bm x}; \bm x) = \tilde{F}( \bar{X}, Y  ) $.
% suppose $\bar{\bm x} = (\bar{X}, \bar{Y})$, and define
% Then both $\psi_1$ and $\psi_2$ are convex over $\bm x = (X,Y)$ and obviously they satisfy \eqref{BSUM requirement b}.
}
The proof of Proposition \ref{prop: K(delta) condition} is given in Appendix \ref{appen: proof of K(d) condition}.
%------------------------------------------
\begin{prop}\label{prop: K(delta) condition}
Suppose the sample set $\Omega$ satisfies \eqref{RSC of P_Omega} and $\delta, \delta_0$ are defined by \eqref{delta definition throughout}. Consider an algorithm that starts from a point $\bm x_0 = (X_0, Y_0)$ and generates a sequence $\{ \bm x_t \} = \{(X_t, Y_t)\}$.
Suppose $\bm x_0$ satisfies
\begin{equation}\label{initial point condition}
\bm x_0 \in (\sqrt{\frac{2}{3}} K_1) \cap (\sqrt{\frac{2}{3}} K_2) \cap K(\delta_0),
\end{equation}
 and $\{\bm x_t\}$ satisfies either of the following three conditions:  % = \{ \bm x_t^1, \dots, \bm x_t^B \}
\begin{subequations}\label{conditions of algorithms}
\begin{align}
& 1) \quad\quad \; \tilde{F}(\bm x_t + \lambda \bm \Delta_t )   \leq 2 \tilde{F}(\bm x_0), \forall \ \lambda \in [0,1],
\nonumber \\
& \text{ where } \bm \Delta_t = \bm x_{t+1} - \bm x_t, \;\; \forall \ t;
 \label{condtion of gradient method}    \\
& 2) \quad\quad
{\black   1 =   \arg \min_{ \bm \lambda \in \dR }  \psi(\bm x_t, \bm \Delta_t; \lambda)  , }  \nonumber \\
& \text{ where } \psi \text{ satisfies } \eqref{BSUM requirement}, \bm \Delta_t = \bm x_{t+1} - \bm x_t, \;\; \forall \ t;
 \label{condtion of exact min method}  \\
 & {\color{black} 3) \quad\quad   \tilde{F}(\bm x_t )   \leq 2 \tilde{F}(\bm x_0), \quad d(\bm x_t, \bm x_0) \leq \frac{5}{6}\delta,  \ \forall \  t. }
 \label{condtion for nonconvex direction}
\end{align}
\end{subequations}
%\begin{enumerate}
%  \item daa
%  \item daa
%\end{enumerate}
 Then $\bm x_t = (X_t, Y_t) \in K_1 \cap K_2 \cap K(2\delta/3),$ for all $ t \geq 0$.  % \frac{5}{6}\delta
\end{prop}

{\color{black}
The first condition means that $\tilde{F}$ is bounded above by $2 \tilde{F}(\bm x_0)$ over the line segment between $\bm x_t$ and $\bm x_{t+1}$ for any $t$.
This condition holds for gradient descent or SGD with small enough stepsize (see Claim \ref{Algo 1-3 satisfy conditions}).
The second condition means that the new point $\bm x_{t+1}$ is the minimum of a convex tight upper bound of the original function along
the direction $\bm x_{t+1} - \bm x_t$, and holds for BCD type methods such as Algorithm 2 and Algorithm 3 (see Claim \ref{Algo 1-3 satisfy conditions}).
Note that the gradient descent method with exact line search stepsize does not satisfy this condition since $\tilde{F}$ is not jointly convex
in the variable $(X,Y)$.
The third condition means that $\tilde{F}(\bm x_t)$ is bounded above and $\bm x_t$ is not far from $\bm x_0$ for any $t$.
 For standard nonlinear optimization algorithms, it is not easy to prove that $\bm x_t$ is not far from $\bm x_0$. However,
 as done by Algorithm 1 with restricted Armijo rule or restricted line search, we can force $d(\bm x_t, \bm x_0) \leq \frac{5}{6}\delta$ to hold when computing the new point $\bm x_{t}$.  }
   % This condition has been used in \cite{keshavan2010matrix}.  then this condition automatically holds

The following claim shows that each of Algorithm 1-4 satisfies one of the three conditions in \eqref{conditions of algorithms}.
% To finish the proof of Lemma \ref{lemma of incoherent neighborhood}, we only need to prove the following claim,
The proof of Claim \ref{Algo 1-3 satisfy conditions} is given in Appendix \ref{appen: prove Claim of algorithm property}.
\begin{claim}\label{Algo 1-3 satisfy conditions}
{\color{black} The sequence $\{ \bm x_t \}$ generated by Algorithm 1 with either restricted Armijo rule or restricted line search satisfies \eqref{condtion for nonconvex direction}.} The sequence $\{ \bm x_t \}$ generated by either Algorithm 2 or Algorithm 3 satisfies  \eqref{condtion of exact min method}.
Suppose the sample set $\Omega$ satisfies \eqref{RSC of P_Omega}, then the sequence $\{ \bm x_t \}$ generated by either Algorithm 1 with constant stepsize or Algorithm 4 satisfies \eqref{condtion of gradient method}. %---TO RECOVER LATER: with constant stepsize
\end{claim}
% Algorithm 1 with exact line search stepsize rule,

To put things together, Claim \ref{claim: stationary point convergence} shows Algorithms 1-4 satisfy Property (a), and
Proposition \ref{prop: K(delta) condition} together with Claim \ref{claim: initial point properties} and Claim \ref{Algo 1-3 satisfy conditions}   shows that Algorithms 1-4 satisfy Property (b). Therefore, we have proved Lemma \ref{lemma of incoherent neighborhood}.

% Now we can replace Property (b) and (c) of Lemma \ref{lemma of incoherent neighborhood} by the conditions of Claim \ref{prop: K(delta) condition}
% to obtain the following result.

%By Claim \ref{P Omega and P has relation}, we have
%$\tilde{F}(\hat{X}_0,\hat{Y}_0,\hat{Z}_0) = \frac{1}{2}\|\bP_{\Omega}(M-\hat{X}_0\hat{Y}_0^T) \|_F^2 \leq p \|M-\hat{X}_0\hat{Y}_0^T \|_F^2$, and
%$ \tilde{F}(\hat{X}_k,\hat{Y}_k,\hat{Z}_k) \geq \frac{1}{2}\|\bP_{\Omega}(M-\hat{X}_0\hat{Y}_0^T) \|_F^2 \geq \frac{1}{18}  p \|M-\hat{X}_k\hat{Y}_k^T \|_F^2 $.
%Since $\tilde{F}$ is nonincreasing, we have $   p \|M-\hat{X}_0\hat{Y}_0^T \|_F^2 \geq \frac{1}{18} p \|M-\hat{X}_k\hat{Y}_k^T \|_F^2 $, thus $\|M-\hat{X}_k\hat{Y}_k^T \|_F  \leq  \sqrt{18} \|M-\hat{X}_0\hat{Y}_0^T \|_F < \delta $.

%%%%%%%%%%%%%%%%%%%%%%%%%%%%%%%%%%%%%%%%%%%%%%%%%%%%%%%%%%%%%%%%%%%%%%%%%%%%%%%%%%%%%%%%%%%%%%%%%%%%%%%%%%%%%%%%%%
%%%%%%%%%%%%%%%%%%%%%%%%%%%%%%%%%%%%%%%%%%%%%%%%%%%%%%%%%%%%%%%%%%%%%%%%%%%%%%%%%%%%%%%%%%%%%%%%%%%%%%%%%%%%%%%%%%
%%%%%%%%%%%%%%%%%%%%%%%%%%%%%%%%%%%%%%%%%%%%%%%%%%%%%%%%%%%%%%%%%%%%%%%%%%%%%%%%%%%%%%%%%%%%%%%%%%%%%%%%%%%%%%%%%%

%%%%%%%%%%%%%%%%%%%%%%%%%%%%%%%%%%%%%%%%%%%%%%%%%%%%%%%%%
%%%%%%%%%%% section   Proof of Proposition 1 %%%%%%%%%%%%%%
%%%%%%%%%%%%%%%%%%%%%%%%%%%%%%%%%%%%%%%%%%%%%%%%%%%%%%%%%%

\appendix

\section{Supplemental Material for Section \ref{sec£ºform and algorithms}} %  and Section \ref{sec: main result} }
\subsection{Proof of Claim \ref{claim: Lip constant}}\label{appen: proof of Claim of Lip constant}
This proof is quite straightforward and we mainly use the triangular inequalities and the boundedness of the considered region $\Gamma(\beta_0)$.
In this proof, $f'(x)$ denotes the derivative of a function $f$ at $x$.
%  and consists mainly of triangle inequalities.

Since $(X,Y), (U,V)$ belong to $\Gamma(\beta_0)$, we have
\begin{equation}\label{beta_0 bound all}
 \| X\|_F \leq \beta_0,   \| Y\|_F \leq \beta_0,
  \| U\|_F \leq \beta_0,  \| V\|_F \leq \beta_0.
\end{equation}
We first prove
\begin{equation}\label{Lip F part}
\| \nabla F(X,Y) - \nabla F( U, V) \|_F \leq  4 \beta_0^2 \| (X,Y) - (U,V) \|_F.
\end{equation}
By the triangular inequality, we have
\begin{equation}\label{nabla_X F bound 1}
\begin{split}
\| \nabla_X F(X,Y) - \nabla_X F( U, V) \|_F \leq \| \nabla_X F(X,Y) - \nabla_X F( U, Y) \|_F   \\
 + \|\nabla_X F(U,Y) - \nabla_X F( U, V) \|_F.
  \end{split}
\end{equation}
The first term of \eqref{nabla_X F bound 1} can be bounded as follows
\begin{equation}\nonumber %\label{nabla_X F term 1 bound}
\begin{split}
\| \nabla_X F(X,Y) - \nabla_X F( U, Y) \|_F
& = \| \bP_{\Omega}(XY^T - M) Y - \bP_{\Omega}(UY^T - M) Y \|_F    \\
&  \leq \| \bP_{\Omega}(XY^T - M)  - \bP_{\Omega}(UY^T - M) \|_F \| Y\|_F    \\
&  = \|  \bP_{\Omega}[ (X-U)Y^T ] \|_F \| Y\|_F      \\
 &  \leq \| (X-U)Y^T  \|_F \| Y\|_F        \\
 &    \leq \| X-U \|_F \| Y\|_F^2          \\
 &      \leq \| X-U \|_F  \beta_0^2.
\end{split}
\end{equation}
The second term of \eqref{nabla_X F bound 1} can be bounded as
\begin{equation}\nonumber %\label{nabla_X F term 2 bound}
\begin{split}
& \| \nabla_X F(U,Y) - \nabla_X F( U, V) \|_F  \\
& = \| \bP_{\Omega}(UY^T - M) Y - \bP_{\Omega}(UV^T - M) V \|_F     \\
& \leq \| \bP_{\Omega}(M) (V - Y) \|_F + \| \bP_{\Omega}(UY^T)Y -  \bP_{\Omega}(UV^T) V \|_F  \\
& \leq \| \bP_{\Omega}(M) (V - Y) \|_F + \| \bP_{\Omega}(UY^T)Y -  \bP_{\Omega}(UY^T) V \|_F   \\
& \quad \quad
+ \| \bP_{\Omega}(UY^T)V -  \bP_{\Omega}(UV^T) V \|_F  \\
& \leq \| \bP_{\Omega}(M) \|_F \|V - Y \|_F + \| \bP_{\Omega}(U Y^T ) \|_F \|Y - V \|_F
 \\ & \quad \quad
+ \| \bP_{\Omega}[U(Y-V)^T]  \|_F \| V\|_F
 \\
& \leq  \| M \|_F \|V - Y \|_F + \| U  \|_F \| Y\|_F \|Y - V \|_F +  \| U \|_F \| Y-V  \|_F \| V\|_F \\
& \leq  3\beta_0^2 \|Y - V \|_F,
\end{split}
\end{equation}
where the last inequliaty follows from \eqref{beta_0 bound all} and the fact that
$ \| M\|_F \leq \sqrt{r} \Sigma_{\rmax} \overset{\eqref{beta 1 beta T def} }{=} \frac{1}{C_T \sqrt{r}}\beta_T^2 \leq \beta_T^2 \leq \beta_0^2 ${\black (here the second last inequality
follows from the fact that the numerical constant $C_T \geq 1$, and the last inequality follows from the assumption of Claim \ref{claim: Lip constant})}.

Plugging the above two bounds into \eqref{nabla_X F bound 1}, we obtain  % \eqref{nabla_X F term 1 bound} and \eqref{nabla_X F term 2 bound}
\begin{equation}\nonumber %\label{nabla_X F final bound}
\| \nabla_X F(X,Y) - \nabla_X F( U, V) \|_F \leq \beta_0^2 ( \| X-U \|_F + 3 \|Y - V \|_F ).
\end{equation}
Similarly, we have
\begin{equation}\nonumber %\label{nabla_Y F final bound}
\| \nabla_Y F(X,Y) - \nabla_Y F( U, V) \|_F \leq \beta_0^2 ( 3\| X-U \|_F + \|Y - V \|_F ).
\end{equation}
% Denote $$ \omega_1 \triangleq  \|X-U \|_F, \omega_2 \triangleq  \|Y - V \|_F .$$
%{\blue Remark: You ask to define $\omega_1, \omega_2$ at the beginning and use it throughout; however, these definition are only used in the following inequality for ease of computation and nowhere else. }
Combining the above two relations, we have (denote $\omega_1 \triangleq  \|X-U \|_F, \omega_2 \triangleq  \|Y - V \|_F$) %\eqref{nabla_X F final bound} and \eqref{nabla_Y F final bound}, we have
\begin{equation}\nonumber
\begin{split}
& \| \nabla F(X,Y) - \nabla F( U, V) \|_F  \\
& = \sqrt{ \| \nabla_X F(X,Y) - \nabla_X F( U, V) \|_F^2 + \| \nabla_Y F(X,Y) - \nabla_Y F( U, V) \|_F^2 }   \\
& \leq  \beta_0^2 \sqrt{ (\omega_1 + 3 \omega_2)^2 + (3\omega_1 +  \omega_2)^2 }    \\
& \leq 4 \beta_0^2 \sqrt{\omega_1^2 + \omega_2^2} \\
& = 4 \beta_0^2 \| (X,Y) - (U,V) \|_F,
\end{split}
\end{equation}
which proves \eqref{Lip F part}.

Next we prove
\begin{equation}\label{Lip G part}
\| \nabla G(X,Y) - \nabla G( U, V) \|_F \leq 54 \rho \frac{\beta_0^2}{\beta_1^4}  \| (X,Y) - (U,V) \|_F.
\end{equation}  % 21\rho ( \frac{1}{\beta_1^2} +  \frac{1}{\beta_0^2} )
Denote
\begin{equation}\label{G1 G2 def}
  G_{1i} ( X )  \triangleq G_0 \left( \frac{3\| X^{(i)} \|^2  }{2\beta_1^2} \right) , \quad  G_2 (X)
   \triangleq G_0 \left(\frac{3\| X\|_F^2 } {2\beta_T^2}  \right) ,
\end{equation}
then we have
\begin{equation}\label{nabla G1 G2 def}
  \nabla G_{1i} ( X ) = G_0^{\prime} \left(\frac{3\| X^{(i)} \|^2  } {2\beta_1^2} \right) \frac{3 \bar{X}^{(i)} }{\beta_1^2},\quad
 \nabla G_2 (X)    = G_0^{\prime} \left(\frac{3\| X\|_F^2 } {2\beta_T^2}  \right) \frac{3 X }{\beta_T^2},
\end{equation}
 where $ G_0^{\prime}(z) =I_{[1,\infty]}(z) 2(z-1)  $  %  e^{(z-1)^2}
 and $\bar{X}^{(i)}$ denotes a matrix with the $i$-th row being $X^{(i)} $ and the other rows being zero.
 Obviously $ G_{1i} ( X )$ is a matrix with all but the $i$-th row being zero.
Recall that
$$
  G(X,Y) =  \rho \sum_i G_{1i}( X ) +   \rho G_2( X ) + f_0(Y),
$$
where $f_0(Y)$ is a certain function of $Y$ which we can ignore for now.
Then we have
\begin{equation}\label{grad G decompose}
\begin{split}
\nabla_X G(X,Y) =  \rho \sum_i \nabla G_{1i}( X ) +   \rho \nabla G_2( X )  \\
= \rho \sum_{i=1}^m G_0^{\prime} \left(\frac{3\| X^{(i)} \|^2  } {2\beta_1^2}  \right) \frac{3 \bar{X}^{(i)}  }{\beta_1^2} +
 \rho G_0^{\prime}\left(\frac{3\| X\|_F^2 } {2\beta_T^2}  \right) \frac{3 X }{\beta_T^2},
 \end{split}
\end{equation}
and, similarly,
 $$
\nabla_X G(U,V) =  \rho \sum_i \nabla G_{1i}( U ) +   \rho G_2( U ).
  $$
Therefore, we have
\begin{equation}\label{nabla_X G diff bound 1}
\begin{split}
& \| \nabla_X G(X,Y) - \nabla_X G(U,V) \|_F \\
& = \| \rho \sum_i [ \nabla G_{1i}(X  ) - \nabla G_{1i}( U ) ]   +   \rho [ \nabla G_2( X ) - \nabla G_2( U )  ] \|_F   \\
&  \leq \| \rho \sum_i [  \nabla G_{1i}(X  ) - \nabla G_{1i}( U )] \|_F + \rho \| \nabla G_2( X ) - \nabla G_2( U )   \|_F   \\
&  = \rho \sqrt{ \sum_i \|  \nabla G_{1i}(X  ) - \nabla G_{1i}( U ) \|_F^2 } + \rho \| \nabla G_2( X ) - \nabla G_2( U )   \|_F,
\end{split}
\end{equation}
where the last equality is due to the fact that each $ \nabla G_{1i}(X  ) - \nabla G_{1i}( U ) $ is a matrix with all but the $i$-th row being zero.
 % Now let us provide an upper bound of the second term  $ \| \nabla G_2( X ) - \nabla G_2( U )   \|_F$.
Denote
\begin{equation}\label{z_1,z_2 def}
 z_1 \triangleq \frac{3\| X\|_F^2 } {2\beta_T^2}, z_2 \triangleq \frac{3\| U\|_F^2 } {2\beta_T^2}.
 \end{equation}
Then by \eqref{z_1,z_2 def}, \eqref{nabla G1 G2 def} and the triangle inequality we have
\begin{equation}\label{G_2 diff bound 1}
\begin{split}
& \frac{\beta_T^2}{3}  \| \nabla G_2( X ) -  \nabla G_2( U ) \|_F  \\
  &  {=} \| G_0^{\prime}(z_1 ) X -  G_0^{\prime}(z_2 ) U  \|_F  \\
          &  \leq  |G_0^{\prime}(z_1 )| \|X-U\|_F  +   | G_0^{\prime}(z_1 ) - G_0^{\prime}(z_2 )| \| U \|_F.
\end{split}
\end{equation}
By the definitions of $z_1,z_2$ in \eqref{z_1,z_2 def} and using $\| X\|_F \leq \beta_0, \| Y\|_F \leq \beta_0$, we have
\begin{equation}\label{z_1 - z_2 bound}
\begin{split}
|z_1 - z_2| & = \frac{3}{2\beta_T^2} (\| X\|_F^2 - \| U\|_F^2)  \\
 & = \frac{3}{2\beta_T^2} (\| X\|_F + \| U\|_F) (\| X\|_F - \| U\|_F)    \\
 & \leq \frac{3 \beta_0}{\beta_T^2}  \| X - U\|_F .
% \\ & = \frac{3}{\beta_0} \| X - U\|_F.
\end{split}
\end{equation}
According to \eqref{beta_0 bound all} and the definitions of $z_1, z_2$ in \eqref{z_1,z_2 def}, we have
\begin{equation}\label{z_1,z_2 range}
 \max\{ z_1, z_2 \} \leq \frac{3}{2} \frac{\beta_0^2}{\beta_T^2}.
\end{equation}
% When $0\leq z \leq 1$, we have $G_0^{\prime}(z ) = 0 $  and thus $ \frac{\beta_0^2}{3}  \| \nabla G_2( X ) -  \nabla G_2( U ) \|_F  = 0$.
% When $ 0 \leq z \leq \frac{3}{2}\frac{\beta_0^2}{\beta_T^2}$,
We can bound the first and second order derivative of $G_0$ as follows:
\begin{align}
G_0^{\prime}(z) = I_{[1,\infty]}(z) 2 (z-1) \leq 3 \frac{\beta_0^2}{\beta_T^2}, \;\; \forall z\in [0, \frac{3}{2}  \frac{\beta_0^2}{\beta_T^2}],  \label{1st order derivative of G_0}   \\
G_0^{\prime \prime}(z ) =2 I_{[1,\infty]}(z)  \leq 2,  \;\;  \forall z\in [0, \infty)  . \label{2nd order derivative of G_0}
\end{align}
%\begin{align}
%G_0^{\prime}(z ) = I_{[1,\infty]}(z) 2 (z-1) e^{(z-1)^2} \leq \frac{1}{2} e^{1/4} \leq 1, \;\; \forall z\in [0,\frac{3}{2}]  \label{1st order derivative of G_0}   \\
%G_0^{\prime \prime}(z ) = I_{[1,\infty]}(z) (4 (z-1) + 2) e^{(z-1)^2} \leq 3 e^{1/4} \leq 4,  \;\;  \forall z\in [0,\frac{3}{2}]  . \label{2nd order derivative of G_0}
%\end{align}
By the mean value theorem and \eqref{2nd order derivative of G_0}, we have  %and \eqref{z_1,z_2 range}
\begin{equation}\label{G_0(z) diff bound}
| G_0^{\prime}(z_1 ) - G_0^{\prime}(z_2 )| \leq 2 |z_1 - z_2| \overset{\eqref{z_1 - z_2 bound}}{\leq} \frac{6\beta_0}{\beta_T^2} \| X - U\|_F.
\end{equation}

%Plugging \eqref{z_1 - z_2 bound} into \eqref{G_0(z) diff bound} we get
%\begin{equation}\label{G_0(z) diff bound}
%| G_0^{\prime}(z_1 ) - G_0^{\prime}(z_2 )| \leq \frac{12}{\beta_0} \| X - U\|_F.
%\end{equation}

Plugging \eqref{1st order derivative of G_0} (with $z = z_1$) and \eqref{G_0(z) diff bound} into \eqref{G_2 diff bound 1}, we obtain
\begin{align}
  \frac{\beta_T^2}{3}  \| \nabla G_2( X ) -  \nabla G_2( U ) \|_F
        &    \leq 3 \frac{\beta_0^2}{\beta_T^2} \|X-U\|_F  +   \frac{6\beta_0}{\beta_T^2} \| X - U\|_F  \| U \|_F \nonumber \\
        &  \leq 9  \frac{\beta_0^2}{\beta_T^2} \|X-U\|_F   \nonumber  \\
\Longrightarrow \quad  \| \nabla G_2( X )   -  \nabla G_2( U ) \|_F  &  \leq  27 \frac{\beta_0^2}{\beta_T^4} \|X-U\|_F.  \label{G2 diff bound final}
\end{align}

% According to the definition of $K_2$,
Since $
 \| X^{(i)} \|_F \leq \| X\|_F \leq \beta_0,   \| U^{(i)}\| \leq \| U \|_F \leq \beta_0,$
%which implies inequalities similar to \eqref{z_1,z_2 range}:
%$$
% \frac{3\| X^{(i)}\|_F^2 } {2\beta_1^2}  \leq \frac{3}{2} \frac{\beta_0^2}{\beta_1^2}, \frac{3\| U^{(i)}\|_F^2 } {2\beta_1^2}  \leq \frac{3}{2}
%$$
by an argument analogous to that for \eqref{G2 diff bound final}, we can prove
\begin{equation}\nonumber%\label{G_1 diff bound per i}
  \| \nabla G_{1i}( X ) -  \nabla G_{1i}( U ) \|_F \leq 27 \frac{\beta_0^2}{\beta_1^4}  \|X^{(i)} - U^{(i)}\|, \;\; \forall \ i,
\end{equation}
which further implies
\begin{equation}\label{G_1 diff bound final}
\begin{split}
& \sqrt{ \sum_i \|  \nabla G_{1i}(X  ) - \nabla G_{1i}( U ) \|^2 }   \\
& \leq 27 \frac{\beta_0^2}{\beta_1^4}  \sqrt{ \sum_i \| X^{(i)} - U^{(i)} \|^2 }
 = 27 \frac{\beta_0^2}{\beta_1^4} \|X - U \|_F.
 \end{split}
\end{equation}
Plugging \eqref{G2 diff bound final} and \eqref{G_1 diff bound final} into \eqref{nabla_X G diff bound 1}, we obtain
\begin{equation}\nonumber% \label{nabla_X G diff bound final}
\begin{split}
\| \nabla_X G(X,Y) - \nabla_X G(U,V) \|_F \leq 54 \rho \frac{\beta_0^2}{\beta_1^4} \|X - U \|_F.
\end{split}
\end{equation}
Similarly, we can prove
\begin{equation}\nonumber%\label{nabla_Y G diff bound final}
\| \nabla_Y G(X,Y) - \nabla_Y G(U,V) \|_F % \leq  21\rho ( \frac{1}{\beta_2^2} +  \frac{1}{\beta_0^2} ) \| Y - V \|_F
\leq  54 \rho \frac{\beta_0^2}{\beta_2^4} \| Y - V \|_F \leq 54 \rho \frac{\beta_0^2}{\beta_1^4} \| Y - V \|_F,
\end{equation}
where the last inequality is due to $ \beta_1 = \beta_T \sqrt{ \frac{3\mu r}{m}  } \leq \beta_T \sqrt{ \frac{3\mu r}{n}   } = \beta_2 $.
Combining the above two relations yields \eqref{Lip G part}. %\eqref{nabla_X G diff bound final} and \eqref{nabla_Y G diff bound final}

Finally, we combine \eqref{Lip F part} and \eqref{Lip G part} to obtain
\begin{equation}\nonumber
\begin{split}
& \| \nabla \tilde{F}(X,Y) - \nabla \tilde{F}(U,V) \|_F  \\
 & \leq \| \nabla F(X,Y) - \nabla F(U,V) \|_F + \| \nabla G(X,Y) - \nabla G(U,V) \|_F \\
 & \leq \left( 4\beta_0^2 + 54 \rho \frac{\beta_0^2}{\beta_1^4}  \right) \|(X,Y) - (U,V) \|_F,
 \end{split}
\end{equation}
which finishes the proof of Claim \ref{claim: Lip constant}.  $\Box$

Remark: If we further assume that the norm of each $X^{(i)}$ (resp.\ $Y^{(j)}$) is bounded by $O( \beta_1 )$ (resp.\ $O( \beta_2)$), the Lipschitz constant can be improved to $ 4\beta_0^2 + 54 \rho \frac{\beta_0^2}{\beta_T^4} $.

\subsection{Solving the Subproblem of Algorithm 3}\label{appen: Algorithm 3 subproblem}
The subproblem of Algorithm 3 for the row vector $X^{(i)}$ is
$$ \min_{X^{(i)}} \tilde{F}(X_{k}^{(1)}, \dots,X_{k}^{(i-1)}, X^{(i)}, X_{k-1}^{(i+1)}
  \dots, X_{k-1}^{(m)}, Y_{k-1}) + \frac{\lambda_0}{2} \| X^{(i)} - X_{k-1}^{(i)} \|^2. $$
For simplicity, denote $X^{(i)} = x_i, X_{k-1}^{(i)} = \bar{x}_i$, $X_{k}^{(j)} = x_j, 1 \leq j \leq i-1$, $X_{k-1}^{(j)} = x_j, i+1 \leq j \leq m $,
and  $Y_{k-1}^{(j)} = y_j, 1 \leq j \leq n$. Then the above problem becomes
$$ \min_{ x_i } \tilde{F}(x_1,\dots, x_{i-1}, x_i, x_{i+1},
  \dots, x_m, y_1, \dots, y_n) + \frac{\lambda_0}{2} \| x_i - \bar{x}_i \|^2. $$
% $$ \min_{X^{(i)}} \tilde{F}(X^{(1)}, \dots,X^{(i-1)}, X^{(i)}, X^{(i+1)},\dots, X^{(m)}, Y) + \frac{\lambda_0}{2} \| X^{(i)} - X_{k-1}^{(i)} \|^2. $$
The optimal solution $x_i^*$ to this subproblem satisfies the equation
$ \nabla_{x_i} \tilde{F} = 0 $, i.e.\ \begin{equation}\label{grad x_i =0}
 A x_i - b + g(\|x_i \|) x_i = 0,
\end{equation}
where $ A = \sum_{j \in \Omega_{i}^{x} } y_j y_j^T + \lambda_0 I $ is a symmetric PD (positive definite) matrix,
$ b  = \sum_{j\in \Omega_i^{x}} M_{ij} y_j + \lambda_0 \bar{x}_i$, and $g$ is a function defined as
\begin{equation}\nonumber
g(z) = \rho  \frac{3}{\beta_1^2} G_0^{\prime}( \frac{3 z^2}{2 \beta_1^2} ) +
 \rho \frac{3}{\beta_T^2} G_0^{\prime}( \frac{3 ( z^2 + \xi_i )}{2 \beta_T^2} ) ,
\end{equation}
 in which $\xi_i = \sum_{j \neq i} \|x_j\|^2 $ is a constant. Note that $g$ has the following properties:
 a) $g(z) = 0 $ when $z^2 \leq \min \{ \frac{2 \beta_1^2}{3}, \frac{2 \beta_T^2}{3} - \xi_i \}$ ; b) $g$ is an increasing function in $[0, \infty)$.
 The equation \eqref{grad x_i =0} is equivalent to
  \begin{equation}\label{LS solution}
    x_i = (A + g(\|x_i \|) I)^{-1} b.
\end{equation}
Suppose the eigendecomposition of $A$ is $B \Lambda B^T$ and let $ \Phi = B^T b b^T B $,
then \eqref{LS solution} implies
\begin{align}
    \|x_i\|^2 =   \|(A + g(\|x_i \|) I)^{-1} b\|^2 & = \tr( (A + g(\|x_i \|) I)^{-2} bb^T  ) \nonumber  \\
    = \tr( (\Lambda + g(\|x_i \|) I)^{-2} \Phi  )  &
    = \sum_{k=1}^r \frac{\Phi_{kk}}{ (\Lambda_{kk} + g(\|x_i \|) )^2 }, \nonumber \\
   \Longrightarrow  \quad \quad 1    = \frac{1}{\|x_i \|^2 }  \sum_{k=1}^r  & \frac{\Phi_{kk}}{ ( \Lambda_{kk} + g(\|x_i \|) )^2 },
   \label{grad x_i = 0 transformed equation}
\end{align}
where $ Z_{kk} $ denotes the $(k,k)$-th entry of matrix $Z$.
Since $A$ and $\Phi$ are PSD (positive semidefinite) matrices, we have $ \Phi_{kk} \geq 0, \Lambda_{kk} \geq 0 $.
The righthand side of \eqref{grad x_i = 0 transformed equation} is a decreasing function of $\| x_i\|$,
thus the equation \eqref{grad x_i = 0 transformed equation} can be solved via a simple bisection procedure.
After obtaining the norm of the optimal solution $z^* = \|x_i^* \|$, the optimal solution $x_i^*$ can be obtained by \eqref{LS solution}, i.e.\   \begin{equation}\label{LS solution, optimal}
    x_i^* = (A + g( z^*) I)^{-1} b.
\end{equation}

Similarly, the subproblem for $Y^{(j)}$ can also be solved by a bisection procedure. % simple binary search method.

\section{ Proof of Proposition \ref{prop1}}\label{appen: proof of Prop 1}

\subsection{Matrix norm inequalities}
{\black We first prove some basic inequalities related to the matrix norms.
These simple results will be used in the proof of Propositions \ref{prop1} and \ref{prop2}.  }
% These simple results are certainly not novel, but
% for completeness we will provide proofs for all of them .}
\begin{prop}\label{matrix prop: |A-B| bound}
If $A, B \in \dR^{n_1 \times n_2}$, then
 \begin{equation}\label{|A-B| bound ineq.}
  \|A-B \|_2 \geq \sigma_{\rmin}(A) - \sigma_{\rmin}(B).
  \end{equation}
\end{prop}
\emph{Proof:}
$ \sigma_{\rmin}(A) = \min_{\| v\| = 1} \| A v\| \leq \min_{\| v\| = 1} ( \| B v \| + \|(A-B) v \|)
\leq \min_{\| v\| = 1} \| B v \| + \| A-B\| = \sigma_{\rmin}(B) + \| A-B\| .  $   % $\Box$

%--------------------------------
\begin{prop}\label{matrix prop: sigma_min(AB) upper bound}
For any $A \in \dR^{n_1 \times n_2}, B \in \dR^{n_2 \times n_3}$, we have
\begin{equation}\label{sigma_min(AB) upper bound}
\sigma_{\rmin}(AB) \leq \sigma_{\rmin}(A) \| B \|_2.
\end{equation}
\end{prop}

\emph{Proof: }
$ \sigma_{\rmin}(AB) =  \min_{v \in \dR^{n_1 \times 1}, \|v \| = 1 } \|v^T AB \| \leq \min_{v \in \dR^{n_1 \times 1}, \|v \| = 1 } \|v^T A \| \|B \|_2
=  \sigma_{\rmin}(A) \| B \|_2. $    % $\Box$

%----------------------------------------------------------------

\begin{prop}\label{matrix prop: |(A)|>|B| with smaller row norm}
Suppose $A, B \in \dR^{n_1 \times n_2} $ and{\black $ c_i A^{(i)} =  B^{(i)}$, where $c_i \in \dR$ and $|c_i| \leq 1$}, for $i =1, \dots, n_1 $
(recall that $Z^{(i)}$  denotes the $i$-th row of $Z$).
Then
\begin{equation}\nonumber%\label{submatrix has smaller spectral norm}
 \| B \|_2 \leq  \| A \|_2.
\end{equation}
\end{prop}

\emph{Proof: } For simplicity, denote $ a_i \triangleq (A^{(i)})^T, b_i \triangleq  (B^{(i)})^T $. Then
\begin{align*}
 \|B \|_2^2 =  \max_{\|v \|=1} \|B v \|^2 = \max_{\|v \|=1} \sum_i (b_i^T v)^2
 \\
= \max_{\|v \|=1} \sum_i c_i^2 (a_i^T v)^2
\leq \max_{\|v \|=1} \sum_i ( a_i^T v)^2
= \|A \|_2^2.
\end{align*}
  % $\Box$

\begin{coro}\label{matrix prop: submatrix |B|<|A|}
Suppose $B \in \dR^{n_1 \times n_2} $ is a submatrix of $A \in \dR^{m_1 \times m_2}$, then
\begin{equation}\label{submatrix has smaller spectral norm}
 \| B \|_2 \leq  \| A \|_2.
\end{equation}
\end{coro}
\emph{Proof:}{\black By Proposition \ref{matrix prop: |(A)|>|B| with smaller row norm}, we have
\begin{equation}\nonumber% \label{(X_1, X_2) norm > X_1 norm}
\|(X_1, X_2)\|_2  \geq \| (X_1, 0)\|_2 = \|X_1 \|_2.
\end{equation} }
%In fact, \eqref{(X_1, X_2) norm > X_1 norm} can be proved as follows:
% $$ \|(X_1, X_2)\|_2^2 = \max_{\|v \|=1} v^T (X_1, X_2) (X_1, X_2)^T  v = \max_{\|v \|=1} v^T (X_1 X_1^T + X_2 X_2^T ) v
%\geq \max_{\|v \|=1} v^T X_1 X_1^T  v = \| X_1\|_2^2. $$
 Without loss of generality, suppose $A = \begin{bmatrix}
  B  & B_1 \\
  B_2 & B_3
\end{bmatrix}$.
Applying the above inequality twice, we get
$$ \| A\|_2 \geq \|(B, B_1)\|_2 \geq \| B\|_2. $$     % $\Box$

%----------------------------------------------------------------

\begin{prop}\label{matrix prop: |AB| bound}
For any $A \in \dR^{n_1 \times n_2}, B \in \dR^{n_2 \times n_3}$, we have
 \begin{subequations}
 \begin{align}
   \|AB \|_F \leq \|A \|_2 \|B \|_F, \label{AB_F upper bound}\\
   \|AB \|_2 \leq \|A \|_2 \|B \|_2. \label{AB spectral upper bound}
  \end{align}
  \end{subequations}
Further, if $n_1 \geq n_2$, then % (i.e.\ $A \in \dR^{n_1 \times n_1}, B \in \dR^{n_1 \times n_3}$), then
  \begin{subequations}
 \begin{align}
     \sigma_{\rmin}(A) \|B \|_F \leq \|AB \|_F,  \label{AB_F lower bound}  \\
  \sigma_{\rmin}(A) \|B \|_2 \leq  \|AB \|_2 .  \label{AB spectral lower bound}
  \end{align}
  \end{subequations}
\end{prop}
\emph{Proof:} Assume the SVD of $A$ is $A_1 D A_2,$ where $A_1 \in \dR^{n_1 \times n_1}, A_2\in \dR^{n_2 \times n_2}$ are orthonormal matrices
 and $D \in \dR^{n_1 \times n_2}$ has nonzero entries $D_{ii}, i=1,\dots, \min \{ n_1, n_2\}$.
 Note that
 $$
   \sigma_{\rmin}(A) \leq D_{ii} \leq \| A\|_2, \forall \ i.
 $$
Let $B^{\prime} = A_2 B$ and suppose the $i$-th row of $B^{\prime}$ is $b_i, i=1,\dots, n_2$, then
\begin{equation}\label{AB norm intermediate 1}
 \|AB \|_F^2 = \|D A_2 B \|_F^2 = \|D B^{\prime} \|_F^2 = \sum_{i=1}^{\min\{n_1, n_2\}} D_{ii}^2 \|b_i\|^2.
 \end{equation}
The the RHS (right hand side) can be bounded from above as
\begin{align*}
\sum_{i=1}^{\min\{n_1, n_2\}} D_{ii}^2 \|b_i\|^2 \leq \|A \|_2^2 \sum_{i=1}^{\min\{n_1, n_2\}} \|b_i\|^2 \\
\leq \|A \|_2^2 \sum_{i=1}^{n_2} b_i^2 = \|A \|_2^2 \|B^{\prime} \|_F^2 =   \|A \|_2^2 \|B \|_F^2   .
\end{align*}

Combining the above relation and \eqref{AB norm intermediate 1}  leads to \eqref{AB_F upper bound}. %we have $ \|AB \|_F \leq \|A_1 \|_2 \|B \|_F$.
% \eqref{AB norm intermediate 2}

If $n_1 \geq n_2$, then $\min\{n_1, n_2\} = n_2$, and the RHS of \eqref{AB norm intermediate 1} can be bounded from below as
\begin{align*}
\sum_{i=1}^{\min\{n_1, n_2\}} D_{ii}^2 \|b_i\|^2 = \sum_{i=1}^{n_2} D_{ii}^2 \|b_i\|^2
\geq \sigma_{\rmin}(A)^2 \sum_{i=1}^{n_2} \|b_i\|^2  \\
= \sigma_{\rmin}(A)^2 \|B^{\prime} \|_F^2 =  \sigma_{\rmin}(A)^2 \|B \|_F^2  .
\end{align*}
Combining the above relation and \eqref{AB norm intermediate 1}  leads to \eqref{AB_F lower bound}.
% \eqref{AB_F lower bound inter}

Next we prove the inequalities related to the spectral norm. We have
\begin{equation}\label{AB spectral interm 1}
 \|AB \|_2 = \|D A_2 B \|_2 = \|D B^{\prime} \|_2 = \max_{\|v \| \leq 1, v \in \dR^{  n_1 \times 1} } \| v^T D B^{\prime} \|.
 \end{equation}
Note that
 $\{ v^T D \mid \|v \| \leq 1, v \in \dR^{ n_1 \times 1} \} \subseteq \{ u^T \mid u \in \dR^{ n_2 \times 1}, \| u\| \leq \|A \|_2  \} $,
 thus the RHS of \eqref{AB spectral interm 1} can be bounded from above as
\begin{align*}
 \max_{\|v \| \leq 1, v \in \dR^{  n_1 \times 1 } } \| v^T D B^{\prime} \| \leq \max_{ u \in \dR^{  n_2 \times 1}, \| u\|
  \leq \|A \|_2 } \|u^T B^{\prime}\|
  \\
 = \|A \|_2 \| B^{\prime} \|_2 = \|A \|_2 \| B \|_2.
\end{align*}
Combining the above relation and \eqref{AB spectral interm 1} leads to \eqref{AB spectral upper bound}.

If $n_1 \geq n_2$, then
$\{ u^T \mid u \in \dR^{ n_2 \times 1}, \| u\| \leq \sigma_{\min}(A)  \}    \subseteq \{ v^T D \mid \|v \| \leq 1, v \in \dR^{ n_1 \times 1} \} $
(in fact, for any $\| u\|\leq \sigma_{\min}(A) $, let $v_i = u_i/D_{ii}, i=1,\dots, n_2 $ and $v_i = 0, n_2 < i \leq n_1$, where $v_i$ denotes the $i$-th entry of $v$, then $v^T D = u^T$ and $\|v \| \leq 1 $). Thus the RHS of  \eqref{AB spectral interm 1} can be bounded from below as
\begin{align*}
 \max_{\|v \| \leq 1, v \in \dR^{ n_1 \times 1} } \| v^T D B^{\prime} \| \geq \max_{ u \in \dR^{ n_2 \times 1}, \| u\| \leq \sigma_{\min}(A) } \|u^T B^{\prime}\|  \\
 = \sigma_{\min}(A) \| B^{\prime} \|_2 = \sigma_{\min}(A) \| B \|_2.
 \end{align*}
Combining the above relation and \eqref{AB spectral interm 1}  leads to \eqref{AB spectral lower bound}. $\Box$

{\black
\subsection{Proof of Proposition \ref{prop1}}
Let $M,X,Y$ satisfy the condition \eqref{cond of X,Y, Prop 1}.}
 First, we specify the choice of $U,V$.
Suppose the SVD of $M$ is $M = \hat{U}\Sigma \hat{V} = Q_1 \tilde{\Sigma} Q_2^T$, where $Q_1\in \mathcal{R}^{m\times m},Q_2\in \mathcal{R}^{n\times n}$ are unitary matrices, and $\tilde{\Sigma} =  \left(  \begin{array}{cc}
 \Sigma & 0 \\
 0 & 0 \\
\end{array}
\right) $. Suppose $Q_1 = (Q_{11},Q_{12})$, $Q_2 = (Q_{21},Q_{22})$, where $Q_{11} = \hat{U}\in \mathcal{R}^{m\times r},Q_{21}=\hat{V}\in \mathcal{R}^{n\times r}$ are incoherent matrices, and $Q_{12} \in \dR^{ m \times (m-r) }, Q_{22} \in \dR^{ n \times (n-r) } $.
% Then $\|M-XY^T \|_F = \|\tilde{\Sigma} - (Q_1^TX)(Q_2^TY)^T \|_F.$ Let $X' = Q_1^TX, Y' = Q_2^TY$.
Let us write $X,Y$ as
\begin{equation}\label{X = Q times X prime}
X=Q_1\left( \begin{array}{c} X_1^{\prime} \\ X_2^{\prime} \\ \end{array}\right) ,
\quad Y = Q_2 \left( \begin{array}{c} Y_1^{\prime} \\  Y_2^{\prime} \\ \end{array}\right) ,
\end{equation}
where $  X_1^{\prime},  Y_1^{\prime} \in \dR^{r \times r} , X_2^{\prime} \in \dR^{ (m-r) \times r}, Y_2^{\prime} \in \dR^{ (n-r) \times r}.$
Define
\begin{equation}\label{def of U,V}
U \triangleq Q_1\left( \begin{array}{c} U_1' \\0 \\ \end{array}\right) ,
\quad V \triangleq Q_2 \left( \begin{array}{c} V_1' \\0 \\ \end{array}\right) ,
\end{equation}
where
$$
U_1^{\prime} = (1 - \bar{\eta} ) X_1^{\prime}, \;\;  V_1^{\prime} = \frac{1}{ 1 - \bar{\eta}} \Sigma (X_1^{\prime})^{-T},
$$
in which
$$
 \bar{\eta} \triangleq \frac{d}{\Sigma_{\min}} \leq \frac{1}{11}.
$$
% In most places of the proof, we will use a weaker bound $\bar{\eta} \leq 1/6$.
The definition of $  V_1^{\prime}$ is valid since $X_1^{\prime}$ is invertible (otherwise, $\mathrm{rank}(X_1^{\prime}(Y_1^{\prime})^T) \leq \mathrm{rank}(X_1^{\prime}) \leq r-1$,
  thus $d \geq \|\Sigma - X_1^{\prime}(Y_1^{\prime})^T \|_F \overset{\eqref{|A-B| bound ineq.}}{\geq}
    \Sigma_{\rmin} - \sigma_{\rmin}( X_1^{\prime}(Y_1^{\prime})^T ) = \Sigma_{\rmin} $, which contradicts \eqref{cond a) of X,Y, Prop 1}.% \eqref{cond a) of X,Y}).
    By this definition, we have
  \begin{equation}\label{U1 V1 = Sigma}
     U_1^{\prime}(V_1^{\prime})^T = (1 - \bar{\eta}) X_1^{\prime} ( V_1^{\prime})^T  =  \Sigma .
  \end{equation}

Now, we prove that $U,V$ defined in \eqref{def of U,V} satisfy the requirement \eqref{req of U,V, Prop 1}. %the requirements (a),(b),(c),(d) of Proposition \ref{prop1}.
The requirement \eqref{req a) of U,V, Prop 1} $UV^T = M$ follows from \eqref{U1 V1 = Sigma} and \eqref{def of U,V}.
The requirement \eqref{req b) of U,V, Prop 1} $\| U\|_F \leq (1 - \frac{d}{\Sigma_{\min}} ) \| X\|_F$ can be proved as follows:
 $$ \| U\|_F = \|U_1' \|_F = (1 - \frac{d}{\Sigma_{\min}} ) \|X_1^{\prime} \|_F \leq (1 - \frac{d}{\Sigma_{\min}} ) \| X\|_F.$$
As a side remark, the following variant of the requirement \eqref{req b) of U,V, Prop 1} also holds:
\begin{equation}\label{complement: U spectral norm smaller}
 \| U\|_2 \leq (1 - \frac{d}{\Sigma_{\min}} ) \| X\|_2.
\end{equation}
In fact, $\| U\|_2 = \| U_1^{\prime} \|_2 = (1 - \frac{d}{\Sigma_{\min}} ) \|X_1^{\prime} \|_2
 \overset{\eqref{submatrix has smaller spectral norm}}{\leq} (1 - \frac{d}{\Sigma_{\min}} )
 \left\| \left( \begin{array}{c} X_1^{\prime} \\ X_2^{\prime} \\ \end{array}\right) \right\|_2
= (1 - \frac{d}{\Sigma_{\min}} ) \| X \|_2 $.

To prove the requirement \eqref{req c) of U,V, Prop 1},  % $\|U-X \|_F \leq \frac{2\beta_T}{\Sigma_{\min}}d, \|V-Y \|_F \leq \frac{4\beta_T}{\Sigma_{\min}}d$.
we first provide the bounds on $\|X_2^{\prime} \|_F,\|V_1'-Y_1^{\prime} \|_F,\|Y_2^{\prime} \|_F.$ Note that
\begin{equation}\label{expansion of d^2}
\begin{split}
d^2 = & \|M-XY^T \|_F^2 \\
    = &  \left\|\left(  \begin{array}{cc}\Sigma & 0 \\ 0 & 0 \\\end{array} \right)
             - Q_1^T X Y^T Q_2 \right\|_F^2 \\
    \overset{ \eqref{X = Q times X prime} }{=}   & \left\|\left(  \begin{array}{cc}\Sigma & 0 \\ 0 & 0 \\\end{array} \right)
             - \left(  \begin{array}{cc} X_1^{\prime}(Y_1^{\prime})^T & X_1^{\prime}(Y_2^{\prime})^T \\ X_2^{\prime}(Y_1^{\prime})^T & X_2^{\prime}(Y_2^{\prime})^T \\\end{array} \right) \right\|_F^2 \\
    =   &   \| \Sigma - X_1^{\prime}(Y_1^{\prime})^T \|_F^2 + \|X_1^{\prime}(Y_2^{\prime})^T \|_F^2 + \|X_2^{\prime}(Y_1^{\prime})^T \|_F^2 + \|X_2^{\prime}(Y_2^{\prime})^T \|_F^2. \\
    \overset{ \eqref{U1 V1 = Sigma} }{=}  &   \| X_1^{\prime}( (1 - \bar{\eta} )V_1' - Y_1^{\prime})^T \|_F^2 + \|X_1^{\prime}(Y_2^{\prime})^T \|_F^2 + \|X_2^{\prime}(Y_1^{\prime})^T \|_F^2   \\
 & \quad\quad\quad    + \|X_2^{\prime}(Y_2^{\prime})^T \|_F^2.
\end{split}
\end{equation}
Intuitively, since $\|X_1^{\prime}\|_F,\|Y_1^{\prime}\|_F$ are $O(1)$, we can upper bound $\|(1 - \bar{\eta} )V_1'-Y_1^{\prime} \|_F,\|Y_2^{\prime} \|_F,\|X_2^{\prime} \|_F$ as $O(d)$.
More rigorously, it follows from (\ref{expansion of d^2}) that $d \geq  \|X_1^{\prime}( (1 - \bar{\eta} ) V_1' - Y_1^{\prime})^T \|_F
\overset{\eqref{AB_F lower bound}}{\geq} \sigma_{\rmin}(X_1^{\prime}) \| (1 - \bar{\eta} ) V_1' - Y_1^{\prime} \|_F$ and,
% Here, we use the fact $\|AB \|_F \geq \sigma_{\rmin}(A) \|B \|_F$ for $r \times r$
  %  matrix $A,B$.
similarly, $d \geq \sigma_{\rmin}(X_1^{\prime})\|(Y_2^{\prime})^T \|_F$, $d \geq \sigma_{\rmin}(Y_1^{\prime})\|(X_2^{\prime})^T \|_F.$
These three inequalities imply
\begin{equation}\label{Y2,V1-Y1,X2 lower bound}
\begin{split}
\| (1 - \bar{\eta} ) V_1' - Y_1^{\prime} \|_F \leq \frac{d}{\sigma_{\rmin}(X_1^{\prime})}, \\
 \|Y_2^{\prime} \|_F \leq \frac{d}{\sigma_{\rmin}(X_1^{\prime})}, \;\;  \|X_2^{\prime} \|_F \leq \frac{d}{\sigma_{\rmin}(Y_1^{\prime})}.  %\frac{2\beta_T}{\Sigma_{\min}}d.$
\end{split}
\end{equation}
% It remains to lower bound
We can lower bound $\sigma_{\rmin}(X_1^{\prime})$ and $\sigma_{\rmin}(Y_1^{\prime})$ as
\begin{equation}\label{bound sigma min (X1)}
  \sigma_{\rmin}(X_1^{\prime}) \geq \frac{ 10 \Sigma_{\min}}{11 \beta_T}, \; \; \sigma_{\rmin}(Y_1^{\prime}) \geq \frac{10 \Sigma_{\min}}{ 11 \beta_T}.
\end{equation}
To prove \eqref{bound sigma min (X1)}, notice that (\ref{expansion of d^2}) implies that $d \geq \|\Sigma - X_1^{\prime}(Y_1^{\prime})^T \|_F \geq  \|\Sigma - X_1^{\prime}(Y_1^{\prime})^T \|_2
\overset{\eqref{|A-B| bound ineq.}}{\geq}
\Sigma_{\rmin} - \sigma_{\rmin}(X_1^{\prime}(Y_1^{\prime})^T)$,
% where the last inequliaty follows from Proposition \ref{matrix prop: |A-B| bound}.
 which further implies
 \begin{equation}\nonumber %\label{X_1Y_1 sigma min}
 \sigma_{\rmin}(X_1^{\prime}(Y_1^{\prime})^T) \geq \Sigma_{\rmin} - d \geq \frac{ 10 }{ 11}\Sigma_{\rmin}.
 \end{equation}
 According to Proposition \ref{matrix prop: sigma_min(AB) upper bound}, we have $\sigma_{\rmin}(X_1^{\prime}(Y_1^{\prime})^T) \leq \sigma_{\rmin}(X_1^{\prime}) \|Y_1^{\prime} \|_2$.
Combining this inequality with the above relation, we get  $\sigma_{\rmin}(X_1^{\prime})  \|Y_1^{\prime} \|_2 \geq \sigma_{\rmin}(X_1^{\prime}(Y_1^{\prime})^T) \geq 5\Sigma_{\rmin}/6$, which further implies
\begin{equation}\label{bound sigma min (X1) by spectral}
\sigma_{\rmin}(X_1^{\prime}) \geq \frac{10 \Sigma_{\min}}{ 11 \|Y_1^{\prime} \|_2}.
\end{equation}
Similarly, we have
\begin{equation}\label{bound sigma min (Y1) by spectral}
\sigma_{\rmin}(Y_1^{\prime}) \geq \frac{ 10 \Sigma_{\min}}{ 11 \|X_1^{\prime} \|_2}.
\end{equation}
Plugging $\|Y_1^{\prime} \|_2 \leq \|Y_1^{\prime} \|_F \leq \|Y \|_F \leq \beta_T$ and similarly $\|X_1^{\prime} \|_2 \leq \beta_T$
into \eqref{bound sigma min (X1) by spectral} and \eqref{bound sigma min (Y1) by spectral}, we obtain (\ref{bound sigma min (X1)}).
% $\sigma_{\rmin}(X_1^{\prime}) \geq \frac{4\Sigma_{\min}}{ 5\beta_T } $ and $ \sigma_{\rmin}(Y_1^{\prime}) \geq \frac{4\Sigma_{\min}}{5\beta_T} .$
% Thus (\ref{bound sigma min (X1)}) is proved.
% Now (\ref{U'-X',V'-Y' is upper bounded}) follows by plugging \eqref{bound sigma min (X1)} into (\ref{Y2,V1-Y1,X2 lower bound}).

Combining \eqref{bound sigma min (X1)} and \eqref{Y2,V1-Y1,X2 lower bound}, we obtain
\begin{equation}\label{one bound for three}
 \max \{ \| (1 - \bar{\eta} ) V_1' - Y_1^{\prime} \|_F, \|X_2^{\prime} \|_F, \|Y_2^{\prime} \|_F  \}
 \leq \frac{11}{10} \frac{d}{ \Sigma_{\min} } \beta_T \leq  \frac{1}{10} \beta_T.
\end{equation}

We can bound the norm of $V_1'$ as
\begin{equation}\label{bound V1'}
\begin{split}
  \| V_1'\|_F = \frac{ 1 }{ 1 - \bar{\eta} }   \| (1 - \bar{\eta} )V_1'\|_F \leq
   \frac{ 1 }{ 1 - \bar{\eta} } (   \| (1 - \bar{\eta} )V_1' - Y_1'\|_F +  \| Y_1'\|_F)  \\
    \overset{ \eqref{one bound for three} }{\leq} \frac{11}{10} \left( \frac{1}{10} \beta_T + \beta_T \right)
    \leq \left(\frac{11}{10}\right)^2 \beta_T .
    \end{split}
\end{equation}
Combining this relation with \eqref{one bound for three}, we have
\begin{align*}
 \|V_1'-Y_1^{\prime} \|_F \leq   \| (1 - \bar{\eta} ) V_1'-Y_1^{\prime}\|_F + \bar{\eta} \| V_1'\|_F    \\
 \leq    \frac{11}{10} \frac{d}{ \Sigma_{\min} } \beta_T  +  \bar{\eta} \left(\frac{11}{10}\right)^2 \beta_T
 \leq \frac{ 7 \beta_T}{ 3 \Sigma_{\min}}d .
\end{align*}
From \eqref{one bound for three} and the above relation we obtain
\begin{equation}\nonumber%\label{U'-X',V'-Y' is upper bounded}
\begin{split}
\|U-X \|_F   =  &    \|X_2^{\prime} \|_F \leq \frac{ 11 \beta_T}{ 10 \Sigma_{\min}}d  \leq \frac{6 \beta_T}{5 \Sigma_{\min}}d, \\
\|V-Y \|_F   =  &    \sqrt{ \|V_1'-Y_1^{\prime} \|_F^2 + \|Y_2^{\prime} \|_F^2 }   \\
& \leq \sqrt{ \left(\frac{7}{3}\right)^2 + \left(\frac{11}{10} \right)^2 } \frac{ \beta_T}{ \Sigma_{\min}}d
  \leq \frac{ 3 \beta_T}{  \Sigma_{\min}}d,
\end{split}
\end{equation}
which finishes the proof of the requirement \eqref{req c) of U,V, Prop 1}.
% as mentioned before, proving (\ref{U'-X',V'-Y' is upper bounded}) finishes the proof of the requirement (b).

As a side remark, the requirement \eqref{req c) of U,V, Prop 1} can be slightly improved to
\begin{equation}\label{complement: U'-X',V'-Y' upper bounded by spectral norm}
\begin{split}
\|U-X \|_F  \leq \frac{ 6 \| Y\|_2 }{ 5 \Sigma_{\min}}d, \quad
\|V-Y \|_F \leq \frac{ 3 \|X \|_2}{ \Sigma_{\min}}d.
\end{split}
\end{equation}
In fact, plugging
$ \| X_1^{\prime} \|_2  \overset{\eqref{submatrix has smaller spectral norm}}{\leq} \| \left( \begin{array}{c} X_1^{\prime} \\ X_2^{\prime} \\ \end{array}\right)\|_2
= \| X \|_2 $ and similarly $ \| Y_1^{\prime} \|_2 \leq \| Y \|_2  $ into
\eqref{bound sigma min (X1) by spectral} and \eqref{bound sigma min (Y1) by spectral}, we obtain
$   \sigma_{\rmin}(X_1^{\prime}) \geq \frac{ 5\Sigma_{\min}}{ 6 \| Y \|_2 }, \; \; \sigma_{\rmin}(Y_1^{\prime}) \geq \frac{ 5 \Sigma_{\min}}{ 6 \| X\|_2}. $
Combining with (\ref{Y2,V1-Y1,X2 lower bound}), we obtain \eqref{complement: U'-X',V'-Y' upper bounded by spectral norm}.
This inequality will be used in the proof of Claim \ref{claim: initial point properties} in Appendix \ref{appen: initialization proof}.

At last, we prove the requirement \eqref{req d) of U,V, Prop 1}.
%\begin{equation}\label{boudn on U^{i},V^{j}}
%\| U^{(i)}\|^2 \leq \frac{3r\mu }{2m}\beta_T^2, \| V^{(j)}\|^2 \leq \frac{3r\mu }{2n}\beta_T^2.
%\end{equation}
By the definitions of $U,V$ in \eqref{def of U,V}, we have
\begin{equation}
\begin{split}
 U = (Q_{11},Q_{12}) \left(\begin{array}{cc} U_1^{\prime} \\ 0  \end{array} \right) = Q_{11} U_1^{\prime} , \\
   V  = (Q_{21},Q_{22}) \left(\begin{array}{cc} V_1' \\ 0  \end{array} \right) =  Q_{21} V_1' .
\end{split}
 \end{equation}
The assumption that $M$ is $\mu$-incoherent implies
$$
   \| Q_{11}^{(i)} \|^2 = \| \hat{U}^{(i)}  \|^2  \leq \frac{r\mu }{m}, \;\; \| Q_{21}^{(i)} \|^2 = \| \hat{V}^{(j)}  \|^2  \leq \frac{r \mu }{n}, \;\; \forall \ i,j.
$$
 Notice the following fact: for any matrix $A \in \dR^{ K \times r}, B \in \dR^{ r \times r} $, where $K \in \{ m, n\}$, we have
\begin{equation}\nonumber % \label{row of product estimation}
\| (AB)^{(i)} \|^2 = \| A^{(i)} B\|^2  \leq \|A^{(i)} \|^2 \|B \|_F^2.
\end{equation}
 Therefore, we have (using the fact $\|U_1^{\prime} \|_F \leq \| X_1^{\prime} \|_F \leq \| X\|_F \leq \beta_T$ and \eqref{bound V1'})
\begin{equation}\label{incoherent of U,V}
\begin{split}
\|U^{(i)} \|^2 = \|(Q_{11} U_1^{\prime})^{(i)}\|^2 &
\leq \|Q_{11}^{(i)}\|^2 \| U_1^{\prime} \|_F^2 \leq \frac{r\mu }{m} \beta_T^2; \\
 \|V^{(j)} \|^2 = \|(Q_{21} V_1')^{(j)} \|^2 \leq \frac{r\mu }{n} \|V_1'\|_F^2 &
 \overset{ \eqref{bound V1'}}{\leq} \left(\frac{11}{10}\right)^4 \frac{r\mu }{ n} \beta_T^2 \leq  \frac{3}{2} \frac{r\mu }{ n } \beta_T^2,
\end{split}
 \end{equation}
which finishes the proof the requirement \eqref{req d) of U,V, Prop 1}.
\section{Proof of Proposition \ref{prop2}}\label{section of proof of Prop 2}

%%---key result 2: prop2-------------------------
%\begin{prop}\label{prop2}
%There exists numerical constant $C_d,C_T$ such that: if
%\newline  (1) $M$ is $(\mu ,\mu_1)$ incoherent;
%\newline (2) $\|M-XY^T \|_F = d < \frac{\Sigma_{\min}}{C_d r^2 \sqrt{r} \kappa^2};$
%\newline (3)$\sqrt{\frac{2}{3}}\beta_T \leq \| X\|_F  \leq \beta_T, \sqrt{\frac{2}{3}}\beta_T \leq \| Y\|_F \leq \beta_T$, where $\beta_T = \sqrt{C_T r\Sigma_{\rmax}}$;
%\newline then there exists $U \in \mathbb{R}^{m\times r}, V\in \mathbb{R}^{n \times r}$ such that
%\newline (a) $UV^T = M$;
%\newline (b) $\| U-X\|_F \leq 6C_T \frac{\beta}{\Sigma_{\min}} r \sqrt{r} \kappa d , \| V-Y\|_F \leq 6C_T^2 \frac{\beta}{\Sigma_{\min}} r^2 \sqrt{r} \kappa^2 d ;$
%\newline (c) $\| U^{(i)}\|^2 \leq \frac{r\mu }{m}\beta_T^2, \| V^{(j)}\|^2 \leq \frac{r\mu }{n}\beta_T^2 $;
%\newline (d) $\|U \|_F \leq \|X \|_F$, $\|V \|_F \leq \|Y \|_F$.
%\end{prop}

% \subsection{Preliminary Analysis}

%We need to find $U,V$ such that
%\newline (a) $UV^T = M$;
%\newline (b) $ \| U-X\|_F \| V-Y\|_F   \leq  25 \sqrt{r} \frac{\beta_T^2}{\Sigma_{\rmin}^2}  d^2. $
%%% $\| U-X\|_F \leq 11 \frac{\beta_T}{\Sigma_{\min}}  \sqrt{r} \kappa d , \| V-Y\|_F \leq 11 \frac{\beta_T}{\Sigma_{\min}} r  \kappa d ;$
%\newline (c) $\| U^{(i)}\|^2 \leq \frac{r\mu }{m}\beta_T^2, \| V^{(j)}\|^2 \leq \frac{r\mu }{n}\beta_T^2 $;
%\newline (d) $\|U \|_F \leq \|X \|_F$, $\|V \|_F \leq \|Y \|_F$.

{\black

We will first reduce Proposition  \ref{prop2} to Proposition \ref{prop2'} for $r \times r$ matrices in Section \ref{appen: Prop 2, tranform}.
This reduction is rather trivial, and the major difficulty lies in Proposition \ref{prop2'}.
% When $r = 2$, the proposition is not hard but requires some analysis.
For general $r$, the proof of Proposition \ref{prop2'} is rather involved.
We will give the overview of the main proof ideas in Section \ref{appen: analysis of prop2', simplfied}.
 Most readers can skip Section \ref{appen: Prop 2, tranform}.
% Proposition \ref{prop2'} is only meaningful for $r \geq 2$ since when $r=1$ the conditions cannot hold.

\subsection{Transformation to a simpler problem}\label{appen: Prop 2, tranform}
% We first show that to prove Proposition \ref{prop2} we only need to prove a simpler Proposition \ref{prop2'}.
We first transform the problem to a simpler problem that only involves $ r \times r$ matrices.
In particular, we will show that to prove Proposition \ref{prop2} we only need to prove Proposition \ref{prop2'}.  }

Similar to the proof of Proposition \ref{prop1}, we use $Q_1 \in \dR^{m \times m}, Q_2 \in \dR^{n \times n} $ to denote the SVD factors of $M$ ($Q_1$
and $Q_2$ are unitary matrices),
and write $X,Y$ as %$X' = \left( \begin{array}{c} X_1^{\prime} \\ X_2^{\prime} \\ \end{array}\right) = \left( \begin{array}{c} Q_{11}^TX \\ Q_{12}^TX \\ \end{array}\right)$, $Y' = \left( \begin{array}{c} Y_1^{\prime} \\ Y_2^{\prime} \\ \end{array}\right) = \left( \begin{array}{c} Q_{21}^TY \\ Q_{22}^TY\\ \end{array}\right)$.
\begin{equation}\nonumber
X=Q_1\left( \begin{array}{c} X_1^{\prime} \\ X_2^{\prime} \\ \end{array}\right) ,
\quad Y = Q_2 \left( \begin{array}{c} Y_1^{\prime} \\  Y_2^{\prime} \\ \end{array}\right) .
\end{equation}
Define
\begin{equation}\label{U,V def in Prop 2}
U=Q_1\left( \begin{array}{c} U_1' \\0 \\ \end{array}\right) ,
\quad V = Q_2 \left( \begin{array}{c} V_1' \\0 \\ \end{array}\right) ,
\end{equation}
 where $U_1^{\prime} \in \dR^{r \times r} $ and $V_1^{\prime} \in \dR^{r \times r}$ are to be determined.

We can convert the conditions on $U,V$ to the conditions on $U_1^{\prime} , V_1^{\prime}$.
As proved in Appendix \ref{appen: proof of Prop 1} (combining \eqref{Y2,V1-Y1,X2 lower bound} and \eqref{bound sigma min (X1)}),
\begin{equation}\label{bound on X2' and Y2'}
\|X_2^{\prime} \|_F \leq \frac{6\beta_T}{5\Sigma_{\min}}d, \;\; \|Y_2^{\prime} \|_F \leq \frac{6\beta_T}{5\Sigma_{\min}}d.
\end{equation}
Obviously, the condition \eqref{cond a) of X,Y, Prop 2} implies the following condition on $X_1^{\prime},Y_1^{\prime}$:
\begin{equation}\label{cond a) of X,Y}
 d^{\prime} \triangleq \| \Sigma - (X_1^{\prime})(Y_1^{\prime})^T \| \leq \frac{ \Sigma_{\rmin}}{ C_d r }.
\end{equation}
Using (\ref{bound on X2' and Y2'}) and the facts $\| X\|_F = \sqrt{ \| X_1^{\prime}\|_F^2 + \| X_2^{\prime}\|_F^2}$ and
 $\| Y \|_F = \sqrt{ \| Y_1^{\prime}\|_F^2 + \| Y_2^{\prime}\|_F^2} $, the condition \eqref{cond b) of X,Y, Prop 2} implies the following condition on $X_1^{\prime},Y_1^{\prime}$:
\begin{equation}\label{cond b) of X,Y}
\sqrt{\frac{3}{5}}\beta_T \leq \| X_1^{\prime} \|_F  \leq \beta_T, \; \; \sqrt{\frac{3}{5}}\beta_T \leq \| Y_1^{\prime} \|_F \leq \beta_T.
\end{equation}

We have the following proposition.
%The following proposition shows that under the above two conditions on $X_1^{\prime}, Y_1^{\prime} $, there exist $U_1^{\prime}, V_1^{\prime}$ that satisfy certain properties. The proof of this proposition will be
% given in Appendix \ref{appen: proof of prop 2'}.
%------------------prop 2'----------------------------
\begin{prop}\label{prop2'}
% Suppose $\Sigma = \mathrm{Diag}\{\Sigma_1,\dots, \Sigma_r \},$ where $\Sigma_i \in [\Sigma_{\rmin}, \Sigma_{\rmax}],i=1,\dots,r.$
There exist numerical constants $C_d,C_T$ such that: if $X_1^{\prime}, Y_1^{\prime} \in \dR^{r \times r} $ satisfy
\eqref{cond a) of X,Y} and \eqref{cond b) of X,Y},
%\begin{subequations}\label{cond of X,Y}
%\begin{align}
%   d^{\prime} \triangleq \| \Sigma - (X_1^{\prime})(Y_1^{\prime})^T \|  & \leq \frac{ \Sigma_{\rmin}}{ C_d r }, \label{cond a) of X,Y} \\
%\sqrt{\frac{3}{5}}\beta_T \leq \| X_1^{\prime} \|_F  \leq \beta_T, &  \; \; \sqrt{\frac{3}{5}}\beta_T \leq \| Y_1^{\prime} \|_F \leq \beta_T, \label{cond b) of X,Y}
%\end{align}
%\end{subequations}
where $\beta_T = \sqrt{C_T r\Sigma_{\rmax}}$, then there exist $U_1^{\prime} \in \mathbb{R}^{r \times r}, V_1^{\prime} \in \mathbb{R}^{ r \times r}$ such that
\begin{subequations}\label{req of U,V}
\begin{align}
  U_1^{\prime} (V_1^{\prime})^T  & = \Sigma,    \label{req a) of U,V} \\
  \|U_1^{\prime} \|_F \leq \|X_1^{\prime} \|_F, \;   \; \|V_1^{\prime} \|_F & \leq (1 - \frac{d}{\Sigma_{\min} } )\|Y_1^{\prime} \|_F,  \label{req b) of U,V} \\
 \| U_1^{\prime} - X_1^{\prime} \|_F \| V_1^{\prime} - Y_1^{\prime} \|_F
 & \leq   63  \sqrt{r} \frac{\beta_T^2}{\Sigma_{\rmin}^2}  d^2 ,  \nonumber \\
 \max \{ \| U_1^{\prime} - X_1^{\prime} \|_F,\| V_1^{\prime} - Y_1^{\prime} \|_F \}   &  \leq  \frac{58 }{7}  \sqrt{r} \frac{\beta_T}{\Sigma_{\min}}   d .
 \label{req c) of U,V}
\end{align}
\end{subequations}
  % without linear convergence, 63--> 23;  58/7 --> 36/7
\end{prop}

We claim that Proposition \ref{prop2'} implies Proposition \ref{prop2}.
Since we have already proved that the conditions of Proposition \ref{prop2} imply the conditions of Proposition \ref{prop2'},
we only need to prove that the conclusion of Proposition \ref{prop2'} implies the conclusion of Proposition \ref{prop2}.
In other words, we only need to show that if $U_1^{\prime}, V_1^{\prime}$ satisfy \eqref{req of U,V},
then they satisfy the requirements \eqref{req of U,V, Prop 2}.

The requirement \eqref{req a) of U,V, Prop 2} $UV^T = M$ follows directly from \eqref{req a) of U,V} and the definition of $U,V$ in \eqref{U,V def in Prop 2}.
The requirement \eqref{req b) of U,V, Prop 2} can be proved as
$\| V\|_F = \|V_1'\|_F \leq  (1 - \frac{d}{\Sigma_{\min} } ) \|Y_1^{\prime}\|_F \leq (1 - \frac{d}{\Sigma_{\min} } ) \|Y \|_F$
and $\|U \|_F = \|U_1'\|_F \leq \|X \|_F$.
Analogous to \eqref{incoherent of U,V}, the requirement \eqref{req d) of U,V, Prop 2} can be proved as
 $ \|V^{(j)} \|^2 = \|(Q_{21} V_1')^{(j)} \|^2 \leq \frac{r\mu }{n} \|V_1'\|_F^2 \leq \frac{r\mu }{n}\beta_T^2 $
and, similarly, $\| U^{(i)}\|^2 \leq \frac{r\mu }{m}\beta_T^2.$
At last, we prove the requirement \eqref{req c) of U,V, Prop 2}. The first relation in \eqref{req c) of U,V, Prop 2} can be proved as
% is a consequence of
%\begin{equation}\label{transform U-X to U1'-X1'}
% \| U_1^{\prime} - X_1^{\prime} \|_F \| V_1^{\prime} - Y_1^{\prime} \|_F  \leq  23 \sqrt{r} \frac{\beta_T^2}{\Sigma_{\rmin}^2}  (d^{\prime})^2 ,    \quad
% \max \{ \| U_1^{\prime} - X_1^{\prime} \|_F,\| V_1^{\prime} - Y_1^{\prime} \|_F \} \leq \frac{36}{7} \sqrt{r} \frac{\beta_T}{\Sigma_{\min}} d^{\prime} .
%\end{equation}
%In fact, if \eqref{transform U-X to U1'-X1'} holds, then
\begin{equation}\nonumber
\begin{split}
& \| U-X\|_F \| V-Y\|_F  \\
&  = \sqrt{\| U_1' - X_1^{\prime}\|_F^2 + \|X_2^{\prime} \|_F^2} \sqrt{\| V_1' - Y_1^{\prime}\|_F^2 + \|Y_2^{\prime} \|_F^2}    \\
& = \left( \| \right. U_1' - X_1^{\prime}\|_F^2 \| V_1' - Y_1^{\prime}\|_F^2  + \|X_2^{\prime} \|_F^2 \| V_1' - Y_1^{\prime}\|_F^2    \\
 & \quad \quad  + \| U_1' - X_1^{\prime}\|_F^2 \|Y_2^{\prime} \|_F^2  + \|X_2^{\prime} \|_F^2 \|Y_2^{\prime} \left. \|_F^2 \right)^{1/2} \\
& \overset{\eqref{bound on X2' and Y2'}, \eqref{req c) of U,V}}{\leq} \sqrt{r} \frac{\beta_T^2}{\Sigma_{\rmin}^2}  d^2 \sqrt{ 63^2 + (\frac{6}{5})^2 (\frac{58 }{7})^2 + ( \frac{58 }{7})^2 (\frac{6}{5})^2 +  (\frac{6}{5})^4  },  \\
& < 65 \sqrt{r} \frac{\beta_T^2}{\Sigma_{\rmin}^2}  d^2,
\end{split}
\end{equation}
where in the second last inequality we also use the fact $d^{\prime} \leq d$.
The second relation in \eqref{req c) of U,V, Prop 2} can be proved by
\begin{align*}
 \| U- X \|_F = \sqrt{ \| U_1' - X_1^{\prime}\|_F^2 + \|X_2^{\prime} \|_F^2  }   \\
  \overset{\eqref{bound on X2' and Y2'}, \eqref{req c) of U,V}}{\leq}  \sqrt{ (\frac{6}{5})^2 + (\frac{58}{7})^2 }  \sqrt{r} \frac{\beta_T}{\Sigma_{\min}}   d
  \leq  \frac{17 }{ 2} \sqrt{r} \frac{\beta_T}{\Sigma_{\min}}   d
\end{align*}
and a similar inequality for $\| V - Y\|_F$.

\subsection{Preliminary analysis for the proof of Proposition \ref{prop2'}}\label{appen: analysis of prop2', simplfied}

%----------------Simplified statement of Prop 2'--------------------------
% and we will prove the following proposition that is essentially equivalent to Proposition \eqref{prop2'}
% in Appendix \ref{appen: proof of prop 2'}.
%\begin{prop}[Restatement of Proposition \eqref{prop2'}]\label{prop2', simplifed notation}
%Suppose $\Sigma = \mathrm{Diag}\{\Sigma_1,\dots, \Sigma_r \} \in \dR^{r \times r} $, where $\Sigma_i \in [\Sigma_{\rmin}, \Sigma_{\rmax}],i=1,\dots,r. $
%There exists numerical constant $C_d,C_T$ such that the following holds. If $X,Y \in \dR^{r \times r}$ satisfy
%\begin{subequations}\label{cond of X,Y}
%\begin{align}
%d \triangleq  \|\Sigma - XY^T \|_F   & \leq \frac{\Sigma_{\min}}{C_d r }, \label{cond a) of X,Y} \\
%\sqrt{\frac{3}{5}}\beta_T \leq \| X\|_F  \leq \beta_T,  &  \;\; \sqrt{\frac{3}{5}}\beta_T \leq \| Y\|_F \leq \beta_T, \label{cond b) of X,Y}
%\end{align}
%\end{subequations}
%where $\beta_T = \sqrt{C_T r\Sigma_{\rmax}}$, then there exists $U \in \mathbb{R}^{m\times r}, V\in \mathbb{R}^{n \times r}$ such that
%\begin{subequations}\label{req of U,V}
%\begin{align}
%  UV^T  & = \Sigma,    \label{req a) of U,V} \\
%  \|U \|_F \leq \|X \|_F, \; &  \; \|V \|_F \leq \|Y \|_F,  \label{req b) of U,V} \\
% \| U-X\|_F \| V-Y\|_F  \leq  23 \sqrt{r} \frac{\beta_T^2}{\Sigma_{\rmin}^2}  d^2 ,   & \quad
% \max \{ \| U-X\|_F,\| V-Y\|_F \} \leq \frac{36}{7} \sqrt{r} \frac{\beta_T}{\Sigma_{\min}}   d .
% \label{req c) of U,V}
%\end{align}
%\end{subequations}
%\end{prop}
%------------------------------------------
We first give a more intuitive explanation of what we want to prove, by relating the result to ``preconditioning''.
{\black Then we analyze two simple examples for $r=2$ to get some ideas on how to approach the problem.
Next we discuss how to extend the ideas to general $r$.
To simplify the notations, from now on, we use $X,Y,U,V ,d $ to replace $X_1^{\prime},Y_1^{\prime}, U_1^{\prime}, V_1^{\prime}, d^{\prime} $
in Proposition \eqref{prop2'}.

\subsubsection{Perturbation Analysis for Preconditioning.}\label{appen: precond relation}

We claim that Proposition \ref{prop2'} is closely related to ``preconditioning'', which refers to reducing the condition number (by preprocessing) in numerical linear algebra.
\begin{prop}(Informal)\label{prop2' simple, precond}
 Suppose $X \in \dR^{r \times r}$ is non-singular and $\|X \|_F = \|X^{-1} \|_F \geq C \sqrt{r} $ where $C \geq 10$ is a constant.
 For any $d' \leq \mathcal{O}(1/r^{1.5})$,
  there exists $U \in \dR^{r \times r}$ such that $ \| U \|_F = \|X \|_F $, $\| U^{-1} \|_F \leq (1 - d') \| X^{-1} \|_F$
 and $\max\{ \| U - X \|_F, \|U^{-1} - X^{-1} \|_F \} \leq \mathcal{O}(d' r^{1.5}) $.
\end{prop}
We will argue later that Proposition \ref{prop2' simple, precond} is a simple version of Proposition \ref{prop2'}.

We explain why this proposition can be understood as perturbation analysis for perconditioning.
Assume $X$ has singular values $\sigma_1 \geq \dots \geq \sigma_r > 0$, then
$\| X\|_F^2 = \sum_i \sigma_i^2$ and $\| X^{-1}\|_F^2 = \sum_i \frac{1}{\sigma_i^2}$.
By Cauchy-Schwartz inequality $\| X\|_F^2 \| X^{-1}\|_F^2 \geq r^2 $, and the equality holds iff $\sigma_1 = \dots = \sigma_r$, i.e.,
$X$ has a  condition number $1$.
In other words, if $\| X\|_F = \| X^{-1}\|_F = \sqrt{r}$, then $X$ has the minimal condition number $1$.
In the assumption $\|X \|_F = \|X^{-1} \|_F \geq C \sqrt{r} $, % where $C \geq 10$ means that $X$ is not well-conditioned,
 $C$ can be viewed as a measure of the ill-conditioned-ness of $X$ (different from the condition number $\sigma_1/\sigma_r$ but related).
Prop. \ref{prop2' simple, precond} simply says that we can perturb $X$ to make $X$ better-conditioned.

Prop. \ref{prop2' simple, precond} itself is not difficult to prove. In fact, without loss of generality we can assume
$X$ is a diagonal matrix (by left and right multiplying $X$ by its singular vector matrices).
Then the problem reduces to the following problem: assume $ \sum_i \sigma_i^2 = \sum_i \frac{1}{\sigma_i^2} \geq C^2 r $,
perturb $\sigma_i$'s so that the $ \sum_i \sigma_i^2$ does not change while $  \sum_i \frac{1}{\sigma_i^2}$ increases.
This is a rather easy problem.
Nevertheless, for the original desired result Prop. \ref{prop2'} we cannot assume $X$ is diagonal.
In Section \ref{appen: 2 examples} we will analyze the problem without assuming $X$ is diagonal.

To show the connection of Prop. \ref{prop2' simple, precond} and Prop. \ref{prop2'}, we first simplify the statement of Prop. \ref{prop2'}.
\begin{prop}(Simpler version of Proposition \ref{prop2'})\label{prop2' simple}
 Suppose $X,Y , \Sigma \in \dR^{r \times r}$ are non-singular and $\Sigma$ is diagonal.
 If $\| XY^T - \Sigma \|_F = d \leq \mathcal{O}(\Sigma_{\min}/r) $  and $ \|X \|_F = \| Y\|_F = \beta \geq C \sqrt{r \Sigma_{\max}} $, % \in [0.8 \beta_T, \beta_T]$ where $\beta_T   $,
 then we can find a factorization $\Sigma = UV^T$ such that
 $ \max \{ \| U-X\|_F, \| V - Y\|_F \}  \leq \mathcal{O}(\sqrt{r} d \beta / \Sigma_{\min} )$ and
 % $(U,V)$ is close to $(X,Y)$ and
  $\| U \|_F \leq \|X \|_F, \|V \|_F \leq \| Y\|_F $.
\end{prop}
There are a few differences with Prop. \ref{prop2'}:
i) In Prop. \ref{prop2'} we assume $\| X\|_F, \| Y\|_F \in [\sqrt{0.6} \beta_T, \beta_T]$,
but by simply scaling $X,U,Y,V$ we can assume $\| X\|_F = \| Y\|_F $ as in the above proposition;
 ii) here we only require $ \| V\|_F \leq \| Y\|_F $, instead of $\|V \|_F \leq (1 - d/\Sigma_{\min})\| Y\|_F$ in \eqref{req b) of U,V};
 iii) in Prop. \ref{prop2'} there is an extra bound of $\| U-X\|_F \| V - Y\|_F$. % has a lower dependency on $r$ in the exponent.
%require $(U,V)$  close to $(X,Y)$, meaning $\|U-X\|_F$ and $\| V-Y\|_F$ are small, but \eqref{req c) of U,V} has slightly different requirements.
Nevertheless, these differences are not essential and do not affect the proof too much.

Now let us consider a special case and show how to reduce Prop. \ref{prop2' simple} to Prop. \ref{prop2' simple, precond}.
This part is mainly for the purpose of rigorous derivation and we suggest first-time readers jump to Section \ref{appen: 2 examples}.
The special case we consider is $ \Sigma = I $ and $XY^T = (1 - d/\sqrt{r}) I$, where $d \leq \mathcal{O}(1/r)$.
Let  $d' = d/\sqrt{r} \leq \mathcal{O}(1/r^{1.5})$, then  $Y^T = (1 - d/\sqrt{r}) X^{-1} = (1 - d') X^{-1} $.
The condition of Prop. \ref{prop2' simple} becomes
\begin{equation}\label{condition new}
 \| X\|_F = \| X^{-1}\|_F (1 - d') = \beta  .
 \end{equation}
One requirement of Prop. \ref{prop2' simple} becomes $\| U\|_F \leq \| X\|_F, \| U^{-1} \|_F \leq \| Y \|_F
= \| X^{-1}\|_F (1 - d')  $.
The distance bound in Prop. \ref{prop2' simple} is $ \mathcal{O}(\sqrt{r} d \beta / \Sigma_{\min} ) $, which becomes
$ \mathcal{O}(d' r^{1.5})$ under the new parameter setting.
By a similar scaling technique, i.e. scaling $X,U$ by $ 1/\sqrt{1 - d' } $ and $Y,V$ by $\sqrt{1 - d'}$,
 we can replace the condition \eqref{condition new} by
 $$ \| X\|_F = \| Y\|_F = \beta \sqrt{\frac{1}{1 - d'}}  \geq C \sqrt{r}. $$
 Note that rigorously speaking the bound should be $ C \sqrt{r}/\sqrt{1 - d'} , $
 but since $ 1/(1-d') \leq 1/(1- 1/r^{1.5}) \in [1/(1- 1/2^{1.5}), 1]  $, the contribution of $1 /\sqrt{1 - d'}$ is just a numerical constant
 which can be absorbed into $C$.
Now the problem becomes:
assume $\| X\|_F = \| X^{-1} \|_F  \geq C \sqrt{r}$,
find $U$ such that $ \| U\|_F \leq \|X \|_F $, $\| U^{-1} \|_F \leq \| X^{-1}\|_F (1 - d') $ and $\max\{ \| U - X \|_F, \|U^{-1} - X^{-1} \|_F \} \leq \mathcal{O}(d' r^{1.5}) $, where $d' \leq \mathcal{O}(1/r^{1.5})$.
By slightly strengthening the requirement $ \| U\|_F \leq \|X \|_F $ to $ \| U\|_F = \|X \|_F $, we obtain Prop. \ref{prop2' simple, precond}.

\subsubsection{Two Motivating Examples}\label{appen: 2 examples}
We denote the $i$-th row of $X,Y$ as $x_i, y_i$, respectively.
In the first example (see Figure \ref{FigDiagonal}), we set $r = 2$, $\Sigma = I$ (which implies $\Sigma_{\rmin} = \Sigma_{\rmax} = 1$), $d = 1/(C_d r)$ and
\begin{equation}\label{X,Y example}
\begin{split}
X = \mathrm{Diag}\left( x_{11}, x_{22} \right) = \mathrm{Diag}\left( C , \frac{1 - d/\sqrt{2} }{ C } \right), \quad  \\
   Y = \mathrm{Diag}\left( y_{11}, y_{22} \right) = \mathrm{Diag}\left( \frac{1 - d/\sqrt{2}}{ C } , C  \right),
   \end{split}
\end{equation}
where $C > 1$ is to be determined, and $ \mathrm{Diag}( w_1, w_2 )  $ denotes a $2 \times 2$ diagonal matrix with diagonal entries $w_1, w_2$.
In this setting $\beta_T =\sqrt{r C_T \Sigma_{\rmax}} =  \sqrt{2 C_T}$ is a large constant.
Condition \eqref{cond a) of X,Y} holds since $\| XY^T - \Sigma \|_F = \|( 1 - d/\sqrt{2})I  - I \|_F = d =  1/(C_d r)$.
%\mathrm{Diag}\left( 1 - d/\sqrt{2}, 1 - d/\sqrt{2} \right)
Note that $ \| X \|_F = \| Y \|_F = \sqrt{ C^2 +  \frac{ (1 - d/\sqrt{2})^2}{ C^2 } } \approx C $, thus there exists
$C \in [\sqrt{3/5}\beta_T , \beta_T]$ so that \eqref{cond b) of X,Y} holds.

  \begin{figure}[ht]
  \centering
{ \includegraphics[width=4.5cm,height=3.5cm]{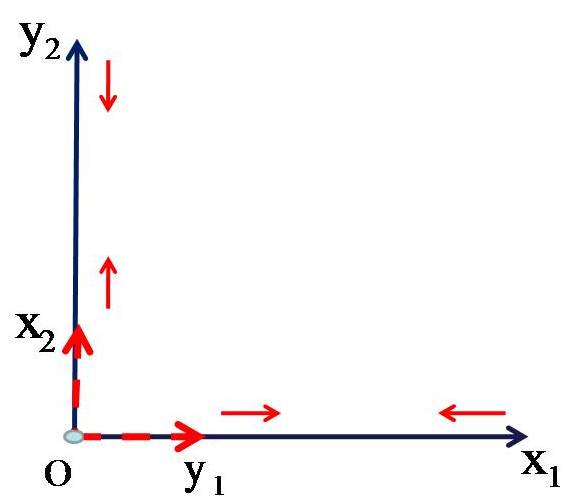} }
  \vspace{-.2cm}
   \caption{ {\small Illustration of the first example. $X = (x_1^T, x_2^T) = \mathrm{Diag}\left( x_{11}, x_{22} \right)$,
   $Y = (y_1^T, y_2^T) = \mathrm{Diag}\left( y_{11}, y_{22} \right)$, where $ x_{11} = y_{22} \gg x_{22} = y_{11} $ and
   $ x_{11}y_{11} = x_{22}y_{22} = 1 - d/\sqrt{2} $.
  We use the following operation to define $U,V$: shrink $x_1$ and extend $x_2$ to obtain $U$, while keeping the norm invariant (i.e.\ $\| U\|_F = \| X\|_F$); shrink $y_2$ and extend $y_1$ to obtain $V$, while keeping the norm invariant (i.e.\ $\|V\|_F = \| Y\|_F$). We can prove that there exists an operation such that $u_{ii}v_{ii} = 1 > x_{ii}y_{ii}, i=1,2$.  }
   }  \label{FigDiagonal} \vspace{-.3cm}
\end{figure} %   $\angle (x_i, \overrightarrow{H_i K_i} ) $ is less than $\frac{\pi}{2}$.

How should we define $U = \mathrm{Diag}(u_{11}, u_{22})  , V = \mathrm{Diag}(v_{11}, v_{22})$ so that \eqref{req of U,V} holds?
% The naive strategy $U = X, V = X^{-1}$ (similar to the proof of Proposition \ref{prop1}) no longer works since
% $ v_{ii} = 1/x_{ii} > y_{ii}/(1-d) > y_{ii}, i=1,2$, implying $\| V\|_F > \| Y \|_F$.
%The difficulty lies in the following ``contradiction'': \eqref{req a) of U,V} says $UV^T = I $, which is ``larger than''
%$XY^T = I - \mathrm{Diag}(d, d)/\sqrt{2}$,
%but \eqref{req b) of U,V} says that $U, V$ should be ``smaller than'' $X,Y$.
% To resolve this difficulty, we need to notice that $\|U \|_F \leq \| X\|_F$ does not mean each entry of $U$ is smaller than
% $X$; it is possible that $\| U\|_F \leq \| X\|_F, \| V\|_F \leq \|Y \|_F$ while $UV^T > XY^T $, if the entries of $U,V$
% are chosen carefullly. Below we explain how this is achieved.
Due to the ``symmetry'' of $X$ and $Y$ in this example (by symmetry we mean $x_{11} = y_{22}, x_{22} = y_{11}$), we choose $U,V$ such that
$ u_{11} = v_{22}, u_{22} = v_{11} $. Then the requirements \eqref{req a) of U,V} and \eqref{req b) of U,V} reduce to:
\begin{equation}\label{simple req}
\begin{split}
  u_{11} u_{22} = 1 = \frac{ x_{11} x_{22} }{1 - d/\sqrt{2}},  \\
  u_{11}^2 +  u_{22}^2 \leq x_{11}^2 +  x_{22}^2.
\end{split}
\end{equation}
It can be easily shown that there exist $u_{11}, u_{22}$ satisfying \eqref{simple req}.
In fact, define $R = \| X\|_F = \sqrt{x_{11}^2 +  x_{22}^2}$ and let a point $(w_1,w_2)$ move along the circle $\{ (w_1, w_2) \mid w_1^2 + w_2^2 = R^2  \}$ from $(x_{11}, x_{22})$ to $ (R/\sqrt{2}, R/\sqrt{2}) $. During this process, the norm of $(w_1, w_2)$ does not change and the product $w_1 w_2$ monotonically increases from $x_{11} x_{22}$ to $R^2/2$.
Therefore, there exist $u_{11}, u_{22}$ satisfying \eqref{simple req} as long as $R^2/2 >  x_{11} x_{22} /(1 - d/\sqrt{2}) $.
% The last inequality
This inequality is equivalent to $(1 - d/\sqrt{2})(x_{11}^2 + x_{22}^2 )/2 > x_{11} x_{22} $, which can be simplified to
 $(1-d/\sqrt{2})(x_{11} - x_{22})^2 > \sqrt{2}d x_{11} x_{22} = \sqrt{2}d(1-d/\sqrt{2})$, or equivalently, $(x_{11} - x_{22})^2  > \sqrt{2} d  $.
 The last inequality holds when $x_{11} - x_{22} = C - (1 - d/\sqrt{2})/C $ is large enough (i.e.\ $C$ is large enough).

To summarize, we will increase the small entry $x_{22}$ (resp.$\ y_{11}$) and decrease the large entry $x_{11}$ (resp.$\ y_{22}$) to obtain a more balanced diagonal matrix $U$ (resp.$\ V$), which has the same norm as $X$ (resp.$\ Y $).
The percentage of increase in the small entry $x_{22}$ (resp.$\ y_{11}$) will be much larger than the percentage of decrease in the large entry $x_{11}$ (resp.$\ y_{22}$), thus the products $x_{22} y_{22} $ and $x_{11}y_{11}$ will increase; in other words, the product $UV^T$ of the more balanced matrices $U,V$ will have larger entries than $XY^T$.
 % the product $UV^T$ is larger than $XY^T$.

% The above idea can be generalized to deal with the case that $X$ and $Y$ are diagonal matrices.
 Note that the above idea of shrinking/extending works when there is a large imbalance in the lengths of the rows of $X,Y$, regardless of
 whether $X,Y$ are diagonal matrices or not.
By the assumption that $\| X\|_F $ and $\| Y\|_F$ are large, we know that there must be a row of $X$ (resp.$\ Y$) that has large norm (here ``large'' means
much larger than $1/\sqrt{r}$); however, it is possible that all rows of $X$ and $Y$ have large norm and there is no imbalance in terms
of the lengths of the rows. See below for such an example.

%or an even general case that the angle between the corresponding rows of $X$ and $Y$ are small (see Appendix \ref{case 1 proof, small angle}
% for a proof of the case of ``small angles'').

 In the second example (see Figure \ref{FigBigAngle}), we still set $r = 2$, $\Sigma = I$, $d = 1/(C_d r)$.
 Suppose $X = (x_1^T, x_2^T) $,   $Y = (y_1^T, y_2^T) $. We define $x_1 = (C,0), x_2 = (-C \sin \alpha, C \cos \alpha )$ and
 $y_1 = (C \cos \alpha, C \sin \alpha ), y_2 = (0, C)$,
 where $C$ is a large constant, and $\alpha \in (0, \pi/2)$ is chosen so that
 \begin{equation}\label{C square cos alpha}
   C^2 \cos \alpha = 1 - d/\sqrt{2}.
 \end{equation}
 When $C$ is large, $\alpha \approx \arccos( 1/C^2 )$ is also large (i.e.\ close to $\pi/2$).
Condition \eqref{cond a) of X,Y} holds since $\| XY^T - \Sigma \|_F = \| C^2 \cos \alpha I - I \|_F =  \|( 1 - d/\sqrt{2})I - I \|_F = d  = 1/(C_d r)$.
Note that $ \| X \|_F = \| Y \|_F = \sqrt{2} C $, so we can choose
$C =\beta_T/\sqrt{2} = \sqrt{ 2C_T}/\sqrt{2} = \sqrt{C_T}$ so that \eqref{cond b) of X,Y} holds.

  \begin{figure}[ht]
  \centering
{ \includegraphics[width=7cm,height=4cm]{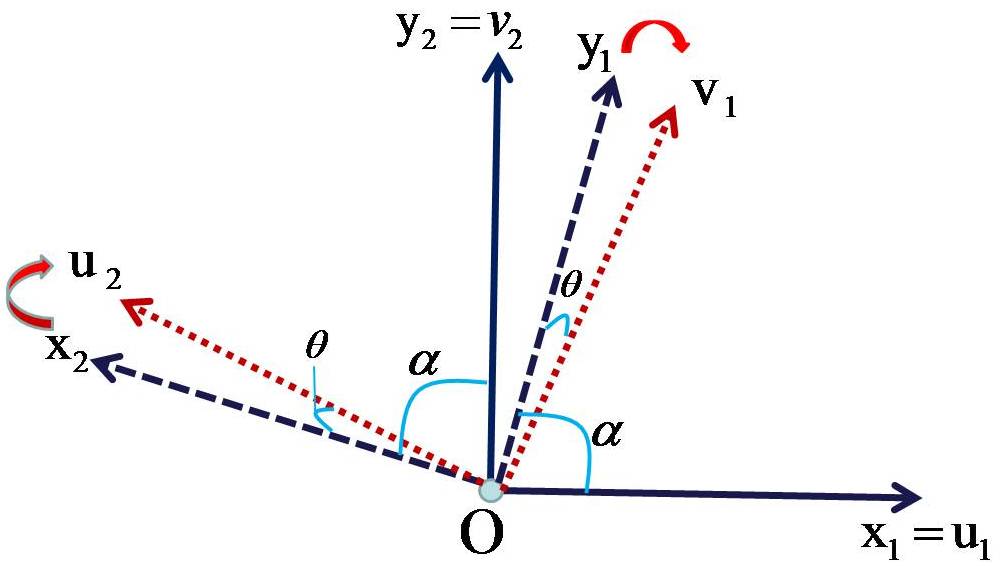} }
  \vspace{-.2cm}
   \caption{ {\small Illustration of the second example.
   $X = (x_1^T, x_2^T) $,   $Y = (y_1^T, y_2^T) $, where $ x_1 = (C,0), x_2 = (-C \sin \alpha, C \cos \alpha )$ and
 $y_1 = (C \cos \alpha, C \sin \alpha ), y_2 = (0, C)$,
 where $C$ is a large constant. Choose $\alpha$ so that $C^2 \cos \alpha = 1 - d/\sqrt{2}$.
  We use the following operation to define $U = (u_1^T, u_2^T) $,   $V = (v_1^T, v_2^T) $: rotate $y_1$ (resp.$\ x_2$) by angle $\theta$ to obtain $v_1$(resp.$\ u_2$),   and let $u_2 = x_2, v_1 = y_1$. Here the angle of rotation $\theta$ is chosen so that
  $ \langle u_1, v_1 \rangle = \langle u_2, v_2 \rangle = 1 $.  }
   }  \label{FigBigAngle} \vspace{-.3cm}
\end{figure} %   $\angle (x_i, \overrightarrow{H_i K_i} ) $ is less than $\frac{\pi}{2}$.

How should we choose $U = (u_1^T, u_2^T) $,   $V = (v_1^T, v_2^T) $ so that \eqref{req of U,V} holds?
The idea for the first example no longer works since it requires that the difference of $\|x_1\|$ and $\|x_2\|$ (resp.$\ \|y_1\|$ and $\|y_2\|$)
is large; however, in this example, $\|x_1\| - \|x_2 \| = \|y_1 \| - \| y_2\| = 0$.
The key idea  for this example is to use rotation. Rotating a vector does not change the norm, so requirement \eqref{cond b) of X,Y} will not
be violated if $u_i$ (resp.$\ v_i$) is obtained by rotating $x_i$(resp.$\ y_i$).
For simplicity, we rotate $y_1, x_2$ to obtain $v_1, u_2$ respectively and let $u_1 = x_1, v_2 = y_2$ (see Figure \ref{FigBigAngle}).
Note that $y_1$ and $x_2$ should be rotated by the same angle as $v_1 $ should be orthogonal to $u_2$ (since the off-diagonal entries
of $UV^T$ are zero).
To increase the inner product $\langle x_i , y_i \rangle$ from $1 - d/\sqrt{2}$ to $1$, we need to decrease the angle of $x_i$ and $y_i$,
thus $y_1$ (resp.$\ x_2$) should be rotated towards $x_1$(resp.$\ y_2$).
Finally, let us specify the angle of rotation $\theta \triangleq \angle(y_1 , v_1) = \angle(x_2, u_2)$.
%More precisely, suppose we rotate $y_1$ to obtain $v_1$ and let $u_1 = x_1$
The requirement $\langle u_1, v_1 \rangle = 1 $ is equivalent to % (using the fact $u_1 =  x_1$)
% $\langle x_1, v_1 - y_1 \rangle = d  $ (here we have
% used $u_1 = x_1$). Using the fact $\langle v_1 - y_1, x_1 \rangle = \pi/2 - (\alpha - \theta/2)$ and $\|v_1 - y_1 \| = 2\| y_1\| \sin(\theta/2)$, the above relation can be simplified to
$
1= \| u_1\| \|v_1\| \cos \angle(u_1, v_1) = \|x_1 \| \|y_1 \| \cos(\alpha - \theta) , % \|x_1 \| \|y_1 \| \cos(\alpha) + d,
$
which can be rewritten as
\begin{equation}\label{example 2, theta choice}
 1=  C^2 \cos(\alpha - \theta) . % C^2 \cos(\alpha) + d.
 \end{equation}
The right-hand side of \eqref{example 2, theta choice} is an increasing function of $\theta$, ranging from $C^2 \cos(\alpha)
\overset{\eqref{C square cos alpha}}{=} 1 - d/\sqrt{2}$
to $ C^2$ for $\theta \in [0, \alpha]$. Since $1$ lies in the range $[1 - d/\sqrt{2}, C^2]$, there exists a unique $\theta$ so that \eqref{example 2, theta choice} holds. One can further verify the requirement \eqref{req c) of U,V}, i.e.\ the difference of $X$ (resp.$\ Y$) and $U$ (resp.$\ V$) is small.
As a rough summary, we rotate $x_i,y_i$ to obtain $u_i, v_i$ when the angle of $x_i$ and $y_i$ is large.
This operation does not change the norm and can increase the inner product $\langle x_i, y_i \rangle$ to the desired amount ($1$ in this case).

\subsubsection{Proof Ideas of Proposition \ref{prop2'}}
In the above two examples, we have used two different operations: one is based on shrinking/extending, and the other is based on rotation.
As we mentioned before, the first operation cannot deal with the second example; also, it is obvious that the second operation cannot deal with
the first example (the angle between $x_i$ and $y_i$ is zero, so rotation only decreases the inner product).
Therefore, both operations are necessary.

Are these two operations sufficient? Fortunately, the answer is yes for the case that $XY^T$ is diagonal and $\langle x_i, y_i \rangle \leq \Sigma_i$
(we need extra effort to reduce the general problem to this case).
When all the angles between $x_i$ and $y_i$  are smaller than a constant $\bar{\alpha}$,
there must be some kind of imbalance in the lengths of $x_i, y_i$'s (to illustrate this, if all $\| x_i\| = \| y_i\|$, then
$\| x_i\|^2 = \| x_i\| \|y_i \| \approx \Sigma_i/\cos\angle(x_i, y_i) \leq \Sigma_i/\cos(\bar{\alpha})$, which implies $\| X\|_F^2 \lesssim r \Sigma_{\rmax}/\cos(\bar{\alpha}) \ll \frac{3}{5}C_T r \Sigma_{\rmax} = \frac{3}{5} \beta_T^2$ for large enough $C_T$, a contradiction to \eqref{cond a) of X,Y}).
 Thus we can use the first operation (i.e.\ shrinking/extending the vectors $x_i,y_i$'s) to obtain the desired $U,V$.
When all the angles between $x_i$ and $y_i$ are larger than a constant $\bar{\alpha}$,
we can use the second operation (i.e.\ rotating the vectors $x_i,y_i$'s) to obtain the desired $U,V$.
In general, some angles may be larger than $\bar{\alpha}$ and others may be smaller, then
a natural solution is to use the two operations \emph{simultaneously}: use the first operation for the pairs $(x_i, y_i)$ with small angles and the second operation for those with large angles.

We had a proof using the two operations simultaneously, but the bounds on $\| U-X\|_F, \|V-Y \|_F$ have a large exponent of $r$.
In the following subsection, we present a different proof that does not use the two operations simultaneously, but only use one of the two operations.
 % for two different cases.  % rather to use only
The basic proof framework is summarized as follows.
We first define $\hat{Y}$ so that $X \hat{Y} = \Sigma$; in other words, we try to satisfy the requirement \eqref{req a) of U,V} first.
Then we try to modify $\hat{Y}$ to satisfy the requirement \eqref{req b) of U,V}.
In particular, we need to reduce the norm of $\hat{Y}$ and keep the norm of $X$ unchanged, while maintaining the relation $X \hat{Y}^T = \Sigma$.
We consider two cases: in Case 1, ``most'' angles between $X$ and $\hat{Y}$ are smaller than $\bar{\alpha}$, and using the first operation (shrinking/extending) can obtain the desired $U,V$; in Case 2, ``most'' angles between $X$ and $\hat{Y}$ are larger than $\bar{\alpha}$, and using the second operation (rotation)
can obtain the desired $U,V$ (see \eqref{three cases} for a precise definition of Case 1 and Case 2).
The difference of this proof framework and the previous one is the following.
 In our previous proof framework, we need to take into account every pair $x_i, y_i$ so that
its inner product is modified to $\Sigma_i$, thus two operations have to be applied simultaneously. In contrast, in this new proof framework, $\langle x_i, \hat{y}_i \rangle$ is already $\Sigma_i$, and
we only need to worry about the ``overall'' requirement that $\|\hat{Y}\|_F$ should be reduced, thus dealing only with the pairs with small angles
(or only with the pairs with large angles) is enough to satisfy the requirement.

{\black
Finally, we would like to mention that when $\Sigma$ is an identity matrix, the proof can be rather simple.
In fact, in this case one can assume $X$ to be diagonal by proper orthonormal transformation, and then assume $Y$ to be diagonal
since the off-diagonal entries are small. By just using the first operation (scaling of the diagonal entries),
we can construct the desired $U,V$ and the proof is similar to that in Appendix \ref{case 1 proof, small angle}.
When $\Sigma$ is not a diagonal matrix, we can replace $X,Y$ by $X Q, Q^{-1}Y$ where $Q$ is orthonormal,
but that only simplifies $X$ to a upper triangular matrix, a condition seems not very helpful.
It seems that the second operation has to be used and the proof becomes more involved. }
%much longer.  }
}

\subsection{Proof of Proposition \ref{prop2'}}\label{appen: proof of prop 2'}  % \ref{prop2', simplifed notation}
As mentioned earlier, to simplify the notations, we use $X,Y,U,V ,d $ to replace $X_1^{\prime},Y_1^{\prime}, U_1^{\prime}, V_1^{\prime}, d^{\prime} $
in Proposition \eqref{prop2'}.
Throughout the proof, we choose
 \begin{equation}\label{CT value}
  C_T = 20,   %% REMARK: can be improved to 10
\end{equation}
and $C_d = 108 $, which implies
\begin{equation}\label{d over Sigma min bound}  % \label{Cd value 37}
\frac{d}{\Sigma_{\rmin}}   \leq \frac{1}{ 108 r }  .
\end{equation}

There are two ``hard'' requirements on $U,V$: \eqref{req a) of U,V} and \eqref{req b) of U,V}.
Our construction of $U,V$ can be viewed as a two-step approach, whereby we satisfy one requirement in each step.
In Step 1, we construct
\begin{equation}\nonumber
\hat{Y} = \Sigma (\Sigma + D)^{-T} Y, \quad \text{where }  D \triangleq XY^T - \Sigma,
\end{equation}
then
\begin{equation}\nonumber% \label{X hat(Y) = Sigma}
X \hat{Y}^T  = (XY^T) (XY^T)^{-1} \Sigma = \Sigma,
\end{equation}
 i.e.\ the first requirement is satisfied.
Since the new $\hat{Y}$ may have higher norm than $\| Y\|_F$, in Step 2 we modify $X, \hat{Y}$ to $U, V$ so that the product does not change,
 and $\| V\|_F \leq \| Y \|_F, \| U\|_F \leq \|X \|_F$.
%----------------
% Note that due to the following Claim \ref{claim of step 1, orthogonization},
% we only need to guarantee $\| V\|_F \leq (1 - \eta)\| \hat{Y} \|_F  $ (which equals $\| Y \|_F$) if $\eta >0$.
% Claim \ref{claim of step 1, orthogonization} is given below
%----------------
 \begin{claim}\label{claim of step 1, orthogonization}
Let $\hat{Y} = \Sigma (\Sigma + D)^{-T} Y$, then
\begin{subequations}
\begin{align}
\eta \triangleq 1 - \frac{ \| Y\|_F }{  \| \hat{Y}\|_F } \leq \frac{d}{ \Sigma_{\rmin} }, \label{eta bound}  \\
 \|Y - \hat{Y} \|_F \leq  \frac{d}{\Sigma_{\rmin}-d}\| Y \|_F.  \label{hat(Y) - Y bound}
\end{align}
\end{subequations}
\end{claim}
\emph{Proof of Claim \ref{claim of step 1, orthogonization}}: By the definition of $\hat{Y}$ we have $Y = (\Sigma + D)^{T}\Sigma^{-1} \hat{Y} $, then we have
\begin{equation}\label{hat(Y) - Y bound interm}
\begin{split}
 \|Y - \hat{Y} \|_F =  \| (\Sigma + D)^{T}\Sigma^{-1} \hat{Y} - \hat{Y}  \|_F = \| D^T \Sigma^{-1}  \hat{Y} \|_F    \\
 \leq \|D^T \Sigma^{-1} \|_F \| \hat{Y} \|_F  \leq  \|D^T \|_F \Sigma_{\rmin}^{-1} \| \hat{Y} \|_F  = \frac{d}{\Sigma_{\rmin}}  \| \hat{Y} \|_F .
\end{split}
\end{equation}
Using the triangular inequality and \eqref{hat(Y) - Y bound interm}, we have
\begin{align}
 &  \| \hat{Y}\|_F \leq  \|Y - \hat{Y} \|_F +  \|Y  \|_F  \leq \frac{d}{\Sigma_{\rmin}}  \| \hat{Y} \|_F +  \|Y  \|_F ,  \nonumber \\
 \Longrightarrow  \quad   &    \|Y  \|_F \geq (1 -\frac{d}{\Sigma_{\rmin}} ) \| \hat{Y}\|_F . \label{hat(Y) and Y ratio}
\end{align}
The first desired inequality \eqref{eta bound} follows immediately from \eqref{hat(Y) and Y ratio},
and the second desired inequality \eqref{hat(Y) - Y bound} is proved by combining \eqref{hat(Y) and Y ratio} and \eqref{hat(Y) - Y bound interm}. $\Box$

Combining \eqref{eta bound} and \eqref{d over Sigma min bound}, we obtain
\begin{equation}\label{eta direct bound by 1 over r}
  \eta \leq \frac{1}{ 108 r}.
\end{equation}

If $ \eta \leq 0$, i.e.\ $  \| \hat{Y}\|_F \leq \| Y\|_F$, then $U = X, V = \hat{Y}$ already satisfy \eqref{req of U,V}.
From now on, we assume $ \eta > 0, $ i.e.\ $  \| \hat{Y}\|_F > \| Y\|_F$.
% Let $\hat{Y} = \Sigma (\Sigma - D)^{-T} Y$, then $X \hat{Y}^T  = \Sigma.$
% \newline Step 2: Modify $X, \hat{Y}$ to $U, V$ so that $UV^T = X \hat{Y}^T = \Sigma$, $\| U\|_F \leq \| X\|_F$ and $\| V\|_F \leq \| Y \|_F$.
Denote $x_i^T, \hat{y}_i^T, u_i^T, v_i^T$ as the $i$-$th$ row of $X,\hat{Y}, U,V$, respectively.
Denote $ \alpha_i \triangleq \angle(x_i, \hat{y}_i) $, i.e.\ the angle between the two vectors $x_i$ and $\hat{y}_i$.
Since $ \langle x_i, \hat{y}_i \rangle = \Sigma_i > 0 $, we have $\alpha_i \in [0, \frac{\pi}{2})$.
Without loss of generality, assume
\begin{equation}\label{angle threshold, for s}
\alpha_1,\dots, \alpha_s >  \frac{3}{8}\pi, \quad \alpha_{s+1},\dots, \alpha_{r} \leq \frac{3}{8}\pi,
\end{equation}
where $s \in \{ 0, 1, \dots, r \}$.
We consider three cases and construct $U,V$ that satisfy the desired properties in the subsequent three subsections.
\begin{subequations}\label{three cases}
\begin{align}
   \text{Case 1}: & \quad \sum_{i=s+1}^r \| \hat{y}_i \|^2 \geq \frac{2}{3} \| \hat{Y} \|_F^2 , \;\; \sum_{i=s+1}^r \| x_i \|^2 \geq \frac{2}{3} \| X \|_F^2.  \label{category 1}\\
      \text{Case 2a}: & \quad \sum_{i=1}^s  \| \hat{y}_i \|^2 > \frac{1}{3} \| \hat{Y} \|_F^2. \label{category 2a} \\
   \text{Case 2b}: & \quad \sum_{i=1}^s \| x_i \|^2 > \frac{1}{3} \| X \|_F^2.  \label{category 2b}
\end{align}
\end{subequations}
% We prove the three cases separately in the following three subsections.

\subsubsection{Proof of Case 1}\label{case 1 proof, small angle}
%Define
%\begin{equation}
%  a_i \triangleq \| x_i\|, i=1,\dots, r,
%\end{equation}
Without loss of generality, assume
\begin{equation}\label{order x_i norm}
   \| x_{s+1}\| \leq \| x_{s+2} \| \leq \dots \leq \|x_r \|.
\end{equation}
Let $K $ be the smallest integer in $\{s+1,s+2,\dots, r \}$ so that
\begin{equation}\label{S1 > 2 S2}
  \sum_{i=s+1}^K \| \hat{y}_i \|^2 \geq 2 \sum_{j=K+1}^r \| \hat{y}_j \|^2.
\end{equation}
By this definition of $K$, we have
\begin{equation}\label{K-1 does not satisfy S1 > 2S2}
  \sum_{i = s+1}^{K-1} \| \hat{y}_i \|^2 < 2 \sum_{j=K}^r \| \hat{y}_j \|^2.
\end{equation}

% For a given integer $K \in \{ 1,2,\dots, s\}$, we
We will shrink and extend $x_i, \hat{y}_i$ to obtain $U,V$. The precise definition of $U = (u_1, u_2, \dots, u_r)^T, V = (v_1, \dots, v_r)^T$
is given in Table \ref{table of Operation 1}.
% In specific, define  as
%\begin{equation}\label{eps and u,v def}
%  u_i \triangleq \frac{x_i}{1 - \epsilon_i }, v_i \triangleq \hat{y}_i(1 - \epsilon_i),  \text{ where }  \epsilon_i = \begin{cases}
%4 \eta &  i \leq K,    \\
%-3 \eta  &  i \geq K+1,
%\end{cases} \quad i=s+1, s+2, \dots, r,
%\end{equation}
%and
%\begin{equation}
%   u_j = x_j, \; v_j = \hat{y}_j, \;\;  j=1, \dots, s.
%\end{equation}
\begin{table*}[htbp]\caption{ Operation 1 }\label{table of Operation 1}
\begin{tabular}{p{460pt}}
\hline
Operation 1: Shrinking and Extending \\
\hline
 \textbf{Input}: $x_k, \hat{y}_k, k=1,\dots, r$.
\\
 \textbf{Output}: $u_k, v_k, k=1,\dots, r$.
\\ \textbf{Procedure}:
\\  $\quad \quad $   (i) For each $j \leq s$, keep $x_j, \hat{y}_j$ unchanged, i.e.\ \begin{equation}
   u_j \triangleq x_j, \; v_j \triangleq \hat{y}_j, \;\;  j=1, \dots, s.
   \vspace{-0.6cm}
\end{equation}
\\  $\quad \quad $   (ii) For each $i \in \{ s+1, \dots, K \} $, extend $x_i$ to obtain $u_i$ and shrink $\hat{y}_i$ to obtain $v_i$.
For each $i \geq K+1 $, shrink $x_i$ to obtain $u_i$ and extend $\hat{y}_i$ to obtain $v_i$.
 More specifically,
\begin{equation}\label{eps and u,v def}
  u_i \triangleq \frac{x_i}{1 - \epsilon_i }, v_i \triangleq \hat{y}_i(1 - \epsilon_i),  \text{ where }  \epsilon_i = \begin{cases}
7 \bar{\eta} &  i \leq K,    \\
- 4.5 \bar{\eta} &  i \geq K+1,
\end{cases} \quad i=s+1, s+2, \dots, r, \vspace{-0.4cm}
\end{equation}
in which
\begin{equation}
              \bar{\eta} \triangleq \frac{d}{\Sigma_{\min}} \geq \eta.
\end{equation}
\\ \hline
\end{tabular}
\end{table*}

We will show that such $U,V$ satisfy the requirements \eqref{req of U,V}.  % there exists some $K$ so that
The requirement \eqref{req a) of U,V} follows directly from the definition of $U,V$ and the fact $ X \hat{Y}^T = \Sigma$.

We then prove the requirement \eqref{req c) of U,V}.
We can bound $\| U-X\|_F$ as
\begin{equation}\label{U - X bound, Case 1}
\begin{split}
 \|U - X \|_F = \sqrt{ \sum_{i > s} \| \frac{1}{1 - \epsilon_i} x_i - x_i \|^2  } = \sqrt{ \sum_{i > s}  \left( \frac{\epsilon_i }{1 - \epsilon_i}\right)^2 \| x_i \|^2  }  \\
 \leq \frac{7  \bar{\eta} }{1 - 7 \bar{\eta}} \sqrt{ \sum_{i > s}   \| x_i \|^2  } \leq \frac{ 7 \bar{\eta} }{1 - 7 \bar{\eta}} \|X \|_F
 \leq \frac{15}{2 }  \bar{\eta} \beta_T.
 \end{split}
\end{equation}
The bound of $ \| V - \hat{Y}\|_F$ is given as
$$ \| V - \hat{Y}\|_F = \sqrt{ \sum_{i > s} \| (1 - \epsilon_i) \hat{y}_i - \hat{y}_i \|^2  }
\leq  \sqrt{ \sum_{i > s}  \epsilon_i^2 \| \hat{y}_i \|^2  }  \leq 7 \bar{\eta} \| \hat{Y}\|_F.
$$
Combining with the bound \eqref{hat(Y) - Y bound interm}, we can bound $ \| V - Y\|_F$ as
\begin{equation}\label{V - Y bound, Case 1}
\begin{split}
 \| V - Y \|_F \leq \| V - \hat{Y}\|_F + \| \hat{Y} - Y \|_F  \leq 7 \bar{\eta} \| \hat{Y}\|_F + \frac{d}{\Sigma_{\rmin}} \| \hat{Y}\|_F   \\
= 8 \bar{\eta} \| \hat{Y}\|_F \overset{\eqref{eta bound}}{\leq } \frac{ 8 \bar{\eta} }{1 - \bar{\eta} }  \| Y \|_F
\leq  \frac{58}{7}   \bar{\eta} \beta_T.
\end{split}
\end{equation}
The first part of the requirement \eqref{req c) of U,V} now follows by multiplying \eqref{U - X bound, Case 1} and \eqref{V - Y bound, Case 1},
and the second part of the requirement \eqref{req c) of U,V} follows directly from \eqref{U - X bound, Case 1} and \eqref{V - Y bound, Case 1}.  %According to \eqref{eta direct bound by 1 over r}, we have $4/(1- 4\eta) < 36/7$ and $5/(1-\eta) \leq 36/7  $. Plugging these two inequalities
%and \eqref{eta bound}
%into \eqref{U - X bound, Case 1} and \eqref{V - Y bound, Case 1}, we obtain
%$ \max \{ \|U - X \|_F ,\| V - Y \|_F   \} \leq \frac{36}{7} \frac{\beta_T}{\Sigma_{\min}}   d \leq \frac{36}{7} \sqrt{r} \frac{\beta_T}{\Sigma_{\min}}   d,$
%which proves the second part of \eqref{req c) of U,V}.
%Multiplying \eqref{U - X bound, Case 1} and \eqref{V - Y bound, Case 1}, we get
%$$
% \| U-X\|_F \| V- Y\|_F \leq \frac{ 7 \eta }{1 - 7 \eta} \frac{ 5d }{\Sigma_{\rmin}} \frac{1}{1 - \eta} \|X \|_F \| Y \|_F
%\overset{\eqref{eta bound}}{\leq} \frac{20}{(1- 4 \eta)(1 - \eta)} \frac{d^2}{\Sigma_{\rmin}^2} \beta_T^2
%\leq  \frac{ 23 \beta_T^2}{\Sigma_{\rmin}^2}d^2 ,
%$$
%which proves the first part of \eqref{req c) of U,V}.
%Here the last inequliaty follows from $ \frac{20}{(1-4 \eta)(1 - \eta)} \leq \frac{20}{(1- 4/54)(1 - 1/54)} \leq 23 $ when $\eta \leq 1/54$.

At last, we prove that $U,V$ satisfy the requirement \eqref{req b) of U,V}.
Let
$$
S_1 \triangleq  \sum_{i=s+1}^{K} \| \hat{y}_i \|^2 , \; S_2 \triangleq  \sum_{j = K+1}^r \| \hat{y}_j \|^2,  \;
S_3 \triangleq  \sum_{k=1}^s \| \hat{y}_k \|^2,
$$
then \eqref{S1 > 2 S2} and \eqref{category 1} imply
\begin{equation}\label{S_1, S_2 bound}
 S_2 \leq S_1/2, \;\;  S_3 \leq (S_1 + S_2)/2 \leq 3S_1/4 .
\end{equation}
Since $\bar{\eta} =  d/\Sigma_{\min} \geq  \eta$, we have $ (1 - \eta)^2 (1 - \bar{\eta} )^2
\geq (1 - 2 \eta) (1 - 2\bar{\eta}  ) \geq (1 - 2 \bar{\eta})^2 $. Then
\begin{equation}\label{(1-eta)hat Y > V}
\begin{split}
   & (1 - \eta)^2 (1 - \bar{\eta} )^2 \| \hat{Y} \|_F^2  - \| V \|_F^2    \\
   & \geq (1 - 2 \bar{\eta} )^2 \| \hat{Y} \|_F^2  - \| V \|_F^2 \\
& = \sum_{i \geq s+1} ( (1 -  2 \bar{\eta} )^2 \| \hat{y}_i\|^2 - \| v_i\|^2 ) + \sum_{k \leq s}  ( (1 - 2 \bar{\eta} )^2 \| \hat{y}_k\|^2 - \| v_k\|^2  )   \\
& =  \sum_{i \geq s+1} ( (1 - 2 \bar{\eta} )^2 \| \hat{y}_i\|^2 - (1 - \epsilon_i)^2\| \hat{y}_i\|^2 )  \\
& \quad \quad + \sum_{k \leq s}  ( (1 - 2 \bar{\eta} )^2 \| \hat{y}_k\|^2 - \| \hat{y}_k\|^2  )  \\
& = \sum_{i \geq s+1}(\epsilon_i - 2 \bar{\eta}) (2 - \epsilon_i - 2 \bar{\eta})\| \hat{y}_i\|^2  - \sum_{k \leq s} 4 \bar{\eta}
   ( 1-\bar{\eta}) \| \hat{y}_k\|^2 \\
& \overset{\eqref{eps and u,v def}}{=} \sum_{s+1 \leq i \leq K} 5 \bar{\eta} (2 - 5 \bar{\eta} - 2 \bar{\eta})\| \hat{y}_i\|^2
\\
& \quad \quad + \sum_{K < j \leq r} (- 6.5 \bar{\eta}) (2 + 4.5 \bar{\eta} - \bar{\eta}) \| \hat{y}_j\|^2
- \sum_{k \leq s} 4 \bar{\eta} (1 -\bar{\eta}) \| \hat{y}_k\|^2     \\
& = 5 \bar{\eta} (2 - 7 \bar{\eta}) S_1
-  6.5 \bar{\eta} (2 + 2.5 \bar{\eta} ) S_2
-  4 \bar{\eta} (1 -\bar{\eta}) S_3  \\
& \overset{\eqref{S_1, S_2 bound}}{\geq}  5 \bar{\eta} (2 - 7 \bar{\eta}) S_1
-   6.5 \bar{\eta} (2 +  2.5 \bar{\eta} ) \frac{1}{2}S_1 -  4 \bar{\eta} (1-\bar{\eta}) \frac{3}{4} S_1  \\
& \geq ( 0.5 -  41 \bar{\eta}) \bar{\eta} S_1 \geq 0,
%& = (c_1-1) \eta (2 - c_1\eta  - \eta) S_1
%- (c_2 + 1) \eta (2 + c_2 \eta - \eta) S_2
%-  \eta (2-\eta) S_3
\end{split}
\end{equation}
where the last inequliaty follows from \eqref{d over Sigma min bound}. Note that $ (1 - \eta) \| \hat{Y} \|_F = \| Y \|_F$, thus \eqref{(1-eta)hat Y > V} implies % the fact $\eta \leq 1/37.$
$$
 \| V\|_F \leq (1 - \eta)(1 - \bar{\eta} ) \| \hat{Y} \|_F = (1 - \frac{ d }{ \Sigma_{\min}}) \| Y\|_F,
$$
which proves the second part of \eqref{req b) of U,V}.

We then prove the first part of \eqref{req b) of U,V}, i.e.\ $\| U\|_F \leq \| X\|_F $.
Let $$ T_1 \triangleq  \sum_{i = s+1 }^{K }  \| x_i\|^2 ,  \; \;  T_2 \triangleq \sum_{j = K }^{ r}  \| x_j\|^2 .$$
We claim that
\begin{equation}\label{T1 > 0.3 T2}
 T_2 \geq 2 T_1.
\end{equation}
We prove \eqref{T1 > 0.3 T2} by contradiction. Assume the contrary that $T_2 < 2 T_1$,
then $ \frac{1}{3} (T_2 + T_1) < T_1 $, i.e.\ \begin{equation}\label{T2 bound}
\frac{1}{3} \sum_{k = s + 1}^r  \| x_k\|^2 <   \sum_{i = s+1}^{K }  \| x_i\|^2  \overset{\eqref{order x_i norm}}{\leq} (K-s) \| x_K\|^2.
\end{equation}
%%%%%%-------------------------------------------------------------------
%%%%%%%%%% Replacement using weaker inequal., 4 times worse constant  %%%%
%%%%%%-------------------------------------------------------------------
Plugging the second inequality of \eqref{category 1}, i.e.\  $\sum_{k = s + 1}^r  \| x_k\|^2  \geq \frac{2}{3}\| X\|_F^2$, into the above relation, we obtain
\begin{equation}\label{xK bound}
\| X\|_F^2 \leq \frac{9}{2}(K-s) \| x_K\|^2 \leq \frac{9}{2} K \| x_K\|^2.
\end{equation}
When $j \in \{ K, K+1, \dots, r \}$, we have
$$ \Sigma_{\rmax} \geq \Sigma_j = \langle x_j, \hat{y}_j \rangle = \| x_j\| \|\hat{y}_j \| \cos(\alpha_j)\overset{\eqref{order x_i norm},\eqref{angle threshold, for s}}{\geq}
  \| x_K\| \|\hat{y}_j \| \cos( 3\pi/8). $$  % \| x_K\| \|\hat{y}_j \| \cos(\alpha_j)  \overset{}{\geq}  \geq \frac{2}{5}\| x_K\| \|\hat{y}_j \| ,
which implies
\begin{equation}\label{omega bound hat yj}
 \| \hat{y}_j\| \leq  \omega, \text{ where } \omega \triangleq \frac{1}{ \cos( 3\pi/8)} \frac{\Sigma_{\rmax}}{ \|x_K \| }, \;\; j=K, K+1, \dots, s.
\end{equation}
Therefore,
\begin{equation}\label{omega bound}
\|\hat{Y} \|_F^2  \overset{\eqref{category 1}}{\leq} \frac{3}{2} \sum_{j=s+1}^r \| \hat{y}_j\|^2
 \overset{\eqref{K-1 does not satisfy S1 > 2S2}}{\leq} \frac{9}{2} \sum_{j=K}^r \| \hat{y}_j\|^2
\overset{\eqref{omega bound hat yj}}{\leq}
  \frac{9}{2} (r - K +1) \omega^2.
\end{equation}

Combining \eqref{xK bound} and \eqref{omega bound}, and using $ K  (r-K + 1) \leq \frac{1}{4}(r+1)^2 \leq  r^2 $,
we get
\begin{equation}\label{|X| |Y| bound}
\begin{split}
 \|X\|_F^2 \|\hat{Y} \|_F^2 \leq \frac{81}{4} r^2 \| x_K\|^2 \omega^2   \\
\overset{\eqref{omega bound hat yj}}{=} \frac{81}{4} r^2  \| x_K\|^2  \frac{1}{ \cos( 3\pi/8)^2} \frac{\Sigma_{\rmax}^2}{\|x_K \|^2 } < 140 r^2 \Sigma_{\rmax}^2.
\end{split}
\end{equation} % \frac{25}{4}
According to \eqref{cond b) of X,Y}, we have
$ \|X\|_F^2 \|\hat{Y} \|_F^2 \geq \|X\|_F^2 \| Y \|_F^2 \geq  (\frac{3}{5})^2 \beta_T^4 = \frac{9}{25}  C_T^2 r^2 \Sigma_{\rmax}^2
$; combining with \eqref{|X| |Y| bound}, we get
$ 140 >\frac{9}{25}  C_T^2 $, which implies $ C_T^2 < 389$. This contradicts the definition \eqref{CT value} that $C_T = 20$,
thus \eqref{T1 > 0.3 T2} is proved.

Now we are ready to prove the first part of \eqref{req b) of U,V} as follows:
\begin{equation}\nonumber% \label{X > U}
\begin{split}
  \| X\|_F^2 - \| U\|_F^2 & = \sum_{i \geq s+1 } (\| x_i\|^2 - \| u_i\|^2 ) + \sum_{k \leq s} (\| x_k\|^2 - \| u_k\|^2 )  \\
 & =  \sum_{i \geq s+1} (\| x_i\|^2 - \frac{1}{(1 - \epsilon_i)^2 }\| x_i\|^2 )  + 0  \\
 & = \sum_{i \geq s+1} \frac{ \epsilon_i(\epsilon_i - 2) }{ (1 - \epsilon_i)^2 } \| x_i\|^2    \\
 & = \sum_{K < j \leq r} \frac{ 4.5 \bar{\eta}( 4.5 \bar{\eta}+2)}{ (1 + 4.5 \bar{\eta})^2}   \| x_j\|^2
   - \sum_{s+1 \leq i \leq K } \frac{ 7 \bar{\eta}(2 - 7 \bar{\eta})} {  (1 - 7 \bar{\eta})^2}   \| x_i\|^2  \\
& \overset{\eqref{T1 > 0.3 T2}}{\geq} T_2 \bar{\eta} \left[ \frac{ 4.5 (4.5 \bar{\eta}+2)}{ (1 + 4.5 \bar{\eta})^2}  -
  \frac{1}{2} \frac{ 7(2 - 7 \bar{\eta})} {  (1 - 7 \bar{\eta})^2}    \right] \\
& \geq T_2 \bar{\eta} \left[ \frac{ 9 }{ (1 + 4.5 \bar{\eta})^2} - \frac{ 7 } { (1 - 7 \bar{\eta} )^2} \right]  \geq 0 ,
\end{split}
\end{equation}
where the last inequality is because
$  \frac{ (1 - 7 \bar{\eta} )^2}{ (1 + 4.5 \bar{\eta})^2} > 0.79 > \frac{7}{9} $
%$ \frac{ (1 + 2 \bar{\eta})^2}{ (1 - 9 \bar{\eta}/2)^2 } \leq \frac{ (1 + 1/18)^2}{ (1 - 1/8)^2} \leq 1.46 \leq \frac{40}{27} $
when $\bar{\eta} \leq 1/(108r) < 1/100 $.
Thus the first part of \eqref{req b) of U,V} is proved.

\subsubsection{Proof of Case 2a}
Denote
\begin{equation}
X^{0}=X, Y^{0}=\hat{Y}, x_k^{0} = x_k, y_k^{0}= \hat{y}_k, \alpha_k^{0} = \alpha_k, \; k=1,\dots, r.
\end{equation}
% D_k^{[0]} = D_k, D_k^{-([0])} = D_k^{-}, D_k^{+[0]}=D_k^{+}.
We will define $X^i = (x_1^i, \dots, x_r^i)^T, Y^i = (y_1^i, \dots, y_r^i)^T$ recursively.
In specific, at the $i$-th iteration, we will adjust $X^{i-1}, Y^{i-1}$ to $X^i, Y^i$ so that
 $ \|X^i \|_F \leq \|X^{i-1} \|_F, \|Y^i \|_F < \|Y^{i-1} \|_F $ while keeping the first requirement satisfied, i.e.\ $X^i (Y^i)^T = \Sigma$.
% to increase $x_i^T y_i$ to at least $\Sigma_i$;
% in other words, $x_i^i, y_i^i$ are chosen so that $(x_i^i)^T y_i^i \geq \Sigma_i$, and other $x_j^i, y_j^i$ will be specified later.
%For each $i,$ the process of  is called
%one \emph{iteration}. At the $i-th$ iteration, we will define $X^{i},Y^{i}$ by adjusting $X^{i-1},Y^{i-1}.$
 % Other notations $ \alpha_k^{i} $ are defined accordingly based on $x_k^i, y_k^i$.
The angle $ \alpha_k^{i} $ is defined accordingly, i.e.\ $\alpha_k^{i} \triangleq \langle x_k^i, y_k^i \rangle $.

To adjust $X^{i-1}, Y^{i-1}$ to $X^i, Y^i$, we will define an operation that consists of rotation and shrinking.
 The basic idea is the following: since the angle between $x_i^{i-1}$ and $y_i^{i-1}$ is large, we can rotate $x_i^{i-1}$ to $x_i^i$ and shrink $y_i^{i-1}$
  to $y_i^{i}$ to keep the inner product invariant, i.e.\ $\langle x_i^{i-1},   y_i^{i-1} \rangle = \langle x_i^i,  y_i^{i} \rangle$.
%  without increasing the length of $x_i$ or $y_i.$
However, rotating $x_i^{i-1}$ may destroy the orthogonal relationship between $x_i^{i-1}$ and $y_j^{i-1}, \forall j\neq i$, thus we further rotate and shrink $y_j^{i-1}$ to $y_j^{i}$ for all  $j \neq i$ so that $y_j^i$ is orthogonal to the new vector $x_i^i$.
Fortunately, we can prove that using such an operation we still have $\langle x_j^{i-1}, y_j^{i} \rangle = \Sigma_j, \forall j\neq i $.

% A formal statement will be given after the following summary of Operation 3.
% we change $X^{i-1},Y^{i-1}$ so that the new $X^i, Y^i$ satisfy some requirements; see the formal statement below.

%  rotate and shrink $y_j^{i-1}$ in plane $ \rspan\{ y_j^{i-1}, y_i^{i-1} \} = \rspan_{k\neq i,j}\{x_k^{i-1} \}^{\bot}$ to get $y_j^i$
%\textbf{Assumptions}:  \\
%  $\quad \quad $  (a)   $ i   \geq s+1; $ \\
%  $\quad \quad $  (b) $ \langle  x_i^{i-1}, y_i^{i-1}\rangle   < \Sigma_i; $ \\
%  $\quad \quad $  (c) $ \frac{ \pi}{3}   \leq \alpha_i^{i-1}  \leq \alpha_j^{i-1}, \forall j > i. $  \\
A complete description of this operation is given in Table \ref{table of Operation 2}.
Without loss of generality, we can make the assumption \eqref{assumption of alpha i being smallest}.
In fact, if \eqref{assumption of alpha i being smallest} does not hold, we can switch $i$
and $m_i \triangleq \arg \min_{k \in \{ i, i+1, \dots, s\}}  \alpha_k^{i-1}$ and then apply Operation 2.
%Rigorously speaking, these $r$ iterations should be performed as follows:
%in the $i$-th iteration, define $m_i \triangleq \arg \min_{k \in \{ 1,2, \dots, s\}\backslash\{ m_{1}, \dots, m_{i-1}  \} } \alpha_k^{i-1}$
%and adjust $ \langle x_{m_i}, y_{m_i} \rangle$ to be at least $\Sigma_{m_i}$.
%In an effort to simplify the notations, we just assume $m_i = i, \forall i > s$ (i.e.\ \eqref{assumption of alpha i being smallest} holds for all $i$) and apply Operation 3 to adjust $ \langle x_{i}, y_{i} \rangle$ to be at least $\Sigma_{i}$.
% so that the new $\langle x_i, y_i \rangle$ is at least $\Sigma_i$ (i.e.\ the original $\langle x_{m_i}, y_{m_i} \rangle $ is at least $\Sigma_{m_i}$).

\begin{table*}[htbp]\caption{ Operation 2 that defines $X^i, Y^i$, where $i \in \{ 1,\dots, s \}$ }\label{table of Operation 2}
\begin{tabular}{p{460pt}}
\hline
Operation 2: Rotation and Shrinking \\
\hline
 \textbf{Input}: $x_k^{i-1}, y_k^{i-1}, \alpha_k^{i-1} \triangleq \angle (x_k^{i-1}, y_k^{i-1}), k=1,\dots, r$ and $D_i$. % \leq \eta \Sigma_i$.
\\
%  \textbf{Assumption}:
% \end{split}
% \end{equation}
 \textbf{Output}: $x_k^{i}, y_k^{i}, k=1,\dots, r$ and $\alpha_k^i \triangleq \angle (x_k^{i}, y_k^{i})$.
\\ \textbf{Procedure}:
\\  $\quad \quad $   (1) Rotate $x_i^{i-1}$ in $\rspan \{x_i^{i-1},y_i^{i-1}\}$ to get $x_i^{i}$, such that $\langle x_i^i, y_i^{i-1} \rangle = \Sigma_i + D_i.$
\\  $\quad \quad $   (2) Shrink $y_i^{i-1}$ to get $y_i^i$ such that $ \langle x_i^i, y_i^{i} \rangle = \Sigma_i . $
\\   $\quad \quad $  (3) For all $j\neq i$, find $y_j^{i}$ in $ \rspan\{ y_j^{i-1}, y_i^{i-1} \} = \rspan_{k\neq i,j}\{x_k^{i-1} \}^{\bot}$ such that $y_j^i \bot x_i^i$ and $\langle x_j^{i-1}, y_j^{i} \rangle = \langle x_j^{i-1}, y_j^{i-1} \rangle.$
\\  $\quad \quad $   (4) Define $x_j^i \triangleq x_j^{i-1}, \forall j \neq i$.
\\ \hline
\end{tabular}
\end{table*}

% The assumption  is added for technical reason (in order to prove the property (d) in Claim \eqref{Claim 3}).

We will prove that Operation 2 is valid (for $D_i$ that is small enough), i.e.\ $X^i, Y^i$ defined in Operation 2 indeed exist.
The properties of $X^i, Y^i$ obtained by Operation 2 are summarized in the following claim, which will be proved in Appendix \ref{appen: proof of key Claim of Operation 2}.
\begin{claim}\label{claim: property of Operation 2}
% Suppose $i >s,$ i.e.\ $\alpha_i^{[0]} \geq \frac{2}{5}\pi$; and $\langle x_i^{i-1}, y_i^{i-1}\rangle < \Sigma_i.$ Then, there exist $X^{i}, Y^{i}$ such that
Consider $i \in \{1, 2,\dots, s \}$. Suppose
\begin{equation}\label{assumption of alpha i being smallest}
 \alpha_i^{i-1}  \leq \alpha_j^{i-1}, \; \forall \ j \in \{ i+1, i+2, \dots, s\},
 \end{equation}
 and $D_i >0$ satisfies
\begin{equation}\label{D_i bound by Sigma_i/10}
   \frac{D_i}{\Sigma_i } \leq \frac{1}{12r},
\end{equation}
 then $X^i = (x_1^i, \dots, x_r^i )^T, Y^i = (y_1^i, \dots, y_r^i )^T  $ described in Operation 2 exist and satisfy the following properties:
\begin{subequations}\label{official property of Xi,Yi}
\begin{align}
  X^i (Y^i)^T & = \Sigma ,    \label{official property a) of Xi,Yi} \\
  \quad  \|x_k^i \| = \|x_k^{i-1} \|, \forall k, \;\;
  \|  Y^{i} - Y^{i-1} \|_F^2 &
\leq \frac{4}{5} \frac{D_i}{\Sigma_i} ( \| Y^{i-1}\|_F^2 - \|Y^{i} \|_F^2 ), \label{official property b) of Xi,Yi}
   \\
  \|X^{i} - X^{i-1} \|_F = \| x_i^i - x_i^{i-1}\| \leq  &  \frac{1}{\sqrt{3}}\frac{D_i}{\Sigma_{i}}\|x_i^{i-1} \|
  \nonumber
\\   \;\;
 \|Y^{i} - Y^{i-1} \|_F & \leq \frac{2}{\sqrt{3}} \frac{ D_i}{\Sigma_{i}}\|Y^{i-1} \|_F ,  \label{official property c) of Xi,Yi} \\
 \alpha_l^{i}  \geq \alpha_l^{i-1} - \frac{1}{r}\frac{\pi}{24}   \geq \frac{1}{3}\pi, & \;  l=i, i+1, \dots, s.  \label{official property d) of Xi,Yi} \\
 \| y_k^{i-1} \| \geq \| y_k^{i}\|  \geq \| y_k^{i-1} \|    - \frac{1}{10 r}\| & y_k^{i-1} \| ,   \;  k=1, 2,\dots, s.  \label{official property e) of Xi,Yi} \\
\| Y^{i-1}\|_F^2 - \|Y^{i} \|_F^2 \geq  & \frac{5}{3} \frac{D_i}{\Sigma_i} \| y_i^i \|^2. \label{official property f) of Xi,Yi}
\end{align}
\end{subequations}
% \newline (e)$ \quad X^{i}(Y^{i})^T$ \text{is diagonal}.  \frac{4}{5} \frac{\beta_T}{\Sigma_{\min}}D_i^{i-1},  \|Y^{i-1} \|_F^2 - \|Y^{i} \|_F^2 \geq \frac{ \|Y^{i-1} - Y^i \|_F^2}{3 \eta},  \frac{2}{\sqrt{3}}\frac{D_i}{\Sigma_i}  \geq \frac{1}{\sqrt{3}}\|y_k^0 \|,
\end{claim}

We continue to prove Proposition \ref{prop2'} using Claim \ref{claim: property of Operation 2}.
Given any $D_1, \dots, D_s$ that satisfy \eqref{D_i bound by Sigma_i/10}, we can apply a sequence of Operation 2 for $i=1,2,\dots, s$
to define two sequences of matrices $Y^1, \dots, Y^s$ and $X^1,\dots, X^s$.
Since $Y^1, \dots, Y^s$ depend on $D_1, \dots, D_s$, thus we can use $Y^s(D_1,\dots, D_s)$ to denote the obtained $Y^s$ by applying Operation 2
for $D_1,\dots, D_s$.
Obviously $Y^s(0,\dots, 0) = Y^0$.
We can also view $ \|Y^s \|_F^2  $ as a function of $D_1,\dots, D_s$, denoted as
\begin{equation}\label{f def}
f(D_1, \dots, D_s) \triangleq \|Y^s(D_1, \dots, D_s) \|_F^2.
\end{equation}
It can be easily seen that $f$ is a continuous function with respect to $D_1,\dots, D_s$.

Define\footnote{In the first version of the paper, we define $\bar{D}_i \triangleq \frac{9}{2} \eta \Sigma_i \leq \frac{9}{2} \bar{\eta} \Sigma_i
\leq 9 \frac{d}{\Sigma_{\min}} \Sigma_i$,
which is enough for proving Theorem \ref{major theorem}.
Here we use a slightly different definition of $\bar{D}_i$ for the purpose of proving Theorem \ref{theorem 2: linear convergence} (linear convergence of the algorithm.) }
\begin{equation}\label{D_i = eta Sigma_i}
 \bar{\eta} \triangleq \frac{d}{\Sigma_{\min}} \overset{\eqref{eta bound}}{\geq} \eta, \quad \bar{D}_i \triangleq 9 \bar{\eta} \Sigma_i,  \; i=1,\dots, s .
\end{equation}
We prove that
\begin{equation}\label{Y0- Ys bound by eta Y0}
 f(\bar{D}_1,\dots, \bar{D}_s ) \leq (1 - 4 \bar{\eta} ) \| \hat{Y} \|_F^2.
\end{equation}

Suppose $\bar{X}^i, \bar{Y}^i, i=1,\dots, s$ are recursively defined by Operation 2 for the choices of $D_i = \bar{D}_i$
and denote $\bar{X}^0 = X, \bar{Y}^0 = \hat{Y}$.
% where $\bar{Y}^s$ is the simplified notation for $Y^s(\bar{D}_1,\dots, \bar{D}_s)$.
 % as in \eqref{D_i = eta Sigma_i}.
Since
\begin{equation}\nonumber
\bar{\eta} = d/\Sigma_{\rmin} \overset{\eqref{d over Sigma min bound}}{\leq} 1/(108r),
\end{equation}
 we know that $D_i = \bar{D}_i, i=1,\dots, s$ as defined in \eqref{D_i = eta Sigma_i} satisfy the condition \eqref{D_i bound by Sigma_i/10},
thus the property \eqref{official property of Xi,Yi} holds for $\bar{X}^i, \bar{Y}^i$. % in Claim \ref{claim: property of Operation 2} hold.
% \|Y^s \|_F^2
Suppose the $k$-th row of $\bar{Y}^i$ is $(\bar{y}_k^i)^T$, $k=1,\dots, r$.
By \eqref{official property f) of Xi,Yi} and the fact $\hat{Y} = \bar{Y}^0$, we have
\begin{equation}\label{Y0- Ys bound by eta Y0 interm 1}
\begin{split}
 \| \hat{Y}\|_F^2 -  f(\bar{D}_1,\dots, \bar{D}_s )  = \| \bar{Y}^0\|_F^2 - \|\bar{Y}^s \|_F^2    \\
 = \sum_{i=1}^s  (\| \bar{Y}^{i-1}\|_F^2 - \|\bar{Y}^i \|_F^2)
\geq \sum_{i=1}^s \frac{5}{3} \frac{\bar{D}_i}{\Sigma_i} \| \bar{y}_i^i \|^2.
\end{split}
\end{equation}
We can bound $\| \bar{y}_i^i\|$ according to \eqref{official property e) of Xi,Yi} as
\begin{align*}
  \| \bar{y}_i^{i}\|  \geq \| \bar{y}_i^{i-1} \|    - \frac{1}{10 r}\| \bar{y}_i^{i-1} \|
\geq \| \bar{y}_i^{i-1} \|    - \frac{1}{10 r}\| \bar{y}_i^{0} \|   \\
\geq \dots \geq \| \bar{y}_i^{0} \| - \frac{i}{10 r} \| \bar{y}_i^{0} \|
\geq  \frac{9}{10} \| \bar{y}_i^{0} \| .
\end{align*}
Plugging into \eqref{Y0- Ys bound by eta Y0 interm 1}, we get
\begin{equation}\nonumber
\begin{split}
 \| \hat{Y}\|_F^2 -  f(\bar{D}_1,\dots, \bar{D}_s )
\geq \sum_{i=1}^s \frac{5}{3} \frac{\bar{D}_i}{\Sigma_i} (\frac{9}{10})^2 \| \bar{y}_i^{0} \|^2    \\
\overset{\eqref{D_i = eta Sigma_i}}{=} 15 \frac{81}{100} \bar{\eta} \sum_{i=1}^s \| \hat{y}_i\|^2
\overset{\eqref{category 2a}}{>} 12 \bar{\eta} \frac{1}{3}\|\hat{Y} \|_F^2 = 4 \bar{\eta} \|\hat{Y} \|_F^2,
\end{split}
\end{equation}
which immediately leads to \eqref{Y0- Ys bound by eta Y0}.

Combining  \eqref{Y0- Ys bound by eta Y0} and the fact $f(0,\dots, 0) = \| Y^0\|_F^2 = \| \hat{Y}\|_F^2$, we have
$$
f(0,\dots, 0) =  \| \hat{Y} \|_F^2  > (1 - 4 \bar{\eta} ) \| \hat{Y} \|_F^2 =  f(\bar{D}_1,\dots, \bar{D}_s ).
 $$
Since $f$ is continuous (in the proof of Claim \ref{claim: property of Operation 2} in Appendix \ref{appen: proof of key Claim of Operation 2},
all new vectors depend continuously on $D_i$),
and notice that $ 1 - 4 \bar{\eta}  <  (1 - \bar{\eta} )^4 \leq (1 - \bar{\eta} )^2 (1 - \eta )^2 \leq 1 $,
 there must exist
\begin{equation}\label{D_i bound by bar D}
 0 \leq D_i \leq \bar{D}_i = 9 \bar{\eta} \Sigma_i, \;\; i=1,\dots, s
 \end{equation}
such that
\begin{equation}\label{f(D_1,...,D_s) = (1-2 eta)Y}
 f(D_1,\dots, D_s) =  (1 - \bar{\eta} )^2 (1 - \eta )^2 \| \hat{Y} \|_F^2.
\end{equation}
Suppose $X^i, Y^i, i=1,\dots, s$ are recursively defined by Operation 2 for these choices of $D_i$,
where $Y^s$ is the simplified notation for $Y^s(D_1,\dots, D_s)$.
Define
\begin{equation}\label{case 2a def of U,V}
  V \triangleq Y^s, \;\;  U \triangleq X^s,
\end{equation}
By this definition of $V$ and \eqref{f def}, the relation \eqref{f(D_1,...,D_s) = (1-2 eta)Y} can be rewritten as
\begin{equation}\label{V = (1-2 eta)Y hat}
 \| V\|_F^2  =  (1 - \bar{\eta} )^2 (1 - \eta )^2 \| \hat{Y} \|_F^2.
\end{equation}

We show that $U,V$ defined by \eqref{case 2a def of U,V} satisfy the requirements \eqref{req of U,V}.
The requirement \eqref{req a) of U,V} follows by the property \eqref{official property a) of Xi,Yi} for $i=s$. % and \eqref{X hat(Y) = Sigma}.
The requirement \eqref{req b) of U,V} is proved as follows.
Combining \eqref{V = (1-2 eta)Y hat} with \eqref{eta bound} leads to {\black
\begin{equation}\label{|V| = |Y|}
 \| V\|_F  =  (1 - \bar{\eta} ) (1 - \eta ) \| \hat{Y} \|_F = (1 - \bar{\eta} ) \| Y\|_F = (1 - \frac{d}{\Sigma_{\min}} ) \| Y\|_F.
\end{equation}  }
According to the property \eqref{official property b) of Xi,Yi}, we have
$ \| X^i \|_F = \| X^{i-1}\|_F,  i=1,\dots, s. $ Thus
$ \| X^s \|_F = \| X^{s-1}\|_F = \dots = \|X^0 \|_F = \|X \|_F$, which implies
\begin{equation}\label{|U| = |X|}
 \| U\|_F  =  \| X\|_F.
\end{equation}
Combining \eqref{|U| = |X|} and \eqref{|V| = |Y|} leads to the requirement \eqref{req b) of U,V} . % according to  $U,V$ satisfy the requirement

It remains to show that $U,V$ satisfy the requirement \eqref{req c) of U,V}.
By the property \eqref{official property b) of Xi,Yi}, we have $\| x_k^{i-1}\| = \| x_k^i\|, \forall 1\leq k \leq r, 1\leq i \leq s$,
which implies
\begin{equation}\label{x_k norm invariant}
\| x_k^i\| = \|x_k^0 \| = \| x_k\|, \;\; \forall 1\leq k \leq r, 1\leq i \leq s.
\end{equation}
Note that $X^i$ differs from $X^{i-1}$ only in the $i$-th row (according to \eqref{official property c) of Xi,Yi}), thus
\begin{equation}\label{U - X bound, interm}
\begin{split}
  \|U - X \|_F  = \| X^s - X^0 \|_F = \sqrt{  \sum_{i=1}^s \| x_i^i - x_i^{i-1}  \|^2 } \\
\overset{\eqref{official property c) of Xi,Yi}}{\leq}  \frac{1}{\sqrt{3}}\frac{D_i}{\Sigma_{i}} \sqrt{ \sum_{i=1}^s \|x_i^{i-1} \|^2 }
\overset{\eqref{x_k norm invariant}}{=}  \frac{1}{\sqrt{3}}\frac{D_i}{\Sigma_{i}} \sqrt{ \sum_{i=1}^s \|x_i \|^2 }   \\
\leq  \frac{1}{\sqrt{3}}\frac{D_i}{\Sigma_{i}} \| X\|_F
\overset{\eqref{D_i bound by bar D}}{\leq} 3 \sqrt{3} \bar{\eta} \| X\|_F.  % true value 2.59 \frac{45}{16}
\end{split}
\end{equation}
% {\black DDDDDDDDDDDDDDDDDDDDDDDD   can improve to $\frac{13}{5}$ }
Plugging $ \bar{\eta} = d/\Sigma_{\rmin}$ and $\|X \|_F \leq \beta_T$ into the above inequality, we get
\begin{equation}\label{U - X bound}
\|U - X \|_F \leq 3\sqrt{3} \frac{\beta_T }{\Sigma_{\rmin}}  d.
\end{equation}  % \overset{\eqref{eta bound}}{\leq}

We then bound $\| V - \hat{Y} \|_F^2$ as
\begin{equation}\nonumber%\label{V - Y bound interm}
\begin{split}
&  \|V - \hat{Y} \|_F^2  = \| Y^s - Y^0 \|_F^2 \\
& \leq s  \sum_{i=1}^s \| Y^i - Y^{i-1}  \|_F^2
\overset{\eqref{official property b) of Xi,Yi}}{\leq} s \frac{4}{5}\frac{D_i}{\Sigma_i}  \sum_{i=1}^s (  \| Y^{i-1}  \|_F^2 - \| Y^i \|_F^2 )   \\
& = s \frac{4}{5}\frac{D_i}{\Sigma_i} ( \| Y^0 \|_F^2 -  \| Y^s \|_F^2 )
= s \frac{4}{5}\frac{D_i}{\Sigma_i} ( \|\hat{Y} \|_F^2 -  \| V\|_F^2 )  \\
& \overset{\eqref{D_i bound by bar D}}{ \leq } \frac{ 36 }{5} s  \bar{\eta}  ( \|\hat{Y} \|_F^2 -  \| V\|_F^2 )  % true vale 3.6 \frac{63}{16}
\overset{\eqref{V = (1-2 eta)Y hat}}{ \leq } \frac{ 36 }{5} s \bar{\eta}  (2 \eta + 2 \bar{\eta} ) \|\hat{Y} \|_F^2  % \frac{63}{16}
\leq \frac{ 144 }{5} r  \bar{\eta}^2  \|\hat{Y} \|_F^2  ,
  % \frac{63}{8}
%= s \frac{4}{5}\frac{D_i}{\Sigma_i} (\| Y^0 \|_F^2 -  \| Y^s \|_F^2 )
%= s \frac{4}{5}\frac{D_i}{\Sigma_i} (\|\hat{Y} \|_F^2 -  \| V\|_F^2)
%\overset{\eqref{V = (1-2 eta)Y hat}}{=} (2 \eta - \eta^2 ) \|\hat{Y} \|_F^2
\end{split}
\end{equation}
% {\black DDDDDDDDDDDDDDDDDDDDDDDDDDDDd can changed to 7.2 }
which leads to
\begin{equation}\label{V - hat Y bound}
\begin{split}
 \|V - \hat{Y} \|_F \leq  \frac{12}{\sqrt{5}} \bar{\eta} \sqrt{r} \|\hat{Y} \|_F.  % \frac{\sqrt{21}}{2}
\end{split}
\end{equation}
% Combining with $ \| \hat{Y}\|_F = \| Y\|_F/(1-\eta) \leq \| Y\|_F/(1-1/54) $ and \eqref{hat(Y) - Y bound interm}, %  $ \|\hat{Y} - Y\|_F \leq d \| \hat{Y} \|_F/\Sigma_{\rmin} $,
Then we can bound $\| V - Y \|_F$ as
\begin{equation}\label{V - Y bound}
\begin{split}
& \|V - Y \|_F \leq \|V - \hat{Y} \|_F + \| Y - \hat{Y} \|    \\
&  \overset{\eqref{V - hat Y bound}, \eqref{hat(Y) - Y bound interm}}{\leq} \frac{12}{\sqrt{5}}   \bar{\eta} \sqrt{r}\|\hat{Y} \|_F + \frac{d}{\Sigma_{\rmin}} \| \hat{Y}\|_F
= (\frac{12}{\sqrt{5}} + 1) \frac{d}{\Sigma_{\rmin}} \sqrt{r} \|\hat{Y} \|_F    \\
& \overset{\eqref{eta bound}}{=} (\frac{12}{\sqrt{5}} + 1) \frac{d}{\Sigma_{\rmin}} \sqrt{r} \| Y \|_F \frac{1}{ 1 - \eta}
<   \frac{ 13 d}{ 2\Sigma_{\rmin}} \sqrt{r} \| Y \|_F  % \frac{17}{5}
\leq   \frac{ 13 \beta_T }{ 2 \Sigma_{\rmin} } \sqrt{r} d,
\end{split}
\end{equation}
where the second last inequality is due to $ (\frac{12}{\sqrt{5}} + 1)/(1 - \eta ) \overset{\eqref{d over Sigma min bound}}{\leq} (\frac{12}{\sqrt{5}} + 1)/(1 - \frac{1}{108} ) < 6.5$.
The first part of the requirement \eqref{req c) of U,V} now follows by multiplying  \eqref{U - X bound} and \eqref{V - Y bound},
and the second part of the requirement \eqref{req c) of U,V} follows directly from \eqref{U - X bound} and \eqref{V - Y bound}.

\subsubsection{Proof of Case 2b}
Similar to Case 2a, denote
\begin{equation}\nonumber
X^{0}=X, Y^{0}=\hat{Y}, x_k^{0} = x_k, y_k^{0}= \hat{y}_k, \alpha_k^{0} = \alpha_k.
\end{equation}
By a symmetric argument to that for Case 2a (switch the role of $U, X^j, j=0,\dots, s$ and $
V, Y^j,j=0,\dots, s$), we can prove that there exist $\bar{U},\bar{V}$
that satisfy properties analogous to \eqref{req a) of U,V}, \eqref{|V| = |Y|}, \eqref{|U| = |X|}, \eqref{U - X bound, interm} and \eqref{V - hat Y bound}, i.e.\ \begin{subequations}\label{bar U, V req}
\begin{align}
  \bar{U} \bar{V}^T  & = \Sigma,    \label{bar U, V req a)} \\
  \|\bar{U} \|_F = (1 - \eta)(1 - \bar{\eta}) \|X^0 \|_F, \; &  \; \|\bar{V} \|_F = \| Y^0 \|_F,   \label{bar U, V req b)} \\
 \| \bar{V} - Y^0 \|_F \leq 3 \sqrt{3} \bar{\eta}  \| Y^0\|_F ,  \;\;\;  &
 \| \bar{U}- X^0 \|_F \leq  \frac{12}{\sqrt{5}} \bar{\eta} \sqrt{r} \| X^0 \|_F.    \label{bar U, V req c)}  % \frac{\sqrt{21}}{2}
\end{align}
\end{subequations}

We will show that the following $U,V$ satisfy the requirements \eqref{req of U,V}:
\begin{equation}\label{U,V def based on bar}
  U \triangleq \frac{\bar{U}}{ ( 1 - \eta)(1 - \bar{\eta} ) }, \;\; V \triangleq \bar{V} ( 1 - \eta)(1 - \bar{\eta} ).
\end{equation}
The requirement \eqref{req a) of U,V} follows directly from \eqref{bar U, V req a)} and \eqref{U,V def based on bar}.
% The requirement \eqref{req b) of U,V} follows directly from
According to \eqref{bar U, V req b)}, \eqref{U,V def based on bar}
and the facts $X^0 = X$, $\| Y^0\|_F = \|\hat{Y}\|_F = \|Y\|_F/(1-\eta)$,
we have $ \| U\|_F = \frac{\|\bar{U} \|_F}{(1 - \eta)(1 - \bar{\eta} ) } = \|X^0 \|_F = \| X\|_F, $
$ \| V\|_F = \|\bar{V} \|_F(1 - \eta)(1 - \bar{\eta} ) = \|Y^0 \|_F (1-\eta) (1 - \bar{\eta} ) = \| Y\|_F (1 - \bar{\eta} ) $, thus the requirement \eqref{req b) of U,V} is proved.

It remains to prove the requirement \eqref{req c) of U,V}.
We bound $\| U-X\|_F$ as
\begin{equation}\label{U - X bound, Case 2b}
\begin{split}
&  \| U - X \|_F \leq \| U - \bar{U} \|_F + \|\bar{U} - X \|_F   \\
& \overset{\eqref{U,V def based on bar}}{ \leq } 2 \bar{\eta} \|U \|_F + \|\bar{U} - X^0 \|_F     \\
& \overset{\eqref{req b) of U,V}, \eqref{bar U, V req c)}}{\leq} 2 \bar{\eta} \|X \|_F +   \frac{12}{\sqrt{5}} \bar{\eta} \sqrt{r} \| X^0 \|_F
\leq \frac{15}{2} \bar{\eta} \sqrt{r} \| X \|_F  \\
& \leq  \frac{15}{2} \frac{\beta_T }{ \Sigma_{\rmin} } \sqrt{r} d.
\end{split}
\end{equation}
Using the fact $\hat{Y} = Y^0$, we bound $\| V-Y\|_F $ as
\begin{equation}\label{V - Y bound, Case 2b}
\begin{split}
 & \| V - Y \|_F \leq \| V - \bar{V} \|_F + \|\bar{V} - \hat{Y} \|_F + \| \hat{Y} - Y \|_F  \\
& \overset{\eqref{U,V def based on bar}, \eqref{hat(Y) - Y bound interm}}{\leq} 2 \bar{\eta} \| \bar{V} \|_F + \|\bar{V} - Y^0 \|_F + \frac{d}{\Sigma_{\rmin}} \|\hat{Y} \|_F    \\
& \overset{\eqref{bar U, V req b)}, \eqref{bar U, V req c)}}{\leq} 2 \bar{\eta} \| \hat{Y} \|_F +  3\sqrt{3} \bar{\eta}  \| Y^0\|_F + \frac{d}{\Sigma_{\rmin}} \|\hat{Y} \|_F \\
% \leq \frac{d}{\Sigma_{\min}}  \| \hat{Y} \|_F +   3 \frac{d}{\Sigma_{\min}}  \| \hat{Y}\|_F + \frac{d}{\Sigma_{\rmin}} \|\hat{Y} \|_F  \\
 & = (3 + 3 \sqrt{3}) \frac{d}{\Sigma_{\min}}  \| \hat{Y} \|_F \\
 & \overset{\eqref{eta bound}}{=} \frac{3 + 3 \sqrt{3} }{1 - \eta} \frac{d}{\Sigma_{\min}}  \| Y \|_F
 \leq  \frac{ 58 \beta_T }{ 7\Sigma_{\rmin} }  d.
\end{split}
\end{equation}
The first part of the requirement \eqref{req c) of U,V} now follows by multiplying   \eqref{U - X bound, Case 2b} and \eqref{V - Y bound, Case 2b},
and the second part follows directly from  \eqref{U - X bound, Case 2b} and \eqref{V - Y bound, Case 2b}.

\subsection{Proof of Claim \ref{claim: property of Operation 2}}\label{appen: proof of key Claim of Operation 2}
Suppose Claim \ref{claim: property of Operation 2} holds for $1,2,\dots, i-1$, we prove Claim \eqref{claim: property of Operation 2} for $i$.
By the property \eqref{official property a) of Xi,Yi} and \eqref{official property d) of Xi,Yi} of Claim \ref{claim: property of Operation 2} for $i-1$, we have
\begin{subequations}
\begin{align}
X^{i-1} (Y^{i-1})^T   & = \Sigma.  \label{XY = Sigma}   \\
\alpha_i^{i-1} \geq \alpha_i^{[0]} -   \frac{i-1}{r}\frac{1}{24}\pi \geq
&  \frac{3}{8} \pi - \frac{1}{24} \pi + \frac{1}{24r}\pi =
\frac{1}{3}\pi + \frac{1}{24 r}\pi \geq \frac{1}{3} \pi .  \label{alpha_i > 60 degree}
\end{align}
\end{subequations}
%Applying Claim \ref{claim: property of Operation 2} for $1,\dots, i-1,$ we have
%\begin{equation}\label{y_k (i-1) upper bound by ini}
% \| y_k^{i-1}\| \leq  \| y_k^{i-2}\| \leq \dots \leq \|y_k^{1} \| \leq \|y_k^0 \|, \;\;  k=1,\dots, s.
%\end{equation}

To simplify the notations, throughout the proof of Claim \ref{claim: property of Operation 2}, we denote $X^{i-1},Y^{i-1}$ as $X,Y$ and
denote $X^{i},Y^{i}$ as $X^{\prime},Y^{\prime}.$ The notations $\alpha_k^{i-1}, \alpha_k^i$ are changed accordingly to $\alpha_k, \alpha_k^{\prime}$.
% for instance,
Then \eqref{XY = Sigma} and (\ref{alpha_i > 60 degree})
 become  % and \eqref{y_k (i-1) upper bound by ini} imply
 \begin{subequations}
\begin{align}
& XY^T  = \Sigma, \\
\alpha_i  &  \geq \frac{1}{3}\pi + \frac{1}{24 r}\pi \geq \frac{1}{3}\pi.   \label{alpha_i bound, Claim 3}
% \quad \| y_k \| \leq \| y_k^0 \|, &  \;\; k=1,\dots, s. \label{y_k upper bound by ini}
\end{align}
\end{subequations}
We need to prove that $X^{\prime}, Y^{\prime}$ exist and satisfy the properties in Claim \eqref{claim: property of Operation 2},
i.e.\ (with the simplification of notations)
\begin{subequations}\label{Operation 2: property of X,Y prime}
\begin{align}
  X^{\prime} (Y^{\prime})^T & = \Sigma .    \label{property a) of X,Y prime} \\
 \|x_k^{\prime} \| = \|x_k \|, \forall k, \;  \;  \|  Y^{\prime} - Y \|_F^2
\leq \frac{4}{5} &
\frac{D_i}{\Sigma_i} ( \| Y\|_F^2 - \|Y^{\prime} \|_F^2 ) .
  \label{property b) of X,Y prime}   \\
  \|X^{\prime} - X  \|_F = \| x_i^{\prime} - x_i \| \leq  \frac{1}{\sqrt{3}} \frac{D_i}{\Sigma_{i}} \| x_i\| ,
\; & \; \|Y^{\prime} - Y  \|_F \leq \frac{2}{\sqrt{3}} \frac{D_i}{\Sigma_{i}} \|Y \|_F  .  \label{property c) of X,Y prime}  \\
 \alpha_l^{\prime}  \geq \alpha_l  - \frac{1}{r}\frac{\pi}{24} &  \geq \frac{1}{3}\pi, \;  l=i, i+1 ,\dots, s.  \label{property d) of X,Y prime} \\
\| y_k\| \geq \| y_k^{\prime}\|  \geq \| y_k \| &  - \frac{1}{10 r}\| y_k \| , \;\;  k=1, 2,\dots, s. \label{property e) of X,Y prime}  \\
\| Y\|_F^2 - \|Y^{\prime} \|_F^2 \geq  & \frac{5}{3} \frac{D_i}{\Sigma_i} \| y_i \|^2.    \label{property f) of X,Y prime}
\end{align}
\end{subequations}
% \|Y  \|_F^2 - \|Y^{\prime} \|_F^2 \geq \frac{ \|Y  - Y^{\prime}  \|_F^2}{3 \eta},
% Rotate $x_i$ in 2 dimensional subspace $span\{x_i,y_i\}$ to get $x_i'$, such that  $\langle x_i', y_i \rangle = \Sigma_i.$
% Consider the following process: starting from $q = x_i$, rotate $q$ in the 2-dimensional subspace $\rspan\{x_i,y_i\}$ towards
% $y_i$ until $q$ has the same direction as $y_i$. During this process, $\langle q, y_i \rangle $ starts from $\langle x_i, y_i \rangle < \Sigma_i$
% and gradually increases to $\|x_i \| \| y_i\|$.
% \frac{2}{\sqrt{3}}\frac{D_i}{\Sigma_i} \| y_k^0 \| &  \geq \frac{1}{\sqrt{3}}\|y_k^0 \|,

% The idea of the proof
{\black
\subsubsection{Ideas of the proof of Claim \eqref{claim: property of Operation 2}}
Before presenting the formal proof, we briefly describe its idea. % of this proof.
The goal of Operation 2 is to reduce the norm of $Y$ while keeping $\langle X, Y \rangle $ and $\| X\|_F$ invariant,
by rotating and shrinking $x_i$, $y_k, k=1,\dots, K$ (note that $x_j, \forall j\neq i,$ do no change).
% Operation 2 can be summarized as follows.
 We first rotate $x_i$ and shrink $y_i$ at the same time so that the new inner product $\langle x_i^{\prime}, y_i^{\prime} \rangle$ equals the previous
 one $\langle x_i, y_i \rangle$ (this step can be viewed as a combination of two steps: first rotate $x_i$ to increase the inner product, then
 shrink $y_i$ to reduce the inner product).
 In order to preserve the orthogonality of $X$ and $Y$, we need to rotate $y_j, \forall j \neq i,$ so that the new $y_j^{\prime}$
is orthogonal to $x_i^{\prime}$.

Although the above procedure is simple, there are two questions to be answered.
% A natural question arises:
The first question is: will the inner product $\langle x_j , y_j \rangle$ increase as we rotate $y_j$, for all $j\neq i$? % be larger than $\langle x_j, y_j \rangle$?
If yes, we could first rotate and then shrink $y_j$ to obtain $y_j^{\prime}$ so that the new inner product $\langle x_j, y_j^{\prime} \rangle $
equals $\langle x_j, y_j \rangle$, which achieves the goal of Operation 2.
% To answer this question is probably the most nontrivial and challenging part of the proof.
% Fortunately,
By resorting to the geometry (in a rigourous way) we are able to provide an affirmative answer to the above question.
To gain an intuition why this is possible,  % answer should be affirmative,
 we use Figure \ref{FigRotate} to illustrate. Consider the case $i=2$
and rotate $x_2$ towards $y_2$ to obtain $x_2^{\prime}$, then $y_1$ has to be rotated so that $y_1^{\prime} $ is orthogonal to $x_2^{\prime}$.
It is clear from this figure that the angle between $y_1$ and $x_1$ also decreases, or equivalently, the inner product $\langle x_1, y_1 \rangle$
also increases. One might ask whether we have utilized additional assumptions on the relative positions of $x_i, y_i$'s. In fact,
we do not utilize additional assumptions; what we implicitly utilize is the fact that $\langle x_i, y_i \rangle > 0, \forall i$ (see
Figure \ref{Fig.2}, Figure \ref{Fig.2'} and the
paragraph after \eqref{yj-yj' bot xk} for detailed explanations).
% (which holds due to the assumption $D_i < \Sigma_{\rmin} \leq \Sigma_i$)
  \begin{figure}[ht]
  \centering
{ \includegraphics[width=6.5cm,height=3.5cm]{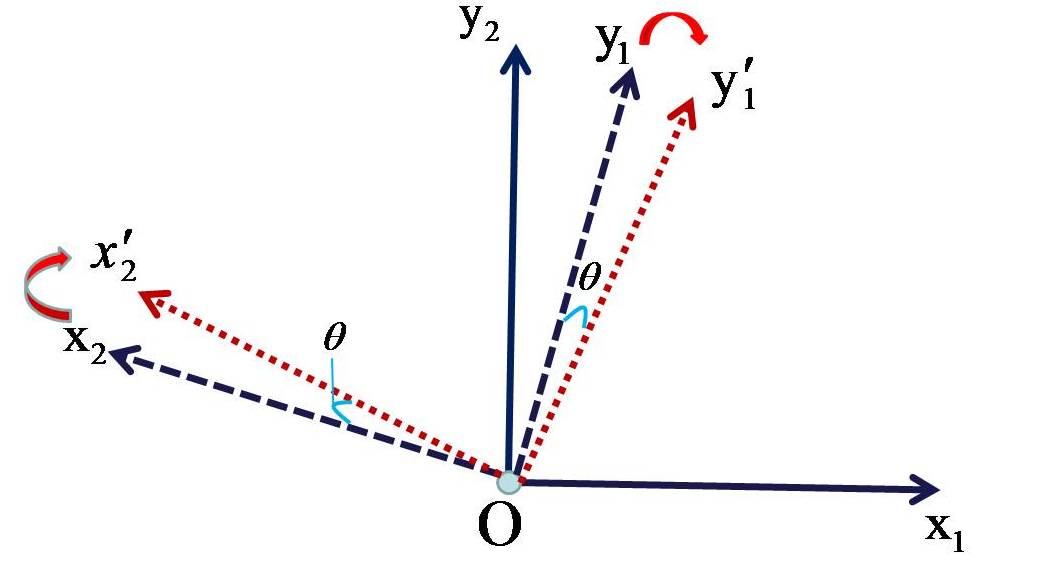} }
  \vspace{-.2cm}
   \caption{
   }  \label{FigRotate} \vspace{-.3cm}
\end{figure}

The second question is: will the angle $\alpha_j^{\prime} = \angle(x_j, y_j^{\prime})$ still be larger than, say, $\frac{1}{3}\pi$, for all $ j > i$?
If yes, then we can apply Operation 2 repeatedly for all $i =1, 2, \dots, s$.
To provide an affirmative answer, we should guarantee that each angle decreases at most $ \frac{1}{s}(\frac{3}{8}\pi - \frac{1}{3}\pi) = \frac{1}{24s}\pi  $,
i.e.\ $\angle(x_j, y_j^{\prime}) \geq \angle(x_j, y_j) - \frac{1}{24s }\pi , \forall \ i < j \leq s$.
% The amount of decrease in $\angle(x_i, y_i)$ is directly related to $D_i$, thus should be
Unlike the first question which can be answered by reading Figure \ref{Fig.2} and Figure \ref{Fig.2'}, this question cannot be answered by just
reading figures.
 We make some algebraic computation to obtain the following result: under the assumption that $\alpha_i$ is no less than $\alpha_j$,
 during Operation 2 the amount of decrease in $\alpha_j$ is upper bounded by the amount of decrease in $\alpha_i$, which can be further bounded above by $\frac{1}{24s }\pi$.
This result explains why our proof requires the assumption $\alpha_i \geq \alpha_j, \forall \ i < j \leq s$, i.e.\ \eqref{assumption of alpha i being smallest}.
% How to guarantee $\alpha_i \geq \alpha_j, \forall i < j \leq s$?
% One simple way (possibly the only way) is the following: after applying Operation 2 for $1, 2, \dots, i-1$,
% pick the pair with the largest angle and switch it with the $i$-th pair.
% This explains why we have the assumption \eqref{assumption of alpha i being smallest}
% By just looking at the figures, it is not
% It is not clear at this point why would all angles d
% $\angle(x_j, y_j^{\prime})$

% According to Figure \ref{FigRotate}, it

% To better understand the result, notice that when we rotate $x_i$ in the plane $\mathrm{span}(x_i, y_i)$, all other $y_j$'s should be rotated
% We prove the following fact: rotating $x_i$ towards $y_i$ will push all other $y_j$ to be closer to $x_j$.

% Our goal is to rotate $x_i$ and $y_j$'s so  (and a bit surprisingly)
}

\subsubsection{Formal proof of Claim \eqref{claim: property of Operation 2}}
We first show how to define $x_i^{\prime}$ and $y_i^{\prime}$. Note that
\begin{equation}\label{x_i y_i bound}
\|x_i \| \|y_i \| = \frac{\langle x_i, y_i \rangle }{\cos \alpha_i}
 \geq \frac{\Sigma_i}{\cos(\frac{\pi}{3})} = 2 \Sigma_i .
 \end{equation}
Since \eqref{x_i y_i bound} implies $ \frac{ \Sigma_i + D_i }{  \|x_i \| \|y_i \| } \leq \frac{ 2 \Sigma_i  }{  \|x_i \| \|y_i \| } \leq 1$, we can define
 \begin{equation}\nonumber%\label{def of alpha'}
   \alpha_i^{\prime} \triangleq \arccos( \frac{ \Sigma_i + D_i }{  \|x_i \| \|y_i \| } ) \in [0, \frac{\pi}{2}].
 \end{equation}
There is a unique $x_i^{\prime}$ in the plane $\rspan\{x_i,y_i\}$ which satisfies
\begin{equation}\label{x_i norm invariant}
\|x_i^{\prime} \| = \|x_i \|
\end{equation}
 and $  \angle (x_i^{\prime} , y_i) = \alpha_i^{\prime}$.
 By the definition of $\alpha_i^{\prime}$ above, we have %in \eqref{def of alpha'}
 \begin{equation}\nonumber%\label{<x_i,y_i> = Sigma_i + D_i}
    \langle x_i^{\prime}, y_i \rangle = \Sigma_i + D_i.
 \end{equation}
 The existence of $x_i^{\prime}$ is proved.
We define
\begin{equation}\label{y_i prime def}
y_i^{\prime} \triangleq \frac{ \Sigma_i }{ \Sigma_i + D_i }  y_i,
\end{equation}
then
\begin{equation}\label{<x_i',y_i'> = Sigma_i}
 \langle x_i^{\prime}, y_i^{\prime} \rangle =  \frac{ \Sigma_i }{ \Sigma_i + D_i } \langle x_i^{\prime}, y_i \rangle = \Sigma_i.
 \end{equation}
  The existence of $y_i^{\prime}$ is also proved.

Since $ 0< \langle x_i, y_i\rangle = \Sigma_i < \langle x_i', y_i \rangle$, we have $\frac{\pi}{2} > \alpha_i > \alpha_i' >0, $ thus we can define
 \begin{equation}\label{theta definition}
 \theta \triangleq \alpha_i - \alpha_i' = \angle (x_i^{\prime}, x_i ) \in (0, \alpha_i).
 \end{equation}

Fix any $j \neq i$, we then show how to define $ y_j^{\prime}. $
Define $$ A_i \triangleq \rspan_{j\neq i}\{x_j\} \bot y_i, \; \; B_i \triangleq \rspan_{j\neq i}\{y_j\} \bot x_i, \; \; T_i \triangleq A_i \cap B_i.$$
Let $\overrightarrow{OY_j} = y_j$, $K_j \triangleq \mathcal{P}_{A_i}(Y_j), H_j \triangleq \mathcal{P}_{T_i}(Y_j).$
Then $\angle Y_jH_jK_j = \min \{  \angle(x_i,y_i) , \pi - \angle(x_i,y_i) \} = \angle(x_i,y_i) = \alpha_i.$
Since $\alpha_i > \theta$, there exists a unique point $Y_j'$ in the line segment $Y_j K_j$ such that
\begin{equation}\label{Y_j' position with angle theta}
\angle Y_jH_jY_j' = \theta.
\end{equation}
Since $K_j = \mathcal{P}_{A_i}(Y_j) $ and $x_k \in A_i, \forall k\neq i$, we have $ \overrightarrow{Y_j K_j} \bot x_k, \forall k\neq i $, thus
\begin{equation}\label{yj-yj' bot xk}
\overrightarrow{ Y_j Y_j'} \bot x_k, \;\; \forall k\neq i.
 \end{equation}

\begin{figure}[ht]
  \centering
{ \includegraphics[width=10cm,height=5cm]{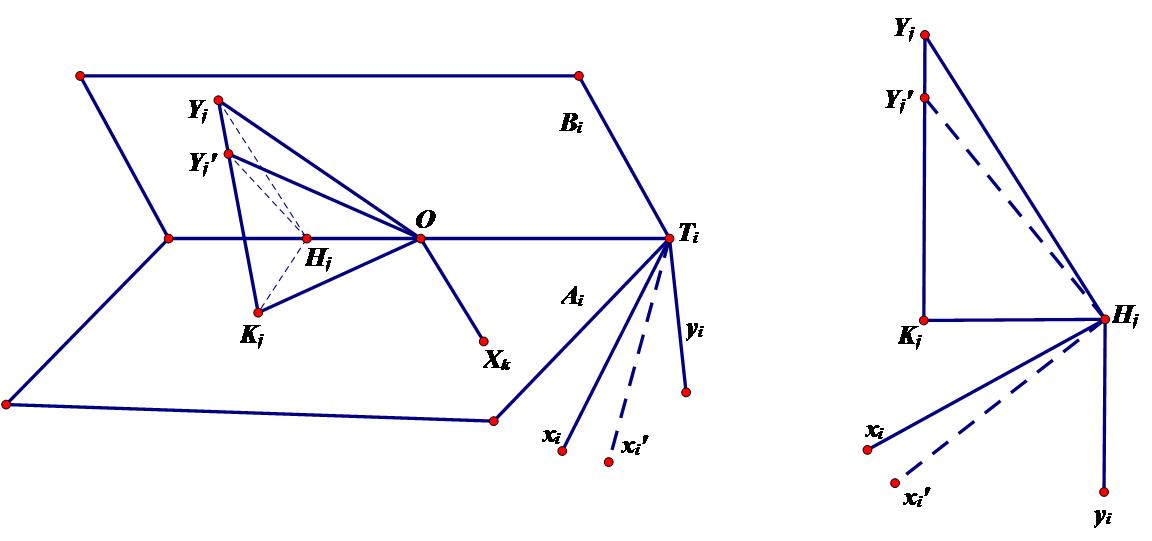} }
  \vspace{-.2cm}
   \caption{ Left: Space $A_i,B_i, T_i$, vectors $x_i,y_i, x_i^{\prime}, x_k$ and some points related to $y_j$.
  Right: Some points and vectors in plane $ H_j Y_j K_j =  T_i^{\bot} = \rspan \{x_i,y_i \}$.
 This figure shows the first possibility: $x_i$ and $K_j$ lie in the same side of line $H_j Y_j$.
   }  \label{Fig.2} \vspace{-.3cm}
\end{figure}

\begin{figure}[ht]
  \centering
{ \includegraphics[width=10cm,height=5cm]{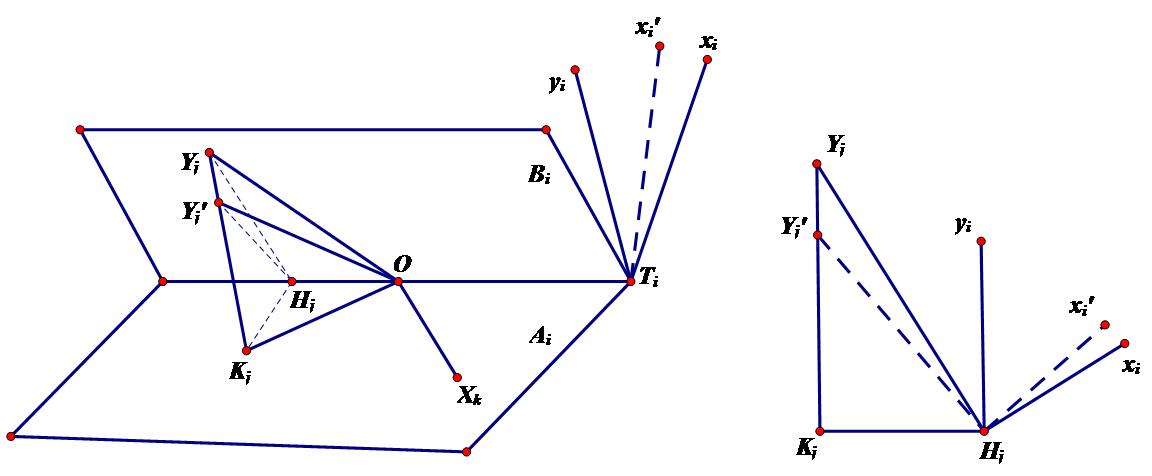} }
  \vspace{-.2cm}
   \caption{ Same objects as in Figure \ref{Fig.2}, but for the second possibility: $x_i$ and $K_j$ lie in different sides of line $H_j Y_j$.    }  \label{Fig.2'} \vspace{-.3cm}
\end{figure}

See Figure \ref{Fig.2} and Figure \ref{Fig.2'} for the geometrical interpretation;
note that $T_i$ in general is not a line but a $r-2$ dimensional space.
 The righthand side subfigures represents the 2 dimensional subspace $T_i^{\bot}$;
since $\rspan \{H_jY_j, H_j K_j \} = T_i^{\bot} = \rspan \{x_i,y_i \}, $ we can draw $x_i, y_i, y_i^{\prime}$ as the vectors starting from $H_j$ and lying in the plane $ H_j Y_j K_j =  T_i^{\bot} $ in the figures.
 Figure \ref{Fig.2} and Figure \ref{Fig.2'} differ in the relative position of $x_i$ and $K_j$:
 $x_i$ and $K_j$ lie in the same side of line $H_j Y_j$ in Figure \ref{Fig.2} but in different sides in Figure \ref{Fig.2'}.
  Given the positions of $x_i$ and $H_j,Y_j,K_j$, the position of $y_i$ is determined since $y_i \bot \overrightarrow{H_j K_j}$ and $\angle (x_i, y_i) < \frac{\pi}{2} $.
%$\angle (x_i, \overrightarrow{H_i K_i} ) $ is less than $\frac{\pi}{2}$ in Figure \ref{Fig.2} and greater than $\frac{\pi}{2}$ in Figure \ref{Fig.2'}.

   %   $\angle (x_i, \overrightarrow{H_i K_i} ) $ is less than $\frac{\pi}{2}$.

% \overrightarrow{H_iK_i}

In both figures, we have
\begin{align}
  & \angle (\overrightarrow{ H_j Y_j'}, x_i^{\prime} ) = \angle (\overrightarrow{ H_j Y_j}, x_i ) - \angle (x_i^{\prime}, x_i )
+ \angle Y_j H_j Y_j' \nonumber \\
& \xlongequal{\eqref{theta definition},\eqref{Y_j' position with angle theta} } \frac{\pi}{2} - \theta + \theta = \frac{\pi}{2},  \nonumber  \\
 \Longrightarrow  & \quad \quad \overrightarrow{ H_j Y_j'} \bot x_i'.   \label{H_j Y_j' ortho x_i'}
 \end{align}

Now we are ready to define $y_j^{\prime} $ and establish its properties. Define
\begin{equation}\label{def of y_j'}
\begin{split}
y_j^{\prime} \triangleq \overrightarrow{OY_j^{\prime}}.
\end{split}
\end{equation}
Since $Y_j^{\prime}$ lies in the line segment $K_jY_j$ and $\angle Y_j K_j O = \pi/2$, we have
\begin{equation}\label{y_j' < y_j}
\| y_j^{\prime} \| \leq \| y_j \|.
\end{equation}
We also have
\begin{equation}\label{y_j prime ortho x_k}
  y_j^{\prime} = y_j + \overrightarrow{ Y_j Y_j'} \in \rspan \{ y_j, y_i \} \; \bot x_k, \; \forall k\neq i,j.
\end{equation}
According to the fact $\overrightarrow{OH_j} \bot x_i^{\prime}$ and (\ref{H_j Y_j' ortho x_i'}), we have
\begin{equation}\label{y_j prime ortho x_i'}
  y_j^{\prime} = \overrightarrow{OH_j} + \overrightarrow{ H_j Y_j'} \; \bot \; x_i'.
\end{equation}
Let $k =j$ in \eqref{yj-yj' bot xk}, we obtain
\begin{equation}\label{inner product of xj,yj stay the same}
 0 = \langle \overrightarrow{ Y_j Y_j'}, x_j \rangle = \langle  y_j^{\prime} - y_j, x_j \rangle = 0 \Longrightarrow \langle x_j, y_j^{\prime} \rangle =
  \langle x_j, y_j \rangle.
\end{equation}
We have shown that $y_j^{\prime}$ defined in \eqref{def of y_j'} satisfies \eqref{y_j prime ortho x_k}, \eqref{y_j prime ortho x_i'} and \eqref{inner product of xj,yj stay the same}, thus the existence of $y_j^{\prime}$ in Operation 2 is proved.
% which means that this $y_j^{\prime}$ is the one desired in Operation 3. T
%According to the fact $y_j \bot x_k, \forall k\neq i,j$ and \eqref{yj-yj' bot xk}, we have
%\begin{equation}\label{y_j prime ortho x_i}
%  y_j^{\prime} = y_j + \overrightarrow{ Y_j Y_j'} \; \bot \; x_k, \forall k\neq i,j.
%\end{equation}

% Let $X' = (x_1',\dots,x_r)^T, Y' = (y_1',\dots,y_r')^T.$

Having defined $x_i^{\prime}, y_i^{\prime}$ and $y_j^{\prime}, \forall j \neq i$, we further define
\begin{equation}\label{other x_j stay, Claim 3 proof}
x_j^{\prime} \triangleq x_j, \forall j \neq i,
\end{equation}
which completes the definition of $X^{\prime},Y^{\prime}$.
In the rest, we prove that $X^{\prime},Y^{\prime}$ satisfy the desired property \eqref{Operation 2: property of X,Y prime}.

The property \eqref{property a) of X,Y prime} can be directly proved by the definitions of $X^{\prime},Y^{\prime}$.
In specific, according to \eqref{<x_i',y_i'> = Sigma_i}, \eqref{inner product of xj,yj stay the same} and the definition \eqref{other x_j stay, Claim 3 proof},
we have $\langle x_k^{\prime}, y_k^{\prime} \rangle = \Sigma_k, \forall k$.
According to the definitions \eqref{other x_j stay, Claim 3 proof}, \eqref{y_i prime def} and the fact $ y_i \bot x_j, \forall j\neq i$,
we have $ y_i^{\prime} \bot x_j^{\prime}, \forall j \neq i$.
Together with  \eqref{y_j prime ortho x_k} and \eqref{y_j prime ortho x_i'},
we obtain $ \langle x_k^{\prime}, y_l^{\prime} \rangle = 0, \forall k \neq l$.
Thus $  X^{\prime} (Y^{\prime})^T = \Sigma. $

% Property (c) is due to the definition \eqref{other x_j stay, Claim 3 proof}, $\|x_i' \| = \|x_i \|$ and
%\begin{equation}\label{y_j' smaller than y_j}
%\|y_j' \| \leq \|y_j\|, \forall j\neq i,
%\end{equation}
%where \eqref{y_j' smaller than y_j} is because $Y_j^{\prime}$ lies in the line segment $Y_j K_j $ and $Y_j K_j \bot OK_j$.
%Property (e) is due to
Next, we prove the property \eqref{property d) of X,Y prime}. % in \eqref{propertis of Claim 3, simplified form}.
We first prove % $\alpha_i' \geq \alpha_i - \frac{1}{r}\frac{\pi}{24}$, which can be rewritten as
\begin{equation}\label{theta bound we want}
\alpha_i' - \alpha_i = \theta \leq \frac{1}{r}\frac{\pi}{24}.
\end{equation}
Define $h_i \triangleq x_i' - x_i,$ then
\begin{equation}\label{h_i expression 1}
\|h_i \| = 2 \|x_i \|\sin(\frac{\theta}{2}).
\end{equation}
From $\langle x_i', y_i \rangle = \Sigma_i + D_i = \langle x_i,y_i \rangle + D_i,$ we obtain
$\langle h_i, y_i \rangle = D_i.$ Note that $\langle h_i, y_i \rangle = \|h_i \| \|y_i \| \cos(\angle(h_i,y_i))$ and $\angle(h_i,y_i)= \frac{\pi}{2} - \alpha_i + \frac{\theta}{2},$ thus
\begin{equation}\label{h_i expression 2}
\|h_i \| = \frac{D_i}{\|y_i \| \sin( \alpha_i - \frac{\theta}{2}) }.
\end{equation}
According to (\ref{h_i expression 1}) and (\ref{h_i expression 2}), we have
\begin{align*}
\frac{D_i}{\| x_i\| \| y_i\|} = 2 \sin(\alpha_i - \frac{\theta}{2} ) \sin (\frac{\theta}{2}) \geq  2 \sin(\frac{\alpha_i}{2})\sin(\frac{\theta}{2}) \\
\geq 2 \sin(\frac{\pi}{6})\sin (\frac{\theta}{2})= \sin (\frac{\theta}{2}) \geq \frac{\theta}{\pi},
\end{align*}
where the last equality follows from the fact that $\frac{\sin(t)}{t}$ is decreasing in $t \in (0,\frac{\pi}{2}]$.
Note that $\frac{D_i}{\| x_i\| \| y_i\|} $ can be upper bounded as
\begin{equation}\nonumber % \label{theta bound 2}
\frac{D_i}{\| x_i\| \| y_i\|} \overset{\eqref{x_i y_i bound}}{\leq} \frac{ D_i}{ 2 \Sigma_i} \overset{\eqref{D_i bound by Sigma_i/10}}{\leq} \frac{1}{24r}.  % \overset{\eqref{alpha_i and D_i bound, Claim 3}}{\leq} \frac{\sqrt{r}d }{\Sigma_{\mathrm{min}}}
\end{equation}
% Remark: Could be strengthend to 1/20r, and further to 1/(20 $\sqrt{3}$ r)}
% where the last inequliaty follows from
% {\black Need to cite, d $<$ ?? }
Combining the above two relations, we get (\ref{theta bound we want}).  % (\ref{theta bound 1}) and (\ref{theta bound 2})

To prove
\begin{equation}\label{alpha_j diff bound}
\alpha_j - \alpha_j^{\prime} \leq \frac{\pi}{24 r}, \forall j \in \{ i+1, \dots, s\} ,
\end{equation}
 we only need to prove
\begin{equation}\label{theta_j <= theta}
 \theta_j \triangleq \alpha_j - \alpha_j^{\prime}  \leq \theta, \; \; \forall j  \in \{ i+1, \dots, s\}
\end{equation}
and then use \eqref{theta bound we want}.
%Define $ \varphi_j \triangleq \angle(x_j, \overrightarrow{OK_j}) $. Since $OK_j = \mathcal{P}_{A_i}(Y_j)$, we have
%$ \langle y_j, x_j \rangle = \langle \overrightarrow{OK_j}, x_j \rangle$ %= \| \overrightarrow{OK_j} \| \|x_j \| \cos(\varphi_j) $.
%and
The equality \eqref{inner product of xj,yj stay the same} implies that $ \|x_j \| \|y_j \| \cos(\alpha_j) = \|x_j \| \|y_j^{\prime} \| \cos(\alpha_j^{\prime})$,
which leads to
\begin{equation}\nonumber % \label{cos/cos 1st}
  \frac{\cos(\alpha_j)}{\cos(\alpha_j- \theta_j)} = \frac{\cos(\alpha_j)}{\cos(\alpha_j^{\prime})} = \frac{\|y_j^{\prime} \|}{\|y_j \|}.
\end{equation}
For any two points $P_1, P_2$, we use $|P_1 P_2|$ to denote the length of the line segment $P_1P_2$. Since $\overrightarrow{OH_j} $ is orthogonal to plane $H_j K_j Y_j$,
we have
\begin{equation}\nonumber % \label{cos/cos 2nd}
  \frac{ \|y_j^{\prime} \|^2 }{ \|y_j \|^2 } = \frac{ |OH_j|^2 + |H_j Y_j^{\prime}|^2  }{ |OH_j|^2 + |H_j Y_j|^2 }
  \geq \frac{ |H_j Y_j^{\prime}|^2  }{  |H_j Y_j|^2 } ,
\end{equation}
where the last inequality follows from the fact that $|H_j Y_j^{\prime}| \leq |H_j Y_j|  $.
Since $\angle Y_j H_j K_j = \alpha_i, \angle Y_j^{\prime}H_j K_j = \alpha_i^{\prime}$ and $\angle Y_j K_j H_j = \frac{\pi}{2}$,
 we have
 \begin{equation}\nonumber % \label{cos/cos 3rd}
 \frac{|H_j Y_j^{\prime}|}{ |H_j Y_j|} = \frac{\sin \angle Y_j^{\prime} Y_j H_j }{\sin \angle Y_j Y_j^{
 \prime} H_j } = \frac{\sin(\pi/2 - \alpha_i)}{ \sin(\pi/2 + \alpha_i^{\prime} ) }  = \frac{\cos(\alpha_i)}{\cos(\alpha_i^{\prime})}.
\end{equation}
 According to the assumption \eqref{assumption of alpha i being smallest} and $i < j \leq s $, we have $0 \leq \alpha_i \leq \alpha_j \leq \frac{\pi}{2}$.
 Since $\cos( x) / \cos(x - \theta) $ is decreasing in $[0, \frac{\pi}{2}]$, we can get
\begin{equation}\nonumber % \label{cos/cos 4th}
\frac{\cos(\alpha_i)}{\cos(\alpha_i^{\prime})} = \frac{\cos(\alpha_i )}{\cos(\alpha_i -\theta)} \geq \frac{\cos(\alpha_j )}{\cos(\alpha_j -\theta)}.
\end{equation}
Combining the above four relations, we get % \eqref{cos/cos 1st},\eqref{cos/cos 2nd},\eqref{cos/cos 3rd},\eqref{cos/cos 4th}, we get
\begin{equation}\nonumber %\label{cos/cos 5th}
  \frac{\cos(\alpha_j)}{\cos(\alpha_j- \theta_j)} \geq \frac{\cos(\alpha_j )}{\cos(\alpha_j -\theta)},
\end{equation}
which implies $\cos(\alpha_j -\theta) \geq \cos(\alpha_j- \theta_j) $ that immediately leads to \eqref{theta_j <= theta}.
Thus we have proved \eqref{alpha_j diff bound}, which combined with \eqref{theta bound we want} establishes the property \eqref{property d) of X,Y prime}.

Then we prove the property \eqref{property c) of X,Y prime}.
Since $x_j^{\prime} = x_j, \forall j \neq i$, we have $ \|X' - X  \|_F = \|x_i^{\prime} - x_i\| $, which can be bounded as
\begin{align*}
\|x_i^{\prime} - x_i\|  = \|h_i \| \overset{\eqref{h_i expression 2}}{=} \frac{D_i}{\|y_i \| \sin( \alpha_i - \frac{\theta}{2}) }
\leq \frac{\|x_i \| D_i }{\|x_i \|\|y_i \| \sin( \frac{\pi}{3} ) }   \\
 \overset{\eqref{x_i y_i bound}}{\leq} \frac{\|x_i \| D_i}{ 2 \Sigma_i \sin(\frac{\pi}{3})  } < \frac{1}{\sqrt{3}} \frac{\|x_i \|}{\Sigma_{i}}D_i,
% {\blue \leq \frac{1}{\sqrt{3}}\frac{\beta_T}{\Sigma_{\min}}D_i}.   \overset{\eqref{theta bound we want}}{\leq}   \frac{1}{ \sqrt{3}}
\end{align*}
where the first inequality is due to
\begin{equation}\label{alpha - theta bound}
\alpha_i - \theta/2 \geq \alpha_i - \theta \overset{\eqref{alpha_i bound, Claim 3}}{\geq} \pi/3 + \pi/24 - \theta \overset{\eqref{theta bound we want}}{\geq} \pi/3 .
\end{equation}
Thus the first part of \eqref{property c) of X,Y prime} is proved.

According to (\ref{h_i expression 1}) and (\ref{h_i expression 2}), we have
\begin{equation}\label{sin theta/2 expression}
 2 \sin(\frac{\theta}{2}) = \frac{D_i}{\| x_i\| \| y_i\| \sin(\alpha_i - \frac{\theta}{2} )}
\end{equation}
Now we upper bound $\| y_j^{\prime} - y_j\|$ as
% Using $\frac{D_i}{\| x_i\| \| y_i\|} = 2 \sin(\alpha_i - \frac{\theta}{2} ) \sin (\frac{\theta}{2}) $, we have
\begin{equation}\label{diff y_j bound by HY}
\begin{split}
& \|y_j' - y_j \| = |Y_j' Y_j| \\
 & = \frac{\sin(\theta)}{\cos(\alpha_i - \theta )} |H_j Y_j|  \\
% & \leq \frac{\sin(\theta)}{\cos(\alpha_i - \theta )} \|y_j \|  \\
 &   =   2 \sin(\frac{\theta}{2})\cos(\frac{\theta}{2})\frac{ 1 }{\cos(\alpha_i - \theta )}  |H_j Y_j|  \\
  &  \overset{\eqref{sin theta/2 expression}}{=}  \frac{D_i}{\| x_i\| \| y_i\| \sin(\alpha_i - \frac{\theta}{2} )}  \cos(\frac{\theta}{2})\frac{ 1 }{\cos(\alpha_i - \theta )}  |H_j Y_j|   \\
 % \overset{\eqref{h_i expression 1}}{=}  \frac{ \|h_i\|}{ \| x_i\| } \cos(\frac{\theta}{2}) \frac{ 1 }{\cos(\alpha_i - \theta )}\|y_j \|
 % \leq 2 \frac{ \|h_i\|}{ \| x_i\| } \|y_j \|
 % \overset{\eqref{h_i upper bound}}{\leq} \frac{1}{\sqrt{3}}\frac{\|y_j \|}{\Sigma_{i}}D_i,
 & \leq  \frac{D_i}{\| x_i\| \| y_i\| \sin(\alpha_i - \frac{\theta}{2} )} \frac{ 1 }{\cos(\alpha_i )}  |H_j Y_j|  \\
 &  \overset{ \eqref{alpha - theta bound} }{\leq}  \frac{D_i}{\sin(\frac{\pi}{3}) \langle x_i, y_i \rangle }  |H_j Y_j|  \\
 % &  \leq \frac{2}{\sqrt{3}} \frac{D_i}{\Sigma_i} |H_j Y_j|    \\
  &  \leq \frac{2}{\sqrt{3}} \frac{D_i}{\Sigma_i}  |H_j Y_j| ,
\end{split}
\end{equation}
where the last inequality is due to the fact $ \langle x_i, y_i \rangle = \Sigma_i$.
Using $|H_j Y_j| \leq \|y_j \| $, we obtain
\begin{equation}\label{y_j diff bound}
\|y_j' - y_j \| \leq  \frac{2}{\sqrt{3}} \frac{D_i}{\Sigma_i}  \|y_j \|.
\end{equation}
% , and the second last inequality is due to \eqref{alpha - theta bound}.
%
%% If we do not use the inequality $|H_j Y_j| \leq \|y_j \| $, we can strengthen \eqref{y_j diff bound} to
%The following intermediate result will be useful later:
%\begin{equation}\label{diff y_j bound by HY}
%\|y_j' - y_j \| \leq \frac{2}{\sqrt{3}} \frac{D_i}{\Sigma_i} |H_jY_j|. % = \frac{2}{\sqrt{3}} \frac{D_i}{\Sigma_i} \| y_j \| \sin(\gamma_j).
%% \frac{\|y_j' - y_j \|}{ \|y_j \| } = \frac{\|y_j' - y_j \|}{ |H_j Y_j| }   \frac{ |H_j Y_j| }{ \|y_j \| }
%\end{equation}
According to the definition \eqref{y_i prime def}, we have
\begin{equation}\label{y_i diff bound}
 \| y_i - y_i^{\prime} \|
=  (1 - \frac{\Sigma_i}{\Sigma_i + D_i}) \|y_i \|
= \frac{ D_i }{ \Sigma_i + D_i }\| y_i \| \leq  \frac{D_i}{\Sigma_i} \| y_i \|.
\end{equation}
% where the last inequliaty follows from \eqref{D_i bound by Sigma_i/10}.
According to \eqref{y_j diff bound} (which holds for any $j \in \{1,\dots, r \}\backslash \{i \}$) and \eqref{y_i diff bound},
we get
\begin{equation}\nonumber % \label{Y diff bound}
 \| Y - Y^{\prime} \|_F
 = \sqrt{ \sum_{k=1}^r \| y_k - y_k^{\prime} \|^2 }
  \leq \frac{2}{\sqrt{3}} \frac{D_i}{\Sigma_i} \sqrt{ \sum_{k=1}^r  \| y_k \|^2} =  \frac{2}{\sqrt{3}} \frac{D_i}{\Sigma_i}  \| Y\|_F,
\end{equation}
which proves the second part of \eqref{property c) of X,Y prime}.

The property \eqref{property e) of X,Y prime} can be proved as follows.
%\eqref{y_j' < y_j}, \eqref{y_i prime def}, \eqref{y_j diff bound} and \eqref{y_i diff bound}. In fact,
By the definition \eqref{y_i prime def}, we have $ \| y_i^{\prime}\| \leq \| y_i\|$, which combined with \eqref{y_j' < y_j} (for all $j \neq i$) leads to
\begin{equation}\nonumber % \label{y_k prime upper bound}
 \| y_k^{\prime} \| \leq \|y_k \|, \;\; k=1,\dots, s.
\end{equation}
According to \eqref{y_j diff bound} (for all $j\neq i$) and \eqref{y_i diff bound}, we have
 $\|y_k^{\prime} - y_k \| \leq \frac{2}{\sqrt{3}}\frac{D_i}{\Sigma_i} \| y_k \|, \forall k$, which implies
\begin{equation}\nonumber % \label{y_k prime lower bound}
  \|y_k^{\prime}\| \geq \| y_k\| - \|y_k^{\prime} - y_k \|  \geq \| y_k\| - \frac{2}{\sqrt{3}}\frac{D_i}{\Sigma_i} \| y_k \| \overset{\eqref{D_i bound by Sigma_i/10}}{\geq} \| y_k\| - \frac{1}{10 r} \|y_k \|, \; \forall k.
\end{equation}
Combining the above two relations we obtain the property \eqref{property e) of X,Y prime}. %\eqref{y_k prime upper bound} and \eqref{y_k prime lower bound}

The property \eqref{property f) of X,Y prime} can be easily proved by \eqref{y_i prime def}.
In fact, we have % it follows from \eqref{y_i prime def} that
\begin{equation}\label{square diff}
\begin{split}
 \| y_i\|^2 -  \| y_i^{\prime} \|^2 = (\| y_i\| -  \| y_i^{\prime} \|)(\| y_i\| +  \| y_i^{\prime} \|)  \\
\geq 2 \| y_i^{\prime}\|(\| y_i\| -  \| y_i^{\prime} \|) \overset{\eqref{y_i prime def}}{=}  2 \| y_i^{\prime}\| (\frac{ \Sigma_i + D_i}{\Sigma_i} - 1) \| y_i^{\prime} \|   \\
= 2 \frac{D_i}{\Sigma_i} \| y_i^{\prime}\|^2
\geq 2 \frac{D_i}{\Sigma_i}  (\frac{11}{12 })^2 \|y_i \|^2
\geq \frac{5}{3}\frac{D_i}{\Sigma_i}  \|y_i \|^2 .
\end{split}
\end{equation}
where the second last inequliaty follows from $ \| y_i^{\prime} \| \geq \| y_i\| - \| y_i - y_i^{\prime}\| \overset{\eqref{y_i diff bound}}{\geq}  \| y_i\| - D_i \|y_i \|/\Sigma_i \overset{\eqref{D_i bound by Sigma_i/10}}{\geq} 11 \| y_i\|/12. $
According to \eqref{y_j' < y_j} (for all $j \neq i$), we have $ \| Y\|_F^2 -  \| Y^{\prime} \|_F^2 \geq  \| y_i\|^2 -  \| y_i^{\prime} \|^2$,
which combined with \eqref{square diff} leads to the property \eqref{property f) of X,Y prime}.

% Here, in the last inequality we use
% $ \frac{2}{\sqrt{3}} \frac{D_i}{\Sigma_i} \| y_k \| \overset{\eqref{D_i bound by Sigma_i/10}}{\leq} \frac{2}{\sqrt{3}} \frac{2}{24r} \| y_k\| = \frac{1}{10 r} \|y_k \| .$
%Plugging this inequality into \eqref{y_k prime lower bound interm}, we get
%\begin{equation}\label{y_k prime lower bound final}
%  \|y_k^{\prime}\|
%\end{equation}

\begin{figure*}[htbp]
  \centering
{ \includegraphics[width=12cm,height=5cm]{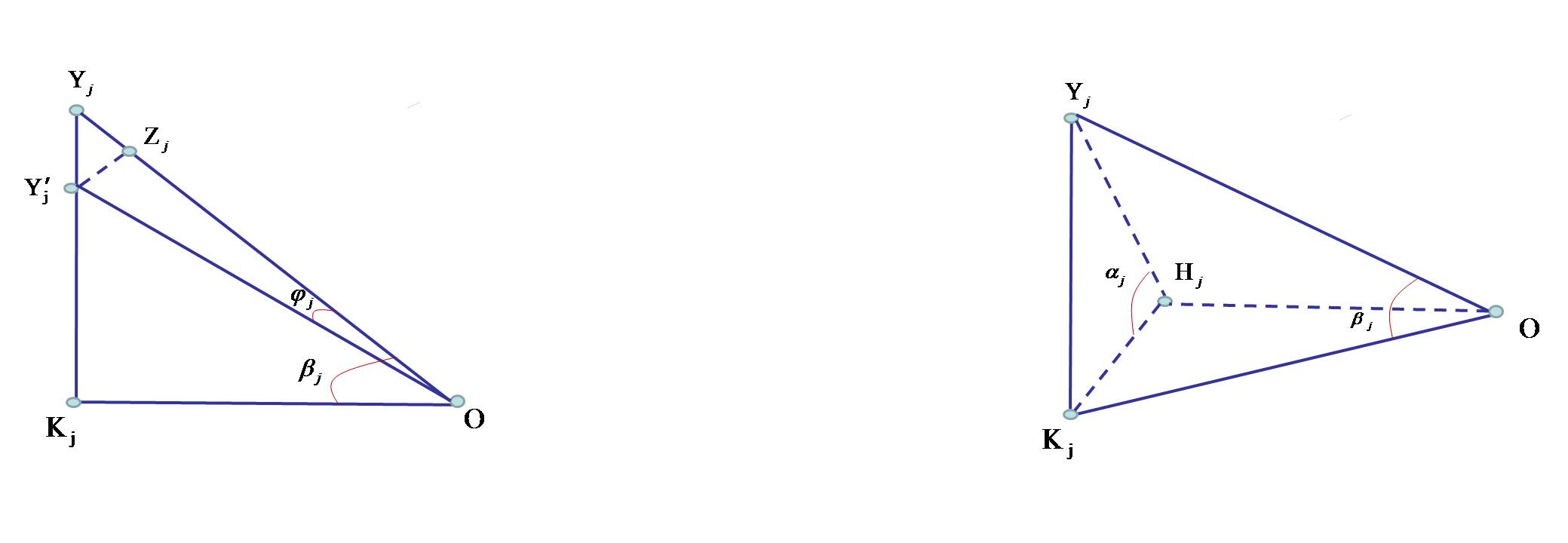} }
  \vspace{-.2cm}
   \caption{ Illustration for the proof of the property \eqref{property b) of X,Y prime}
   }  \label{Fig.3} \vspace{-.3cm}
\end{figure*}

At last, we prove the property \eqref{property b) of X,Y prime}.
The first part $ \| X^{\prime}\|_F = \| X\|_F $ follows from \eqref{x_i norm invariant} and \eqref{other x_j stay, Claim 3 proof},
thus it remains to prove the second part.
Denote $\varphi_j \triangleq \angle Y_j O Y_j^{\prime} , \beta_j \triangleq \angle Y_j O K_j$ as shown in Figure \ref{Fig.3}.
Pick a point $Z_j$ in the line segment $OY_j$ so that $|OZ_j| = |OY_j^{\prime}|$, then $|Y_j Z_j| = \| y_j\| - \|y_j^{\prime} \|$.
Thus we have
\begin{equation}\label{diff of y_j and diff of |y_j|}
\begin{split}
 \frac{ \| y_j - y_j^{\prime} \| }{  \| y_j\| - \|y_j^{\prime} \| } = \frac{|Y_j Y_j^{\prime}|   }{ |Y_j Z_j| }
 = \frac{ \sin (\angle Y_j Z_j Y_j^{\prime}) }  { \sin(\angle Y_j Y_j^{\prime} Z_j) }
 \\
 = \frac{ \sin (\pi/2 - \varphi_j/2) }  { \sin( \beta_j - \varphi_j/2 ) }
\leq \frac{1}{ \sin(\beta_j - \varphi_j ) }.
% \leq \frac{ 1 }  { \sin( \beta_j /2 ) }
% = \frac{2\cos(\beta_j/2) }{ \sin(\beta_j)}
% \leq \frac{2 }{ \sin(\beta_j) }.
\end{split}
\end{equation}
In order to bound $ 1/\sin(\beta_j - \varphi_j)$ \footnote{The part from \eqref{diff of y_j and diff of |y_j|} to \eqref{sin(beta) bound}
can be replaced by a simpler bound $ \sin(\beta_j - \varphi_j)\geq \sin(\beta_j/2) \geq \sin(\beta_j)/2  $ and we can still obtain a similar bound
as \eqref{diff y and diff |y| ratio 2nd bound}; however, by using this simpler yet looser bound,
the constant coefficient $7/8$ will be replaced by a larger constant. }, we use the following bound:
$$
 \frac{ \sin \beta_j }{ \sin(\beta_j - \varphi_j )} = \frac{|Y_j K_j|}{ \|y_j \|} \frac{ \|y_j^{\prime} \|}{ |Y_j^{\prime} K_j| }
\leq \frac{|Y_j K_j|}{|Y_j^{\prime} K_j |} = \frac{\tan \alpha_i }{ \tan (\alpha_i - \theta)}.
$$
Then  we have
\begin{equation}\label{sin(beta) bound interm}
\begin{split}
  \frac{ \sin \beta_j }{ \sin(\beta_j - \varphi_j )} \frac{ \sin (\alpha_i - \theta)}{ \sin(\alpha_i) }
\frac{ \cos (\alpha_i - \theta)}{ \cos(\alpha_i) } \\
= \frac{ \cos \alpha_i \cos \theta + \sin \alpha_i \sin \theta }{ \cos(\alpha_i) }
\leq \frac{ \sin(\theta)}{\cos(\alpha_i)}  + 1.
\end{split}
\end{equation}
According to \eqref{sin theta/2 expression} and the fact $\cos(\alpha_i) = \langle x_i, y_i \rangle / (\| x_i\| \| y_i\|) = \Sigma_i / (\| x_i\| \| y_i\|)$, we have
\begin{equation}\nonumber
\begin{split}
 \frac{ \sin(\theta)}{\cos(\alpha_i)} \leq \frac{2 \sin(\theta/2)}{ \cos(\alpha_i)} = \frac{D_i}{ \|x_i \| \| y_i \| \sin(\alpha_i - \theta/2) }
 \frac{\| x_i\| \| y_i\|}{\Sigma_i} \\
 = \frac{D_i}{\Sigma_i} \frac{1}{\sin(\alpha_i - \theta/2)}  \overset{\eqref{D_i bound by Sigma_i/10},\eqref{alpha - theta bound}}{\leq} \frac{1}{12} \frac{1}{\sin(\pi/3)}
 = \frac{1}{6 \sqrt{3}}.  % \leq \frac{1}{8},
 \end{split}
\end{equation}
% where the last inequality is due to \eqref{alpha - theta bound}.
Plugging the above relation into \eqref{sin(beta) bound interm}, we obtain  % \eqref{sin(theta)/cos(alpha) bound}
\begin{equation}\label{sin(beta) bound}
  \frac{ \sin \beta_j }{ \sin(\beta_j - \varphi_j )} \frac{ \sin (\alpha_i - \theta)}{ \sin(\alpha_i) } \leq \frac{6\sqrt{3}+1}{6\sqrt{3}}.
% \frac{ \cos (\alpha_i - \theta)}{ \cos(\alpha_i) }
\end{equation}

% Note that $ \sin(\beta_j - \varphi_j/2) \geq \sin(\beta_j - \varphi_j)   $
Combining  \eqref{diff of y_j and diff of |y_j|} and \eqref{diff y_j bound by HY}, we obtain
\begin{equation}\label{diff y and diff |y| ratio 1st bound}
\begin{split}
&  \frac{ \| y_j - y_j^{\prime} \| }{  \| y_j\| - \|y_j^{\prime} \| } \frac{\| y_j - y_j^{\prime} \| }{ \|y_j \| }  \\
& \leq \frac{1 }{ \sin(\beta_j - \varphi_j)  } \frac{2}{\sqrt{3}} \frac{D_i}{\Sigma_i} \frac{|H_jY_j|}{\| y_j\|}   \\
& \overset{\eqref{sin(beta) bound}}{\leq} \frac{2}{\sqrt{3}} \frac{D_i}{\Sigma_i} \frac{6\sqrt{3}+1}{6\sqrt{3}} \frac{|H_jY_j|}{\| y_j\|} \frac{\sin(\alpha_i)}{\sin(\beta_j)}
 \frac{1}{\sin (\alpha_i - \theta)}    \\
& = \frac{ 6\sqrt{3} +1  }{ 9 } \frac{D_i}{\Sigma_i} \frac{1}{\sin (\alpha_i - \theta)}
\overset{\eqref{alpha - theta bound}}{\leq} \frac{ 6\sqrt{3} +1  }{ 9 } \frac{2}{\sqrt{3}} \frac{D_i}{\Sigma_i}
\leq \frac{3}{2}\frac{D_i}{\Sigma_i},  %  real value: 1.461  \frac{5\sqrt{3} +1 }{6}
\end{split}
\end{equation}
where the last equality is due to $ |H_jY_j| \sin(\alpha_i) = |Y_j K_j| = \| y_j\| \sin(\beta_j) $.
% the last inequliaty follows from $\alpha_i - \theta \geq \pi/3  $ (see ), and

%%%----------Simpler way to bound----------------------
% \leq \frac{2 }{ \sin(\beta_j )\| y_j\|  } \frac{2}{\sqrt{3}} \frac{D_i}{\Sigma_i} |H_jY_j|
%= \frac{5}{2} \frac{D_i}{\Sigma_i} \frac{|H_jY_j| }{|K_j Y_j|}
%= \frac{5}{2} \frac{D_i}{\Sigma_i} \frac{1}{\sin(\alpha_i)}
%\leq \frac{5}{2} \frac{D_i}{\Sigma_i}  \frac{1}{\sin(\pi/3)}
%=  \frac{5}{\sqrt{3}} \frac{D_i}{\Sigma_i} .
%%%--------------------------------

According to \eqref{y_j diff bound} and \eqref{D_i bound by Sigma_i/10}, we obtain that $ \| y_j - y_j^{\prime}\| \leq \frac{2}{\sqrt{3}}\frac{1}{12} \|y_j \|
\leq \frac{1}{8} \|y_j \|$,
which further implies $\|y_j^{\prime} \| + \|y_j \| \geq 2 \|y_j \|  - \| y_j - y_j^{\prime}\| \geq \frac{15}{8} \|y_j \| $.
Then by \eqref{diff y and diff |y| ratio 1st bound} we have
\begin{equation}\label{diff y and diff |y| ratio 2nd bound}
\begin{split}
\| y_j - y_j^{\prime} \|^2 \leq  \frac{5\sqrt{3} +1 }{6} \frac{D_i}{\Sigma_i} ( \| y_j\| - \|y_j^{\prime} \|) \|y_j \| \\
\leq  \frac{ 3 }{2} \frac{D_i}{\Sigma_i} ( \| y_j\| - \|y_j^{\prime} \|) (\|y_j^{\prime} \| + \|y_j \| )\frac{8}{15}
= \frac{4}{5} \frac{D_i}{\Sigma_i} ( \| y_j\|^2 - \|y_j^{\prime} \|^2 ).
\end{split}
\end{equation}

According to the definition \eqref{y_i prime def}, we have
\begin{align*}
\frac{  \| y_i\|^2 - \|y_i^{\prime} \|^2 }{ \| y_i - y_i^{\prime} \|^2 }
= \frac{ 1 - (\Sigma_i)^2/(\Sigma_i + D_i)^2 }{ [1 - \Sigma_i/(\Sigma_i + D_i)]^2 }  \\
= \frac{ (\Sigma_i+D_i)^2 - \Sigma_i^2 }{ D_i^2 } = \frac{ D_i^2 + 2 D_i \Sigma_i }{ D_i^2} \geq 2 \frac{\Sigma_i}{D_i},
\end{align*}
which implies
\begin{equation}\label{diff y_i and diff |y_i| ratio bound}
\| y_i - y_i^{\prime} \|^2   \leq \frac{1}{2} \frac{D_i}{\Sigma_i}  (\| y_i\|^2 - \|y_i^{\prime} \|^2 ).
\end{equation}
Summing up \eqref{diff y and diff |y| ratio 2nd bound} for $j \in \{1,\dots, r \}\backslash \{ i\} $ and \eqref{diff y_i and diff |y_i| ratio bound}, we obtain
\begin{equation}\nonumber % \label{diff Y and diff |Y| ratio final bound}
\| Y - Y^{\prime} \|_F^2
\leq \frac{4}{5} \frac{D_i}{\Sigma_i} ( \| Y\|_F^2 - \|Y^{\prime} \|_F^2 ),
\end{equation}
which proves the second part of \eqref{property b) of X,Y prime}.

%%%%%%%%%%%%%%%%%%%%%%%%%%%%%%%%%%%%%%%%%%%%%%%%%%%%%%%%%%%%%%%%%%%%%%%%%%%%%%%%%%%%%%%%%%%%%%%%%%%%%
%%%%%%%%%%%%%%%%%%%%%%%%   Proof of lemmas for Lemma 2   %%%%%%%%%%%%%%%%%%%%%%%%%%%%%%%%%%%%%%%%%%%%%%%%%%%%%%%%%%%
%%%%%%%%%%%%%%%%%%%%%%%%%%%%%%%%%%%%%%%%%%%%%%%%%%%%%%%%%%%%%%%%%%%%%%%%%%%%%%%%%%%%%%%%%%%%%%%%%%%%%
\section{Proofs of the results in Section \ref{sec: proof of lemma 2}}\label{appen: Lemma 2 additional proof}
%%%%%%%%%%%%%%%%%%%%%%%%%%%%%%%%%%%%%%%%%%%%%%%%%%%%%%%%%%%%%%%%%%%%%%%%%%%%%%%%%%%%%%%%%%%%%%%%%%%%%
%%%%%%%%%%%%%%%%%%%%%%%%   Proof of Initialization    %%%%%%%%%%%%%%%%%%%%%%%%%%%%%%%%%%%%%%%%%%%%%%%%%%%%%%%%%%%
%%%%%%%%%%%%%%%%%%%%%%%%%%%%%%%%%%%%%%%%%%%%%%%%%%%%%%%%%%%%%%%%%%%%%%%%%%%%%%%%%%%%%%%%%%%%%%%%%%%%%
\subsection{Proof of Claim \ref{claim: initial point properties}}\label{appen: initialization proof}

%Now, we state the modified method to choose the initial point as in \cite{KMO09}.
%For any matrix, define the degree of a row (column) as the number of nonzero entries in the row (column).
%Set to zero all columns in  $ \bP_{\Omega}(M)$ with degree larger than $2|\Omega|/m$, and then set to zero all rows with degree larger than
%$2|\Omega|/n$. This step is called ``trimming".
%
%Denote the resulting matrix as $T_m(M)$. Let the SVD of $T_m(M)$ be $T_m(M) = \sum_{i=1}^n \sigma_i x_i y_i^T$, where $\sigma_1\geq \sigma_2
% \geq \dots \geq \sigma_{n}$. Set to zero all but the $r$ largest singular values, we get a matrix $M_0 = \sum_{i=1}^r \sigma_i x_i y_i^T$.

The proof of this claim consists of two parts: first, by a classical result we have that $M_0$, the best rank-$r$ approximation of $\frac{1}{p} \bP_{\Omega}(M)$, is close to $M$; second, show that the scaling does not change the closeness.

We first present the following result. % \cite[Theorem 1.1]{keshavan2010matrix}, which is restated below.
\begin{lemma}\label{lemma of KMO}% {\cite{keshavan2010matrix}}
Assume $M$ is a rank $r$ matrix of dimension $m \times n$ with $m \geq n$, and denote $M_{\max} = \| M\|_{\infty}$ as the maximum magnitude of the entries of $M$. Suppose each entry of $M$ is included in $\Omega$ with probability $p \geq C_0 \frac{\log( m + n )}{ m }$, and $M_0 $ is the best rank-r approximation of $\frac{1}{p} \bP_{\Omega}(M)$. Then with probability larger than $1-1/(2n^4)$,
\begin{equation}\label{M_0 - M bounded}
\frac{1}{mn M_{\max}^2 }\|M - M_0 \|_F^2 \leq C_2 \frac{ \alpha^{\frac{3}{2}} {r}}{pm},
\end{equation}
for some numerical constant $C_2$.
\end{lemma}

Remark:
% (note that \cite[Theorem 6.5]{recht2011simpler} is for a resampling-based model of $\Omega$).
Lemma \ref{lemma of KMO} can be found in \cite{keshavan2010matrix}. The original version \cite[Theorem 1.1]{keshavan2010matrix}  holds for $M_0 = \mathrm{P}_r( \mathrm{T}_{\mathrm{r}}(\bP_{\Omega}(M))/p )$, where  $\mathrm{T}_{\mathrm{r}}(\cdot)$ denotes a trimming operator which sets to zero all rows and columns that have too many observed entries, and $\mathrm{P}_r(\cdot)$ denotes the best rank-$r$ approximation.
By standard Chernoff bound one can show that none of the rows and columns have too many observed entries with high probability, thus the conclusion of \cite[Theorem 1.1]{keshavan2010matrix} holds for $M_0 = \mathrm{P}_r( \bP_{\Omega}(M)) /p$.
 The key to establish Lemma \ref{lemma of KMO} is a bound on $ \| M - \frac{1}{p} \bP_{\Omega}(M)\|_2 $, which can be simply proved by matrix concentration inequalities; see \cite[Remark 6.1.2]{keshavan2012efficient}, \cite[Theorem 6.3]{candes2009exact} or \cite[Theorem 3.5]{recht2011simpler}.
The proof of \cite[Theorem 1.1]{keshavan2010matrix} is more complicated than applying matrix concentration inequalities since it holds for a weaker condition $|\Omega| \geq O( n )$.
%one can prove $ \| M - \frac{1}{p} \bP_{\Omega}(M)\|_2 \leq C_2' \sqrt{ \frac{ \alpha^{\frac{3}{2}} n }{ p }  }  M_{\max} $;

Note that $\hat{X}_0, \hat{Y}_0$ defined in Table \ref{table of Initialization} satisfy
\begin{equation}\label{decomposition of M0}
\hat{X}_0 \hat{Y_0}^T = \mathrm{P}_r( \bP_{\Omega}(M)/p )
= M_0.
\end{equation}
% Lemma \ref{lemma of KMO} has the following corollary, which states that if $| \Omega |$ is large enough, then $M_0$ is close to $M.$
%\begin{coro}\label{bound on delta0}
%If $|\Omega| \geq \alpha^{3/2} C_0 \mu_1^2  r^{12} \kappa^{8} n$, then $\|M - M_0 \|_F \leq \sqrt{C_1/C_0} \frac{\Sigma_{\min}}{r^5 \kappa^4}$, where $C_1$ is the numerical constant in Proposition \ref{prop1}.
%\end{coro}
%Plugging the incoherence condition A1: $M_{\rmax} \leq \mu_1 \Sigma_{\rmax}\frac{\sqrt{r}}{\sqrt{mn}} $
% According to the incoherence condition defined in \eqref{incoherence def of M},
Recall that the SVD of $M$ is $M = \hat{U}\Sigma \hat{V}$, where $\hat{U}, \hat{V} $ satisfies \eqref{incoherence cond}.
We have
\begin{equation}\label{Mij bound by mu}
\begin{split}
   |M_{ij}| = \sum_{k=1}^r | \hat{U}_{ik} \hat{V}_{jk} \Sigma_k | \leq \Sigma_{\rmax} \sum_{k=1}^r | \hat{U}_{ik} \hat{V}_{jk} |
   \\
   \leq \Sigma_{\rmax} \sqrt{\sum_{k=1}^r \hat{U}_{ik}^2} \sqrt{\sum_{k=1}^r \hat{V}_{jk}^2} \overset{\eqref{incoherence cond}}{\leq} \Sigma_{\rmax} \frac{\mu r}{\sqrt{mn}}, \ \forall \ i,j.
   \end{split}
\end{equation}   % \|U^{(i)} \| \| V^{(j)}\|
 The above relation implies $M_{\rmax} \leq \Sigma_{\rmax} \frac{\mu r}{\sqrt{mn}}$. Plugging this inequality and $p = |\Omega|/(mn)$ into \eqref{M_0 - M bounded}, we get
\begin{equation}\label{M_0 - M bounded, interm}
   \|M - M_0 \|_F^2 \leq C_2 \frac{mn \alpha^{\frac{3}{2}} r }{pm} \Sigma_{\rmax}^2 \frac{\mu^2 r^2 }{ mn } = C_2 n \frac{\alpha^{\frac{3}{2}} r^3 \kappa^2 \mu^2  }{ |\Omega| } \Sigma_{\rmin}^2.
\end{equation}

 Plugging \eqref{decomposition of M0} and the assumption \eqref{Omega bound} %\eqref{p bound 1st time, claim of initialization}  $ p \geq C_0 \alpha^{2/3} r^{t_1} \kappa^{t_2} \mu_1^{t_3} \frac{1}{m}$
 into \eqref{M_0 - M bounded, interm}, we get
\begin{equation}\label{M_0 - M bounded by hat delta0}
 \hat{\delta}_0  \triangleq \|M - \hat{X}_0 \hat{Y_0}^T \|_F \leq \sqrt{\frac{C_2}{C_0}} \frac{\Sigma_{\min}}{r^{1.5} \kappa^2 }.
 %  \frac{\Sigma_{\min}}{r^{t_1/2 - 1} \kappa^{t_2/2 - 1} \mu_1^{t_3/2 - 1}  }.
\end{equation}

The property (a), i.e.\ $(X_0, Y_0) \in (\sqrt{2/3}K_1)$ follows directly from the definitions of $X_0$ and $Y_0$ in \eqref{initial point rescale}.
 % we have $\|X_0^{(i)}\| \leq \sqrt{2/3} \beta_1$ and $\|Y_0^{(j)}\| \leq \sqrt{2/3} \beta_2, \forall i,j$, thus $(X_0, Y_0) \in \sqrt{2/3}K_1$.
We then prove the property (b), i.e.\ $(X_0, Y_0) \in (\sqrt{2/3}K_2)$. By \eqref{M_0 - M bounded by hat delta0} we have
 $ \| M - M_0\|_F \leq \Sigma_{\rmin}/5 \leq \Sigma_{\rmax}/5$ for large enough $C_0$.
 This inequality combined with $\| M - M_0\|_F \geq \| M - M_0\|_2 \geq \|M_0\|_2 - \Sigma_{\rmax}$ yields
 \begin{equation}\label{M_0 spectral bound}
 \|M_0\|_2 \leq \frac{6}{5}\Sigma_{\rmax}.
 \end{equation}
 {\black By the definitions of $\hat{X}_0, \hat{Y}_0$ (i.e.\ $\hat{X}_0 = \bar{X}_0 D_0^{\frac{1}{2}}$, $\hat{Y}_0 = \bar{Y}_0 D_0^{\frac{1}{2}}$, where
  $\bar{X}_0 D_0\bar{Y}_0^T $ is the SVD of $M_0$), } we have
\begin{equation}\label{hat(X0Y0) bound}
\| \hat{X}_0\|_2 = \| \hat{Y}_0\|_2 = \sqrt{\|M_0 \|_2} \overset{ \eqref{M_0 spectral bound} }{\leq} \sqrt{ \frac{6}{5}} \sqrt{ \Sigma_{\rmax} }.
\end{equation}
 Then we have
 \begin{equation}\label{X_0 norm bound}
 \|\hat{X}_0\|_F^2  \leq r\| \hat{X}_0\|_2^2 \leq  \frac{6}{5}r \Sigma_{\rmax} \overset{\eqref{beta 1 beta T def}}{<} \frac{2}{3}\beta_T^2,
 \end{equation} %% $ \|M_0\|_2 \leq
 where the last inequality follows from {\black $C_T > 9/5.$ }
 By the definition of $X_0$ in \eqref{initial point rescale}, we have $\|X_0 \|_F^2 \leq   \|\hat{X}_0\|_F^2 \leq \frac{2}{3}\beta_T^2$.
 Similarly, we can prove $\|Y_0 \|_F^2 \leq \frac{2}{3}\beta_T^2$. Thus the property (b) is proved.

Next we prove the property (c), i.e.\ $ \|M - X_0 Y_0^T \|_F  \leq \delta_0$.
Since $\hat{X}_0, \hat{Y}_0$ satisfy $\max \{ \| \hat{X}_0 \|_F, \| \hat{Y}_0 \|_F \} \leq \beta_T ${\black (due to \eqref{X_0 norm bound} and the analogous inequality for $\hat{Y}_0$}) and  \eqref{M_0 - M bounded by hat delta0},
it follows from Proposition \ref{prop1} that there exist $U_0,V_0$ such that
\begin{subequations}
\begin{align}
  U_0V_0^T  &  =M;   \\
  \| U_0\|_2   &  \leq \| X_0\|_2 ;     \label{U_0 <= X_0}  \\
  \|U_0-\hat{X}_0 \|_F \leq \frac{ 6 \| \hat{Y}_0 \|_2}{ 5 \Sigma_{\min}}\hat{\delta}_0,  & \;\;
\|V_0-\hat{Y}_0 \|_F \leq \frac{3 \| \hat{X}_0 \|_2}{ \Sigma_{\min}}\hat{\delta}_0;   \label{U0-X0, V0-Y0 bound}  \\
  \| U_0^{(i)}\|^2 \leq \frac{r\mu }{m}\beta_T^2,                     &  \;\;
     \| V_0^{(j)}\|^2 \leq \frac{3r\mu }{2n}\beta_T^2 .  \label{U_0 incoherent}
\end{align}
\end{subequations}
Note that the above inequalities \eqref{U_0 <= X_0} and \eqref{U0-X0, V0-Y0 bound} are not due to \eqref{req b) of U,V, Prop 1}
and \eqref{req c) of U,V, Prop 1} of Proposition \ref{prop1},
but stronger results \eqref{complement: U spectral norm smaller} and \eqref{complement: U'-X',V'-Y' upper bounded by spectral norm}
established during the proof of Proposition \ref{prop1}.

Note that
\begin{equation}\label{M-X_0Y_0 bound interm}
\begin{split}
& \|M - X_0Y_0^T \|_F = \|U_0(V_0 - Y_0)^T + (U_0 - X_0)Y_0^T   \|_F  \\
 & \leq \|U_0(V_0 - Y_0)^T\|_F + \|(U_0 - X_0)Y_0^T   \|_F             \\
 & \leq    \|U_0 \|_2 \|V_0 - Y_0 \|_F + \|U_0 - X_0 \|_F \|Y_0 \|_2,
 \end{split}
\end{equation}
where the last inequality follows from Proposition \ref{matrix prop: |AB| bound}.
Since $X_0^{(i)} $ and $\hat{X}_0^{(i)} $ has the same direction and $\| X_0^{(i)} \| \leq \|\hat{X}_0^{(i)} \|  $, by Proposition \ref{matrix prop: |(A)|>|B| with smaller row norm} we have
\begin{equation}\label{X_0 <= sqrt(Sigma_max)}
 \| X_0 \|_2 \leq \| \hat{X}_0\|_2 \leq \sqrt{ \frac{6}{5}} \sqrt{ \Sigma_{\rmax} } .
\end{equation}
Combining \eqref{U_0 <= X_0} and \eqref{X_0 <= sqrt(Sigma_max)}, we get
\begin{equation}\label{U_0 <= sqrt(Sigma_max)}
\| U_0\|_2 \leq \sqrt{ \frac{6}{5}} \sqrt{ \Sigma_{\rmax} } .
 \end{equation}
Similar to \eqref{X_0 <= sqrt(Sigma_max)}, we have
 \begin{equation}\label{Y_0 <= sqrt(Sigma_max)}
 \|Y_0 \|_2 \leq \sqrt{ \frac{6}{5}} \sqrt{ \Sigma_{\rmax} } .
 \end{equation}

It remains to bound $\|V_0 - Y_0 \|_F$ and  $\|U_0 - X_0 \|_F.$
Let us prove the following inequality:
\begin{equation}\label{U0-X0 bounded}
\|U_0^{(i)} - X_0^{(i)} \| \leq \|U_0^{(i)} - \hat{X}_0^{(i)} \|, \;\; \forall \ i.
\end{equation}
 % Consider two cases $\|\hat{X}_0^{(i)} \| \leq \sqrt{\frac{2}{3 }} \beta_1$ and
If $\|\hat{X}_0^{(i)} \| \leq \sqrt{\frac{2}{3 }} \beta_1$, then (\ref{U0-X0 bounded}) becomes equality since $\hat{X}_0^{(i)} = X_0^{(i)}$.
Thus we only need to consider the case $\|\hat{X}_0^{(i)} \| > \sqrt{\frac{2}{3 }} \beta_1.$
In this case by the definition of $X_0$ in \eqref{initial point rescale} we have $\|X_0^{(i)} \| = \sqrt{\frac{2}{3 }} \beta_1.$
From \eqref{U_0 incoherent}, we get
\begin{equation}\label{u0 < x0}
\| U_0^{(i)}\|^2 < \frac{3}{2} \frac{r\mu }{m}\beta_T^2  \leq \frac{2}{3}\beta_1^2 < \|\hat{X}_0^{(i)} \|^2 .
\end{equation}
% Then $\|\hat{X}_0^{(i)} \| > \sqrt{\frac{2}{3 }} \beta_1 > \| U_0^{(i)}\|.$
For simplicity, denote $u \triangleq U_0^{(i)}, x \triangleq X_0^{(i)}, \tau  \triangleq \frac{\|\hat{X}_0^{(i)} \|}{ \sqrt{2/3 } \beta_1 } = \frac{\|\hat{X}_0^{(i)} \|}{\|x \| } >1.$
Then \eqref{u0 < x0} becomes $\| u\| \leq \|x \|$ and \eqref{U0-X0 bounded} becomes $\|u- x \| \leq \|u- \tau x \|$. The latter can be transformed as follows:
% \begin{subequations} %\nonumber
\begin{align}
\|u- x \| \leq \|u-\tau x \| & \Longleftrightarrow \|x \|^2 -2 \langle u,x\rangle \leq \tau^2 \|x \|^2 -2\tau \langle u,x\rangle \nonumber \\
                             &  \Longleftrightarrow 2(\tau-1)\langle u,x\rangle \leq (\tau^2-1)\|x \|^2  \nonumber \\
                             & \Longleftrightarrow 2\langle u,x\rangle \leq (\tau+1)\|x \|^2. \label{2<u,x> bound}
\end{align}
% \end{subequations}
Since $\langle u,x\rangle \leq \| u\|\|x\| \leq \|x \|^2$ (here we use $\| u\| \leq \| x\|$ which is equivalent to \eqref{u0 < x0})
 and $2< \tau+1,$ the last inequality of \eqref{2<u,x> bound} holds,
which implies that $\|u- x \| \leq \|u- \tau x \|$ holds and, consequently, (\ref{U0-X0 bounded}) holds.

An immediate consequence of (\ref{U0-X0 bounded}) is
\begin{equation}\label{U0-X0 total bound}
\|U_0 - X_0 \|_F \leq \|U_0 - \hat{X}_0 \|_F \overset{\eqref{U0-X0, V0-Y0 bound}}{\leq}  \frac{5 \| \hat{Y}_0 \|_2}{4\Sigma_{\min}}\hat{\delta}_0
\overset{\eqref{hat(X0Y0) bound}}{\leq} \frac{5 }{4} \sqrt{\frac{6}{5}} \sqrt{\Sigma_{\rmax}} \frac{\hat{\delta}_0}{\Sigma_{\rmin}}.
\end{equation}
Similarly, we have
\begin{equation}\label{V0-Y0 total bound}
\|V_0 - Y_0 \|_F \overset{\eqref{U0-X0, V0-Y0 bound}}{\leq} 3  \sqrt{\frac{6}{5}} \sqrt{\Sigma_{\rmax}} \frac{\hat{\delta}_0}{\Sigma_{\rmin}}.
\end{equation}

Plugging \eqref{U_0 <= sqrt(Sigma_max)}, \eqref{Y_0 <= sqrt(Sigma_max)}, \eqref{U0-X0 total bound} and \eqref{V0-Y0 total bound} into
\eqref{M-X_0Y_0 bound interm}, we get
\begin{equation}\nonumber
\begin{split}
& \|M - X_0Y_0^T \|_F \\
  \leq &  \sqrt{ \frac{6}{5}} \sqrt{ \Sigma_{\rmax} }  \frac{5 }{4} \sqrt{\frac{6}{5}} \sqrt{\Sigma_{\rmax}}  \frac{\hat{\delta}_0}{\Sigma_{\rmin}} +   \sqrt{ \frac{6}{5}} \sqrt{ \Sigma_{\rmax} } 3 \sqrt{\frac{6}{5}} \sqrt{\Sigma_{\rmax}}  \frac{\hat{\delta}_0}{\Sigma_{\rmin}}   \\
                    =        &  ( \frac{3}{2} +  \frac{18}{5} ) \kappa \hat{\delta}_0  \\
                    \overset{\eqref{M_0 - M bounded by hat delta0}}{\leq}     &  \frac{51}{10} \sqrt{\frac{C_2}{C_0}} \frac{\Sigma_{\min}}{r^{1.5} \kappa} \\
                    \overset{\eqref{delta definition throughout}}{\leq} &  \delta_0,
                      % \frac{\Sigma_{\min}}{r^{t_1/2 - 1} \kappa^{t_2/2 - 2} \mu_1^{t_3/2 - 1}  }.
\end{split}
\end{equation}
where the last inequality holds for $C_d \geq  \frac{ 5 }{ 153 } \sqrt{\frac{C_0}{C_2}} $. Therefore property (c) is proved.

%With the bounds on $\| U_0\|_2$,$\|Y_0 \|_2$,$\|U_0 - X_0 \|_F$ and $\|V_0 - Y_0 \|_F$,
%we have
%\begin{equation}
%\begin{split}
%\|M - X_0Y_0^T \|_F \leq     &  \|U_0 \|_2 \|V_0 - Y_0 \|_F + \|U_0 - X_0 \|_F \|Y_0 \|_2   \\
%                     \leq    &  2\sqrt{\frac{2}{3}}\beta_T \frac{2\beta_T}{\Sigma_{\min}}\hat{\delta}_0 \\
%                     \leq    & 4 \sqrt{\frac{2}{3}} C_T r \kappa \hat{\delta}_0 \\
%                     \leq    & \frac{\Sigma_{\min}}{5C_d r^4 \kappa^3}.
%\end{split}
%\end{equation}

% [Proposition 1.2.3] convergence to stationary points

\subsection{Proof of Claim \ref{lemma: P Omega and P has relation}}\label{appen: RSC proof}
% This claim is a corollary of Proposition \ref{upper bound of P((U-X)(V-Y))} and Corollary \ref{coro: summary of U,V}.
As mentioned in Section \ref{sec: assumptions}, in this proof we only need to consider the Bernolli model that $\Omega$ includes each entry of $M$ with probability $p$ and the expected size $S$ satisfies \eqref{Omega bound}.
Denote $d \triangleq \|M-XY^T \|_F$.
Let $a = U(V-Y)^T + (U-X)V^T$, $b = (U-X)(V-Y)$, where $U,V$ are defined with the properties in Corollary \ref{coro: summary of U,V}.

According to \eqref{P Omega (a) bound} we have $\| \bP_{\Omega}(a)\|_F^2 \geq  \frac{27}{40} p d^2 $.
According to \eqref{P Omega (b) bound}, we have  $\| \bP_{\Omega}(b)\|_F \leq \frac{1}{5} \sqrt{p} d $.
Therefore, $\|\bP_{\Omega}(M-XY^T) \|_F = \| \bP_{\Omega}(a-b)\|_F \geq \| \bP_{\Omega}(a)\|_F - \|\bP_{\Omega}(b)\|_F \geq \sqrt{ \frac{27}{40}} \sqrt{p} d - \frac{1}{5} \sqrt{p} d  \geq  \frac{3}{5}\sqrt{p}d \geq \frac{1}{\sqrt{3}}\sqrt{p}d$.

According to \eqref{bound b}, we have $ \| b\|_F \leq \frac{1}{10}d$.{\black According to \eqref{P Omega (a) bound, upper and low} (which is a corollary of \cite[Theorem 4.1]{candes2009exact})}, we have $\| \bP_{\Omega}(a)\|_F^2 \leq \frac{7}{6}p\|a \|_F^2 \leq \frac{7}{6}p(\|M-XY^T \|_F + \|b \|_F)^2
\leq  \frac{7}{6}p(1 + \frac{1}{10})^2 d^2 \leq \frac{17}{12} p d^2. $
% 2p\frac{16}{9}d^2$.
Thus, $\| \bP_{\Omega}(a-b)\|_F \leq \|\bP_{\Omega}(a)\|_F + \|\bP_{\Omega}(b)\|_F \leq (\sqrt{\frac{17}{12}} + \frac{1}{5} ) \sqrt{p}d \leq  \sqrt{2p}d$.
$\quad$ $\Box$

\subsection{Proof of Proposition \ref{prop: K(delta) condition}}\label{appen: proof of K(d) condition}
We first provide a general condition for $(X,Y) \in K_1 \cap K_2$ (i.e.\ incoherent and bounded) based on the function value $\tilde{F}(X, Y)$.
% thus by Claim \ref{claim: decreasing implies K_1 intersect K_2} we have $(X_k, Y_k) \in K_1 \cap K_2$.
% (i.e.\ the objective function $\tilde{F}(X_k,Y_k)$ is decreasing).
\begin{prop}\label{prop: decreasing implies K_1 intersect K_2}
Suppose the sample set $\Omega$ satisfies \eqref{RSC of P_Omega} and $\rho = 2 p\delta_0^2/G_0(3/2)$, where $\delta_0$ is defined in \eqref{delta definition throughout}.
Suppose $(X_0, Y_0)$ satisfies \eqref{initial point condition} and  % an algorithm satisfies: \text{The initial point }
\begin{align}
%  (X_0,Y_0) \in (\sqrt{\frac{2}{3}} K_1) \cap (\sqrt{\frac{2}{3}} K_2) \cap K(\delta_0), \label{initial point condition} \\
  \tilde{F}(X,Y) \leq 2 \tilde{F}(X_0, Y_0). \label{F(x) less than 2 F_0}
\end{align}
 Then $(X, Y) \in K_1 \cap K_2$.
% use the initial point $(X_0,Y_0)$ given by the procedure $\textsc{Initialize}$ and
\end{prop}

\emph{Proof of Proposition \ref{prop: decreasing implies K_1 intersect K_2}}:
% \emph{Proof of Claim \ref{prop: decreasing implies K_1 intersect K_2}:}
We prove by contradiction. Assume the contrary that $(X, Y) \notin K_1 \cap K_2$.{\black By the definition of $K_1, K_2$ in \eqref{def of K_1, K_2}, we have either $ \| X^{(i)} \|^2  > \beta_1^2 $ for some $i$, $  \| Y^{(j)} \|^2 > \beta_2^2  $ for some $j$,
$  \| X\|_F^2 > \beta_T^2 $ or $  \| Y \|_F^2  > \beta_T^2 $.
Hence at least one term of   % (\eqref{regularized function}
$ G(X,Y) =  \rho \sum_{i=1}^m G_0( \frac{3\| X^{(i)} \|^2  } {2\beta_1^2} ) + \rho \sum_{j=1}^n G_0(\frac{3 \| Y^{(j)} \|^2 } {2\beta_2^2} ) + \rho G_0(\frac{3\| X\|_F^2 } {2\beta_T^2} ) + \rho G_0( \frac{3\| Y \|_F^2 } {2\beta_T^2} )  $ is larger than $ G_0(\frac{3}{2}) $.
In addition, all the other terms in the expression of $G(X,Y)$ are nonnegative, thus we have $ G(X,Y) >  \rho G_0(\frac{3}{2})$.
Therefore, }
\begin{equation}\label{F_k lower bound}
\tilde{F}(X, Y) \geq G(X, Y) > \rho G_0(\frac{3}{2}) = 2 p \delta_0^2.
\end{equation}
We have
\begin{equation}\label{F_0 upper bound}
\tilde{F}(X_0,Y_0) = \frac{1}{2}\| \bP_{\Omega}(M - X_0 Y_0^T) \|_F^2 \leq p \| M - X_0 Y_0^T \|_F^2 \leq p \delta_0^2,
 \end{equation}
 where the first equality is due to $G(X_0, Y_0) = 0$ which follows from $(X_0,Y_0) \in (\sqrt{\frac{2}{3}} K_1) \cap (\sqrt{\frac{2}{3}} K_2)$,
 the second inequality follows from \eqref{RSC of P_Omega} and the fact $(X_0,Y_0) \in (\sqrt{\frac{2}{3}} K_1) \cap (\sqrt{\frac{2}{3}} K_2) \cap K(\delta_0) \subseteq K_1 \cap  K_2 \cap K(\delta)$, and the last inequality is due to $(X_0,Y_0) \in K(\delta_0) $.
Combining  \eqref{F_k lower bound} and \eqref{F_0 upper bound}, we get
$$
 \tilde{F}(X, Y) > 2 \tilde{F}(X_0,Y_0),
$$
which contradicts \eqref{F(x) less than 2 F_0}. $\Box$

%If the algorithm starts from $\bm x_0$ and the algorithm is decreasing, i.e.\ $ \tilde{F}(X_k,Y_k) \leq \tilde{F}(X_{k-1},Y_{k-1}), \forall k$,
% then $\tilde{F}$
% because in this case $\tilde{F}(X_k,Y_k)  \leq  \tilde{F}(X_0, Y_0) \leq 2 \tilde{F}(X_0, Y_0)$.
%Throughout the paper, $\delta_0$ is defined as
%\begin{equation}\label{delta_0 definition throughout}
% \delta_0 \triangleq \frac{1}{5} \delta,
%\end{equation}
%where $\delta$ is defined in \eqref{delta definition throughout}.

{\black
We can prove that \eqref{conditions of algorithms} implies
 % either \eqref{condtion of gradient method} or \eqref{condtion of exact min method} implies % $\{ \tilde{F}(\bm x_{i}) \}$ is a decreasing sequence,
 % \leq \tilde{F}(\bm x_{i}), \forall i$,
\begin{equation}\label{F(x_i) < F_0}
\tilde{F}(\bm x_{i}) \leq 2 \tilde{F}(\bm x_0), \ \forall \ i.
\end{equation}
In fact, when \eqref{condtion for nonconvex direction} holds, as the first inequality in \eqref{condtion for nonconvex direction} the above relation also holds.
When \eqref{condtion of gradient method} holds, let $\lambda = 0$ in \eqref{condtion of gradient method} we get \eqref{F(x_i) < F_0}.}{\black When \eqref{condtion of exact min method} holds, we have % since $1$ is the minimizer of the problem $\min_{\lambda} \psi(\bm x_t, \Delta_t; \lambda)$, we have
% $\bm x_{i+1}$ is the minimizer of the problem $\min_{ \bm x \in \Psi_i }  \psi(\bm x_i; \bm x)$, we have
\begin{equation}\label{psi(xi,xi+1) < F(xi)}
  \psi(\bm x_i, \bm \Delta_i; 1)
  \overset{\eqref{condtion of exact min method}}{\leq } \psi(\bm x_i,\bm \Delta_i; 0) \overset{\eqref{BSUM requirement b}}{=} \tilde{F}(\bm x_i),
\end{equation}
which implies $ \tilde{F}(\bm x_{i+1}) = \tilde{F}(\bm x_{i} + \bm \Delta_i ) \overset{\eqref{BSUM requirement b}}{\leq} \psi(\bm x_i, \bm \Delta_i; 1)
\leq  \tilde{F}(\bm x_i)$.
This relation holds for any $i$, thus  $ \tilde{F}(\bm x_{i+1}) \leq \tilde{F}(\bm x_i) \leq \dots \leq \tilde{F}(\bm x_0) \leq 2 \tilde{F}(\bm x_0) $. }

% Since either   \eqref{condtion of gradient method} or \eqref{condtion of exact min method}
Since \eqref{conditions of algorithms} implies implies $\tilde{F}( \bm x_t) \leq 2 \tilde{F}( \bm x_0 )$
 (see \eqref{F(x_i) < F_0}), by Proposition \ref{prop: decreasing implies K_1 intersect K_2} we have $ \bm x_t \in K_1 \cap K_2  $.
% the conditions of Proposition \ref{prop: K(delta) condition} are stronger than those of Proposition \ref{prop: decreasing implies K_1 intersect K_2}.
The rest of the proof is devoted to establish
\begin{equation}\label{K(4/5 delta) result}
\bm x_t \in K(\frac{2}{3}\delta), \;\; \forall \ t.
\end{equation}
% \subsubsection{Proof of Proposition \ref{prop: K(delta) condition}}
% \emph{Proof of Claim \ref{prop: K(delta) condition}:}
Define the distance of $\bm x = (X,Y)$ and $\bm u = (U,V)$ as
\begin{equation}\nonumber
  d(\bm x, \bm u) = \| XY^T - UV^T \|_F,
\end{equation}
then  $(X_t, Y_t) \in K(\delta) \Longleftrightarrow \|X_t Y_t^T - M \|_F \leq \delta $ can be expressed as
\begin{equation}\nonumber%\label{eq to prove: dis < delta}
  d(\bm x_t, \bm u^*) \leq \delta.
\end{equation}
We first prove the following result:
\begin{lemma}\label{small lemma, x not in [4/5delta, delta]}
If $\tilde{F}(\bm x) \leq 2 \tilde{F}(\bm x_0) $, then $d(\bm u^*, \bm x) \notin [\frac{2}{3}\delta, \delta]$.
%\begin{equation}\label{F >= 2 F_0, if d>3/5 delta}
% % \tilde{F}(\bm x) \geq \frac{8}{3} \tilde{F}(\bm x_0), \text{ if } d(\bm u^*, \bm x) \in [\frac{2}{3}\delta, \delta] \text{ and } \bm x \in K_1 \cap K_2.
%\end{equation}
\end{lemma}

\emph{Proof of Lemma \ref{small lemma, x not in [4/5delta, delta]}}:
We prove by contradiction. Assume the contrary that
\begin{equation}\label{x range [4/5 delta, delta]}
   d(\bm u^*, \bm x) \in [\frac{2}{3}\delta, \delta].
\end{equation}
Since $\bm x_0$ satisfies \eqref{initial point condition}, according to the proof of Proposition \ref{prop: decreasing implies K_1 intersect K_2}
 we have \eqref{F_0 upper bound}, i.e.\ \begin{equation}\label{F_0 upper bound repeat}
\tilde{F}(\bm x_0) \leq p \delta_0^2.
\end{equation}
 According to Proposition \ref{prop: decreasing implies K_1 intersect K_2} and the assumption $\tilde{F}(\bm x) \leq 2 \tilde{F}(\bm x_0) $, we have
$\bm x \in K_1 \cap K_2. $ Together with \eqref{x range [4/5 delta, delta]} we get $\bm x \in K_1 \cap K_2 \cap K(\delta)$. Then  we have
\begin{equation}\label{F(x) lower bound, intermediate}
\tilde{F}(\bm x ) \geq \frac{1}{2} \|\bP_{\Omega}(M - XY^T) \|^2 \overset{\eqref{RSC of P_Omega}}{\geq} \frac{1}{6} p \|M - XY^T \|^2 = \frac{1}{6} p d(\bm u^*, \bm x)^2.
\end{equation}
Plugging $d(\bm u^*, \bm x)^2 \geq (\frac{2}{3})^2 \delta^2  \overset{ \eqref{delta definition throughout}}{=}  16 \delta_0^2  \overset{\eqref{F_0 upper bound repeat}}{\geq} 16 \tilde{F}(\bm x_0)/p $
into \eqref{F(x) lower bound, intermediate}, we get $ \tilde{F}(\bm x ) \geq \frac{8}{3} \tilde{F}(\bm x_0) $,
which together with the assumption $\tilde{F}(\bm x ) \leq 2 \tilde{F}(\bm x_0)  $ leads to $ \tilde{F}(\bm x ) = \tilde{F}(\bm x_0) =0  $.
Then by \eqref{F(x) lower bound, intermediate} we get $d(\bm u^*, \bm x ) =0 $, which contradicts \eqref{x range [4/5 delta, delta]} since $\delta >0$. Thus Lemma \ref{small lemma, x not in [4/5delta, delta]} is proved.
% Lemma \ref{lemma: P Omega and P has relation} we have
% $0 = 6 \tilde{F}(\bm x ) \geq 3 \|\bP_{\Omega}(M - X Y^T) \|_F^2 \geq p \|M - X Y^T \|_F^2 = p d(\bm x , \bm u^* )^2  $. This

Now we get back to the proof of \eqref{K(4/5 delta) result}.
% Proposition \ref{prop: K(delta) condition}.
% It remains to prove % $\bm x_t = (X_t, Y_t) \in K(2\delta/3)$, % prove a slightly stronger result
We prove \eqref{K(4/5 delta) result} by induction on $t$. The basis of the induction holds due to \eqref{initial point condition} and
the fact $\delta_0 = \delta/6$.
Suppose $\bm x_t  \in K(2\delta/3)$, we need to prove $\bm x_{t+1} \in K(2\delta/3)$.
Assume the contrary that $\bm x_{t+1}\notin K(2\delta/3)$, i.e.\ \begin{equation}\label{contrary, d(u, x_t+1) > 4/5 delta}
 d(\bm u^*, \bm x_{t+1}) > \frac{2}{3}\delta.
\end{equation}

Let $i = t+1$ in \eqref{F(x_i) < F_0}, we get $\tilde{F}(\bm x_{t+1}) \leq 2 \tilde{F}(\bm x_0)$. Then by Lemma \ref{small lemma, x not in [4/5delta, delta]} we have
\begin{equation}\label{not in [4/5, 1] delta}
 d(\bm x_{t+1}, \bm u^*) \notin  [\frac{2}{3}\delta , \delta ] ;
 \end{equation}
Combining \eqref{not in [4/5, 1] delta} and \eqref{contrary, d(u, x_t+1) > 4/5 delta}, we get
\begin{equation}\label{xt+1 distance > delta}
 d(\bm x_{t+1}, \bm u^*)  > \delta .
 \end{equation}

In the rest of the proof, we will derive a contradiction for the three cases \eqref{condtion of gradient method}, \eqref{condtion of exact min method}
and \eqref{condtion for nonconvex direction} separately.

Case 1: \eqref{condtion of gradient method} holds. By the induction hypothesis, $d(\bm x_t , \bm u^*) \leq \frac{2}{3}\delta$.
Since $d(\bm x, \bm u^*)$ is a continuous function over $\bm x$, the relation $d(\bm x_t , \bm u^*) \leq \frac{2}{3}\delta$ and \eqref{xt+1 distance > delta} imply that there must exist some $ \bm x^{\prime} = (1-\lambda)\bm x_{t+1} + \lambda \bm x_t, \lambda \in [0,1]  $
such that
\begin{equation}\label{dis(x', u) = delta}
  d(\bm x^{\prime} , \bm u^*) = \delta.
\end{equation}
 According to \eqref{condtion of gradient method}, we have $ \tilde{F}(\bm x^{\prime}) \leq 2\tilde{F}(\bm x_0) $. By Lemma \ref{small lemma, x not in [4/5delta, delta]}, we have $d(\bm u^*, \bm x^{\prime}) \notin [\frac{2}{3}\delta, \delta]$, which contradicts \eqref{dis(x', u) = delta}.

Case 2: \eqref{condtion of exact min method} holds. Define
{\black
 \begin{equation}\label{x' min constrained problem}
 \bm \lambda^{\prime} =  \arg \min_{ \lambda \in \dR, d(\bm x_t + \lambda \Delta_t, \bm u^*) \leq \delta } \psi(\bm x_t, \bm \Delta_t;  \lambda). % \tilde{F}( \bm x).
 \end{equation}
% \begin{equation}\label{x' min constrained problem}
% \bm x^{\prime} =  \arg \min_{ \bm x \in \Psi_t, d(\bm x, \bm u^*) \leq \delta } \psi(\bm x_t;  \bm x). % \tilde{F}( \bm x).
% \end{equation}
By the induction hypothesis, $d(\bm x_t , \bm u^*) \leq \delta$, thus  $ 0 $ lies in the feasible region of the optimization problem in \eqref{x' min constrained problem}, which implies
% i.e.\ $ \bm x_t \in \Psi_t $ and $ d(\bm x_t, \bm u^*) \leq \delta$ (the induction hypothesis),
% we have
\begin{equation}\label{F(x'') < F(x0)}
   \psi(\bm x_t, \bm \Delta_t ; \lambda^{\prime})  \leq \psi(\bm x_t, \Delta_t; 0) \overset{\eqref{BSUM requirement b}}{=} \tilde{F}(\bm x_t).
 %  \psi(\bm x_t; \bm x^{\prime})  \leq \psi(\bm x_t; \bm x_t) \overset{\eqref{BSUM requirement b}}{=} \tilde{F}(\bm x_t).
   %\tilde{F}(\bm x^{\prime}) \leq \tilde{F}(\bm x_t)
% \overset{\eqref{F(x_i) < F_0}}\leq \tilde{F}(\bm x_0) .
\end{equation}
Define $\bm x^{\prime} = \bm x_t + \lambda^{\prime} \Delta_t$, then the feasibility of $\lambda^{\prime}$ for the optimization problem in \eqref{x' min constrained problem} implies $ \delta \geq d(\bm x^{\prime}, \bm u^*) $.
}
Since $d(\bm x, \bm u^*)$ is a continuous function over $\bm x$ and $d(\bm x^{\prime}, \bm u^*) \leq \delta  \overset{\eqref{xt+1 distance > delta}}{<} d(\bm x_{t+1}, \bm u^*)  $, there must exist some $ \bm x^{\prime \prime} = (1- \epsilon )\bm x_{t+1} + \epsilon \bm x^{\prime}{\black = \bm x_t + ( 1 - \epsilon + \epsilon \lambda^{\prime} ) \bm \Delta_t }, \epsilon \in [0,1]  $
such that
\begin{equation}\label{dis(x'', u) = delta}
  d(\bm x^{\prime \prime} , \bm u^*) = \delta.
\end{equation}
Then we have
 {\black
\begin{align*}
  & \tilde{F}(\bm x^{\prime \prime} ) \overset{\eqref{BSUM requirement b}}{\leq} \psi(\bm x_t, \bm \Delta_t ; 1 - \epsilon + \epsilon \lambda^{\prime} )
  \\ &   \overset{\eqref{BSUM requirement a}}{\leq} (1- \epsilon) \psi(\bm x_t, \bm \Delta_t ; 1 ) + \epsilon  \psi(\bm x_t, \bm \Delta_t; \lambda^{\prime}  )
\\ &  \overset{\eqref{psi(xi,xi+1) < F(xi)}, \eqref{F(x'') < F(x0)}}{\leq} \tilde{F}(\bm x_t)
   \overset{\eqref{F(x_i) < F_0}}{\leq} 2 \tilde{F}(\bm x_0).
\end{align*} }
% ---- DELETE IN 2014.11.02 \overset{\eqref{BSUM requirement b}}{\leq} \psi(\bm x_t; \bm x^{\prime \prime})
%------ DELETE IN 2014.11.02-    \overset{\eqref{BSUM requirement a}}{\leq} (1-\lambda) \psi(\bm x_t; \bm x_{t+1} ) + \lambda  \psi(\bm x_t; \bm x^{\prime} )
%  \tilde{F}(\bm x^{\prime \prime} ) \leq (1-\lambda)  \tilde{F}(\bm x_{t+1} ) + \lambda  \tilde{F}(\bm x^{\prime} )  \leq (1-\lambda) \tilde{F}(\bm x_0 ) + \lambda \tilde{F}(\bm x_0 ) = \tilde{F}(\bm x_0 ),
%where the second inequliaty follows from the convexity of $\psi(\bm x_t; \cdot)$, and the second inequliaty follows from \eqref{F(x_i) < F_0} and \eqref{F(x'') < F(x0)}.
  Again we apply Lemma \ref{small lemma, x not in [4/5delta, delta]} to obtain $d(\bm u^*, \bm x^{\prime \prime}) \notin [\frac{2}{3}\delta, \delta]$, which contradicts \eqref{dis(x'', u) = delta}.

{\color{black}
Case 3:  \eqref{condtion for nonconvex direction} holds. By \eqref{initial point condition} and the fact $\delta_0 = \delta/6$ we get
$d(\bm x_0, \bm u^*) \leq \delta/6$. Then we have % Combined with the second relation of \eqref{condtion for nonconvex direction} we obtain
$$ d(\bm x_{t+1}, \bm u^*) \leq d(\bm x_{t+1}, \bm x_0) + d(\bm x_0, \bm u^*) \overset{\eqref{condtion for nonconvex direction}}{\leq} \frac{5}{6}\delta + \frac{1}{6}\delta = \delta  ,$$
which contradicts \eqref{xt+1 distance > delta}.  }

In all three cases we have arrived at a contradiction, thus the assumption \eqref{contrary, d(u, x_t+1) > 4/5 delta} does not hold, which
finishes the induction step for $t+1$. Therefore, \eqref{K(4/5 delta) result} holds for all $t$.
% proves \eqref{K(4/5 delta) result}.
% $\bm x_t \in K(2\delta/3) $.
% $\bm x_t \in K(2\delta/3) \subseteq K(\delta)$.
% Proposition \ref{prop: K(delta) condition}. $\Box$

% \subsection{Proof of Proposition \ref{prop: K(delta) condition}}\label{appen: proof of K(d) condition}
\subsection{Proof of Claim \ref{Algo 1-3 satisfy conditions}}\label{appen: prove Claim of algorithm property}
{\color{black} The sequence $\{\bm x_t \}$ generated by Algorithm 1 with either restricted Armijo rule or restricted line search satisfies \eqref{condtion for nonconvex direction} because the sequence $\tilde{F}(\bm x_t)$ is decreasing and the requirement $d(\bm x_t, \bm x_0 ) \leq 5\delta/6$ is
enforced throughout computation.
 % forced when computing the new points. }

% Algorithm 1 with exact line search stepsize,
{\black Algorithm 2 and Algorithm 3 satisfy \eqref{condtion of exact min method}
since all of them perform exact minimization of a convex upper bound of the objective function along some directions.}
% In specific, Algorithm 1 (with exact line search) satisfies $\eqref{condtion of exact min method}$ with $\Psi_t = \{ \bm x_t + \lambda^{\prime} \nabla \tilde{F}(\bm x_t)  \mid \lambda^{\prime} \in \dR \}$ and $\psi(\bm x_t; \bm x) = \tilde{F}(\bm x)$.
Note that % in BCD type methods Algorithm 2 and Algorithm 3,
 $\bm x_t $ should be understood as the produced solution after $t$ ``iterations'' (one block of variables is updated in one ``iteration'').
  In contrast, $(X_k, Y_k)$ defined in these algorithms is the produced solution after $k$ ``loops'' (all variables are updated once in one ``loop'').
% With this clarification,
For $(X_k, Y_k)$ generated by Algorithm 2, we define $\bm x_{2k} = (X_k, Y_k ), \bm x_{2k+1} = (X_{k+1}, Y_{k}) $ and{\black
$\psi(\bm x_t, \Delta_t; \lambda ) = \tilde{F}(\bm x_t + \lambda \Delta_t)$, then $\psi \text{ satisfies } \eqref{BSUM requirement} $} and  $\{\bm x_t \}_{t=0}^{\infty}= \{(X_k, Y_k ),  (X_{k+1}, Y_{k}) \}_{k=0}^{\infty} $ satisfies \eqref{condtion of exact min method}.
Similarly, for $(X_k, Y_k)$ generated by Algorithm 3, define
 \begin{equation}\nonumber
 \begin{split}
 \bm x_{(m+n)k + i } = (X_{k+1}^{(1)}, \dots,X_{k+1}^{(i-1)}, X^{(i)}, X_{k}^{(i+1)},\dots, X_{k}^{(m)}, Y_k), \\
  \; i=1,\dots, m ,  \\
  \bm x_{(m+n)k + m + j } = ( X_{k+1}, Y_{k+1}^{(1)}, \dots, Y_{k+1}^{(j-1)}, Y^{(j)}, Y_{k}^{(j+1)},\dots, Y_{k}^{(m)} ), \\
  \; j=1,\dots, n,
 \end{split}
 \end{equation}
{\black and $\psi(\bm x_t, \bm \Delta_t; \lambda) = \tilde{F}(\bm x_t + \lambda \bm \Delta_t ) + \lambda_0  \|  \lambda \bm \Delta_t \|^2/2$,
 then  $ \psi \text{ satisfies } \eqref{BSUM requirement} $ and $\{\bm x_t \}_{t=0}^{\infty}$ satisfies \eqref{condtion of exact min method}. }

We then show that Algorithm 1 with constant stepsize $\eta < \bar{\eta}_1$ satisfies \eqref{condtion of gradient method} for some $\bar{\eta}_1$
when $\Omega$ satisfies \eqref{RSC of P_Omega}.
We prove by induction on $t$.
Define $\bm x_{-1} = \bm x_0$, then \eqref{condtion of gradient method} holds for $ t = 0$.
Assume \eqref{condtion of gradient method} holds for $t-1$, i.e.,
$ \tilde{F}(\bm x_{t-1} + \lambda \bm \Delta_{t-1} ) \leq 2 \tilde{F}(\bm x_0), \forall \lambda \in [0,1], \text{ where } \bm \Delta_t = \bm x_{t} - \bm x_{t-1}$.
In particular, we have $\tilde{F}(\bm x_t) \leq 2 \tilde{F}(\bm x_0)$, which together with the assumption that
$\Omega$ satisfies \eqref{RSC of P_Omega} leads to (by Proposition \eqref{prop: decreasing implies K_1 intersect K_2})
\begin{equation}\nonumber %\label{xt bounded}
  \bm x_t \in K_1 \cap K_2 .
\end{equation}
Thus $ \max \{ \|  X_{t} \|_F,  \|  Y_{t} \|_F \}  \leq \beta_T $, $\|X_t^{(i)} \| \leq \beta_1, \forall i, $ and $ \|Y_t^{(j)} \| \leq \beta_2, \forall j$.
%We show that there exists a constant $c_1$ such that
%$$
%  \| \nabla \tilde{F}(\bm x_t) \|_F \leq c_1.
%$$
%In fact,
Then we have
\begin{equation}\nonumber %\label{F(x_k + lambda D k bound}
\begin{split}
& \| \nabla_X \tilde{F}(\bm x_t) \|_F  \\
 & = \| \nabla_X F(\bm x_t) +  \nabla_X G(\bm x_t) \|_F    \\
% & = \left\|  \bP_{\Omega}(X_t Y_t^T - M) Y_t + \rho \sum_{i=1}^m G_0^{\prime}(\frac{3\|  X_t^{(i)} \|^2  } {2\beta_1^2} ) \frac{3 \mathcal{X}_t^{(i)} } {\beta_1^2} + \rho G_0^{\prime}(\frac{3\| X\|_F^2 } {2\beta_T^2} ) \frac{3 X}{\beta_T^2} \right\|_F \\
 & \leq \|  \bP_{\Omega}(X_t Y_t^T - M)  Y_t\|_F +  \left \|\rho \sum_{i=1}^m G_0^{\prime}(\frac{3\|  X_t^{(i)} \|^2  } {2\beta_1^2} ) \frac{3 \bar{X}_t^{(i)} } {\beta_1^2} \right\|_F
 \\ & \quad\quad\quad + \left\|\rho G_0^{\prime}(\frac{3\| X_t \|_F^2 } {2\beta_T^2} ) \frac{3 X_t }{\beta_T^2} \right\|_F  \\
 & \leq \|  \bP_{\Omega}(X_t Y_t^T - M)\|_F \|  Y_t\|_F + \frac{3 \rho \| X_t\|_F } {\beta_1^2}  +  \frac{3 \rho \| X_t\|_F } {\beta_T^2}  \\
 & \leq  \sqrt{ \tilde{F}(\bm x_t) } \beta_T +  \frac{6 \rho \| X_t\|_F } {\beta_1^2}  \\
 &  \leq \sqrt{ 2\tilde{F}(\bm x_0) } \beta_T +  \frac{6 \rho \beta_T } {\beta_1^2},
\end{split}
\end{equation}
where in the second inequality we use $ G_0^{\prime}(\frac{3\|  X_t^{(i)} \|^2  } {2\beta_1^2} ) \leq G_0^{\prime}(\frac{3} {2} ) = 1 $
 and $ G_0^{\prime}(\frac{3\| X\|_F^2 } {2\beta_T^2} ) \leq G_0^{\prime}(\frac{3} {2} ) = 1 $.
Assume
\begin{equation}\label{eta 1 first bound}
  \bar{\eta}_1 \leq \frac{1}{4 \beta_T^2 }.
\end{equation}
Recall that $\eta \leq \bar{\eta}_1 $, thus we have
\begin{equation}\label{c 1 def}
\begin{split}
 \| X_{t+1}  \|_F \leq \| X_t\|_F + \eta \|\nabla_X \tilde{F}(\bm x_t)  \|_F
\\  \leq \beta_T + \frac{1}{ 4 \beta_T^2 } \left( \sqrt{ 2\tilde{F}(\bm x_0) } \beta_T +   \frac{6 \rho \beta_T } {\beta_1^2} \right)  \\
 \overset{\eqref{F_0 upper bound}}{\leq}  \beta_T + \frac{1}{ 4 \beta_T } \left( \sqrt{2 p  }\delta_0 + \frac{6 \rho  } {\beta_1^2} \right)
  \triangleq c_1.
  \end{split}
 \end{equation}
By a similar argument, we can prove $ \| Y_{t+1}  \|_F \leq c_1$, thus $\bm x_{t+1} = (X_{t+1}, Y_{t+1}) \in \Gamma(c_1)$ (recall the definition of $\Gamma(\cdot)$ in \eqref{Gamma def} is $\Gamma(\beta) = \{(X,Y) \mid \| X\|_F \leq \beta, \| Y\|_F \leq \beta \}$).
Since $(X_{t}, Y_{t}) \in  \Gamma( \beta_T ) \subseteq  \Gamma(c_{1})$
and $\Gamma(c_{1})$ is a convex set, we have that the line segment connecting $\bm x_t $ and $\bm x_{t+1}$, denoted as $[\bm x_t, \bm x_{t+1}]$, lies in $\Gamma(c_{1})$.
Then by Claim \ref{claim: Lip constant} we have that $\nabla \tilde{F}$ is Lipschitz continuous in $[\bm x_t, \bm x_{t+1}]$ with Lipschitz constant
\begin{equation}\label{L1 def}
L_1 = L(c_{1}) = 4 c_1^2 + 54 \rho \frac{c_1^2}{\beta_1^4}  \geq L(\beta_T ) \geq 4 \beta_T^2,
\end{equation}
where the last inequality is due to the fact $c_1 \geq \beta_T$.
Define (note $c_1$ is defined by \eqref{c 1 def})
\begin{equation}\label{eta 1 def}
   \bar{\eta}_1 \triangleq \frac{ 1 }{ L_1} = \frac{ 1 }{ 4 c_1^2 + 54 \rho \frac{c_1^2}{\beta_1^4} } , % \leq \frac{2}{L(\beta_T)},
\end{equation}
then $ \bar{\eta}_1 \leq  \frac{ 1 }{L(\beta_T)} \leq   \frac{ 1 }{ 4 \beta_T^2 } =  \frac{1}{ 4\beta_T^2 } $, which is consistent with \eqref{eta 1 first bound}.

It follows from a classical descent lemma (see, e.g., \cite[Prop. A.24]{bertsekas1999nonlinear}) that
%\begin{equation}
%\tilde{F}(\bm x^{\prime} )
% \leq  \tilde{F}(\bm x ) + \langle \bm x^{\prime} - \bm x , \nabla \tilde{F}(\bm x) \rangle + \frac{L}{2} \|  \bm x^{\prime} - \bm x \|^2, \; \forall \bm x, \bm x^{\prime} \in K_1 \cap K_2.
% \end{equation}
%% According to Proposition \eqref{prop: decreasing implies K_1 intersect K_2}, $\bm x_k , \bm x_{k+1} \in K_1 \cap K_2$.
%Since $K_1 \cap K_2$ is a convex set, $ \bm x^{\prime} = \bm x_{k} - \lambda \eta \nabla \tilde{F}(\bm x_k) =   $
%Let  $\bm x = \bm x_k$ and (where $0 \leq \lambda \leq 1$) in the above inequality
%
%we get
\begin{equation}\label{descent bound for GD}
\begin{split}
& \tilde{F}(\bm x_t - \lambda \eta \nabla \tilde{F}(\bm x_t) ) \\
& \leq  \tilde{F}(\bm x_t ) - \langle \lambda \eta \nabla \tilde{F}(\bm x_t), \nabla \tilde{F}(\bm x_t) \rangle + \frac{L_1}{2} \|  \bm \lambda \eta \nabla \tilde{F}(\bm x_t)\|^2  \\
& =  \tilde{F}(\bm x_t ) + \|\nabla \tilde{F}(\bm x_t)\|^2 ( \frac{L_1}{2} \lambda^2 \eta^2 - \lambda \eta )  \\
& \leq \tilde{F}(\bm x_t )  - \frac{ \lambda \eta  }{2 } \|\nabla \tilde{F}(\bm x_t)\|^2  \\
& \leq \tilde{F}(\bm x_t ) \\
& \leq 2 \tilde{F}(\bm x_0 ), \ \forall \ \lambda \in [0, 1],
\end{split}
\end{equation}
where the second inequality follows from the fact that $\lambda \eta \leq \eta \leq  \bar{\eta}_1 = 1/L_1 $. %, and the last inequality follows from the induction hypothesis.
This finishes the induction step (note that $ \bm \Delta_t = \bm x_{t+1} - \bm x_t  = - \eta \nabla \tilde{F}(\bm x_t)$), thus \eqref{condtion of gradient method} is proved.
%Let $\lambda = 1$ in \eqref{F(x_k + lambda D k bound}, we get $\tilde{F}(\bm x_{k+1} ) \leq \tilde{F}(\bm x_k ), \forall k $,
%which implies $ \tilde{F}(\bm x_k ) \leq \tilde{F}(\bm x_0), \forall k$. Plugging into \eqref{F(x_k + lambda D k bound}, we get
%$ \tilde{F}(\bm x_{k} - \lambda \eta \nabla \tilde{F}(\bm x_k) )  \leq \tilde{F}(\bm x_0)$, which proves \eqref{condtion of gradient method}.

Finally, we show that Algorithm 4 (SGD) satisfies \eqref{condtion of gradient method} with $\bm x_t = (X_k, Y_k)$ representing
the produced solution after the $t$-th loop, provided that $\Omega$ satisfies \eqref{RSC of P_Omega}.
Denote $ N = |\Omega| + m + n  +2$ and $ \bm x_{k,i} = (X_{k,i}, Y_{k,i}), i=1,\dots, N $.
We prove \eqref{condtion of gradient method} by induction on $t$.
Define $\bm x_{-1} = \bm x_0$, then \eqref{condtion of gradient method} holds for $ t = 0$.
Assume \eqref{condtion of gradient method} holds for $0,1,\dots t-1$, i.e.,
$ \tilde{F}(\bm x_k + \lambda \bm \Delta_k ) \leq 2 \tilde{F}(\bm x_0), \forall \lambda \in [0,1], \text{ where } \bm \Delta_k = \bm x_{k+1} - \bm x_k$,
$0 \leq k \leq t-1$.
In particular, we have $\tilde{F}(\bm x_t) \leq 2 \tilde{F}(\bm x_0)$, which together with the assumption that
$\Omega$ satisfies \eqref{RSC of P_Omega} leads to (by Proposition \eqref{prop: decreasing implies K_1 intersect K_2})
\begin{equation}\label{xt bounded}
  \bm x_t \in K_1 \cap K_2 .
\end{equation}
Now we show that there exist constants $c_{1,i}, c_{2,i}, i= 0, 1,\dots, N$ (independent of $t$) so that  % for any $ j =1,2,\dots, N $,
\begin{subequations}\label{component grad bounded}
\begin{align}
 \max \{ \|  X_{t, i} \|_F, &  \|  Y_{t, i} \|_F \}  \leq c_{1,i},  \label{component grad bounded a)}  \\
 \max \{  \|  \nabla_X f_{i+1}(\bm x_{t,i}) \|_F, \| & \nabla_Y f_{i+1}(\bm x_{t,i-1}) \|_F \}   \leq c_{2,i}.  \label{component grad bounded b)}
\end{align}
\end{subequations}
We prove \eqref{component grad bounded} by induction on $i$.
When $i=0$, since by \eqref{xt bounded} we have $ \max \{ \|  X_{t, 0} \|_F,  \|  Y_{t, 0} \|_F \} = \max \{ \|  X_{t} \|_F,  \|  Y_{t} \|_F \}  \leq \beta_T $,
thus \eqref{component grad bounded a)} holds for $c_{1,0} = \beta_T$.

Suppose \eqref{component grad bounded a)} holds for $i$, we prove \eqref{component grad bounded b)} holds for $i$ with suitably chosen $c_{2,i}$.
Note that $f_{i+1}$ can be one of the five different functions in \eqref{component fn def}.
When $f_{i+1}$ equals some $F_{j l} $, we have
\begin{equation}\nonumber
\begin{split}
\|  \nabla_X f_{i+1}(\bm x_{t,i}) \|_F = \|  \nabla_X F_{j,l}(\bm x_{t,i}) \|_F
= |(X_{t,i}^{(j)})^T Y_{t,i}^{(l)} - M_{jl}|  \| Y_{t,i}^{(l)}  \|      \\
\leq ( \|X_{t, i} \|_F \|Y_{t,i} \|_F + M_{\rmax} ) \|Y_{t,i} \|_F
\leq (c_{1,i}^2 + M_{\rmax})c_{1,i}.
\end{split}
\end{equation}
When $f_{i+1}(X,Y)$ equals some $G_{1j}(X)$, we have (see \eqref{grad of F,G expression} for the expression of $\nabla_X G_{1j}$)
\begin{equation}\nonumber
\begin{split}
\|  \nabla_X f_{i+1}(\bm x_{t,i}) \|_F = \|  \nabla_X G_{1j}(X_{t,i}) \|_F
= \rho G_0^{\prime}(\frac{3\|  X_{t,i}^{(j)} \|^2  } {2\beta_1^2} ) \frac{3 \| X_{t,i}^{(j)} \| } {\beta_1^2} \\
\leq \rho G_0^{\prime}(\frac{3 c_{1,i}^2  } {2\beta_1^2} ) \frac{3 c_{1,i} } {\beta_1^2}
\leq \rho G_0^{\prime}(\frac{3 c_{1,i}^2  } {2\beta_T^2} ) \frac{3 c_{1,i} } {\beta_T^2}.
\end{split}
\end{equation}
When $f_{i+1}(X,Y)$ equals some $G_{3}(X)$, we have
\begin{equation}\nonumber
\begin{split}
\|  \nabla_X f_{i+1}(\bm x_{t,i}) \|_F = \|  \nabla_X G_{3}(X_{t,i}) \|_F
= \rho G_0^{\prime}(\frac{3\|  X_{t,i} \|_F^2  } {2\beta_T^2} ) \frac{3 \| X_{t,i} \|_F } {\beta_T^2}   \\
\leq \rho G_0^{\prime}(\frac{3 c_{1,i}^2  } {2\beta_T^2} ) \frac{3 c_{1,i} } {\beta_T^2}.
\end{split}
\end{equation}
When $f_{i+1}(X,Y)$ equals some $G_{2j}(Y) $ or $G_4(Y)$ that only depend on $Y$, we have  $\nabla_X f_{i+1}(\bm x_{t,i}) = 0 .$
Let $$ c_{2,i} \triangleq \max \left\{(c_{1,i}^2 + M_{\rmax})c_{1,i}, \;\;  \rho G_0^{\prime}(\frac{3 c_{1,i}^2  } {2\beta_T^2} ) \frac{3 c_{1,i} } {\beta_T^2} \right\},$$
then no matter what kind of function $f_{i+1}$ is, we always have
$ \|  \nabla_X f_{i+1}(\bm x_{t,i}) \|_F \leq c_{2,i}$.
Similarly, $ \|  \nabla_Y f_{i+1}(\bm x_{t,i}) \|_F \leq c_{2,i}$. Thus \eqref{component grad bounded b)} holds for $i$.

Suppose \eqref{component grad bounded b)} holds for $i-1$, we prove that \eqref{component grad bounded a)} holds for $i$ with suitably chosen $c_{1,i}$.
In fact,
\begin{align*}
  \| X_{t, i} \|_F = \| X_{t, i-1} - \eta_t \nabla_X f_i(\bm x_{t,i-1}) \|_F  \\
  \leq \| X_{t, i-1} \|_F + \eta_t \| \nabla_X f_i(\bm x_{t,i-1}) \|_F
 % \leq c_{1, i-1} + \eta_t c_{2,i-1}
  \leq c_{1, i-1} + \bar{\eta} c_{2,i-1},
\end{align*}
thus \eqref{component grad bounded a)} holds for $c_{1,i} = c_{1, i-1} + \bar{\eta} c_{2,i-1}$.
This finishes the induction proof of \eqref{component grad bounded}.

In Claim \ref{claim: Lip constant}, we have proved that $ \nabla \tilde{F}$ is Lipschitz continuous with
Lipschitz constant $L(\beta_0) = 4\beta_0 + 54 \rho \frac{\beta_0^2}{\beta_1^4}   $ in the set $\Gamma(\beta_0)$ (the definition of $\Gamma(\cdot)$ is given
in \eqref{Gamma def}).
By a similar argument (or set irrelevant rows of $X,Y,U,V$ to zero in the proof of Claim \eqref{claim: Lip constant}), we can prove that each $ \nabla f_i $ is also Lipschitz continuous with
Lipschitz constant $L(\beta_0) = 4\beta_0 + 54 \rho \frac{\beta_0^2}{\beta_1^4}   $ in the set $\Gamma(\beta_0)$. Then we have
\begin{equation}\label{fi grad is Lip cts}
\| \nabla f_i(\bm x_{t, i-1}) - \nabla f_i(\bm x_t) \|_F \leq c_{i-1}^{\prime} \| \bm x_{t, i-1} - \bm x_t \|_F, \quad i=1,\dots, N,
\end{equation}
where $ c_{i-1}^{\prime} = L(c_{1,i-1})  $.

Note that $ \bm x_{t+1} =  \bm x_t  + \sum_{i=1}^{N} (\bm x_{t,i} - \bm x_{t, i-1})
=  \bm x_t  - \eta_t \sum_{i=1}^{N} \nabla f_i(\bm x_{t, i-1}) $.
We can express SGD as an approximate gradient descent method:
\begin{equation}\label{approx GD}
  \bm x_{t+1} = \bm x_t - \eta_t ( \nabla \tilde{F}(\bm x_t) + w_t  ),
\end{equation}
where the error
$$
  w_t = \sum_{i=1}^{N} \nabla f_i(\bm x_{t, i-1}) -  \nabla \tilde{F}(\bm x_t)
   = \sum_{i=1}^{N}( \nabla f_i(\bm x_{t, i-1}) - \nabla f_i(\bm x_t)  ).
$$
Following the analysis in \cite[Lemma 1]{luo1994analysis}, we can bound each term $\nabla f_i(\bm x_{t, i-1}) - \nabla f_i(\bm x_t)$ as
\begin{align*}
 \| \nabla f_i(\bm x_{t, i-1}) - \nabla f_i(\bm x_t) \|_F \overset{\eqref{fi grad is Lip cts}}{\leq}
 c_{i-1}^{\prime} \| \bm x_{t, i-1} - \bm x_t \|_F   \\
 = \eta_t c_{i-1}^{\prime} \| \sum_{l=1}^{i-1} \nabla f_l(\bm x_{t, l-1})   \|_F
 \overset{\eqref{component grad bounded b)}}{\leq} \eta_t c_{i-1}^{\prime} \sum_{l=1}^{i-1} \sqrt{2} c_{2,l}.
\end{align*}
Plugging this inequality for $i=1,\dots, N$ into the expression of $w_t$, we obtain an upper bound of the error $w_t$:
\begin{equation}\label{error control}
  \| w_t\|_F \leq \eta_t c_0,
\end{equation}
where $c_0 \triangleq \sum_{i=1}^{N} (c_{i-1}^{\prime} \sum_{l=1}^{i-1} \sqrt{2}c_{2,l}) $ is a constant.

Applying \eqref{component grad bounded a)} for $i=N$, we get $ \max \{ \| X_{t+1}\|_F , \| Y_{t+1}\|_F \} \leq c_{1,N} $,
thus $ \bm x_{t+1} \in \Gamma(c_{1,N})$. Since $\bm x_t \in  \Gamma( \beta_T ) \subseteq  \Gamma(c_{1,N})$
and $\Gamma(c_{1,N})$ is a convex set, we have that the line segment connecting $\bm x_t $ and $\bm x_{t+1}$ lies in $\Gamma(c_{1,N})$.
Then by Claim \ref{claim: Lip constant} we have that $\nabla \tilde{F}$ is Lipschitz continuous over this line segment with Lipschitz constant $L^{\prime} = L(c_{1,N})$.
It follows from a classical descent lemma (see, e.g., \cite[Prop. A.24]{bertsekas1999nonlinear}) that
\begin{equation}\nonumber
\begin{split}
  \tilde{F}(\bm x_{t+1}) \leq \tilde{F}(\bm x_{t})  + \langle \bm x_{t+1} - \bm x_t, \nabla \tilde{F}(\bm x_t) \rangle + \frac{L^{\prime} }{2}\|  \bm x_{t+1} - \bm x_t \|_F^2.
\end{split}
\end{equation}
Using the expression \eqref{approx GD}, the above relation becomes
\begin{equation}\label{Ft+1 - Ft interm bound}
\begin{split}
\tilde{F}(\bm x_{t+1}) - \tilde{F}(\bm x_{t}) \leq
 - \eta_t \langle \bm \nabla \tilde{F}(\bm x_t) + w_t, \nabla \tilde{F}(\bm x_t) \rangle
+ \frac{ L^{\prime} }{2} \eta_t^2 \| \nabla \tilde{F}(\bm x_t) + w_t \|_F^2.
\end{split}
\end{equation}
Plugging
\begin{align*}
-\eta_t \langle  w_t, \nabla \tilde{F}(\bm x_t) \rangle \leq \eta_t \| w_t\|_F \|\nabla \tilde{F}(\bm x_t) \|_F   \\
\overset{\eqref{error control}}{\leq} \eta_t^2 c_0 \| \nabla \tilde{F}(\bm x_t) \|_F \leq \frac{1}{2} \eta_t^2 c_0 ( 1 + \| \nabla \tilde{F}(\bm x_t) \|_F^2 )
\end{align*}
and
$$ \frac{1}{2}\| \nabla \tilde{F}(\bm x_t) + w_t \|_F^2 \leq  \| \nabla \tilde{F}(\bm x_t)  \|_F^2 + \|  w_t \|_F^2
\overset{\eqref{error control}}{\leq} \| \nabla \tilde{F}(\bm x_t)  \|_F^2 + \eta_t^2 c_0^2
 $$
 into \eqref{Ft+1 - Ft interm bound}, we get
\begin{equation}\label{Ft+1 - Ft interm bound 2}
\begin{split}
 &  \tilde{F}(\bm x_{t+1}) - \tilde{F}(\bm x_{t}) \\
   & \leq  -\eta_t \| \nabla \tilde{F}(\bm x_t)  \|_F^2 + \frac{1}{2} \eta_t^2 c_0 ( 1 + \| \nabla \tilde{F}(\bm x_t) \|_F^2 )
   + L^{\prime}  \eta_t^2 ( \| \nabla \tilde{F}(\bm x_t)  \|_F^2 + \eta_t^2 c_0^2 )   \\
   & = (\frac{1}{2} \eta_t^2 c_0 + \eta_t^2 L^{\prime} - \eta_t )  \| \nabla \tilde{F}(\bm x_t)  \|_F^2 + \eta_t^2 (\frac{1}{2}c_0 + L^{\prime} \eta_t^2 c_0^2).
\end{split}
\end{equation}
 Pick $$ \bar{\eta} \triangleq  \frac{1}{c_0 + 2 L^{\prime}} .$$
Since $\eta_t \leq \bar{\eta}$, we have $ \frac{1}{2} \eta_t^2 c_0 + \eta_t^2 L^{\prime} - \eta_t \leq -\eta_t/2 $ and
$L^{\prime} \eta_t^2 c_0^2 \leq L^{\prime}c_0^2 \frac{1}{( c_0 + 2 L^{\prime} )^2 }  \leq \frac{c_0}{8} $ (the last inequality follows from
$ (c_0 + 2 L^{\prime})^2 \geq 8 c_0 L^{\prime} $).
Plugging these two inequalities into \eqref{Ft+1 - Ft interm bound 2}, we obtain
\begin{equation}\label{Ft+1 - Ft interm bound 3}\nonumber
  \tilde{F}(\bm x_{t+1}) - \tilde{F}(\bm x_{t})  \leq \eta_t^2 c_0.
\end{equation}
By the same argument we can prove
\begin{equation}\nonumber
  \tilde{F}(\bm x_{k+1}) - \tilde{F}(\bm x_{k})  \leq \eta_k^2 c_0, \;\; k=0,1, \dots, t.
\end{equation}
Summing up these inequalities, we get
\begin{equation}\nonumber
  \tilde{F}(\bm x_{t+1})  \leq  \tilde{F}(\bm x_{0}) +  \sum_{k= 0}^t \eta_k^2 c_0 \leq \tilde{F}(\bm x_{0}) +  \eta_{\mathrm{sum}} c_0.
\end{equation}
where the last inequality follows from the assumption $\sum_{k=0}^{\infty} \eta_k^2 \leq \eta_{\mathrm{sum}}$. Pick
$$ \eta_{\mathrm{sum}} \triangleq \frac{  \tilde{F}(\bm x_{0})  }{c_0} ,$$ the above relation becomes
\begin{equation}\nonumber
  \tilde{F}(\bm x_{t+1})  \leq  2 \tilde{F}(\bm x_{0}) .
\end{equation}
By a similar argument, we can prove
\begin{equation}\nonumber
  \tilde{F}(\bm x_{t} + \lambda (\bm x_{t+1} - \bm x_t ) )  \leq  2 \tilde{F}(\bm x_{0}), \ \forall \ \lambda \in [0,1],
\end{equation}
which completes the induction.
Thus we have proved that Algorithm 4 (SGD) satisfies \eqref{condtion of gradient method} with suitably chosen $\bar{\eta}$ and $\eta_{\mathrm{sum}}$.

\subsection{Proof of Claim \ref{claim: stationary point convergence}}\label{appen: local convergence proof}
% --TO BE RECOVERED According to \cite[Proposition 1.2.1]{bertsekas1999nonlinear}, each limit point of the sequence generated by Algorithm 1 with the stepsize chosen
% --TO BE RECOVERED  by the exact minimization rule is a stationary point of problem (P1).
 For Algorithm 1 with constant stepsize $\eta < \bar{\eta}_1$ (defined in \eqref{eta 1 def}), since the objective value $\tilde{F}(\bm x_t)$
 is decreasing, we have $\tilde{F}(\bm x_t) \leq \tilde{F}(\bm x_0) $. By Proposition \ref{prop: decreasing implies K_1 intersect K_2}
 this implies that the algorithm generates a sequence in $K_1 \cap K_2$.%when the stepsize $\eta < \bar{\eta}_1$.
 {\black By Claim \ref{claim: Lip constant} and the fact $K_2 = \Gamma(\beta_T)$ (see the definitions of $K_2$ in \eqref{def of K_1, K_2}
and the definition of $\Gamma(\cdot)$ in \eqref{Gamma def}), $\nabla \tilde{F}$ is Lipschitz continuous with Lipschitz constant
$ L(\beta_T) $ over the set $K_2 $.} According to \cite[Proposition 1.2.3]{bertsekas1999nonlinear}, each limit point of the sequence generated by Algorithm 1 with constant stepsize
 $\eta < \bar{\eta}_1 \overset{\eqref{eta 1 def} }{ \leq } 2 /L(\beta_T)$ is a stationary point of problem (P1).
% Algorithm 1 with constant stepsize converges to the stationary points of problem (P1).
% (or more precisely, each limit point of the sequence generated by this method is a stationary point).

{\color{black}
We then consider Algorithm 1 with stepsize chosen by the restricted Armijo rule.
 The proof of \cite[Proposition 1.2.1]{bertsekas1999nonlinear} for the standard Armijo rule can not be directly applied, and some extra effort is needed.
% Below we present a new proof.
%  still applies since that proof
For the restricted Armijo rule,
 the procedure of picking the stepsize $\eta_k$ can be
viewed as a two-phase approach. In the first phase, we find the smallest nonnegative integer so that the distance requirement is fulfilled, i.e.\ \begin{equation}\label{i1 def}
 i_1 \triangleq \min \{ i \in \mathbb{Z}^+ \mid  d(\bm x_k(\xi^{i} s_0 ) , \bm x_0 ) \leq \frac{5}{6}\delta \},
\end{equation}
where $\mathbb{Z}^+$ denotes the set of nonnegative integers, and let $\bar{s}_k = \xi^{i_1 } s_0 $.
Since
\begin{equation}\label{3delta/2 bound}
d(\bm x_k(0), s_0 ) = d(\bm x_{k-1}, \bm x_0) \leq \frac{2}{3}\delta,
 \end{equation}
 (according to Proposition \ref{prop: K(delta) condition} and Claim \ref{Algo 1-3 satisfy conditions}), such an integer $i_1 $ must exist.
 In the second phase, find the smallest nonnegative integer
so that the reduction requirement is fulfilled, i.e.\ \begin{equation}\label{i2 def}
 i_2 \triangleq \min \{ i \in \mathbb{Z}^+ \mid  \tilde{F}(\bm x_k( \xi^{i} \bar{s}_k ) ) \leq \tilde{F}(\bm x_{k-1} ) - \sigma \xi^{i} \bar{s}_k  \| \nabla \tilde{F}(\bm x_{k-1})\|_F^2 \} ,
 \end{equation}
and let $\eta_k = \xi^{i_2 } \bar{s}_k= \xi^{i_1 + i_2 } s_0$.

Note that the second phase follows the same procedure as the standard Armijo rule  (see (1.11) of \cite{bertsekas1999nonlinear}).
Hence the difference between the standard Armijo rule and the restricted Armijo rule can be viewed as the following: in each iteration the former starts from a fixed initial stepsize $s$ while the latter starts from a varying initial stepsize $ \bar{s}_k $. We notice that the proof of \cite[Proposition 1.2.1]{bertsekas1999nonlinear} does not require the initial stepsizes to be constant, but rather the following property:
if the final stepsize $\eta_k$ goes to zero for a subsequence $k \in \mathcal{K}$, then for large enough $k \in  \mathcal{K}$
the initial stepsize must be reduced at least once (see the remark after (1.17) in \cite{bertsekas1999nonlinear}).
This property also holds when the initial stepsize is lower bounded (asymptotically).
%thus the the proof of \cite[Proposition 1.2.1]{bertsekas1999nonlinear} can be applied to the restricted Armijo rule.
In the following, we will prove that for the restricted Armijo rule the initial stepsize $ \bar{s}_k $ is lower bounded (asymptotically),
and  then show how to apply the proof of \cite[Proposition 1.2.1]{bertsekas1999nonlinear} to the restricted Armijo rule.

% in fact, as long as the sequence $\{ \bar{s}_k \} $
%Assume the conclusion does not hold, i.e.\ there exists a cluster point $\bar{\bm x}$ of the sequence $\{\bm x_k \}$ such that
%$ \nabla \tilde{F}(\bar{\bm x} ) >0 $. Assume the subsequence $\{\bm x_{k} \}_{k \in \mathcal{K}} $ converges to $\bar{\bm x}$.
%If the set  $\{ \bar{s}_k \mid k \in \mathcal{K}, \bar{s}_k < s_0  \} $ is a finite set,
%then there exists an integer $N$ so that $ \bar{s}_k = s_0  $
% is just the whole sequence $\{ \bm x_k \}$.  Without loss of generality, we
% the norm of the gradients at these points are lower bounded.
We first prove that the sequence $\{ \bar{s}_k \}$ is lower bounded (asymptotically), i.e.\  \begin{equation}\label{s lower bounded}   % , k\in \mathcal{K}
\liminf_{k \rightarrow \infty } \bar{s}_k > 0 .
 \end{equation}
Assume the contrary that $\liminf_{k \rightarrow \infty} \bar{s}_k = 0  $, i.e.\ there exists a subsequence
$ \{ \bar{s}_k \}_{k\in \mathcal{K} } $ that converges to zero. % , where $\mathcal{K}_1 \subseteq \mathcal{K}$.
Since $s_0$ is a fixed scalar, we can assume $ \bar{s}_k < s_0, \forall k \in \mathcal{K} $, thus the corresponding $i_1 > 0 $ for all $k \in \mathcal{K}$.
By the definition of $i_1$ in \eqref{i1 def}, we know that $i_1 - 1$ does not satisfy the distance requirement; in other words, we have
$$ d(\bm x_k( \xi^{-1} \bar{s}_k ) , \bm x_0 ) > \frac{5}{6}\delta . $$
Denote $g_{k-1} \triangleq \nabla \tilde{F}(\bm x_{k-1} )$, then the above relation becomes
\begin{align*}
   \frac{5}{6}\delta < d(\bm x_{k-1} - \xi^{-1} \bar{s}_k g_{k-1} , \bm x_0 )
 \leq d(\bm x_{k-1} , x_0 ) + \xi^{-1} \bar{s}_k \| g_{k-1} \|_F  \\
       \overset{\eqref{3delta/2 bound}}{\leq} \frac{2}{3}\delta  + \xi^{-1} \bar{s}_k \| g_{k-1} \|_F ,
\end{align*}
       implying
       $$
 \quad\quad\quad  \quad  \frac{1}{6 }\xi \delta  \leq   \bar{s}_k \| g_{k-1} \|_F .
$$
Since $ \frac{1}{6  }\xi \delta $ is a constant and $\{ \bar{s}_k \}_{k\in \mathcal{K} } $ converges to zero,
the above relation implies that $\{ \| g_{k-1} \|_F \}_{k\in \mathcal{K} }  $ goes to infinity.
However, it is easy to verify that $ \| g_{k-1} \|_F = \| \nabla \tilde{F}(\bm x_{k-1} ) \|_F $ is bounded above by a universal constant
when $\| \bm x_{k-1} \|_F \leq \beta_T $ (note that $\| \bm x_{k-1} \|_F \leq \beta_T $ holds due to Proposition \ref{prop: K(delta) condition} and Claim \ref{Algo 1-3 satisfy conditions})), which is a contradiction. Therefore,  \eqref{s lower bounded} is proved.

Now we prove that each limit point of the sequence $\{ \bm x_k \}$ generated by Algorithm 1 with restricted Armijo rule is a stationary point.
Assume the contrary that there exists a limit point $\bar{\bm x}$ with $\nabla \tilde{F}(\bar{\bm x}) \neq 0$, and
suppose the subsequence $\{ \bm x_k\}_{k \in \mathcal{K}}$ converges to $\bar{\bm x}$. By the same argument as that for \cite[Proposition 1.2.1]{bertsekas1999nonlinear}, we can prove that the subsequence of final stepsizes $\{ \eta_k \}_{ k \in \mathcal{K} } \rightarrow 0$ (see the inequality before (1.17) in \cite{bertsekas1999nonlinear}).
Since $\{ \bar{s}_k \}$ is lower bounded (asymptotically), we must have that
$ \bar{s}_k > \eta_k , \ \forall \ k \in  \mathcal{K}, k \geq \bar{k} $ for large enough $ \bar{k} $. Thus the corresponding $i_2 > 0 $ for all $k \in \mathcal{K}, k \geq \bar{k}$.
By the definition of $i_2$ in \eqref{i2 def}, we know that $i_2 - 1$ does not satisfy the reduction requirement; in other words, we have
$
\tilde{F}(\bm x_k( \eta_k \xi^{-1} ) ) > \tilde{F}(\bm x_{k-1} ) - \sigma \eta_k \xi^{-1}  \| \nabla \tilde{F}(\bm x_{k-1})\|_F^2,
$
or equivalently,
\begin{align*}
 \tilde{F}(\bm x_{k-1} ) - \tilde{F}(\bm x_{k-1} -  \eta_k \xi^{-1} \nabla \tilde{F}(\bm x_{k-1})  ) ) <  \\
   \sigma \eta_k \xi^{-1}  \| \nabla \tilde{F}(\bm x_{k-1})\|_F^2, \ \forall \ k \in  \mathcal{K}, k \geq \bar{k} .
\end{align*}
This relation is the same as (1.17) in \cite{bertsekas1999nonlinear} (except that (1.17) in \cite{bertsekas1999nonlinear} considers a more general descent direction), and the rest of the proof is also the same as \cite{bertsekas1999nonlinear}
and is omitted here.

% In fact, in the proof of \cite[Proposition 1.2.1]{bertsekas1999nonlinear} the initial stepsize is a constant that
% This implies that in the first phase the stepsize is reduced at least once.
% Therefore, each limit point of the sequence generated by Algorithm 1 with the second stepsize rule is a stationary point.

For Algorithm 1 with stepsize chosen by the restricted line search rule, since it ``gives larger reduction in cost at each iteration'' than
the restricted Armijo rule, it ``inherits the convergence properties'' of the restricted Armijo rule
(as remarked in the last paragraph of the proof of \cite[Proposition 1.2.1]{bertsekas1999nonlinear}).
The rigorous proof is similar to that in the second last paragraph of the proof of \cite[Proposition 1.2.1]{bertsekas1999nonlinear}) and is omitted here.  }

Algorithm 2 is a two-block BCD method to solve problem (P1). According to \cite[Corollary 2]{grippo2000convergence}, each limit point of the sequence generated by Algorithm 2 is a stationary point of problem (P1).

Algorithm 3 belongs to the class of BSUM methods \cite{razaviyayn2013unified}.
According to Proposition \ref{prop: decreasing implies K_1 intersect K_2}, the level set $\mathcal{X}^0 = \{ \bm x \mid \tilde{F}(\bm x) \leq \tilde{F}(\bm x_0) \}$ is a subset of the  bounded set $K_1 \cap K_2$, thus $\mathcal{X}^0$ is bounded. Moreover, $\mathcal{X}^0$ is a closed set, thus $\mathcal{X}^0$ is compact.{\black It is easy to verify that the objective function of each subproblem in Algorithm 3 is a convex tight upper bound of $\tilde{F}(\bm x)$ (more precisely, satisfies Assumption 2 in \cite{razaviyayn2013unified}).
It is also obvious that the objective function of each subproblem is strongly convex, thus each subproblem of Algorithm 3 has a unique solution.}
Based on these facts, it follows from \cite[Theorem 2]{razaviyayn2013unified} that each limit point of the sequence generated by Algorithm 3 is a stationary point.
% the upper bound function $\psi(\bm x_t , \bm \Delta_t ; \lambda ) = \tilde{F}(\bm x + \lambda \bm \Delta_t ) + \lambda_0  \lambda^2 \| \bm \Delta_t \|^2/2$

Algorithm 4 is a SGD method (or more precisely, incremental gradient method) with a specific stepsize rule.
According to \eqref{approx GD} and \eqref{error control} in Appendix \eqref{appen: prove Claim of algorithm property},
Algorithm 4 can be viewed as an approximate gradient descent method with bounded error.
By \cite[Proposition 1]{bertsekas2000gradient}, each limit point of the sequence generated by Algorithm 4 is a stationary point.

\section{Proof of Lemma \ref{lemma: cost-to-go}}\label{appen: cost-to-go lemma proof}

We will prove a statement that is stronger than Lemma \ref{lemma main; about local convexity}:
with probability at least $1-1/n^4$,
for any $(X,Y) \in K_1 \cap K_2 \cap K(\delta)$ and $U,V$ defined in Table \ref{table: def of U,V}, we have
\begin{equation}\label{enhanced local convexity}
\langle \nabla_X \tilde{F}(X,Y), X-U \rangle + \langle \nabla_Y \tilde{F}(X,Y), Y-V \rangle \geq \frac{p}{4} d^2 +  \frac{ 2 \sqrt{\rho} }{\Sigma_{\min}}  d \sqrt{ G(X,Y) },
\end{equation}
where $d = \| M - XY^T \|_F $.

We have already proved \eqref{phi F}, i.e. with probability at least $1-1/n^4$,
\begin{equation}\nonumber
\phi_F = \langle \nabla_X F, X-U \rangle + \langle \nabla_Y F, Y-V \rangle \geq \frac{p}{4} d^2.
\end{equation}
It remains to prove a bound on $\phi_G$, which is stronger than the bound $\phi_G \geq 0$.
Note that  $\phi_F$ depends on the observed set $\Omega$, thus the bound on $\phi_F$ holds with high probability;
in contrast,  $\phi_G$ does not depend on $\Omega$, thus the bound on $\phi_G$ always holds.
\begin{claim}\label{claim of phi >= G}
For any $(X,Y) \in K_1 \cap K_2 \cap K(\delta)$ and $U,V$ defined in Table \ref{table: def of U,V}, we have
\begin{equation}\label{phi G stronger bound}
\phi_G = \langle \nabla_X G, X-U \rangle + \langle \nabla_Y G, Y-V \rangle \geq \frac{ 2 \sqrt{\rho} }{\Sigma_{\min}}  d \sqrt{ G(X,Y) }.
\end{equation}
\end{claim}

%We rewrite \eqref{grad G decompose} and \eqref{nabla G1 G2 def} below:
%the expression of $\nabla_X G(X,Y) $ is given by
%$$
%\nabla_X G(X,Y)  =  \rho \sum_i \nabla G_{1i}( X ) +   \rho \nabla G_2( X ),
%$$
\emph{Proof of Claim \ref{claim of phi >= G}:}
By the definition of $G$ in \eqref{regularized function},
$G(X,Y) = \rho(\sum_i G_{1i}(X) + G_2(X) + \sum_j G_{3j} ( Y )  +  G_4 ( Y ) )$,
where the component functions
 \begin{equation}\label{G1 G2 def}
\begin{split}
  G_{1i} ( X )  = G_0\left(\frac{3\| X^{(i)} \|^2  } {2\beta_1^2} \right) ,\quad
  G_2 (X)    = G_0\left(\frac{3\| X\|_F^2 } {2\beta_T^2}  \right) , \quad  \\
 G_{3j} ( Y )  \triangleq G_0 \left( \frac{3\| Y^{(j)} \|^2  }{2\beta_2^2} \right) ,  \quad
 G_4 ( Y )  \triangleq G_0 \left(\frac{3\| Y \|_F^2 } {2\beta_T^2}  \right) .
\end{split}
\end{equation}
%Then the gradients of the component functions are
% \begin{equation}\label{nabla G1 G2 def, again}
%\begin{split}
%  \nabla G_{1i} ( X )  = G_0^{\prime} \left(\frac{3\| X^{(i)} \|^2  } {2\beta_1^2} \right) \frac{3 \bar{X}^{(i)} }{\beta_1^2}, & \quad
% \nabla G_2 (X)    = G_0^{\prime} \left(\frac{3\| X\|_F^2 } {2\beta_T^2}  \right) \frac{3 X }{\beta_T^2}, \quad  \\
% \nabla G_{3j} ( Y )  \triangleq G_0 \left( \frac{3\| Y^{(j)} \|^2  }{2\beta_1^2} \right) \frac{3 \bar{Y}^{(j)} }{\beta_2^2},  &  \quad
% \nabla G_4 ( Y )  \triangleq G_0 \left(\frac{3\| Y \|_F^2 } {2\beta_T^2}  \right)  \frac{3 Y }{\beta_T^2} ,
%\end{split}
%\end{equation}
% in which $ G_0^{\prime}(z) =I_{[1,\infty]}(z) 2(z-1)  $  and $\bar{X}^{(i)}$ (resp.\ $\bar{Y}^{(j)}$) denotes a matrix with the $i$-th row being $X^{(i)} $
%  (resp.\ the $j$-th row being $Y^{(j)}$) and the other rows being zero.

By the expressions of $\nabla_X G, \nabla_Y G$ in \eqref{grad of F,G expression}, we have
\begin{equation}\label{phi G expression, again}
\begin{split}
\phi_G = \langle \nabla_X G,   X-U \rangle + \langle \nabla_Y G, Y-V \rangle &  = \\
 \rho \sum_{i=1}^m G_0'(\frac{3\| X^{(i)} \|^2 } {2\beta_1^2} )\frac{3}{\beta_1^2}\langle X^{(i)}, X^{(i)}- U^{(i)} \rangle
 & + \rho G_0'(\frac{3\| X\|_F^2 } {2\beta_T^2} ) \frac{3}{\beta_T^2}\langle X, X-U \rangle \\
   + \rho \sum_{j=1}^n G_0'(\frac{3 \| Y^{(j)} \|^2 } {2\beta_2^2} ) \frac{3}{\beta_2^2}\langle Y^{(j)}, Y^{(j)}- V^{(j)} \rangle
          &  + \rho G_0'(\frac{3\| Y \|_F^2 } {2\beta_T^2} )\frac{3}{\beta_T^2}\langle Y, Y-V \rangle,
\end{split}
\end{equation}
where $ G_0^{\prime}(z) =I_{[1,\infty]}(z) 2(z-1) = 2 \sqrt{ G_0(z) } $.

Firstly, we prove
\begin{subequations}\label{h_1, h_2 >=G}
\begin{align}
h_{1i} \triangleq G_0'(\frac{3\| X^{(i)} \|^2 } {2\beta_1^2}) \frac{3}{\beta_1^2} \langle X^{(i)}, X^{(i)}- U^{(i)} \rangle \geq
 \frac{1}{2} \sqrt{ G_{1i}(X) }   , \ \forall \ i,  \label{h_1 >=G}  \\
h_{3 j} \triangleq G_0'(\frac{3\| Y^{(j)} \|^2 } {2\beta_2^2}) \frac{3}{\beta_2^2} \langle Y^{(j)}, Y^{(j)}- V^{(j)} \rangle \geq
\frac{1}{2} \sqrt{ G_{3j}(Y) } , \ \forall \ j.  \label{h_2 >=G}
\end{align}
\end{subequations}
We only need to prove \eqref{h_1 >=G}; the proof of \eqref{h_2 >=G} is similar. We consider two cases.

 Case 1: $ \|X^{(i)} \|^2 \leq \frac{2\beta_1^2}{3}.$ Note that $\frac{3\| X^{(i)} \|^2 } {2\beta_1^2} \leq 1$ implies $
 G_0(\frac{3\| X^{(i)} \|^2 } {2\beta_1^2}) = G_0'( \frac{ 3\| X^{(i)} \|^2 } {2\beta_1^2})=0$, thus $h_{1i} = G_{1i} = 0$,
 in which case \eqref{h_1 >=G} holds.

 Case 2: $\|X^{(i)} \|^2 > \frac{2\beta_1^2}{3}.$ %We need to prove $\langle X^{(i)}, X^{(i)}- U^{(i)} \rangle \geq 0.$
By Corollary \ref{coro: summary of U,V} and the fact that $\beta_1^2 = \beta_T^2  \frac{3\mu r}{m} $, we have
 \begin{equation}\label{Ui<=Xi, introduce 3r in beta1}
 \|U^{(i)} \|^2 \leq \frac{3r\mu }{2m}\beta_T^2  \overset{ \eqref{beta 1 beta T def} }{=}  \frac{3}{4} \frac{2 \beta_1^2}{3} < \frac{3}{4} \|X^{(i)} \|^2.
 \end{equation}
 As a result, $ \frac{\sqrt{3}}{2} \langle X^{(i)}, X^{(i)} \rangle =
 \frac{\sqrt{3}}{2} \|X^{(i)}\| \|X^{(i)}\|  > \|X^{(i)}\| \|U^{(i)}\| \geq \langle X^{(i)}, U^{(i)} \rangle,$ which implies
 $\langle X^{(i)}, X^{(i)}- U^{(i)} \rangle \geq (1 - \frac{\sqrt{3}}{2} ) \|X^{(i)}\|^2
   > (1 - \frac{\sqrt{3}}{2} ) \frac{2}{3} \beta_1^2 > \frac{1}{12} \beta_1^2 $. Combining this inequality with the fact that
   $ G_0'(\frac{3\| X^{(i)} \|^2 } {2\beta_1^2}) = 2 \sqrt{ G_0\left(\frac{3\| X^{(i)} \|^2  } {2\beta_1^2} \right) }
   = 2 \sqrt{ G_{1i} ( X ) }, $ we get \eqref{h_1 >=G}.
    % $G_0'( z ) = 2\sqrt{G_0(z)},$

Secondly, we prove
\begin{equation}\label{h_2, h_4 >=G}
\begin{split}
     h_2 + h_4   \geq \frac{ 2 d}{\Sigma_{\min}} \left( \sqrt{G_2(X) } \right.  + &  \left. \sqrt{G_4(Y)}  \right)  ,  \\
\text{where } \quad   h_2 \triangleq  G_0'(\frac{3\| X\|_F^2 } {2\beta_T^2} )
& \frac{3}{\beta_T^2}\langle X, X-U \rangle,
\\  \quad  h_4 \triangleq  G_0'(\frac{3\| Y \|_F^2 } {2\beta_T^2} )
& \frac{3}{\beta_T^2}\langle Y, Y-V \rangle.
\end{split}
\end{equation}

Without loss of generality, we can assume $ \| Y\|_F \geq \| X\|_F, $ and we will apply Corollary \ref{coro: summary of U,V} to prove \eqref{h_2, h_4 >=G}.
If $\| Y\|_F <\| X\|_F$, we can apply a symmetric result of Corollary \ref{coro: summary of U,V} to prove \eqref{h_2, h_4 >=G}.
We  consider three cases.
% We consider several different cases, similar to Table \ref{table: def of U,V}.

Case 1: $ \| X\|_F \leq \|Y \|_F  \leq \sqrt{\frac{2}{3} } \beta_T.$
In this case $G_0(\frac{3\| X\|_F^2 } {2\beta_T^2} ) = G_0'(\frac{3\| X\|_F^2 } {2\beta_T^2} ) =
G_0(\frac{3\| Y \|_F^2 } {2\beta_T^2} ) = G_0'(\frac{3\| Y \|_F^2 } {2\beta_T^2} ) = 0 $, which implies $h_2 = h_4 = G_2(X) = G_4(Y) = 0$,
 thus $\eqref{h_2, h_4 >=G}$ holds.

Case 2: $ \|X \|_F \leq \sqrt{\frac{2}{3} } \beta_T < \|Y \|_F .$
Then we have $\frac{ 3 \| X \|_F^2 }{2 \beta_T^2} \leq 1$, which implies $h_2 = 0 = G_2(X)$.
By \eqref{summary of U,V (d)} in Corollary \ref{coro: summary of U,V} we have $ \| V \|_F \leq (1 - \frac{d}{\Sigma_{\min}})\| Y \|_F $,
which implies $(1 - \frac{d}{\Sigma_{\min}}) \langle Y , Y \rangle = (1 - \frac{d}{\Sigma_{\min}}) \|Y \|_F^2
\geq \| Y \|_F \| V \|_F \geq \langle Y ,V \rangle$.
This further implies \ $\langle Y, Y - V \rangle \geq \frac{d}{\Sigma_{\min}} \| Y \|_F^2 \geq
 \frac{d}{\Sigma_{\min}} \frac{2\beta_T^2}{3} $. Combined with the fact that
  $G_0'(\frac{3\| Y \|_F^2 } {2\beta_T^2} ) = 2 \sqrt{G_0(\frac{3\| Y \|_F^2 } {2\beta_T^2} ) } = 2 \sqrt{G_4( Y ) } $, we get
  \begin{align*}
  h_4 =  G_0'(\frac{3\| Y \|_F^2 } {2\beta_T^2} )\frac{3}{\beta_T^2}\langle Y, Y-V \rangle   \\
  \geq 2 \sqrt{G_4( Y ) }   \frac{3}{\beta_T^2}   \frac{d}{\Sigma_{\min}} \frac{2\beta_T^2}{3}
   = \frac{ 4 d}{\Sigma_{\min}} \sqrt{G_4(Y)}.
    \end{align*}
Thus $h_2 + h_4  = h_4 \geq \frac{ 4 d}{\Sigma_{\min}} \sqrt{G_4(Y)} = \frac{ 4 d}{\Sigma_{\min}} \left( \sqrt{G_4(Y)} + \sqrt{G_2(X) } \right) \geq \frac{ 2 d}{\Sigma_{\min}} \left( \sqrt{G_4(Y)} + \sqrt{G_2(X) } \right) . $ % \geq -pd^2/8.$

Case 3: $  \sqrt{\frac{2}{3} } \beta_T < \| X\|_F \leq \|Y \|_F $.
Since $\| Y\|_F \geq \| X \|_F $, we have $G_4(Y) = G_0 \left(\frac{3\| Y \|_F^2 } {2\beta_T^2}  \right)  \geq G_0 \left(\frac{3\| X \|_F^2 } {2\beta_T^2}  \right)  = G_2 (X)$.
By Corollary \ref{coro: summary of U,V}, we have $ \| U\|_F \leq \|X \|_F$ and $\| V \|_F \leq (1 - \frac{d}{\Sigma_{\min}})\| Y\|_F$.
Similar to the argument in Case 2 we can prove $h_2 \geq 0, h_4 \geq \frac{ 4 d}{\Sigma_{\min}} \sqrt{G_4(Y)} $;
thus $h_2 + h_4 \geq \frac{ 4 d}{\Sigma_{\min}} \sqrt{G_4(Y)} \geq \frac{ 2 d}{\Sigma_{\min}} \left( \sqrt{G_4(Y)} + \sqrt{G_2(X) } \right) $. % , which proves \eqref{h_2, h_4 >=G}.

In all three cases, we have proved \eqref{h_2, h_4 >=G}, thus \eqref{h_2, h_4 >=G} holds.
% Combining \eqref{h_1, h_2 >=0} and \eqref{h_2, h_4 >=0}, we obtain

We conclude that for $U, V$ defined in Table \ref{table: def of U,V},
\begin{equation}
\begin{split}
 \phi_G & \overset{ \eqref{phi G expression, again}}{=} \rho \left(  \sum_i h_{1i}  + \sum_j h_{3j} + h_2 + h_4 \right)   \\
& \overset{\eqref{h_1, h_2 >=G},\eqref{h_2, h_4 >=G}}{\geq}
\rho \left( \frac{1}{2} \right. \sum_i \sqrt{G_{1i}(X) } + \frac{1}{2} \sum_j \sqrt{G_{2j}(Y) }
\\ & \quad \quad \quad
 + \frac{ 2 d}{\Sigma_{\min}}  \sqrt{G_2(X) }  + \frac{ 2 d}{\Sigma_{\min}}  \left.  \sqrt{G_4(Y)}  \right)    \\
 & \geq \rho \frac{ 2 d}{\Sigma_{\min}}  \left(  \sum_i \sqrt{G_{1i}(X) } + \sum_j \sqrt{G_{2j}(Y) } +  \sqrt{G_2(X) }  + \sqrt{G_4(Y)}  \right)  \\
 & \geq \rho \frac{ 2 d}{\Sigma_{\min}} \sqrt{  \sum_i G_{1i}(X)  +  \sum_j G_{2j}(Y) + G_2(X) + G_4(Y) }  \\
 & = \rho  \frac{ 2 d}{\Sigma_{\min}} \sqrt{\frac{1}{\rho} G(X,Y) }
 =  \frac{ 2 \sqrt{\rho} }{\Sigma_{\min}}  d \sqrt{ G(X,Y) }  .
\end{split}
\end{equation}
which finishes the proof of Claim \ref{claim of phi >= G}.  $ \quad \quad \Box$

Let us come back to the proof of Lemma \ref{lemma: cost-to-go}.
The rest of the proof is just algebraic computation.
According to \eqref{enhanced local convexity}, we have %   according to \eqref{summary of U,V (b)} and Cauchy-Schwartz inequality, we have
\begin{equation}\nonumber
\begin{split}
& \frac{p}{4} d^2 +  \frac{ 2 \sqrt{\rho} }{\Sigma_{\min}}  d \sqrt{ G(X,Y) }  \\
 & \leq      \langle \nabla_X \tilde{F}(X,Y), X-U \rangle + \langle \nabla_Y \tilde{F}(X,Y), Y-V \rangle        \\
 & \leq  ( \| \nabla_X \tilde{F}(X,Y) \|_F + \| \nabla_Y \tilde{F}(X,Y)\|_F) \max \{ \| X-U\|_F, \|Y-V\|_F \}         \\
 & \overset{\eqref{summary of U,V (b)} }{\leq}  \sqrt{2} \sqrt{ \| \nabla_X \tilde{F}(X,Y) \|_F^2 + \| \nabla_Y \tilde{F}(X,Y)\|_F^2 }
 \frac{17}{2} \sqrt{r} \frac{\beta_T}{\Sigma_{\min}}   d    \\
 & = \| \nabla \tilde{F}(X,Y) \|_F  \frac{17}{\sqrt{2} } \sqrt{r} \frac{\beta_T}{\Sigma_{\min}}   d .
\end{split}
\end{equation}
Eliminating a factor of $d$ from both sides and taking square, we get
\begin{equation}\label{temp of cost to go}
\begin{split}
  \| \nabla \tilde{F}(X,Y) \|_F^2 \frac{289}{2} r \frac{ \beta_T^2 }{\Sigma_{\min}^2 }
  \geq \left( \frac{p}{4} d + \frac{ 2 \sqrt{\rho} }{\Sigma_{\min}}   \sqrt{ G(X,Y) } \right)^2
 \\  \geq \frac{pd^2}{16} + \frac{ 4 \rho }{\Sigma_{\min}^2 } G(X,Y) .
  \end{split}
\end{equation}
By the definition of $\beta_T$ in \eqref{beta 1 beta T def}, we have
\begin{equation}\nonumber
   r \frac{ \beta_T^2 }{\Sigma_{\min}^2 } = r \frac{ C_T r \Sigma_{\max} }{ \Sigma_{\min}^2 } = C_T \frac{r^2 \kappa}{ \Sigma_{\min}}.
 \end{equation}
According to Claim \ref{lemma: P Omega and P has relation}, we have
\begin{equation}\nonumber
 pd^2 = p\| M - XY^T \|_F^2  \geq
 \frac{1}{2}\| \bP_{\Omega}(M - XY^T) \|_F^2 = F(X,Y) .
 \end{equation}
By the definition of $\rho$ in \eqref{rho definition throughout} and the definition of $\delta_0$ in \eqref{delta definition throughout}, we have
$$
 \frac{ 4 \rho }{\Sigma_{\min}^2 } = \frac{4 }{\Sigma_{\min}^2 } 8 p \delta_0^2
 = \frac{32p}{\Sigma_{\min}^2} \frac{1}{36} \frac{\Sigma_{\min}^2 }{ C_d^2 r^3 \kappa^2 }
 = \frac{8 }{ 9 } \frac{1}{ C_d^2 r^3 \kappa^2 } p.
$$
Substituting the above three relations into \eqref{temp of cost to go}, we get (when $C_d \geq 32/3$)
\begin{equation}\nonumber
\begin{split}
  \| \nabla \tilde{F}(X,Y) \|_F^2 \frac{289}{2}  C_T \frac{r^2 \kappa}{ \Sigma_{\min}}
  \geq \frac{p}{32} F(X,Y) +  \frac{8 }{ 9 } \frac{1}{ C_d^2 r^3 \kappa^2 } p G(X,Y)   \\
  \geq \frac{8 }{ 9 } \frac{1}{ C_d^2 r^3 \kappa^2 } p ( F(X,Y) + G(X,Y) )
  = \frac{8 }{ 9 } \frac{1}{ C_d^2 r^3 \kappa^2 } p \tilde{F}(X,Y).
\end{split}
\end{equation}
This can be further simplified to
$$
  \| \nabla \tilde{F}(X,Y) \|_F^2  \geq  \frac{\Sigma_{\min} }{C_g r^5 \kappa^3 } p \tilde{F}(X,Y),
$$
where the numerical constant $C_g = \frac{ 2601 }{16} C_T C_d^2  $.
This finishes the proof of Lemma \ref{lemma: cost-to-go}.

% \section{REFERENCES}
\bibliographystyle{IEEEbib}
\vspace{0.3cm}
% \bibliography{refs}

{\footnotesize
\bibliography{refs}
}

\end{document}